\documentclass{svjour3}                   
\smartqed  

\usepackage{amsmath,amssymb,graphicx}
\usepackage{epsfig}
\usepackage{lscape}
\usepackage{footmisc}
\usepackage{rotating}

\begin{document}
\title{Decomposition into Low-rank plus Additive Matrices for Background/Foreground Separation: A Review for a Comparative Evaluation with a Large-Scale Dataset}

\author{Thierry Bouwmans  \and Andrews Sobral \and Sajid Javed \and Soon Ki Jung \and El-Hadi Zahzah}

\institute{T. Bouwmans \at
           Lab. MIA, Univ. La Rochelle, France \\
           Tel.: +05.46.45.72.02\\
           \email{tbouwman@univ-lr.fr}   
           \and
           A. Sobral \at
           Lab. L3i - Lab. MIA, Univ. La Rochelle, France
           \and     
           S. Javed \at
           Kyungpook National University, Republic of Korea
           \and     
           S. Jung \at
           Kyungpook National University, Republic of Korea
           \and 
           E. Zahzah \at
           Lab. L3i, Univ. La Rochelle, France         
}

\date{Received: date / Accepted: date}

\maketitle

\setlength{\skip\footins}{0.01cm}

\begin{abstract}
Background/foreground separation is the first step in video surveillance system to detect moving objects. Recent research on problem formulations based on decomposition into low-rank plus sparse matrices shows a suitable framework to separate moving objects from the background. The most representative problem formulation is the Robust Principal Component Analysis (RPCA) solved via Principal Component Pursuit (PCP) which decomposes a data matrix in a low-rank matrix and a sparse matrix. However, similar robust implicit or explicit decompositions can be made in the following problem formulations:  Robust Non-negative Matrix Factorization (RNMF), Robust Matrix Completion (RMC), Robust Subspace Recovery (RSR), Robust Subspace Tracking (RST) and Robust Low-Rank Minimization (RLRM). The main goal of these similar problem formulations is to obtain explicitly or implicitly a decomposition into low-rank matrix plus additive matrices. These formulation problems differ from the implicit or explicit decomposition, the loss function, the optimization problem and the solvers. As the problem formulation can be NP-hard in its original formulation, and it can be convex or not following the constraints and the loss functions used, the key challenges concern the design of efficient relaxed models and solvers which have to be with iterations as few as possible, and as efficient as possible. In the application of background/foreground separation, constraints inherent to the specificities of the background and the foreground as the temporal and spatial properties need to be taken into account in the design of the problem formulation. Practically, the background sequence is then modeled by a low-rank subspace that can gradually change over time, while the moving foreground objects constitute the correlated sparse outliers. Although, many efforts have been made to develop methods for the decomposition into low-rank plus additive matrices that perform visually well in foreground detection with reducing their computational cost, no algorithm today seems to emerge and to be able to simultaneously address all the key challenges that accompany real-world videos. This is due, in part, to the absence of a rigorous quantitative evaluation with synthetic and realistic large-scale dataset with accurate ground truth providing a balanced coverage of the range of challenges present in the real world. In this context, this work aims to initiate a rigorous and comprehensive review of the similar problem formulations in robust subspace learning and tracking based on decomposition into low-rank plus additive matrices for testing and ranking existing algorithms for background/foreground separation. For this, we first provide a preliminary review of the recent developments in the different problem formulations which allows us to define a unified view that we called Decomposition into Low-rank plus Additive Matrices (DLAM). Then, we examine carefully each method in each robust subspace learning/tracking frameworks with their decomposition, their loss functions, their optimization problem and their solvers. Furthermore, we investigate if incremental algorithms and real-time implementations can be achieved for background/foreground separation. Finally, experimental results on a large-scale dataset called Background Models Challenge (BMC 2012) show the comparative performance of 32 different robust subspace learning/tracking methods.

\keywords{Background Subtraction \and Foreground Detection  \and Robust Principal Component Analysis \and Robust Non-negative Matrix Factorization \and Robust Matrix Completion \and Subspace Tracking \and Low Rank Minimization}
\end{abstract}

\section{Introduction}
\label{sec:Introduction}
\begin{figure}
\begin{center}
\includegraphics[width=2.5cm]{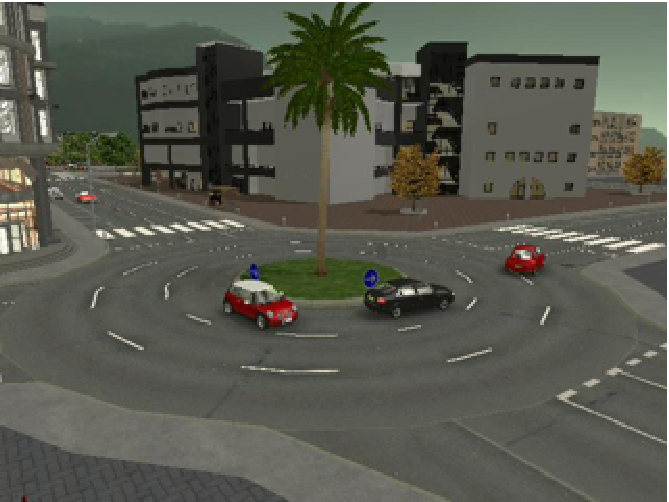}
\includegraphics[width=2.5cm]{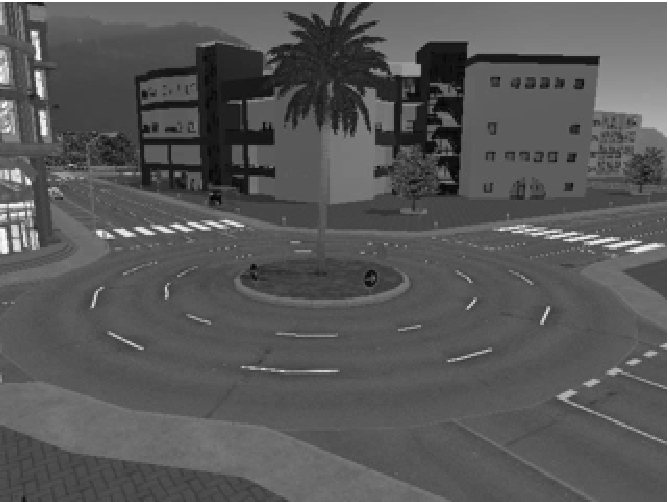}
\includegraphics[width=2.5cm]{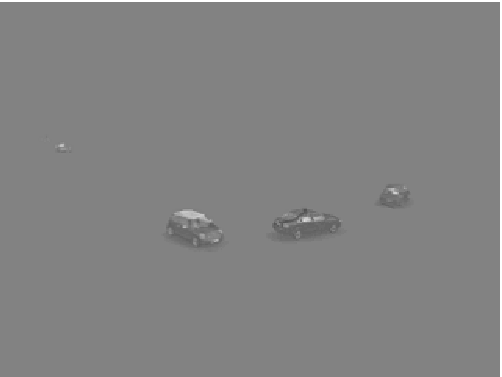}
\includegraphics[width=2.5cm]{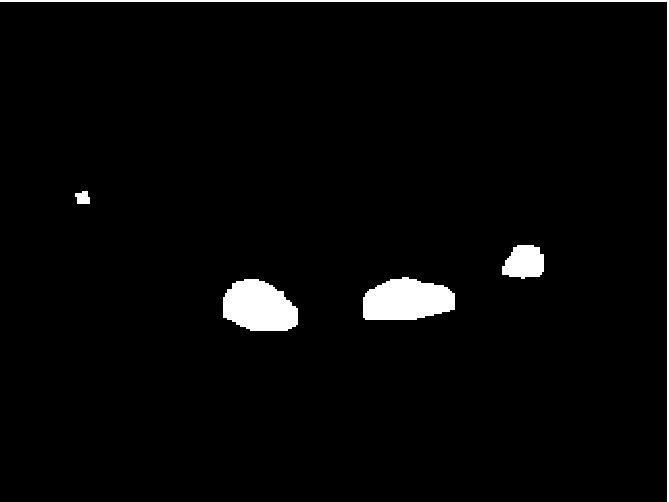} \\
\vspace{0.05cm}
\includegraphics[width=2.5cm]{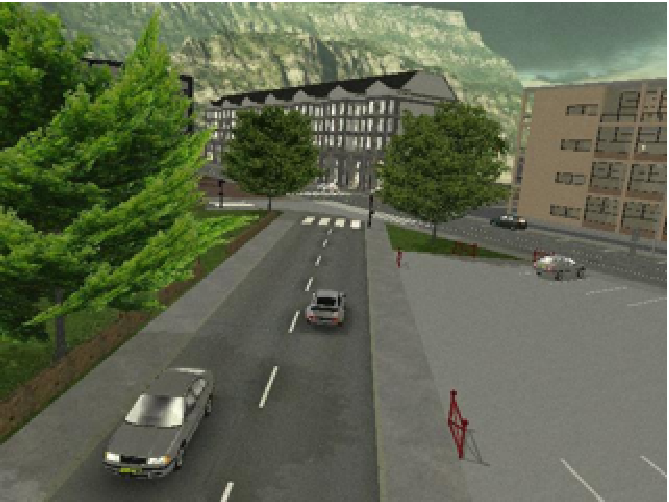}
\includegraphics[width=2.5cm]{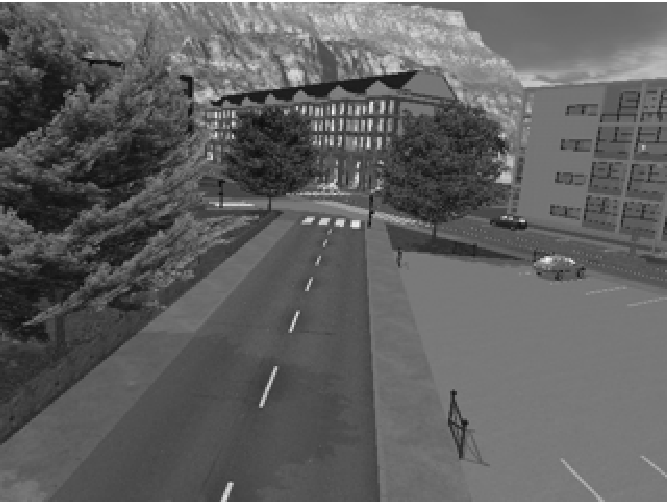}
\includegraphics[width=2.5cm]{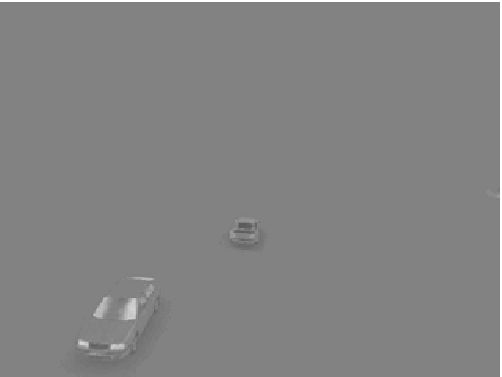}
\includegraphics[width=2.5cm]{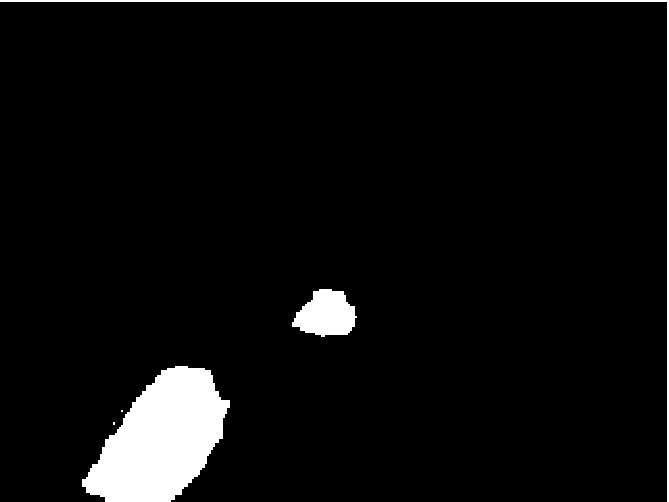} \\
\vspace{0.05cm}
\includegraphics[width=2.5cm]{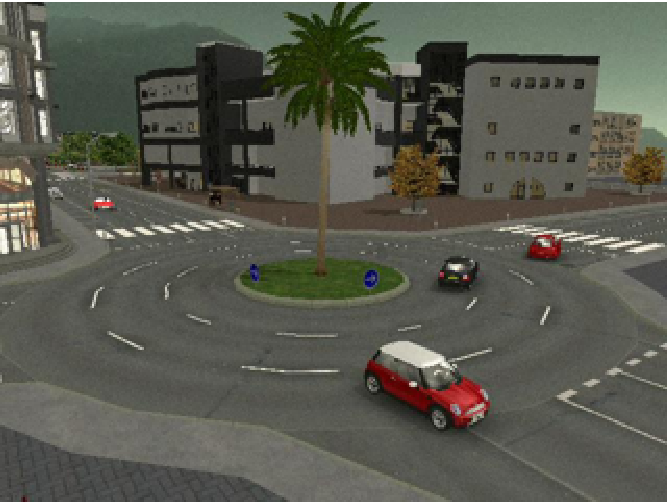}
\includegraphics[width=2.5cm]{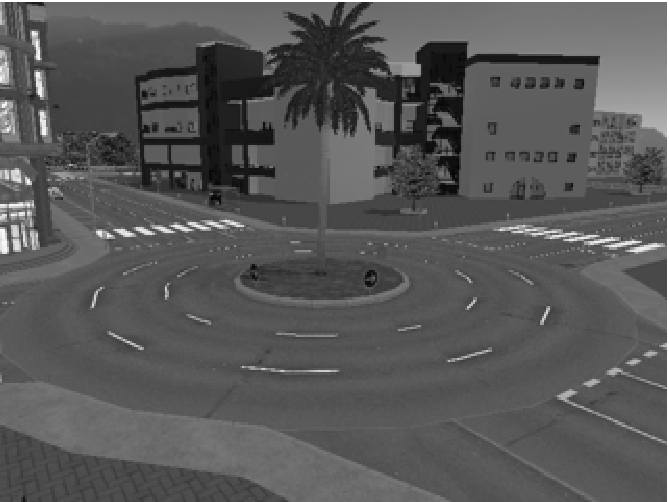}
\includegraphics[width=2.5cm]{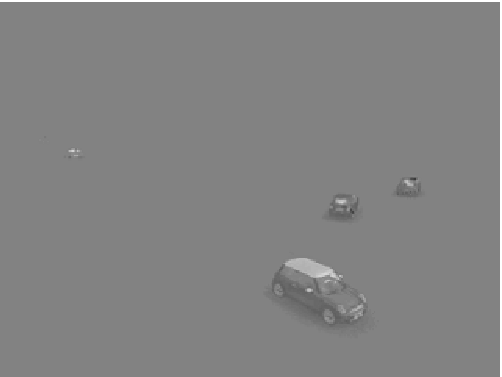}
\includegraphics[width=2.5cm]{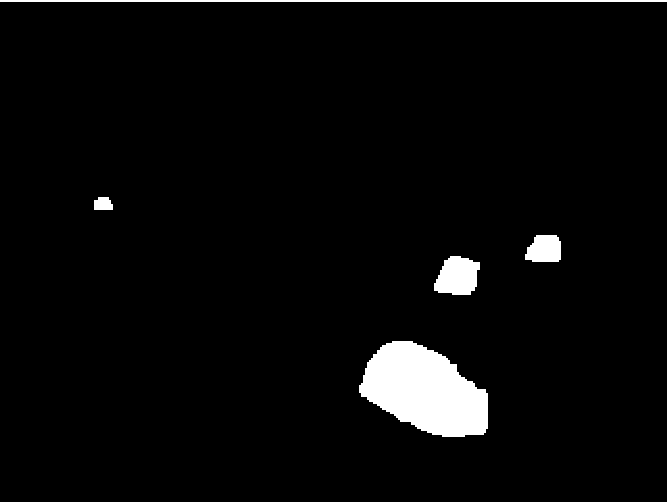} \\
\caption{RPCA via decomposition into low-rank plus sparse matrices in foreground/background separation: Original image (309), low-rank matrix $L$ (background), sparse matrix $S$ (foreground), foreground mask (Sequences from BMC 2012 dataset \cite{205}).} 
\label{BMC2012}
\end{center}
\end{figure}

The detection of moving objects is the basic low-level operation  in video analysis. This detection is usually done by using foreground detection. This basic operation consists of separating the moving objects called "foreground" from the static information called "background". Many foreground detection methods have been developed \cite{300}\cite{301}\cite{302}\cite{303}\cite{304}\cite{6000}, and several implementations are available in the BGS Library \cite{305}. Several foreground detection methods are based on subspace learning models such as Principal Component Analysis (PCA) \cite{301}. In 1999, Oliver et al. \cite{310} were the first authors to model the background by PCA. Foreground detection is then achieved by thresholding the difference between the generated background image and the current image. PCA provides a robust model of the probability distribution function of the background, but not of the moving objects while they do not have a significant contribution to the model. Although there are several PCA improvements \cite{1}\cite{2} that address the limitations of classical PCA with respect to outlier and noise, yielding to the field of robust PCA via outliers suppression, these methods do not possess the strong performance guarantees provided by the recent works on robust PCA via decomposition into low-rank plus sparse matrices \cite{3}\cite{4}\cite{3-1}\cite{1392}. The idea of this recent RPCA approach is that the data matrix $A$ can be decomposed into two components such that $A=L+S$, where $L$ is a low-rank matrix and $S$ is a matrix that can be sparse. The decomposition into low-rank plus additive matrices are used in similar problem formulations such as Robust Non-negative Matrix Factorization (RNMF), Robust Matrix Completion (RMC), Robust Subspace Recovery (RSR), Robust Subspace Tracking (RST) and Robust Low-Rank Minimization (RLRM) \cite{1050}. RNMF assumes that the matrix $L$ is a non-negative matrix. Sparsity constraints are applied on $S$ in RPCA and not in LRM. Furthermore, changes can be tracked in the subspace, that is the field of subspace tracking. Applying RPCA via decomposition into low-rank plus sparse matrices in video-surveillance, the background sequence is modeled by the low-rank subspace that can gradually change over time, while the moving foreground objects constitute the correlated sparse outliers. For example, Fig. \ref{BMC2012} shows original frames of sequences from the BMC dataset \cite{205} and its decomposition into the low-rank matrix $L$ and sparse matrix $S$. We can see that $L$ corresponds to the background whereas $S$ corresponds to the foreground. The fourth image shows the foreground mask obtained by thresholding the matrix $S$. So, the different advances in the different problem formulations of the decomposition into low-rank plus additive matrices are fundamental and can be applied to background modeling and foreground detection in video surveillance \cite{510}\cite{303}.  \\
\indent The rest of this introduction is organized as follows. Firstly, we provide a preliminary overview of the different problem formulations for the robust subspace learning/tracking frameworks which used the decomposition into low-rank plus additive matrices. Then, we present a unified view of decomposition into low-rank plus additive matrices with a discussion about its adequacy for the application of background/foreground separation. Thus, we review quickly similar decompositions such as sparse and mixed decompositions. Finally, we introduce our motivations to provide this review for a comparative evaluation in the application of background/foreground separation. 

\vspace{-0.5cm}
\subsection{Problem Formulations based on Decomposition into Low-rank plus Additive Matrices: A Preliminary Overview}
\label{subsec:Overview}
The aim of this section is to allow the reader to have a quick preliminary overview of the different robust problem formulations that are reviewed in details in the different sections of this paper. These different problem formulations based on an implicit or explicit decomposition into low-rank plus additive matrices are classified in the following categories: robust PCA, robust non-negative matrix factorization, robust subspace recovery, robust subspace tracking, robust matrix completion and robust low-rank minimization.

\subsubsection{Robust Principal Component Analysis (RPCA)}
\label{subsubsec:RPCAFramework}
Recent research in robust PCA is based on the explicit decomposition into low-rank plus sparse matrices which differs from the decomposition, the loss functions, the optimization problem and the solvers used. These different approaches can be classified as follows:  \\
\begin{enumerate}
\item \textbf{RPCA via Principal Component Pursuit (RPCA-PCP):} The first work on RPCA-PCP developed by Candes et al. \cite{3}\cite{4} and by Chandrasekharan et al. \cite{3-1} proposed the robust PCA problem as one of separating a low-rank matrix $L$ (true data matrix) and a sparse matrix $S$ (outliers' matrix) from their sum $A$ (observed data matrix). Thus, a convex optimization allowed them to address the robust PCA problem. Under minimal assumptions, this approach called Principal Component Pursuit (PCP) perfectly recovers the low-rank and the sparse matrices. The background sequence is then modeled by a low-rank subspace that can gradually change over time, while the moving foreground objects constitute the correlated sparse outliers. Therefore, Candes et al. \cite{3} showed visual results on foreground detection that demonstrated encouraging performance but PCP presents several limitations for foreground detection. The first limitation is that it required algorithms to be solved that are computational expensive. The second limitation is that PCP is a batch method that stacked a number of training frames in the input observation matrix. In real-time application such as foreground detection, it would be more useful to estimate the low-rank matrix and the sparse matrix in an incremental way quickly when a new frame comes rather than in a batch way. The third limitation is that the spatial and temporal features are lost as each frame is considered as a column vector. The fourth limitation is that PCP imposed the low-rank component being exactly low-rank and the sparse component being exactly sparse but the observations like in video surveillance are often corrupted by noise affecting every entry of the data matrix. The fifth limitation is that PCP assumed that all entries of the matrix to be recovered are exactly known via the observation and that the distribution of corruption should be sparse and random enough without noise. These assumptions are rarely verified in the case of real applications because of the following main reasons: \textbf{1)} only a fraction of entries of the matrix can be observed in some environments, \textbf{(2)} the observation can be corrupted by both impulsive and Gaussian noise, and \textbf{(3)} the outliers i.e. moving objects are spatially localized. Many efforts have been recently concentrated to develop low-computational algorithms for solving PCP \cite{18}\cite{19}\cite{20}\cite{21}\cite{22}\cite{23}\cite{4}\cite{11}\cite{38}\cite{39}\cite{46}\cite{47}. Other authors investigated incremental algorithms of PCP to update the low-rank and sparse matrix when a new data arrives \cite{14}\cite{15}\cite{16}\cite{17}. Real-time implementations \cite{12}\cite{13}\cite{41} have been developed too. Moreover, other efforts have addressed problems that appear specifically in real applications as background/foreground separation: \\

\begin{enumerate}
\item \textbf{Presence of noise:} Noise in image is due to a poor quality image source such as images acquired by a web cam or images after compression. 
\item \textbf{Quantization of the pixels:} The quantization can induce at most an error of $0.5$ in the pixel value. 
\item \textbf{Spatial and temporal constraints of the foreground pixels:} Low-rank and sparse decomposition is based on the condition that the outlier/noise can be considered as sparsity patterns and are uniformly located at the scene, which is not realistic in realworld applications as foreground moving objects are located in a connexed area. Furthermore, foreground moving objects present a continuous motion through the sequence. These two points need to introduce spatial and temporal constraints on the detection. 
\item \textbf{Local variations in the background:} Variations in the background may be due to a camera jitter or dynamic backgrounds.  \\
\end{enumerate}

To address (a), Zhou et al. \cite{5} proposed a stable PCP (SPCP) that guarantees accurate recovery in the presence of entry-wise noise. Becker et al. \cite{8} proposed an inequality constrained version of PCP to take into account the quantization error of the pixel's value (b). To address (c), Tang and Nehorai \cite{41} proposed a PCP method via a decomposition that enforces the low-rankness of one part and the block sparsity of the other part. Wohlberg et al. \cite{26} used a decomposition corresponding to a more general underlying model consisting of a union of low-dimensional subspaces for local variation in the background (d). Furthermore, RPCA is generally applied in the pixel domain by using intensity or color features but other features can be used such as depth \cite{1010-4}\cite{1542} and motion (optical flow \cite{1065}) features. Furthermore, RPCA can been extended to the measurement domain, rather than the pixel domain, for use in conjunction with compressive sensing \cite{29}\cite{30}\cite{1008}\cite{45}\cite{45-1}\cite{72}\cite{88}\cite{1049}\cite{1049-1}\cite{10490}\cite{10490-1}\cite{1700}. Although experiments show that moving objects can be reliably extracted by using a small amount of measurements, we have limited the investigation and the comparative evaluation in this paper to the pixel domain to compare with the classical background subtraction methods. \\ 

\item \textbf{RPCA via Outlier Pursuit (RPCA-OP):} Xu et al. \cite{56} proposed a robust PCA via Outlier Pursuit to obtain a robust decomposition when the outliers corrupted entire columns, that is every entry is corrupted in some columns. Moreover,
Xu et al. \cite{56} proposed a stable OP (SOP) that guarantee stable and accurate recovery in the presence of entry-wise noise. \\
\item \textbf{RPCA via Sparsity Control (RPCA-SpaCtrl):} Mateos and Giannakis \cite{9}\cite{10} proposed a robust PCA where a tunable parameter controls the sparsity of the estimated matrix, and the number of outliers as a by-product. \\
\item \textbf{RPCA via Sparse Corruptions  (RPCA-SpaCorr):} Even if the matrix $A$ is exactly the sum of a sparse matrix $S$ and a low-rank matrix $L$, it may be impossible to identify these components from the sum. For example, the sparse matrix $S$ may be low-rank, or the low-rank matrix $L$ may be sparse. To address this issue, Hsu et al. \cite{58} imposed conditions on the sparse and low-rank components in order to guarantee their identifiability from $A$. \\
\item \textbf{RPCA via Log-sum heuristic Recovery (RPCA-LHR):} When the matrix has high intrinsic rank structure or the corrupted errors become dense, the convex approaches may not achieve good performances. Then, Deng et al. \cite{55} used the log-sum heuristic recovery to learn the low-rank structure. \\
\item \textbf{RPCA via Iteratively Reweighted Least Squares (IRLS):} Guyon et al. \cite{505} proposed to solve the RPCA problem by using an Iteratively Reweighted Least Squares (IRLS) alternating scheme for matrix low-rank decomposition. Furthermore, spatial constraint can be added in the minimization process to take into account the spatial connexity of pixels \cite{506}. The advantage of IRLS over the classical solvers is its fast convergence and its low computational cost. Furthermore, Guyon et al. \cite{507} improved this scheme by addressing in the minimization the spatial constraints and the temporal sparseness of moving objects. \\
\item \textbf{RPCA via Stochastic Optimization (RPCA-SO):} Goes et al. \cite{1009} proposed a robust PCA via a stochastic optimization. For computer vision, Feng et al. \cite{1010} developed an online Robust PCA (OR-PCA) that processes one sample per time instance and hence its memory cost is independent of the number of samples, significantly enhancing the computation and storage efficiency. The algorithm is equivalent to a reformulation of the batch RPCA \cite{1009}. Therefore, Javed et. al \cite{1010-1} modified OR-PCA via stochastic optimization method to perform it on background subtraction. An initialization scheme is adopted which allows the algorithmt to converge very fastly as compared to the original OR-PCA. Therefore, OR-PCA was further improved to enhance the foreground segmentation using the continuous constraints such as with Markov Random Field (MRF) \cite{1010-2}, and using dynamic feature selection \cite{1010-3}. In an other way, Han et al. \cite{1541}\cite{1911} improved OR-PCA to be robust against camera jitter.\\
\item \textbf{RPCA with Dynamic Mode Decomposition (RPCA-DMD):} Grosek et al. \cite{2021} introduced the use of dynamic mode decomposition (DMD) for robustly separating video frames into background and foreground components in real-time. DMD \cite{2020} is a method used for modeling nonlinear dynamical systems in an equation-free manner by decomposing the state of the system into low-rank terms whose Fourier components in time are known. DMD terms with Fourier frequencies near the origin (zero-modes) are considered as background portions of the given video frames, whereas the terms with Fourier frequencies bounded away from the origin are considered as their sparse counterparts. For the approximation of the low-rank/sparse separation, it is achieved at the computational cost of one singular value decomposition and one linear equation solve. Thus, it produces results orders of magnitude faster than the original RPCA \cite{3}. Kutz et al. \cite{2022}\cite{2023}\cite{2024} improved this approach to robustly separate the background and foreground components into a hierarchy of multi-resolution time-scaled component. In an other way, Tirunagari et al. \cite{2050} applied DMD in the color domain for background initialization on the Scene Background Initialisation (SBI) dataset \cite{210}.\\
\item \textbf{Bayesian RPCA (BRPCA):} Ding et al. \cite{7} proposed a Bayesian framework which infers an approximate representation for the noise statistics while simultaneously inferring the low-rank and sparse components. Furthermore, Markov dependency is introduced spatially and temporarily between consecutive rows or columns corresponding to image frames. This method has been improved in a variational Bayesian framework \cite{24} and a factorized variational Bayesian framework \cite{94}. In a similar manner, Zhao et al. \cite{1024} developed a generative RPCA model under the Bayesian framework by modeling data noise as a mixture of Gaussians (MoG). \\
\item \textbf{Approximated RPCA:} Zhou and Tao \cite{6} proposed an approximated low-rank and sparse matrix decomposition. This method called Go Decomposition (GoDec) produces an approximated decomposition of the data matrix whose RPCA exact decomposition does not exist due to the additive noise, the predefined rank on the low-rank matrix and the predefined cardinality of the sparse matrix. GoDec is significantly accelerated by using bilateral random projection. Furthermore, Zhou and Tao \cite{6} proposed a Semi-Soft GoDec which adopts soft thresholding to the entries of $S$, instead of GoDec which imposes hard thresholding to both the singular values of the low-rank part $L$ and the entries of the sparse part $S$. \\
\item \textbf{Sparse Additive Matrix Factorization:} Nakajima et al. \cite{24-1} \cite{24-2} developed a framework called Sparse Additive Matrix Factorization (SAMF). The aim of SAMF is to handle various types of sparse noise such as row-wise and column-wise sparsity, in addition to element-wise sparsity (spiky noise) and low-rank sparsity (low-dimensional). Furthermore, their arbitrary additive combination is also allowed. In the original robust PCA \cite{3}, row-wise and column-wise sparsity can capture noise observed only in the case when some sensors are broken or their outputs are  unreliable. SAMF due to its flexibility in sparsity design incorporate side information more efficiently. In background/foreground separation, Nakajima et al. \cite{24-1} \cite{24-2} induced the sparsity in SAMF using image segmentation. \\
\item \textbf{Variational Bayesian Sparse Estimator:} Chen et al.  \cite{24-3} proposed a variational Bayesian Sparse Estimator (VBSE) based algorithm for the estimation of the sparse component of an outlier corrupted low-rank matrix, when linearly transformed composite data are observed. It is a generalization of the original robust PCA \cite{3}. VBSE can achieved background/foreground separation in blurred and noisy video sequences.
\end{enumerate}

\subsubsection{Robust Non-negative Matrix Factorization (RNMF)}
\label{subsubsec:RNMFFramework}
Non-negative matrix factorization (NMF) approximates a non-negative matrix $A$ by a product of two non-negative low-rank factor matrices $W$ and $H$. Classical NMF methods minimize either the Euclidean distance or the Kullback-Leibler divergence between $X$ and $W^TH$ to model the Gaussian noise or the Poisson noise. Practically, these methods do not perform well when the noise distribution is heavy tailed like in the background/foreground separation. To address this problem, Guan et al. \cite{761} proposed to minimize the Manhattan distance between $X$ and $W^TH$. This method called Manhattan NMF (MahNMF) robustly estimates the low-rank part and the sparse part of a non-negative matrix and performs effectively when data are contaminated by outliers. MahNMF shows similar qualitative performance as the RPCA solved via IALM \cite{3}. In an other way, Kumar et al. \cite{762-1} proposed a family of conical hull finding procedures called Xray for Near-separable NMF (NS-NMF) problems with Frobenius norm loss. However, the use of Frobenius norm approximations is not very suitable in presence of outliers or different noise characteristics. Then, Kumar and Sindhwani \cite{762} improved Xray to provide robust factorizations with respect to the $l_1$-loss function, and approximations with respect to the family of Bregman divergences. This algorithm is called RobustXray. Quantitative results \cite{762} show that RobustXray outperforms the RPCA solved via IALM \cite{3} in presence of noise. An other approach developed by Woo and Park \cite{765} used a formulation called $l_\infty$-norm based robust asymmetric nonnegative matrix factorization (RANMF) for the grouped outliers and low nonnegative rank separation problems. The main advantage of RANMF is that the denseness of the low nonnegative rank factor matrices can be controlled. Furthermore, RANMF is not sensitive to the nonnegative rank constraint parameter due to the soft regularization method. 

\subsubsection{Robust Matrix Completion (RMC)}
\label{subsubsec:RMCFramework}
The matrix completion aims at recovering a low-rank matrix from partial observations of its entries. Robust matrix completion RMC, also called RPCA plus matrix completion problem can also be used for background/foreground separation or for background initialization. RPCA via principal component pursuit \cite{3} can be considered as RMC using $l_1$-norm loss function. Following this idea, Yang et al. \cite{2001} proposed  a nonconvex relaxation approach to the matrix completion problems when the entries are contaminated by non-Gaussian noise or outliers. A nonconvex loss function based on the $l_\sigma$-norm instead of the $l_1$-norm is used with a rank constrained as well as a nuclear norm regularized model. This method can be  solved via two algorithms based on iterative soft thresholding (IST) and iterative hard thresholding (IHT). A nonconvex loss function used in robust statistics is used with a rank constrained as well as a nuclear norm regularized model. This method called RMC-$l_\sigma$-IHT is also faster than RPCA solved via IALM \cite{3}. In an other way, Shang et al. \cite{1037}\cite{1043} proposed a scalable, provable structured low-rank matrix factorization method to recover low-rank plus sparse matrices from missing and grossly corrupted data. A scalable robust bilinear structured factorization (RBF) method recovered low-rank plus sparse matrices from incomplete, corrupted data or a small set of linear measurements. In a similar way, Shang et al. \cite{2003} proposed a scalable convex model (RMC with convex formulation) and a non-convex model solved with matrix factorization (RMC-MF) in which the desired low-rank matrix $L$ is factorized into two much smaller matrices. In an other way, Mansour and Vetro \cite{2000} proposed a factorized robust matrix completion (FRMC) algorithm with global motion compensation. The algorithm decomposes a sequence of video frames into the sum of a low-rank background component and a sparse motion component. The algorithm alternates between the solution of each component following a Pareto curve trajectory for each subproblem. For videos with moving background, Mansour and Vetro \cite{2000} used the motion vectors extracted from the coded video bitstream to compensate for the change in the camera perspective. This approach is faster than state-of-the-art solvers and results in highly accurate motion segmentation. 
In a similar way, Yang et al. \cite{2002} proposed a Motion-Assisted Matrix Completion (MAMC) which used a dense motion field for each frame. This field is then mapped into a weighting matrix to assign the reliability of pixels that belong to the background. This method is robust to slowly moving objects and camouflage. Yang et al. \cite{2002} extended MAMC to a robust MAMC model (RMAMC) which is robust to noise. In a comprehensive study, Sobral et al. \cite{1710} provided a comparison of several matrix completion algorithms on the SBI dataset \cite{210}. This study was extended by Sobral and Zahzah \cite{1710-1} to tensor completion algorithms.

\subsubsection{Robust Subpace Recovery (RSR)}
\label{subsubsec:RSRFramework}
This category contains the robust decompositions other than RPCA and RNMF decompositions. First, Wang et al. \cite{1002} studied the problem of discovering a subspace in the presence of outliers and corruptions. In this context, additional knowledge is added to relax this problem as a convex programming problem. Thus, Wang et al. \cite{1002} provided a Robust Subspace Discovery (RSD) method solved via an efficient and effective algorithm based on the Augmented Lagrangian Multiplier. Since high dimensional data is supposed to be distributed in a union of low dimensional subspaces, Bian and Krim \cite{1014}\cite{1014-1} proposed a bi-sparse model as a framework to take into account that the underlying structure may be affected by sparse errors and/or outlier. So, Bian and Krim \cite{1014} provided an algorithm called Robust Subspace Recovery via bi-sparsity pursuit (RoSuRe) to recover the union of subspaces in presence of sparse corruptions. Experimental results \cite{1014} show robustness in the case of camera jitter. Conventional robust subspace recovery models address the decomposition problem by iterating between nuclear norm and sparsity minimization. However, this scheme is computationally prohibitive to achieve real time requirements. To solve this problem, Shu et al. \cite{759-1} proposed a Robust Orthogonal Subspace Learning (ROSL) method to achieve efficient low-rank recovery. A rank measure on the low-rank matrix is introduced that imposes the group sparsity of its coefficients under orthonormal subspace. Furthermore, an efficient sparse coding algorithm minimizes this rank measure and recovers the low-rank matrix at quadratic complexity of the matrix size. Finally, Shu et al. \cite{759-1} developed a random sampling algorithm to further speed up ROSL such that its accelerated version (ROSL+) has linear complexity with respect to the matrix size. Experiments \cite{759-1} demonstrate that both ROSL and ROSL+ provide more efficiency against RPCA solved via IALM \cite{3} with the same detection accuracy. In a different manner, She et al. \cite{1042} proposed a robust orthogonal complement principal component analysis (ROC-PCA). The aim is to deal with orthogonal outliers that are not necessarily apparent in the original observation space but could affect the principal subspace estimation. For this, She et al. \cite{1042} introduced a projected mean-shift decomposition and developed a fast alternating optimization algorithm on the basis of Stiefel manifold optimization and iterative nonlinear thresholdings. 

\subsubsection{Robust Subspace Tracking (RST)}
\label{subsubsec:STFramework}
Subspace tracking aims to address the problem when new observations come in asynchronously in the case of online streaming application. The algorithm cannot store in general all the input data in memory. The incoming observations must be immediate processed and then discarded. Furthermore, since the subspace can be identified from incomplete vectors, it can be subsampled in order to improve on computational efficiency, and it still retain subspace estimation accuracy. The involved subspaces can have low-rank and/or sparse structures like in the previous decomposition problem formulations. In this idea, He et al. \cite{27}\cite{28} proposed an incremental gradient descent on the Grassmannian, the manifold of all $d$-dimensional subspaces for fixed $d$. This algorithm called Grassmannian Robust Adaptive Subspace Tracking Algorithm (GRASTA) uses a robust $l_1$-norm cost function in order to estimate and track non-stationary subspaces when the streaming data vectors, that are image frames in foreground detection, are corrupted with outliers, that are foreground objects. This algorithm allows to separate background and foreground online. GRASTA shows high-quality visual separation of foreground from background. Following the idea of GRASTA, He et al. \cite{2801}\cite{2802} proposed  (transformed-GRASTA) which iteratively performs incremental gradient descent constrained to the Grassmannian manifold of subspaces in order to simultaneously estimate a decomposition of a collection of images into a low-rank subspace, a sparse part of occlusions and foreground objects, and a transformation such as rotation or translation of the image. t-GRASTA is four times faster than state-of-the-art algorithms, has half of the memory requirement, and can achieve alignment in the case of camera jitter. Although the $l_1$-norm in GRASTA leads to favorably conditioned optimization problems it is well known that penalizing with non-convex $l_0$-surrogates allows reconstruction even in the case when $l_1$-based methods fail. Therefore, Hage and Kleinsteuber \cite{69}\cite{6901} proposed an improved GRASTA using $l_0$-surrogates solving it by a Conjugate Gradient method. This method called pROST \cite{6901} outperforms GRASTA in the case of multi-modal backgrounds. An other approach developed by Xu et al. \cite{85} consists of a Grassmannian Online Subspace Updates with Structured-sparsity (GOSUS), which exploits a meaningful structured sparsity term to significantly improve the accuracy of online subspace updates. Their solution is based on Alternating Direction Method of Multipliers (ADMM), where most key steps in the update procedure are reduced to simple matrix operations yielding to real-time performance. Finally, Ahn \cite{1035} proposed a fast adapted subspace tracking algorithm which shares the procedure of separating frames into background and foreground with GRASTA, but it uses a recursive least square algorithm for subspace tracking, which makes it fast adapted to dynamic backgrounds.

\subsubsection{Robust Low Rank Minimization (RLRM)} 
\label{subsubsec:LRMFramework}
Low-rank minimization (approximation or representation) is a minimization problem, in which the cost function measures the fit between a given data matrix $A$ and an approximating matrix $L$, subject to a constraint that the approximating matrix $L$ has reduced rank. In the application of background/foreground separation, Zhou et al. \cite{25} proposed a framework called Detecting Contiguous Outliers in the Low-Rank Representation (DECOLOR) and formulated outlier detection in the robust low-rank representation, in which the outlier support and the low-rank matrix are estimated. This approach integrates object detection and background learning into a single process of optimization solved by an alternating algorithm. In a different manner, Xiong et al. \cite{54} proposed a direct robust matrix factorization (DRMF) assuming that a small portion of the matrix $A$ has been corrupted by some arbitrary outliers. The aim is to get a reliable estimation of the true low-rank structure of this matrix and to identify the outliers. To achieve this, the outliers are excluded from the model estimation. Furthermore, Xiong et al. \cite{54} proposed an extension of DRMF to deal with the presence of outliers in entire columns. This method is called DRMF-Row (DRMF-R). In an other way, Wang et al. \cite{5400} proposed a probabilistic method for robust matrix factorization (PRMF) based on the $l_1$-norm loss and $l_2$-regularizer, which bear duality with the Laplace error and Gaussian prior, respectively. 
But, PRMF treats each pixel independently with no clustering effect but the moving objects in the foreground usually form groups with high within-group spatial or temporal proximity. Furthermore, the loss function is defined based on the $l_1$-norm, and its results are not robust enough when the number of outliers is large. To address these limitations, Wang et al. \cite{5401} proposed a full Bayesian formulation called Bayesian Robust Matrix Factorization (BRMF). BRMF used a Laplace mixture with the generalized inverse Gaussian distribution as the noise model to further enhance model robustness. Furthermore, BRMF contained a Markov extension (MBRMF) which assumes that the outliers exhibit spatial or temporal proximity. In a different manner, Zheng et al. \cite{1025} added a convex nuclear-norm regularization term to improve convergence of LRM, without introducing too much heterogenous information. This method is called Practical Low-Rank Matrix Factorization (PLRMF). The previous low-rank factorization used loss functions such as the $l_2$-norm and $l_1$-norm losses. $l_2$-norm is optimal for Gaussian noise, while $l_1$-norm is for Laplacian distributed noise. Since videos are often corrupted by an unknown noise distribution, which is unlikely to be purely Gaussian or Laplacian, Meng et al. \cite{2-1} used a low-rank matrix factorization problem with a Mixture of Gaussians (LRMF-MoG) noise model. Since the MoG model is a suitable approximator for any continuous distribution, it is able to model a wider range of noise. \\

\indent Table \ref{SurveyOverview-1} and Table \ref{SurveyOverview-2} show an overview of the different problem formulations based on decomposition into low-rank plus additive matrices. The first column indicates the name of the different problem formulations and the second column shows the different categories of each problem formulations. The third column indicates the different methods of each category with their corresponding acronym. The fourth column gives the name of the authors and the date of the related publication. The previous surveys in the field are indicated in bold and the reader can refer to them for more references on the corresponding category or sub-category. \\

\indent Furthermore, we present in different tables some quick comparisons on the different key characteristics of these different problem formulations based on the decomposition into low-rank plus additive matrices. Thus, Table \ref{TPCP1Overview-1}, Table \ref{TPCP1Overview-2}, Table  \ref{TPCP2Overview} and Table \ref{TPCP3Overview} show an overview of the different decompositions into low-rank plus additive matrices in terms of minimization, constraints and convexity to allow us to define a unified view that we called Decomposition into Low-rank plus Additive Matrices (DLAM). The key characteristics of this unified view of the different problem formulations are the following ones:
\begin{itemize}
\item \textbf{Decomposition:} The form of the decomposition can be implicit or explicit. Furthermore, this decomposition can be made in two or three terms. \\
\item \textbf{Minimization problem:} The problem of the decomposition seeks to a minimization problem written in its original form or its Lagrangian form. Practically, the minimization problem is viewed as an optimization problem which can be convex or not.  \\
\item \textbf{Loss functions:} Several loss functions can be used to enforce the constraints on each matrix which composes the decomposition. Most of the time, proxy loss functions are used as surrogate of the original loss functions to obtain a solvable problem. \\
\item \textbf{Solvers:} Algorithms which are called solvers are then used to solve the optimization problem. Furthermore, instead of directly solving the original convex optimizations, some authors use their strongly convex approximations in order to design efficient algorithms. \\
\end{itemize}

\newpage
\begin{landscape}
\begin{table}
\scalebox{0.76}{
\begin{tabular}{|l|l|l|l|} 
\hline
\scriptsize{Robust Problem Formulations} &\scriptsize{Categories} &\scriptsize{Sub-categories} &\scriptsize{Authors - Dates}\\
\hline
\hline
\scriptsize{Robust Principal Components Analysis (RPCA)} &\scriptsize{Principal Component Pursuit} &\scriptsize{PCP}  &\scriptsize{Candes et al. (2009) \cite{3}} \\
\scriptsize{(\textbf{Survey Guyon et al. \cite{500}})}   &\scriptsize{(\textbf{Survey Bouwmans and Zahzah\cite{510}})} &\scriptsize{Stable PCP} 
&\scriptsize{Zhou et al. (2010) \cite{5}}\\
\scriptsize{}                         &\scriptsize{} 							&\scriptsize{Quantized PCP}    &\scriptsize{Becker et al. (2011) \cite{8}}     \\
\scriptsize{}                         &\scriptsize{} 							&\scriptsize{Block based PCP}  &\scriptsize{Tang and Nehorai (2011) \cite{41}}  \\
\scriptsize{}                         &\scriptsize{} 							&\scriptsize{Local PCP}        &\scriptsize{Wohlberg et al. (2012) \cite{26}}   \\
\cline{2-4}
\scriptsize{}                         &\scriptsize{Outlier Pursuit} 					 &\scriptsize{OP}   &\scriptsize{Xu et al. (2010) \cite{56}}  \\
\scriptsize{}                         &\scriptsize{} 					                 &\scriptsize{SOP}  &\scriptsize{Xu et al. (2010) \cite{56}}  \\
\cline{2-4}
\scriptsize{}                         &\scriptsize{Sparsity Control} 					 &\scriptsize{SpaCtrl} &\scriptsize{Mateos et al. (2010) \cite{9}}  \\
\scriptsize{}                         &\scriptsize{} 					                 &\scriptsize{SpaCorr} &\scriptsize{Hsu et al. (2011) \cite{58}} \\
\cline{2-4}
\scriptsize{}                         &\scriptsize{Non Convex Heuristic Recovery}    &\scriptsize{$l_p$ HR (pHR)}   &\scriptsize{Deng (2013) \cite{5500}}  \\   
\scriptsize{}                         &\scriptsize{}  &\scriptsize{Log-sum HR (LHR)} &\scriptsize{Deng et al. (2012) \cite{55}}      \\   

\cline{2-4}
\scriptsize{}  &\scriptsize{Iteratively Reweighted Least Square} 	&\scriptsize{IRLS} &\scriptsize{Guyon et al. (2012) \cite{505}}  \\
\scriptsize{}  &\scriptsize{} 					                 &\scriptsize{Spatial IRLS}  &\scriptsize{Guyon et al. (2012) \cite{506}}  \\
\scriptsize{}  &\scriptsize{} 					          &\scriptsize{Spatio-temporal IRLS} &\scriptsize{Guyon et al. (2012) \cite{507}}  \\
\cline{2-4}
\scriptsize{}            &\scriptsize{Stochastic Optimization} 	   &\scriptsize{RPCA-SO}         &\scriptsize{Goes et al. (2014) \cite{1009}}       \\
\scriptsize{}            &\scriptsize{} 					                 &\scriptsize{OR-PCA}         &\scriptsize{Feng et al. (2013) \cite{1010}}       \\
\scriptsize{}            &\scriptsize{} 					                 &\scriptsize{OR-PCA with Markov Random Field}  &\scriptsize{Javed et al. (2014) \cite{1010-2}}  \\
\scriptsize{}            &\scriptsize{} 					                 &\scriptsize{OR-PCA with Dynamic Feature Selection}  &\scriptsize{Javed et al. (2015) \cite{1010-3}}  \\
\scriptsize{}            &\scriptsize{} 					                 &\scriptsize{Depth-extended OR-PCA (DEOR-PCA}  &\scriptsize{Javed et al. (2015) \cite{1010-4}}  \\
\scriptsize{}            &\scriptsize{} 					                 &\scriptsize{OR-PCA with Active Random Field}  &\scriptsize{Javed et al. (2015) \cite{1010-5}}  \\
\scriptsize{}            &\scriptsize{} 					                 &\scriptsize{Max-norm Regularized Matrix Decomposition (MRMD)}  &\scriptsize{Shen et al. (2014) \cite{1110}}  \\
\scriptsize{}            &\scriptsize{} 					                 &\scriptsize{Incremental Nonnegative Matrix Factorization (INMF)}  &\scriptsize{Chen and Li \cite{16050}}  \\
\scriptsize{}            &\scriptsize{} 					                 &\scriptsize{Modified OR-PCA}      &\scriptsize{Han et al. \cite{1911}}  \\
\cline{2-4}
\scriptsize{}            &\scriptsize{RPCA with Dynamic Mode Decomposition} 	&\scriptsize{DMD}     &\scriptsize{Grosek et al. (2014) \cite{2021}}   \\
\scriptsize{}            &\scriptsize{} &\scriptsize{Multi-Resolution Time-Scale DMD}   &\scriptsize{Kutz et al. (2015) \cite{2022}}           \\
\scriptsize{}            &\scriptsize{} &\scriptsize{Multi-Resolution DMD}              &\scriptsize{Kutz et al. (2015) \cite{2023}}           \\
\scriptsize{}            &\scriptsize{} &\scriptsize{Randomized Low-Rank DMD}           &\scriptsize{Erichson and Donovan (2015) \cite{2025}}  \\
\scriptsize{}            &\scriptsize{} &\scriptsize{Compressed DMD (cDMD)}             &\scriptsize{Erichson et al. (2015) \cite{2026}}       \\
\cline{2-4}    
\scriptsize{}            &\scriptsize{Bayesian RPCA} &\scriptsize{Bayesian RPCA (BRPCA)}              &\scriptsize{Ding et al. (2011) \cite{7}}     \\
\scriptsize{}            &\scriptsize{} 					   &\scriptsize{Variational Bayesian RPCA (VBRPCA)} &\scriptsize{Babacan et al. (2012) \cite{24}} \\
\scriptsize{}            &\scriptsize{} 				&\scriptsize{Factorized Variational Bayesian RPCA (FVBRPCA)} &\scriptsize{Aicher (2013) \cite{94}}  \\
\scriptsize{}            &\scriptsize{} 				&\scriptsize{Bayesian RPCA with MoG noise(MoG-BRPCA)}   &\scriptsize{Zhao et al. (2014) \cite{1024}}
\\
\scriptsize{}            &\scriptsize{} 				&\scriptsize{Bayesian-Ising-Signal (BIS)}   &\scriptsize{Huan et al. (2016) \cite{1572}} \\
\cline{2-4}
\scriptsize{}                         &\scriptsize{Approximated RPCA} &\scriptsize{GoDec}   &\scriptsize{Zhou and Tao (2011) \cite{6}}  \\
\scriptsize{}                         &\scriptsize{} 					&\scriptsize{Semi-Soft GoDec} &\scriptsize{Zhou and Tao (2011) \cite{6}}  \\
\cline{2-4}
\scriptsize{}     &\scriptsize{Sparse Additive Matrix Factorization} &\scriptsize{SAMF}     &\scriptsize{Nakajima et al. (2012) \cite{24-1}}  \\
\cline{2-4}
\scriptsize{}     &\scriptsize{Variational Bayesian Sparse Estimator} &\scriptsize{VBSE}    &\scriptsize{Chen et al. (2014) \cite{24-3}}     \\
\hline
\scriptsize{Robust Non-negative Matrix Factorization (RNMF)}   &\scriptsize{Manhattan Non-negative Matrix Factorization}      &\scriptsize{MahNMF} &\scriptsize{Guan et al. (2012) \cite{761}}  \\
\scriptsize{}                                                  &\scriptsize{Near-separable Non-negative Matrix Factorization} & \scriptsize{NS-NMF (Xray-$l_2$)} &\scriptsize{Kumar et al. (2013) \cite{762-1}}  \\
\scriptsize{}                                                  &\scriptsize{} & \scriptsize{NS-NMF (RobustXray)} &\scriptsize{Kumar et al. (2013) \cite{762}}  \\
\scriptsize{}                                                  &\scriptsize{Robust Asymmetric Non-negative Matrix Factorization} &\scriptsize{RANMF}
&\scriptsize{Woo and Park (2013) \cite{765}}  \\
\scriptsize{}                                                  &\scriptsize{Alternating-Updating Nonnegative Matrix Factorization} &\scriptsize{MPI-FAUN}
&\scriptsize{Kannan et al. (2016) \cite{770}}  \\
\hline
\end{tabular}}
\caption{Robust problem formulations based on decomposition into low-rank plus additive matrices: A complete overview (Part 1).} \centering
\label{SurveyOverview-1}
\end{table}
\end{landscape}

\newpage
\begin{landscape}
\begin{table}
\scalebox{0.79}{
\begin{tabular}{|l|l|l|l|} 
\hline
\scriptsize{Robust Problem Formulations} &\scriptsize{Categories} &\scriptsize{Sub-categories} &\scriptsize{Authors - Dates}\\
\hline
\hline
\scriptsize{Robust Matrix Completion (RMC)}                           &\scriptsize{Robust Matrix Completion with $l_{\sigma}$ norm loss function}             & \scriptsize{RMC-$l_{1}$-ADM} &\scriptsize{Candes et al. (2009) \cite{3}}  \\
\scriptsize{}                                                  &\scriptsize{}             & \scriptsize{RMC-$l_{\sigma}$-IST} &\scriptsize{Yang et al. (2014) \cite{2001}}  \\
\scriptsize{}                                                  &\scriptsize{}             & \scriptsize{RMC-$l_{\sigma}$-IHT} &\scriptsize{Yang et al. (2014) \cite{2001}}  \\
\scriptsize{}                                                  &\scriptsize{Robust Matrix Completion with Robust Bilateral Factorization} & 
\scriptsize{RMC-RBF} &\scriptsize{Shang et al. (2014) \cite{1043}}  \\
\scriptsize{}                                                  &\scriptsize{Robust Matrix Completion with Convex Formulation} & \scriptsize{RMC (convex formulation)} &\scriptsize{Shang et al. (2014) \cite{2003}}  \\
\scriptsize{}                                                  &\scriptsize{Robust Matrix Completion with Matrix Factorization} & \scriptsize{RMC-MF (non-convex formulation)} &\scriptsize{Shang et al. (2014) \cite{2003}}  \\
\scriptsize{}   &\scriptsize{Factorized Robust Matrix Completion}  & \scriptsize{FRMC} &\scriptsize{Mansour and Vetro (2014) \cite{2000}}  \\ 
\scriptsize{}   &\scriptsize{}                                     & \scriptsize{FRMC-MV} &\scriptsize{Kao et al. (2016) \cite{2000-1}}  \\      
\scriptsize{}                                                  &\scriptsize{Nearly-optimal Robust Matrix Completion} & \scriptsize{R-RMC} &\scriptsize{Cherapanamjeri et al. (2016) \cite{17150}}  \\
\scriptsize{}                                                  &\scriptsize{}                                        & \scriptsize{PG-RMC} &\scriptsize{Cherapanamjeri et al. (2016) \cite{17150}}  \\
\scriptsize{}                                                  &\scriptsize{Motion-Assisted Matrix Completion} & \scriptsize{MAMC} &\scriptsize{Yang et al. (2014) \cite{2002}}  \\
\scriptsize{}                                                  &\scriptsize{} & \scriptsize{Robust MAMC (RMAMC)} &\scriptsize{Yang et al. (2014) \cite{2002}}  \\
\scriptsize{}                                                  &\scriptsize{} & \scriptsize{Spatiotemporal Lowrank MC (SLMC)} &\scriptsize{Javed et al. (2016) \cite{1010-10}}  \\
\hline
\scriptsize{Robust Subspace Recovery (RSR)}&\scriptsize{Robust Subspace Discovery} &\scriptsize{RSD}  &\scriptsize{Wang et al. (2013) \cite{1002}}   \\
\scriptsize{} 				&\scriptsize{Robust Subspace Recovery via Bi-Sparsity} &\scriptsize{RoSuRe}    &\scriptsize{Bian and Krim (2014) \cite{1014}}  \\
\scriptsize{}          &\scriptsize{Robust Orthonomal Subspace Learning}     &\scriptsize{ROSL}      &\scriptsize{Xu et al. (2014) \cite{759-1}}     \\
\scriptsize{}          &\scriptsize{}                                        &\scriptsize{ROSL+}      &\scriptsize{Xu et al. (2014) \cite{759-1}}    \\
\scriptsize{}          &\scriptsize{Robust Orthogonal Complement Principal Component Analysis}     &\scriptsize{ROC-PCA}  &\scriptsize{She et al. (2014) \cite{1042}}     \\
\scriptsize{}          &\scriptsize{Sparse Latent Low-rank representation} &\scriptsize{SLL}      &\scriptsize{Li et al. (2015) \cite{1397}}     \\
\hline
\scriptsize{Robust Subspace Tracking (RST)}        &\scriptsize{Grassmannian Robust Adaptive Subspace Tracking Algorithm} &\scriptsize{GRASTA}     &\scriptsize{He et al. (2011) \cite{27}}        \\
\scriptsize{}     &\scriptsize{} &\scriptsize{t-GRASTA}   &\scriptsize{He et al. (2013) \cite{2801}}        \\
\scriptsize{}     &\scriptsize{} &\scriptsize{GASG21}     &\scriptsize{He et Zhang (2014) \cite{2803}}      \\
\scriptsize{}   &\scriptsize{$l_p$-norm Robust Online Subspace Tracking} &\scriptsize{pROST}  &\scriptsize{Hage and Kleinsteuber \cite{69}}    \\
\scriptsize{}   &\scriptsize{}                            &\scriptsize{Real Time pROST}       &\scriptsize{Hage and Kleinsteuber \cite{6901}}  \\
\scriptsize{}   &\scriptsize{Grassmannian Online Subspace Updates with Structured-sparsity} &\scriptsize{GOSUS}      &\scriptsize{Xu et al. (2013) \cite{85}}         \\
\scriptsize{}   &\scriptsize{Fast Adaptive Robust Subspace Tracking}  &\scriptsize{FARST}      &\scriptsize{Ahn (2014) \cite{1035}}         \\
\scriptsize{}   &\scriptsize{Robust Online Subspace Estimation and Tracking Algorithm}  &\scriptsize{ROSETA}      &\scriptsize{Mansour and Jiang (2015) \cite{1075}} \\
\scriptsize{}   &\scriptsize{Adaptive Projected Subgradient Method }  &\scriptsize{APSM}      &\scriptsize{Chouvardas et al. (2015) \cite{1396}} \\
\hline
\scriptsize{Robust Low Rank Minimization (RLRM)}  &\scriptsize{Contiguous Outlier Detection} &\scriptsize{DECOLOR} &\scriptsize{Zhou et al. (2011) \cite{25}}  \\
\scriptsize{}   &\scriptsize{Direct Robust Matrix Factorization} &\scriptsize{DRMF}   &\scriptsize{Xiong et al. (2011) \cite{54}}  \\
\scriptsize{}   &\scriptsize{}                                   &\scriptsize{DRMF-R} &\scriptsize{Xiong et al. (2011) \cite{54}}  \\
\scriptsize{}   &\scriptsize{Probabilistic Robust Matrix Factorization} &\scriptsize{PRMF}  &\scriptsize{Wang et al. (2012) \cite{5400}}  \\
\scriptsize{}   &\scriptsize{Bayesian Robust Matrix Factorization}      &\scriptsize{BRMF}  &\scriptsize{Wang et al. (2013) \cite{5401}}  \\
\scriptsize{}   &\scriptsize{}                                          &\scriptsize{MBRMF}  &\scriptsize{Wang et al. (2013) \cite{5401}}  \\
\scriptsize{}   &\scriptsize{Practical Low-Rank Matrix Factorization}   &\scriptsize{PLRMF (RegL1-ALM)} &\scriptsize{Zheng et al. (2012) \cite{1025}}  \\
\scriptsize{}   &\scriptsize{Low Rank Matrix Factorization with MoG noise} &\scriptsize{LRMF-MoG}  &\scriptsize{Meng et al. (2013) \cite{2-1}}  \\
\scriptsize{}   &\scriptsize{Unifying Nuclear Norm and Bilinear Factorization} &\scriptsize{UNN-BF} &\scriptsize{Cabral et al. (2013) \cite{2-2}}   \\
\scriptsize{}   &\scriptsize{Low Rank Matrix Factorization with General Mixture noise} &\scriptsize{LRMF-GM} &\scriptsize{Cao et al. (2015) \cite{2-3}}  \\
\scriptsize{}   &\scriptsize{Robust Rank Factorization}                 &\scriptsize{RRF (LOIRE)}   &\scriptsize{Sheng et al. (2014) \cite{1027}}  \\
\scriptsize{}   &\scriptsize{Variational Bayesian Method}               &\scriptsize{VBMF-$l_1$}    &\scriptsize{Zhao et al. (2015) \cite{1024-1}} \\
\scriptsize{}   &\scriptsize{Robust Orthogonal Matrix Factorization}    &\scriptsize{ROMF}    &\scriptsize{Kim and Oh (2015) \cite{1074}} \\
\scriptsize{}   &\scriptsize{Contiguous Outliers Representation via Online Low-Rank Approximation}   & \scriptsize{COROLA}            &\scriptsize{Shakeri and Zhang (2015) \cite{1077}}  \\
\scriptsize{}  &\scriptsize{Online Low Rank Matrix Completion} & \scriptsize{ORLRMR} &\scriptsize{Guo (2015) \cite{1120}}  \\ 
\scriptsize{}  &\scriptsize{Matrix Factorization - Elastic-net Regularization} & \scriptsize{FactEN}  &\scriptsize{Kim et al. (2015) \cite{1310}}  \\
\scriptsize{}  &\scriptsize{Incremental Learning Low Rank Representation - Spatial Constraint} & \scriptsize{LSVD-LRR}  &\scriptsize{Dou et al. (2015) \cite{1330}}  \\
\scriptsize{}  &\scriptsize{Online Robust Low Rank Matrix Recovery} & \scriptsize{ORLRMR}  &\scriptsize{Guo (2015) \cite{1536}}  \\
\hline
\end{tabular}}
\caption{Robust problem formulations based on decomposition into low-rank plus additive matrices: A complete overview (Part 2).} \centering
\label{SurveyOverview-2}
\end{table}
\end{landscape}

\newpage
\begin{landscape}
\begin{table}
\scalebox{1.0}{
\begin{tabular}{|l|l|l|l|l|}
\hline
\scriptsize{Methods}&\scriptsize{Decomposition} &\scriptsize{Minimization}&\scriptsize{Constraints} &\scriptsize{Convexity} \\
\hline
\hline
\scriptsize{\textbf{PCP}} 
&\scriptsize{$A=L+S$} 
&\scriptsize{$\underset{L,S}{\text{min}} ~~ ||L||_* + \lambda ||S||_{l_{1}}$}
&\scriptsize{$A-L-S=0$}
&\scriptsize{Yes}\\ 
\scriptsize{\textbf{Candes et al. \cite{3}}} 
&\scriptsize{} 
&\scriptsize{}
&\scriptsize{}
&\scriptsize{}\\
\hline
\scriptsize{Modified PCP (Fixed Rank)} 
&\scriptsize{$A=L+S$} 
&\scriptsize{$\underset{L,S}{\text{min}} ~~ ||L||_* + \lambda ||S||_{l_{1}}$}
&\scriptsize{$rank(L)= known ~~r$}
&\scriptsize{Yes}\\
\scriptsize{Leow et al. \cite{78}} 
&\scriptsize{} 
&\scriptsize{}
&\scriptsize{}
&\scriptsize{}\\
\hline
\scriptsize{Modified PCP (Nuclear Norm Free)} 
&\scriptsize{$A=u1^T+S$} 
&\scriptsize{$\underset{u}{\text{min}} ||A-u1^T||_{l_{1}}$}
&\scriptsize{$rank(u1^T)=1$}
&\scriptsize{Yes}\\
\scriptsize{Yuan et al. \cite{75}} 
&\scriptsize{} 
&\scriptsize{}
&\scriptsize{}
&\scriptsize{}\\
\hline
\scriptsize{Modified PCP (Capped Norm)} 
&\scriptsize{$A=L+S$} 
&\scriptsize{$\underset{L,S}{\text{min}} ~~rank(L) + \lambda||S||_{l_{0}}$}
&\scriptsize{$||A-L-S||_F^2\leq \sigma^2$}
&\scriptsize{No}\\
\scriptsize{Sun et al. \cite{82}} 
&\scriptsize{} 
&\scriptsize{}
&\scriptsize{}
&\scriptsize{}\\
\hline
\scriptsize{Modified PCP (Inductive)} 
&\scriptsize{$A=PA+S$} 
&\scriptsize{$\underset{P,S}{\text{min}} ||P||_* + \lambda ||S||_{l_{1}}$}
&\scriptsize{$A-PA-S=0$}
&\scriptsize{Yes}\\
\scriptsize{Bao et al. \cite{67}} 
&\scriptsize{} 
&\scriptsize{}
&\scriptsize{}
&\scriptsize{}\\
\hline
\scriptsize{Modified PCP (Partial Subspace Knowledge)} 
&\scriptsize{$A=L+S$} 
&\scriptsize{$\underset{L,S}{\text{min}} ||L||_* + \lambda ||S||_{l_{1}}$}
&\scriptsize{$L+P_{\Gamma^\perp}S=P_{\Gamma^\perp}A$}
&\scriptsize{Yes}\\ 
\scriptsize{Zhan and Vaswani \cite{17-4}} 
&\scriptsize{} 
&\scriptsize{}
&\scriptsize{}
&\scriptsize{}\\
\hline
\scriptsize{$p$,$q$-PCP (Schatten-$p$ norm, $l_q$ norm)} 
&\scriptsize{$A=L+S$} 
&\scriptsize{$\underset{L,S}{\text{min}} ||L||_{S_p}^p + \lambda ||S||_{l_q}$}
&\scriptsize{$A=L+S$}
&\scriptsize{No}\\
\scriptsize{Wang et al.\cite{1022}} 
&\scriptsize{} 
&\scriptsize{}
&\scriptsize{}
&\scriptsize{}\\
\hline
\scriptsize{Modified $p$,$q$-PCP (Schatten-$p$ norm, $L_q$ seminorm)} 
&\scriptsize{$A=L+S$} 
&\scriptsize{$\underset{L,S}{\text{min}} ||L||_{S_p}^p + \lambda ||S||_{L_q}^q$}
&\scriptsize{$A=L+S$}
&\scriptsize{No}\\
\scriptsize{Shao et al.\cite{1023}} 
&\scriptsize{} 
&\scriptsize{}
&\scriptsize{}
&\scriptsize{}\\
\hline
\scriptsize{Modified PCP (2D-PCA)} 
&\scriptsize{$A=L+S$} 
&\scriptsize{$\underset{U,V}{\text{min}} \frac{1}{T} \sum_{i=1}^T ||A_i- U \Sigma_i V^T||_{F}^2$}
&\scriptsize{$U^TU=I_{r \times r}~,V^TV=I_{c \times c}$}
&\scriptsize{No}\\
\scriptsize{Sun et al.\cite{1005}} 
&\scriptsize{} 
&\scriptsize{}
&\scriptsize{}
&\scriptsize{}\\
\hline
\scriptsize{Modified PCP (Rank-$N$ Soft Constraint)} 
&\scriptsize{$A=L+S$}
&\scriptsize{$\underset{L,S}{\text{min}} \sum_{i=N+1}^{\text{min(m,n)}} |\sigma_i(L)|+ \lambda||S||_{l_{1}}$}
&\scriptsize{$A=L+S$}
&\scriptsize{Yes}\\
\scriptsize{Oh \cite{1011}} 
&\scriptsize{} 
&\scriptsize{}
&\scriptsize{}
&\scriptsize{}\\
\hline
\scriptsize{Modified PCP (JVFSD-RPCA)} 
&\scriptsize{$A=L+S$}
&\scriptsize{$\underset{L,S}{\text{min}} ~~ ||L||_* + \lambda ||S||_{l_{1}}$}
&\scriptsize{$A=L+S$}
&\scriptsize{Yes}\\
\scriptsize{Wen et al. \cite{1026}} 
&\scriptsize{} 
&\scriptsize{}
&\scriptsize{}
&\scriptsize{}\\
\hline
\scriptsize{Modified PCP (NSMP)} 
&\scriptsize{$A=L+S$}
&\scriptsize{$\underset{L,S}{\text{min}} \lambda ||L||_* + \mu||S||_2$}
&\scriptsize{$A=L+S$}
&\scriptsize{Yes}\\
\scriptsize{Wang and Feng \cite{1034}} 
&\scriptsize{} 
&\scriptsize{}
&\scriptsize{}
&\scriptsize{}\\
\hline
\scriptsize{Modified PCP (WNSMP)} 
&\scriptsize{$A=L+S$}
&\scriptsize{$\underset{L,S}{\text{min}} \lambda ||\omega(L)||_* + \mu||\omega^{-1}(S)||_2$}
&\scriptsize{$A=L+S$}
&\scriptsize{Yes}\\
\scriptsize{Wang and Feng \cite{1034}} 
&\scriptsize{} 
&\scriptsize{}
&\scriptsize{}
&\scriptsize{}\\
\hline
\scriptsize{Modified PCP (Implicit Regularizers)} 
&\scriptsize{$A=L+S$}
&\scriptsize{$\underset{L,S}{\text{min}} ~~ \lambda ||L||_* + \varphi(S)$}
&\scriptsize{$A=L+S$}
&\scriptsize{}\\
\scriptsize{He et al. \cite{87}} 
&\scriptsize{} 
&\scriptsize{}
&\scriptsize{}
&\scriptsize{Yes}\\
\hline 
\end{tabular}}
\caption{Decompositions in low-rank plus additive matrices: An homogeneous overview (Part 1)}
\label{TPCP1Overview-1}
\end{table}
\end{landscape}

\newpage
\begin{landscape}
\begin{table}
\scalebox{0.65}{
\begin{tabular}{|l|l|l|l|l|l|}
\hline
\scriptsize{Categories}&\scriptsize{Methods}&\scriptsize{Decomposition} &\scriptsize{Minimization}&\scriptsize{Constraints} &\scriptsize{Convexity} \\
\hline
\hline
\scriptsize{RPCA-PCP} 
&\scriptsize{\textbf{SPCP}} 
&\scriptsize{$A=L+S+E$} 
&\scriptsize{$\underset{L,S}{\text{min}} ~~ ||L||_* + \lambda ||S||_{l_{1}} $}
&\scriptsize{$||A-L-S||_F<\delta$}
&\scriptsize{Yes}\\
\scriptsize{} 
\scriptsize{} 
&\scriptsize{\textbf{Zhou et al. \cite{5}}} 
&\scriptsize{} 
&\scriptsize{}
&\scriptsize{} 
&\scriptsize{}\\
\cline{2-6} 
\scriptsize{} 
&\scriptsize{Modified SPCP (Bilateral Projection)} 
&\scriptsize{$A=UV+S+E$} 
&\scriptsize{$\underset{U,V,S}{\text{min}} ~~ \lambda ||S||_{l_{1}}  + ||A-UV-S||_F^2$}
&\scriptsize{$rank(U)=rank(S)\leq r$}
&\scriptsize{Yes}\\
\scriptsize{} 
\scriptsize{} 
&\scriptsize{Zhou and Tao \cite{601}} 
&\scriptsize{} 
&\scriptsize{}
&\scriptsize{} 
&\scriptsize{}\\
\cline{2-6} 
&\scriptsize{Modified SPCP (Nuclear Norm Free)} 
&\scriptsize{$A=U1^T+S+E$} 
&\scriptsize{$\underset{S \in \mathbf{R}^{m\times n}, u \in \mathbf{R}^m}{\text{min}} ~~ ||S||_{l_{1}}  + \frac{\mu}{2} ||A-u1^T-S||_F^2$}
&\scriptsize{$rank(U1^T)=1$}
&\scriptsize{Yes}\\
\scriptsize{} 
&\scriptsize{Yuan et al. \cite{75}} 
&\scriptsize{} 
&\scriptsize{}
&\scriptsize{}
&\scriptsize{}\\
\cline{2-6} 
&\scriptsize{Modified SPCP (Nuclear Norm Free for blur in video)} 
&\scriptsize{$A=U1^T+S+E$} 
&\scriptsize{$\underset{S \in \mathbf{R}^{m\times n}, u \in \mathbf{R}^m}{\text{min}} ~~ ||S||_{l_{1}} + \frac{\mu}{2} ||A-H(u1^T+S)||_F^2$}
&\scriptsize{$rank(U1^T)=1$}
&\scriptsize{Yes}\\
\scriptsize{} 
&\scriptsize{Yuan et al. \cite{75}} 
&\scriptsize{} 
&\scriptsize{}
&\scriptsize{}
&\scriptsize{}\\
\cline{2-6} 
&\scriptsize{Modified SPCP (Undercomplete Dictionary)} 
&\scriptsize{$A=UT+S+E$} 
&\scriptsize{$\underset{U, T, S}{\text{min}} ~~  \lambda ||S||_{l_{1}} + \frac{\lambda_1}{2} (||U||_F^2 + ||T||_F^2) + \frac{1}{2} ||A-UT-S)||_F^2$}
&\scriptsize{$rank(UT) \leq r$}
&\scriptsize{Yes}\\
\scriptsize{} 
&\scriptsize{Sprechman et al. \cite{68}} 
&\scriptsize{} 
&\scriptsize{}
&\scriptsize{}
&\scriptsize{}\\
\cline{2-6} 
&\scriptsize{Variational SPCP (Huber penalty)} 
&\scriptsize{$A=L+S+E$} 
&\scriptsize{$\text{min}  ~~ \Phi(L,S)$}
&\scriptsize{$\rho(L+S-Y)<\epsilon$}
&\scriptsize{Yes}\\
\scriptsize{} 
&\scriptsize{Aravkin et al. \cite{1029}} 
&\scriptsize{} 
&\scriptsize{}
&\scriptsize{}
&\scriptsize{}\\
\cline{2-6} 
&\scriptsize{Modified SPCP (Three Term Low-rank Optimization)} 
&\scriptsize{$A=L+S+E$}
&\scriptsize{$\underset{A,L,S}{\text{min}} ~~ \lambda ||L||_* + \lambda_1 ||f_{\pi}(S)|| + \lambda_2 ||E||_F^2$}
&\scriptsize{$A=L+S+E$}
&\scriptsize{Yes}\\
&\scriptsize{Oreifej et al. \cite{1031}} 
&\scriptsize{} 
&\scriptsize{}
&\scriptsize{}
&\scriptsize{}\\
\cline{2-6} 
\scriptsize{} 
&\scriptsize{\textbf{QPCP}} 
&\scriptsize{$A=L+S$} 
&\scriptsize{$\underset{L,S}{\text{min}} ~~ ||L||_* + \lambda ||S||_{l_{1}} $}
&\scriptsize{$||A-L-S||_{\infty} < 0.5$}
&\scriptsize{Yes}\\
\scriptsize{} 
&\scriptsize{\textbf{Becker et al. \cite{8}}} 
&\scriptsize{} 
&\scriptsize{}
&\scriptsize{} 
&\scriptsize{}\\
\cline{2-6} 
\scriptsize{} 
&\scriptsize{\textbf{BPCP}} 
&\scriptsize{$A=L+S$} 
&\scriptsize{$\underset{L,S}{\text{min}} ~~ ||L||_* + \kappa (1-\lambda) ||L||_{l_{2,1}} + \kappa \lambda||S||_{l_{2,1}}$}
&\scriptsize{$A-L-S=0$}
&\scriptsize{Yes}\\
\scriptsize{} 
&\scriptsize{\textbf{Tang and Nehorai \cite{41}}} 
&\scriptsize{} 
&\scriptsize{}
&\scriptsize{} 
&\scriptsize{}\\
\cline{2-6}
\scriptsize{} 
&\scriptsize{\textbf{LPCP}} 
&\scriptsize{$A=AU+S$} 
&\scriptsize{$\underset{U,S}{\text{min}} ~~ \alpha||U||_{l_{1}} + \beta ||U||_{l_{2,1}} + \beta ||S||_{l_{1}}$}
&\scriptsize{$A-AU+S=0$}
&\scriptsize{Yes}\\
\scriptsize{} 
&\scriptsize{\textbf{Wohlberg et al. \cite{26}}} 
&\scriptsize{} 
&\scriptsize{}
&\scriptsize{} 
&\scriptsize{}\\
\hline 
\scriptsize{RPCA-OP} 
&\scriptsize{OP} 
&\scriptsize{$A=L+S$} 
&\scriptsize{$\underset{L,S}{\text{min}} ~~ ||L||_* + \lambda ||S||_{l_{1,2}}$}
&\scriptsize{$A-L-S=0$}
&\scriptsize{Yes}\\
\scriptsize{} 
&\scriptsize{Xu et al. \cite{56}} 
&\scriptsize{} 
&\scriptsize{}
&\scriptsize{}
&\scriptsize{} \\
\cline{2-6}
\scriptsize{} 
&\scriptsize{SOP} 
&\scriptsize{$A=L+S+E$} 
&\scriptsize{$\underset{L,S}{\text{min}} ~~ ||L||_* + \lambda ||S||_{l_{1,2}}$}
&\scriptsize{$||A-L-S||_F<\delta$}
&\scriptsize{Yes}\\
\scriptsize{} 
&\scriptsize{Xu et al. \cite{56}} 
&\scriptsize{} 
&\scriptsize{}
&\scriptsize{} 
&\scriptsize{}\\
\hline
\scriptsize{RPCA-SpaCtrl} 
&\scriptsize{Sparsity Control} 
&\scriptsize{$A=M +U^{T}P+S+E$} 
&\scriptsize{$\underset{U,S}{\text{min}} ~~ ||X+1_N M^{T} -P U^{T}-S||_F^2 + \lambda ||S||_{l_{2(r)}} $}
&\scriptsize{$UU^{T}=I_q$}
&\scriptsize{Yes}\\
\scriptsize{} 
&\scriptsize{Mateos et Giannakis \cite{9}\cite{10}} 
&\scriptsize{} 
&\scriptsize{}
&\scriptsize{} 
&\scriptsize{}\\
\hline
\scriptsize{RPCA-SpaCorr}
&\scriptsize{Sparse Corruptions (case 1)} 
&\scriptsize{$A=L+S$} 
&\scriptsize{$\underset{L,S}{\text{min}} ||L||_* + \lambda ||S||_{l_{1}}$}
&\scriptsize{$||A-L-S||_{l_{1}} \leq \epsilon_1$ }
&\scriptsize{Yes}\\
\scriptsize{} 
&\scriptsize{Hsu et al. \cite{58}} 
&\scriptsize{} 
&\scriptsize{}
&\scriptsize{$||A-L-S||_* \leq \epsilon_*$} 
&\scriptsize{}\\
\scriptsize{} 
&\scriptsize{} 
&\scriptsize{} 
&\scriptsize{}
&\scriptsize{$||L||_{\infty} \leq b$} 
&\scriptsize{}\\
\cline{2-6}
&\scriptsize{Sparse Corruptions (case 2)} 
&\scriptsize{$A=L+S$} 
&\scriptsize{$\underset{L,S}{\text{min}} ||L||_* + \lambda ||S||_{l_{1}} + \frac{1}{2 \mu} ||A-L-S||_F^2$}
&\scriptsize{$||A-L-S||_{l_{1}} \leq \epsilon_1$ }
&\scriptsize{Yes}\\
\scriptsize{} 
&\scriptsize{Hsu et al. \cite{58}} 
&\scriptsize{} 
&\scriptsize{}
&\scriptsize{$||A-L-S||_* \leq \epsilon_*$} 
&\scriptsize{}\\
\scriptsize{} 
&\scriptsize{} 
&\scriptsize{} 
&\scriptsize{}
&\scriptsize{$||A-S||_{\infty} \leq b$} 
&\scriptsize{}\\
\hline
\scriptsize{RPCA-LHR}
&\scriptsize{LHR} 
&\scriptsize{$A=L+S$} 
&\scriptsize{$\underset{\hat{X} \in \hat{D}}{\text{min}} \frac{1}{2} (||Diag(Y)||_{L}+||Diag(Z)||_{L})+\lambda||S||_{L}$}
&\scriptsize{$\hat{X}=\left\{Y,Z,L,S\right\}$}
&\scriptsize{No}\\
\scriptsize{} 
&\scriptsize{Deng et al. \cite{55}} 
&\scriptsize{} 
&\scriptsize{}
&\scriptsize{$\hat{D}=\left\{\left(Y,Z,L,S\right):\begin{pmatrix} Y & L \\ L^T & Z \end{pmatrix} \geq 0, (L,S) \in C \right\}$} 
&\scriptsize{}\\
\hline
\scriptsize{RPCA-IRLS}
&\scriptsize{IRLS} 
&\scriptsize{$A=UV+S$} 
&\scriptsize{$\underset{U \in \mathbb{R}^{n \times p}, V \in \mathbb{R}^{p \times m}}{min} \mu ||UV||_* + ||(A-UV)\circ W_1||_{l_{\alpha,\beta}}$}
&\scriptsize{$A-UV-S=0$}
&\scriptsize{Yes}\\
\scriptsize{} 
&\scriptsize{Guyon et al.\cite{505}} 
&\scriptsize{} 
&\scriptsize{}
&\scriptsize{} 
&\scriptsize{}\\
\hline
\scriptsize{RPCA-SO}
&\scriptsize{OR-PCA} 
&\scriptsize{$A=LR+S+E$} 
&\scriptsize{$\underset{L \in \mathbb{R}^{n \times p}, R \in \mathbb{R}^{n \times r}}{\text{min}} \frac{1}{2} ||A-LR^T-S||_F^2+ \frac{\lambda_1}{2} (||L||_F^2+ ||R||_F^2) + \lambda_2 ||S||_{l_{1}}$}
&\scriptsize{$A-LR-S=0$}
&\scriptsize{Yes}\\
\scriptsize{} 
&\scriptsize{Feng et al. \cite{1010}} 
&\scriptsize{} 
&\scriptsize{}
&\scriptsize{} 
&\scriptsize{}\\
\hline
\end{tabular}}
\caption{Decompositions in low-rank plus additive matrices: An homogeneous overview (Part 2)}
\label{TPCP1Overview-2}
\end{table}
\end{landscape}

\newpage
\begin{landscape}
\begin{table}
\scalebox{0.70}{
\begin{tabular}{|l|l|l|l|l|l|}
\hline
\scriptsize{Categories}&\scriptsize{Methods}&\scriptsize{Decomposition} &\scriptsize{Minimization}&\scriptsize{Constraints} &\scriptsize{Convexity} \\
\hline
\hline
\scriptsize{Bayesian RPCA} 
&\scriptsize{BRPCA} 
&\scriptsize{$A=D(ZG)W_2+BX+E$} 
&\scriptsize{$- log~(p(\Theta | A, H))$}
&\scriptsize{Distribution constraints}
&\scriptsize{-}\\
\scriptsize{}
&\scriptsize{Ding et al. \cite{7}} 
&\scriptsize{} 
&\scriptsize{}
&\scriptsize{} 
&\scriptsize{}\\
\cline{2-6}
\scriptsize{} 
&\scriptsize{VBRPCA} 
&\scriptsize{$A=DB^T+S+E$} 
&\scriptsize{$p(A,D,B,S)$}
&\scriptsize{Distribution constraints}
&\scriptsize{-} \\
\scriptsize{}
&\scriptsize{Babacan et al. \cite{24}} 
&\scriptsize{} 
&\scriptsize{}
&\scriptsize{} 
&\scriptsize{}\\
\cline{2-6}
&\scriptsize{MoG-RPCA} 
&\scriptsize{$A=L+S$} 
&\scriptsize{${\text{min}}$ KL divergence}
&\scriptsize{MOG distribution constraints for $S$}
&\scriptsize{-}\\
\scriptsize{} 
&\scriptsize{Zhao et al.\cite{1024}} 
&\scriptsize{} 
&\scriptsize{}
&\scriptsize{}
&\scriptsize{}\\
\hline
\scriptsize{Approximated RPCA} 
&\scriptsize{GoDec} 
&\scriptsize{$A=L+S+E$} 
&\scriptsize{$\underset{L,S}{\text{min}} ~~ ||A-L-S||_F^2$}
&\scriptsize{$rank(L) \leq e , card(S) \leq k$}
&\scriptsize{No}\\
\scriptsize{} 
&\scriptsize{Zhou and Tao \cite{6}} 
&\scriptsize{} 
&\scriptsize{}
&\scriptsize{} 
&\scriptsize{}\\
\cline{2-6}
&\scriptsize{Semi-Soft GoDec} 
&\scriptsize{$A=L+S+E$} 
&\scriptsize{$\underset{L,S}{\text{min}} ~~ ||A-L-S||_F^2$}
&\scriptsize{$rank(L) \leq e , card(S) \leq \tau$}
&\scriptsize{No}\\
\scriptsize{} 
&\scriptsize{Zhou and Tao \cite{6}} 
&\scriptsize{} 
&\scriptsize{}
&\scriptsize{$\tau$ is a soft threshold} 
&\scriptsize{}\\
\hline
\scriptsize{Sparse Additive Matrix Factorization} 
&\scriptsize{SAMF} 
&\scriptsize{$A=\sum_{k=0}^K S + E$} 
&\scriptsize{$r(\Theta)$}
&\scriptsize{Distribution constraints}
&\scriptsize{}\\
\scriptsize{} 
&\scriptsize{Nakajima et al. \cite{24-1}} 
&\scriptsize{} 
&\scriptsize{}
&\scriptsize{} 
&\scriptsize{}\\
\hline
\scriptsize{Variational Bayesian Sparse Estimator} 
&\scriptsize{VBSE} 
&\scriptsize{$A=UV^T+RS+E$} 
&\scriptsize{$\rho(A,U,V,S,\gamma,\alpha,\beta)$}
&\scriptsize{Distribution constraints}
&\scriptsize{-}\\
\scriptsize{} 
&\scriptsize{Chen et al. \cite{24-2}} 
&\scriptsize{} 
&\scriptsize{}
&\scriptsize{} 
&\scriptsize{}\\
\hline
\hline
\scriptsize{Robust Non-negative Matrix Factorization} 
&\scriptsize{Manhattan NMF} 
&\scriptsize{$A=W^TH+S$} 
&\scriptsize{$\underset{W\geq0,H\geq0}{\text{min}} ~~ f(W,H)=||A-W^TH||_M$}
&\scriptsize{$r \ll min(m,n)$}
&\scriptsize{No}\\
\scriptsize{} 
&\scriptsize{(MahNMF) Guan et al.\cite{761}} 
&\scriptsize{} 
&\scriptsize{}
&\scriptsize{} 
&\scriptsize{}\\
\cline{2-6}
&\scriptsize{NS-NMF (Xray-$l_2$)} 
&\scriptsize{$A=WH=A_BH+S$} 
&\scriptsize{$\underset{A_B\geq 0, H\geq 0}{\text{min}} ~~ ||A-A_BH||_F^2$}
&\scriptsize{$W=A_B\geq 0, H\geq 0$}
&\scriptsize{Yes}\\
\scriptsize{} 
&\scriptsize{(Xray-$l_2$) Kumar et al. \cite{762-1}} 
&\scriptsize{} 
&\scriptsize{}
&\scriptsize{} 
&\scriptsize{}\\
\cline{2-6}
&\scriptsize{NS-NMF (RobustXray)} 
&\scriptsize{$A=WH+E=A_BH+S+E$} 
&\scriptsize{$\underset{A_B\geq 0, H\geq 0}{\text{min}} ~~ ||A-A_BH||_{l_{1}}$}
&\scriptsize{$W=A_B \geq 0, H \geq 0$}
&\scriptsize{Yes}\\
\scriptsize{} 
&\scriptsize{Kumar and Sindhwani \cite{762}} 
&\scriptsize{} 
&\scriptsize{}
&\scriptsize{} 
&\scriptsize{}\\
\cline{2-6}
&\scriptsize{RANMF} 
&\scriptsize{$A=L+S= W\Lambda H +S$} 
&\scriptsize{$\underset{L,S, \Phi}{\text{min}} ~~ \Phi(S) + \frac{alpha}{2} ||A-L-S||_F^2 +$}
&\scriptsize{$R(L) \leq \tau, 0<L<B_L$}
&\scriptsize{Yes}\\
\scriptsize{} 
&\scriptsize{Woo and Park  \cite{765}}
&\scriptsize{} 
&\scriptsize{$\beta (\Psi(S, \Phi) + \gamma TV(Q(\Phi))$}
&\scriptsize{} 
&\scriptsize{}\\
\hline
\hline
\scriptsize{Robust Matrix Completion} 
&\scriptsize{RMC-$l_\sigma$-IHT} 
&\scriptsize{$A=L+S$} 
&\scriptsize{$\underset{S \in \mathbf{R}^{m \times n}}{\text{min}} ~~ l_{\sigma}(L) + \lambda ||L||_* $}
&\scriptsize{$rank(L)\leq r$}
&\scriptsize{No}\\
\scriptsize{} 
&\scriptsize{Yang et al.\cite{2001}} 
&\scriptsize{} 
&\scriptsize{} 
&\scriptsize{}
&\scriptsize{}\\
\cline{2-6}
\scriptsize{} 
&\scriptsize{RMC-RBF} 
&\scriptsize{$A=L+S=UV^T+S$} 
&\scriptsize{$\underset{U,V,S}{\text{min}} ~~ ||P_{\Omega}(S)||_1+ \lambda ||V||_*$}
&\scriptsize{$P_{\Omega}(A)=P_{\Omega}(UV^T+S),U^TU=I$} 
&\scriptsize{Yes}\\
\scriptsize{} 
&\scriptsize{Shang et al. \cite{1043}} 
&\scriptsize{} 
&\scriptsize{}
&\scriptsize{} 
&\scriptsize{}\\
\cline{2-6}
&\scriptsize{RMC (convex formulation)} 
&\scriptsize{$A=L+S$} 
&\scriptsize{$\underset{L,S}{\text{min}} ~~ ||L||_*+ \lambda ||P_{\Omega}(S)||_{l_{1}}$}
&\scriptsize{$P_{\Omega}(L+S)=P_{\Omega}(A), E_{\Omega^C}=0$}
&\scriptsize{Yes}\\
\scriptsize{} 
&\scriptsize{Shang et al.  \cite{2003}} 
&\scriptsize{} 
&\scriptsize{}
&\scriptsize{} 
&\scriptsize{}\\
\cline{2-6}
&\scriptsize{RMC-MF (non-convex formulation)} 
&\scriptsize{$A=L+S=UV^T+S$} 
&\scriptsize{$\underset{U,V,S}{\text{min}} ~~ ||V||_*+ \lambda ||P_{\Omega}(S)||_{l_{1}}$}
&\scriptsize{$A=UV^T+S, U^TU=I$}
&\scriptsize{No}\\
\scriptsize{} 
&\scriptsize{Shang et al.  \cite{2003}} 
&\scriptsize{} 
&\scriptsize{}
&\scriptsize{} 
&\scriptsize{}\\
\cline{2-6}
&\scriptsize{Factorized Robust Matrix Completion} 
&\scriptsize{$A=L+S=L_{L}L_{R}^T+S$} 
&\scriptsize{$||L||_*=\underset{L_{L} \in \mathbf{R}^{m,r}, L_{R} \in \mathbf{R}^{n,r}} {\text{min}} ~~\frac{1}{2} (||L_{L}||_F^2+ ||L_{R}||_F^2)$}
&\scriptsize{$L_{L}L_{R}^T=L$}
&\scriptsize{Yes}\\
\scriptsize{} 
&\scriptsize{(FRMC) Mansour and Vetro (2014) \cite{2000}} 
&\scriptsize{} 
&\scriptsize{}
&\scriptsize{} 
&\scriptsize{}\\
\cline{2-6}
&\scriptsize{Motion-Assisted Matrix Completion} 
&\scriptsize{$A=L+S$} 
&\scriptsize{$\underset{L,S}{\text{min}} ~~ ||L||_*+ \lambda ||S||_{l_{1}}$}
&\scriptsize{$W_3 \circ A = W_3 \circ (L+S)$}
&\scriptsize{Yes}\\
\scriptsize{} 
&\scriptsize{(MAMC) Yang et al. \cite{2002}} 
&\scriptsize{} 
&\scriptsize{}
&\scriptsize{} 
&\scriptsize{}\\
\hline
\end{tabular}}
\caption{Decompositions in low-rank plus additive matrices: An homogeneous overview (Part 3)}
\label{TPCP2Overview}
\end{table}
\end{landscape}

\newpage
\begin{landscape}
\begin{table}
\scalebox{0.65}{
\begin{tabular}{|l|l|l|l|l|l|}
\hline
\scriptsize{Categories}&\scriptsize{Methods}&\scriptsize{Decomposition} &\scriptsize{Minimization}&\scriptsize{Constraints} &\scriptsize{Convexity} \\
\hline
\hline
\scriptsize{Robust Subspace Recovery} 
&\scriptsize{RoSuRe} 
&\scriptsize{$A=L+S=LW+S$} 
&\scriptsize{$\underset{W,S}{\text{min}} ~~ ||W||_{l_{1}}+ \lambda ||S||_{l_{1}}$}
&\scriptsize{$A=L+S, L=LW, W_{ii}=0, \forall i$}
&\scriptsize{No}\\
\scriptsize{} 
&\scriptsize{Bian and Krim \cite{1014}} 
&\scriptsize{} 
&\scriptsize{}
&\scriptsize{} 
&\scriptsize{}\\
\cline{2-6}
\scriptsize{} 
&\scriptsize{ROSL} 
&\scriptsize{$A=D\alpha+S$} 
&\scriptsize{$\underset{S,D, \alpha}{\text{min}} ~~ ||\alpha||_{row-1}+ \lambda ||S||_{l_{1}}$}
&\scriptsize{$D \alpha+S=A, D^TD=I_k, \forall i$}
&\scriptsize{No}\\
\scriptsize{} 
&\scriptsize{Xu et al. \cite{759-1}} 
&\scriptsize{} 
&\scriptsize{}
&\scriptsize{} 
&\scriptsize{}\\
\cline{2-6}
\scriptsize{}
&\scriptsize{ROC-PCA} 
&\scriptsize{$AV_{\perp} = L+S+E$} 
&\scriptsize{$\underset{V_{\perp},\mu,S}{\text{min}} ~~ \frac{1}{2}||AV_{\perp}-L-S||_F^2+ \sum_{ij} P(||s_{ij}||_{l^{2}}; \lambda_{i})$}
&\scriptsize{$V_{\perp}^TV_{\perp}=I$}
&\scriptsize{No}\\
\scriptsize{} 
&\scriptsize{She et al. \cite{1042}} 
&\scriptsize{} 
&\scriptsize{}
&\scriptsize{} 
&\scriptsize{}\\
\scriptsize{} 
&\scriptsize{} 
&\scriptsize{} 
&\scriptsize{}
&\scriptsize{} 
&\scriptsize{}\\
\hline
\hline
\scriptsize{Subspace Tracking} 
&\scriptsize{GRASTA} 
&\scriptsize{$A=UW+S+E$} 
&\scriptsize{$\underset{U,w,S}{\text{min}} ~~ ||S||_{l_{1}}$}
&\scriptsize{$A_{\Omega_t}=U_{\Omega_t}w+S ~~,~~U \in G(d,n)$}
&\scriptsize{No}\\
\scriptsize{} 
&\scriptsize{He et al. \cite{27}\cite{28}} 
&\scriptsize{} 
&\scriptsize{}
&\scriptsize{} 
&\scriptsize{}\\
\cline{2-6}
&\scriptsize{t-GRASTA} 
&\scriptsize{$A=UW+S$} 
&\scriptsize{$\underset{U,w,S,\tau}{\text{min}} ~~ ||S||_{l_{1}}$}
&\scriptsize{$A_{\Omega_t} \circ \tau=U_{\Omega_t}w+S ~~,~~U \in G(d,n)$}
&\scriptsize{No}\\
\scriptsize{} 
&\scriptsize{He et al. \cite{2801}\cite{2802}} 
&\scriptsize{} 
&\scriptsize{}
&\scriptsize{} 
&\scriptsize{}\\
\cline{2-6}
&\scriptsize{GASG21} 
&\scriptsize{$A=UW+S$} 
&\scriptsize{$\underset{w}{\text{min}} ~~ ||Uw-A||_{l_{2,1}}- \sum_{j=1}^m||U_jw_j-A_j||_{l_2}$}
&\scriptsize{$A=U_w+S ~~,~~U \in G(d,n)$}
&\scriptsize{No}\\
\scriptsize{} 
&\scriptsize{He and Zhang \cite{2803}} 
&\scriptsize{} 
&\scriptsize{}
&\scriptsize{} 
&\scriptsize{}\\
\cline{2-6}
&\scriptsize{pROST} 
&\scriptsize{$A=UW+S+E$} 
&\scriptsize{$\underset{Rank(L)\leq k}{\text{min}} ~~ ||UW-A||_{L_{p}}$}
&\scriptsize{$A-UW-S=0$}
&\scriptsize{No}\\
\scriptsize{} 
&\scriptsize{Hage and Kleinsteuber \cite{69}\cite{6901}} 
&\scriptsize{} 
&\scriptsize{}
&\scriptsize{} 
&\scriptsize{}\\
\cline{2-6}
&\scriptsize{GOSUS} 
&\scriptsize{$A=UW+S+E$} 
&\scriptsize{$\underset{U^TU=I_d,W,S}{\text{min}} ~~ \sum_{i=1}^{l} \mu_i||D^iS||_{l_{2}}+\frac{\lambda}{2}||UW+S-A||_{l_{2}}^2$}
&\scriptsize{$A-UW-S=0$}
&\scriptsize{No}\\
\scriptsize{} 
&\scriptsize{Xu et al. \cite{85}}
&\scriptsize{} 
&\scriptsize{}
&\scriptsize{} 
&\scriptsize{}\\
\cline{2-6}
&\scriptsize{FARST} 
&\scriptsize{$A=UW+S+E$} 
&\scriptsize{$\underset{W}{\text{min}} ~~||UW-A||_{l_{1}}$}
&\scriptsize{$A-UW-S=0$}
&\scriptsize{No}\\
\scriptsize{} 
&\scriptsize{Ahn \cite{1035}}
&\scriptsize{} 
&\scriptsize{}
&\scriptsize{} 
&\scriptsize{}\\
\hline
\hline
\scriptsize{Robust Low-Rank Minimization} 
&\scriptsize{Contiguous Outlier Detection} 
&\scriptsize{$A=L+S+E$} 
&\scriptsize{$\underset{L,S}{\text{min}} ~~  \alpha ||L||_* +  \beta ||F||_{l_{1}} + \gamma ||C vec(F)||_{l^{1}}$}
&\scriptsize{$rank(L) \leq K$}
&\scriptsize{No}\\
\scriptsize{} 
&\scriptsize{(DECOLOR) Zhou et al. \cite{25}} 
&\scriptsize{} 
&\scriptsize{$+\frac{1}{2}||P_{\bar{F}}(A-L)||_F^2$} 
&\scriptsize{}
&\scriptsize{}\\
\cline{2-6}
\scriptsize{} 
&\scriptsize{Direct Robust Matrix Factorization} 
&\scriptsize{$A=L+S$} 
&\scriptsize{$\underset{L,S}{\text{min}} ||A-S-L||_F$}
&\scriptsize{$rank(L) \leq r$, $||S||_0 \leq p$}
&\scriptsize{No}\\
\scriptsize{} 
&\scriptsize{(DRMF) Xiong et al. \cite{54}} 
&\scriptsize{} 
&\scriptsize{}
&\scriptsize{} 
&\scriptsize{Original formulation PCP \cite{3}}\\
\cline{2-6}
&\scriptsize{Direct Robust Matrix Factorization-Row} 
&\scriptsize{$A=L+S$} 
&\scriptsize{$\underset{L,S}{\text{min}} ||A-S-L||_F$}
&\scriptsize{$rank(L) \leq r$, $||S||_{2,0} \leq p$}
&\scriptsize{No}\\
\scriptsize{} 
&\scriptsize{(DRMF-R) Xiong et al. \cite{54}} 
&\scriptsize{} 
&\scriptsize{}
&\scriptsize{} 
&\scriptsize{Original formulation OP \cite{56}}\\
\cline{2-6}
&\scriptsize{Probabilistic Robust Matrix Factorization} 
&\scriptsize{$A=UV^\prime+S$} 
&\scriptsize{$log~(p(U,V|A,\lambda,\lambda_U,\lambda_V))$}
&\scriptsize{Distribution constraints}
&\scriptsize{No}\\
\scriptsize{} 
&\scriptsize{(PRMF) Wang et al. \cite{5400}} 
&\scriptsize{} 
&\scriptsize{}
&\scriptsize{} 
&\scriptsize{}\\
\cline{2-6}
&\scriptsize{Bayesian Robust Matrix Factorization} 
&\scriptsize{$A=UV^\prime+S$} 
&\scriptsize{$log~(p(U,V|A,\lambda,\lambda_U,\lambda_V))$}
&\scriptsize{Bayesian distribution constraints}
&\scriptsize{No}\\
\scriptsize{} 
&\scriptsize{(BRMF) Wang et al. \cite{5401}} 
&\scriptsize{} 
&\scriptsize{}
&\scriptsize{} 
&\scriptsize{}\\
\cline{2-6}
&\scriptsize{Practical Low-Rank Matrix Factorization} 
&\scriptsize{$A=UV+S$} 
&\scriptsize{$\underset{U,V}{\text{min}} ||W_5 \odot (A-UV) ||_{l_{1}} + \lambda ||V||_*$}
&\scriptsize{$U^TU=I_{r}$}
&\scriptsize{Yes}\\
\scriptsize{} 
&\scriptsize{(PLMR) Zheng et al. \cite{1025}} 
&\scriptsize{} 
&\scriptsize{}
&\scriptsize{} 
&\scriptsize{}\\
\cline{2-6}
&\scriptsize{Low Rank Matrix Factorization with MoG noise} 
&\scriptsize{$A=UV^T+S$} 
&\scriptsize{$\underset{U,V,\Pi,\Sigma}{\text{max}} \sum_{i,j \in \Omega} \sum_{k=1}^K \pi_k \mathbf{N}(x_{ij}|(u^i)^Tv^j, \sigma_k^2)$}
&\scriptsize{MoG distribution constraints on $S$}
&\scriptsize{No}\\
\scriptsize{} 
&\scriptsize{(LRMF-MOG) Meng et al. \cite{2-1}} 
&\scriptsize{} 
&\scriptsize{}
&\scriptsize{} 
&\scriptsize{}\\
\cline{2-6}
&\scriptsize{Unifying Nuclear Norm and Bilinear Factorization} 
&\scriptsize{$A=UV^T+S$} 
&\scriptsize{$\underset{L,U,V}{\text{min}} ||W_5 \odot (A-L) ||_{l_{1}} + \frac{\lambda}{2}(||U||_F^2 + \lambda ||V||_F^2)$}
&\scriptsize{$L=UV^T$}
&\scriptsize{Yes}\\
\scriptsize{} 
&\scriptsize{(UNN-BF) Cabral et al. \cite{2-2}} 
&\scriptsize{} 
&\scriptsize{}
&\scriptsize{} 
&\scriptsize{}\\
\cline{2-6}
&\scriptsize{Robust Rank Factorization} 
&\scriptsize{$A=BX+S+E$} 
&\scriptsize{$\underset{B}{\text{min}}~~\underset{S,X}{\text{min}} ||A-BX-S||_{l_{2}}^2 + \frac{\lambda}{2}||S||_{l_{1}}$}
&\scriptsize{$\lambda>0, A=BX+S+E$}
&\scriptsize{Yes}\\
\scriptsize{} 
&\scriptsize{(RRF) Sheng et al. \cite{1027}} 
&\scriptsize{} 
&\scriptsize{}
&\scriptsize{} 
&\scriptsize{}\\
\hline
\end{tabular}}
\caption{Decompositions in low-rank plus additive matrices: An homogeneous overview (Part 4)}
\label{TPCP3Overview}
\end{table}
\end{landscape}

\subsection{A Unified View of Decomposition into Low-rank plus Additive Matrices} 
\label{subsec:UnifiedFramework}

\subsubsection{Notations}
\label{subsubsec:Notations} 
To provide to the readers an easy comparison, we homogenized all the different notations found on all the papers as follows: \\
\begin{enumerate}
\item \textbf{Matrices:} For the common matrices, $A$ stands for the observation matrix, $L$ is the low-rank matrix, $S$ is the unconstrained (residual) matrix or sparse matrix, and $E$ is the noise matrix. $I$ is the identity matrix. For the specific matrices, the notations are given in the section of the corresponding method. \\
\item \textbf{Indices :} $m$ and $n$ are the number of columns and rows of the observed data matrix $A$. In the case of background/foreground separation, $m$ corresponds to the number of pixels in a frame, and $n$ corresponds to the number of frames in the sequence. $n$ is taken usually to $200$ due to computational and memory limitations. $i$ and $j$ stand for the current indices of the matrix. $r$ is the estimated or fixed rank of the matrix $L$. $p$ stands for the $p^{th}$ largest value in truncated matrix. \\
\end{enumerate}

\subsubsection{Norms}
The different norms used in this paper for vectors and matrices can be classified as follows: \\
\begin{itemize}
\item \textbf{Vector $l^{\alpha}$-norm with $0\leq \alpha \leq 2$ :} $||V||_{l^0}$ is the $l^0$-norm of the vector $V$, and it corresponds to the number of non-zero entries. $||V||_{l^1}=\sum_{i} v_i$ is the $l^1$-norm of the vector $V$, and it corresponds to the sum of the entries.
$||V||_{l^2}=\sqrt{\sum_{i} (v_i)^2}$ is the $l^2$-norm of the vector $V$, and it corresponds to the Euclidean distance \cite{1050}. \\
\item \textbf{Matrix $l_{\alpha}$-norm with $0\leq \alpha \leq 2$ :} $||M||_{l_{0}}$ is the $l_0$-norm of the matrix $M$, and it corresponds to the number of non-zero entries \cite{1050}. $||M||_{l_{1}}=\sum_{i,j}|M_{ij}|$ is the $l_1$-norm of the matrix $M$ \cite{1050}, and its corresponds to the  Manhattan distance \cite{761}. $||M||_{l_{2}}=\sqrt{\sum_{i,j} M_{i,j}^2}$ is the $l_2$-norm of the matrix $M$ also known as the Frobenius norm. \\
\item \textbf{Matrix $l_{\infty}$-norm:} $||M||_{l_{\infty}}=\underset{ij}max~|M_{ij}|$ \cite{56-1} is the $l_\infty$-norm of the matrix $M$. It can be used to capture the quantization error of the observed value of the pixel like in Becker et al. \cite{8}. It is equivalent to the max-norm \cite{1110}.\\ 
\item \textbf{Matrix  $l_{\alpha,\beta}$-norm with $0\leq \alpha, \beta \leq 2$:} $||M||_{l_{\alpha,\beta}}$ is the $l_{\alpha,\beta}$-mixed norm of the matrix $M$, and it corresponds to the $l^\beta$-norm of the vector formed by taking the $l^\alpha$-norms of the columns of the underlying matrix. $\alpha$ and $\beta$ are in the interval $[0,2]$. For example, $||M||_{l_{2,0}}$ corresponds to the number of non-zero columns of the matrix $M$ \cite{1050}. $||M||_{l_{2,1}}$ forces spatial homogeneous fitting in the matrix $M$ \cite{507}, and it is suitable in presence of column outliers or noise \cite{41}\cite{507}\cite{2803}. $||M||_{l_{2,1}}$ is equal to $\sum_j ||M_{:j}||_{l^2}$ \cite{1050}. The influence of $\alpha$ and $\beta$ on the matrices $L$ and $S$ was studied in \cite{507}.  This norm is also called structured norm.\\
\item \textbf{Matrix $L_{\alpha}$-seminorm with $0< \alpha \leq 2$:} $||M||_{L_{\alpha}}=(\sum_{i,j}|M_{ij}|^{\alpha})^{1/ \alpha}$ is the $L_{\alpha}$-seminorm of the matrix $M$ \cite{1023}. The $L_{1}$-seminorm is equivalent to the  $l_{1}$-norm .\\
\item \textbf{Matrix $L_{\alpha}$-quasi-norm with $0< \alpha <1$:} $L_{\alpha}$-quasi-norm is defined by $L_{\alpha}(M)= \sum_{i=1}^m(M_i^2+ \mu)^{\frac{1}{\alpha}}$ \cite{69}\cite{6901}.  \\
\item \textbf{Matrix Frobenius norm:} $||M||_F=\sqrt{\sum_{i,j} M_{i,j}^2}$ is the Frobenius norm \cite{1050}. The Frobenius norm is sometimes also called the Euclidean norm which may cause confusion with the vector $l^2$-norm which is also sometimes known as the Euclidean norm too.  \\
\item \textbf{Matrix nuclear norm:} $||M||_*$ is the nuclear norm of the matrix $M$, and it corresponds to the sum of its singular values \cite{1050}. The nuclear norm is the $l^1$-norm applied on the vector composed with the singular values of the matrix \cite{55}. It is equivalen to the Ky Fan $n$-norm and the Schatten-$1$-norm \cite{1110}. \\
\item \textbf{Matrix dual norm:} $||.||_d$ is the $d$ dual norm of any norm $||.||_{norm}$ previously defined, that is $norm \in \left\{l_{\alpha}, l_{\infty}, l_{\alpha,\beta}, L_{\alpha}, F, * \right\}$. For example, the dual norm of the nuclear norm is $||.||_2$ called spectral norm which corresponds to the largest singular value of the matrix \cite{1034}. \\
\item \textbf{Matrix Schatten-$\alpha$ norm with  $0< \alpha \leq 2$:} The Schatten-$\alpha$ norm $||M||_{S_{\alpha}}=(\sum_{k=1}^{min(m,n)} (\sigma_k(M))^{\alpha})^{1/ \alpha}$  where $\sigma_k(M)$, denotes the $k^{th}$ singular values of $M$, can also be used as a surrogate of the nuclear norm like in \cite{1022}\cite{1023}\cite{1545}\cite{1546}. The Schatten-$1$-norm is equivalent to the nuclear norm \cite{1110}. \\
\item \textbf{Matrix log-sum norm:} The log-sum norm $||M||_L$ is defined as $ \sum_{ij}log~(\left|M_{ij}\right|+\delta)$ with $\delta>0$ is a small regularization constant \cite{55}. \\
\item \textbf{Matrix max-norm:} The max-norm $||M||_{max}$ is defined as $\underset{ij}max~|M_{ij}|$. The max-norm is equivalent to the $l_{\infty}$-norm \cite{1110}. \\
\item \textbf{Matrix $\gamma$-norm:} The norm $||M||_{\gamma}$ \cite{1543} is defined as $\sum_i \frac{(1+\gamma) \sigma_i(M)}{\gamma +  \sigma_i(M)}$ with $\gamma>0$. We can noticed that $\lim\limits_{\gamma \rightarrow 0} ||M||_{\gamma}= rank(M)$ and $\lim\limits_{\gamma \rightarrow \infty} ||M||_{\gamma} =||M||_{*}$. It is not a real norm but it is unitarily invariant. \\
\end{itemize}

\subsubsection{Loss Functions and Regularization Functions}
Loss functions are used for the minimization term, and the regularized functions are used to enforce the low-rank, sparse and noise constraints on $L$, $S$ and $E$, respectively. They are needed to solve the minimization problem (See Section \ref{SectionMinimizationProblem}). In literature, most of the time, the authors do not distinguish "loss functions" and "regularized functions", and they use them indifferently. In this paper, the regularized functions are respectively noted $f_{low}()$, $f_{sparse}()$ and $f_{noise}()$, and we only use the term "loss functions". Most of the time, loss functions are defined on the previous defined norms such as: $l_0$-loss function ($||.||_{l_{0}}$), $l_1$-loss function ($||.||_{l_{1}}$), $l_2$-loss function ($||.||_{l_{2}}$), nuclear norm function, Frobenius loss function and log-sum heuristic function \cite{55}. Other loss functions can be used such as $l_{\sigma}$-loss function \cite{2001}, Least Squares (LS) loss function ($||.||_F^2$), Huber loss function \cite{1029}, $M$-estimator based loss functions \cite{87}, and the generalized fused Lasso loss function \cite{1070}\cite{1575}. Lipschitz loss function can also be used in a two-stage convex relaxation approach \cite{1533} by the majorization for a class of locally Lipschitz continuous surrogates of Equation \ref{EquationPCP2}, which solves the nuclear norm plus $l_1$-norm minimization problem in the first stage and a nuclear semi-norm plus weighted $l_1$-norm minimization problem in the second stage with theoretical guarantee. \\

\indent Practically, proxy loss functions are used as surrogate of the original loss function ($rank(.)$) for the low-rank constraint and $l_0$-loss function for the sparsity constraint to obtain a solvable problem. Table \ref{LossFunctionRPCA} in the column "Surrogate loss functions (for the decomposition only)" shows an overview of the different loss functions used in the different problem formulations. The surrogate lost functions present the main advantage that they allow to reach to a solvable convex problem but their use can present main disadvantages as detailed in the following: \\

\begin{itemize}
\item The nuclear norm loss function is the most tightest convex surrogate of the rank function over the unit spectral norm ball, but it presents a big difference over a general closed convex set since the former is convex whereas the latter is nonconvex even concave \cite{1533}. \\
\item The Frobenius norm loss function is a valid proxy for nuclear norm loss function, but it fails to recover the low-rank matrix without rank estimate \cite{759-1}. \\
\item  The $l_1$-loss function also known as least absolute deviations (LAD) or least absolute errors (LAE) is robust to outliers but it is not stable and do not reach to a unique solution. In addition, it may be suboptimal, since the $l_1$-norm is a loose approximation of the $l_0$-norm and often leads to an over-penalized problem. Furthermore, $l_1$-loss function cannot handle collinearity. \\
\item The $l_2$-loss function is sensitive to outliers and missing data but it is stable and reach always to one solution. \\
\item The Least Squares (LS) loss function is known to be very sensitive to outliers too \cite{10} but the least squares loss function is a suitable solution in applications in which it is needed to take into account any and all outliers. \\
\end{itemize}

\indent For the computational point of view, $l_1$-norm does not have an analytical solution, but $l_2$-norm does. Thus, $l_2$-norm problems can be solved computationally efficiently. $l_1$-norm solutions can take into acount the sparsity properties which allows it to be used for sparse constraints, which makes the computation more efficient than the $l_0$-norm. \\

\begin{landscape}
\begin{table}
\scalebox{0.70}{
\begin{tabular}{|l|l|l||l|} 
\hline
\scriptsize{Functions} &\scriptsize{Original loss function} &\scriptsize{Surrogate loss functions (for the decomposition only)} &\scriptsize{Regularization functions (for background/foreground separation) }\\
\hline
\hline
\scriptsize{$f_{low}()$} &\scriptsize{$rank(.)$ \cite{3}} &\scriptsize{\textbf{PCP/SPCP/QPCP/BPCP/LPCP/OP}}        &\scriptsize{\textbf{Temporal Coherence}}   \\
\scriptsize{Low-rank $L$} &\scriptsize{} &\scriptsize{Nuclear norm \cite{3}}   &\scriptsize{\textit{Linear Operator $T(L)$:} Invariant translational and rotational transformation \cite{26-6}\cite{26-7}}    \\
\scriptsize{} &\scriptsize{} &\scriptsize{Truncated nuclear norm \cite{1540}\cite{17120}\cite{17140}}      &\scriptsize{}   \\                        \scriptsize{} &\scriptsize{} &\scriptsize{Partial sum \cite{1920}}                                         &\scriptsize{}   \\                        
\scriptsize{} &\scriptsize{} &\scriptsize{\textbf{Modified PCP}}                                                   &\scriptsize{-}   \\
\scriptsize{} &\scriptsize{} &\scriptsize{Capped nuclear norm \cite{82}}                                           &\scriptsize{}    \\
\scriptsize{} &\scriptsize{} &\scriptsize{Schatten-$\alpha$ norm  \cite{1022}\cite{1023}, Rank-$N$ \cite{1011}}    &\scriptsize{}    \\
\scriptsize{} &\scriptsize{} &\scriptsize{$\gamma$-norm \cite{1543}}                                               &\scriptsize{}    \\
\scriptsize{} &\scriptsize{} &\scriptsize{\textbf{Heuristic Recovery}}                                             &\scriptsize{-}   \\
\scriptsize{} &\scriptsize{} &\scriptsize{Log-sum heuristic  (LHR) \cite{55}}                                      &\scriptsize{}    \\
\scriptsize{} &\scriptsize{} &\scriptsize{\textbf{Stochastic Optimization}}                                        &\scriptsize{-}   \\
\scriptsize{} &\scriptsize{} &\scriptsize{Max-norm (MRMD) \cite{1110}}                                             &\scriptsize{}    \\
\hline
\scriptsize{$f_{sparse}()$} &\scriptsize{$l_0$-norm \cite{3}} &\scriptsize{\textbf{PCP/Modified PCP}}   &\scriptsize{\textbf{Spatial Coherence}}  \\
\scriptsize{Sparsity $S$} &\scriptsize{} &\scriptsize{$l_1$-norm \cite{3}, capped $l_1$-norm \cite{82}} &\scriptsize{$||S||_{l_{1,2}}$\cite{1920}} \\
\scriptsize{} &\scriptsize{} &\scriptsize{$l_\alpha$-norm \cite{1022}, $L_\alpha$-seminorm \cite{1023}} &\scriptsize{}   \\
\scriptsize{} &\scriptsize{} &\scriptsize{dual norm \cite{1034}, $M$-estimator \cite{87}}               &\scriptsize{}   \\
\scriptsize{} &\scriptsize{} &\scriptsize{\textbf{SPCP}}                                                &\scriptsize{\textbf{Temporal Coherence}}   \\
\scriptsize{} &\scriptsize{} &\scriptsize{Generalized fused Lasso \cite{1070}\cite{1575}}               &\scriptsize{\textit{Linear Operator $\Pi(S)$:}  1) Confidence Map based on Dense Optical Flow \cite{1031}}   \\
\scriptsize{} &\scriptsize{} &\scriptsize{}                           &\scriptsize{2) Confidence Map based on Salient Motion \cite{1711}}   \\
\scriptsize{} &\scriptsize{} &\scriptsize{\textbf{BPCP/LPCP}}                                            &\scriptsize{-}   \\
\scriptsize{} &\scriptsize{} &\scriptsize{$l_{2,1}$-norm \cite{41}\cite{26}}                             &\scriptsize{}    \\
\scriptsize{} &\scriptsize{} &\scriptsize{\textbf{OP}}                                                   &\scriptsize{-}   \\
\scriptsize{} &\scriptsize{} &\scriptsize{$l_{1,2}$-norm  \cite{46}}                                     &\scriptsize{}    \\
\scriptsize{} &\scriptsize{} &\scriptsize{\textbf{LHR}}                                                  &\scriptsize{-}   \\
\scriptsize{} &\scriptsize{} &\scriptsize{Log-sum heuristic \cite{55}}                                   &\scriptsize{}    \\
\scriptsize{} &\scriptsize{} &\scriptsize{\textbf{pROST}}                                                &\scriptsize{-}   \\
\scriptsize{} &\scriptsize{} &\scriptsize{$L_p$-quasi-norm \cite{6901}}                                  &\scriptsize{}    \\
\hline
\scriptsize{$f_{noise}()$} &\scriptsize{\textbf{PCP}}               &\scriptsize{\textbf{Modified PCP}}                     &\scriptsize{-}  \\
\scriptsize{Error $E$}     &\scriptsize{Frobenius norm \cite{3}}    &\scriptsize{Inequality ($||A-L-S||_F^2\leq \sigma^2$)} &\scriptsize{}  \\
\scriptsize{}              &\scriptsize{Equality ($||A-L-S||_F=0$)} &\scriptsize{Frobenius norm \cite{82}}                  &\scriptsize{}  \\
\scriptsize{} &\scriptsize{}                                 &\scriptsize{\textbf{Modified SPCP}}                        &\scriptsize{-}  \\
\scriptsize{} &\scriptsize{}                                 &\scriptsize{Inequality ($\rho(A-L-S)\leq\epsilon$):}       &\scriptsize{}  \\
\scriptsize{} &\scriptsize{}                                 &\scriptsize{Huber penalty \cite{1029}}                     &\scriptsize{}  \\
\scriptsize{} &\scriptsize{}                                 &\scriptsize{\textbf{QPCP}}                                 &\scriptsize{-}  \\
\scriptsize{} &\scriptsize{}                                 &\scriptsize{Inequality ($|||A-L-S||_{l_{\infty}}\leq0.5$)} &\scriptsize{}  \\
\scriptsize{} &\scriptsize{}                                 &\scriptsize{$l_{\infty}$-norm \cite{8}}                    &\scriptsize{}  \\
\cline{2-4}
\scriptsize{} &\scriptsize{\textbf{SPCP}}                                                &\scriptsize{\textbf{SPCP}}     &\scriptsize{\textbf{Temporal Coherence}}  \\
\scriptsize{} &\scriptsize{Frobenius norm \cite{5}}   &\scriptsize{Equality ($||A-L-S-E||_F=0$)}               &\scriptsize{\textit{Weighting matrix $W$:}}    \\
\scriptsize{} &\scriptsize{Inequality ($||A-L-S||_F<\delta$)}             &\scriptsize{+ Frobenius norm \cite{1031} on $E$}        &\scriptsize{1) Dense Optical Flow (RMC) \cite{2002}\cite{2002-1}}  \\
\scriptsize{} &\scriptsize{}             &\scriptsize{}        &\scriptsize{2) Salient Motion (RPCA) \cite{1711}} \\
\scriptsize{} &\scriptsize{}             &\scriptsize{}        &\scriptsize{Equality ($||W\circ A-W\circ (L + S + E)||=0$ \cite{2002}\cite{2002-1}\cite{1711}}  \\
\hline
\hline 
\scriptsize{$f_{back}()$}                           &\scriptsize{-}     &\scriptsize{-}    &\scriptsize{$||L||_{l_{2,1}}$\cite{1900}}    \\
\scriptsize{Low-rank (Background) $L$}              &\scriptsize{}     &\scriptsize{}      &\scriptsize{}               \\
\scriptsize{}                                       &\scriptsize{}     &\scriptsize{}      &\scriptsize{}               \\
\hline
\scriptsize{$f_{fore}()$}                           &\scriptsize{-}     &\scriptsize{-}    &\scriptsize{\textbf{Spatial Coherence}}   \\
\scriptsize{Sparsity (Foreground)  $S$}             &\scriptsize{}     &\scriptsize{}      &\scriptsize{\textit{$||.||_{2,1}$:} \cite{41}\cite{501}\cite{505}\cite{506}\cite{507}\cite{1520}\cite{1522}\cite{1523}\cite{1524}} \\
\scriptsize{}                                       &\scriptsize{}     &\scriptsize{}      &\scriptsize{\textit{Weighted $||.||_{2,1}$:} \cite{1542}}              \\
\scriptsize{}                                       &\scriptsize{}     &\scriptsize{}      &\scriptsize{\textit{Static structured sparsity norm:} $\Omega(S)$  \cite{1901}}   \\
\scriptsize{}                                       &\scriptsize{}     &\scriptsize{}      &\scriptsize{\textit{Dynamic tree structured sparsity norm:} $\Phi(S)$ \cite{1524}\cite{1525}\cite{1527}}   \\
\scriptsize{}                                       &\scriptsize{}     &\scriptsize{}      &\scriptsize{\textit{Total Variation:} $TV(S)$ \cite{506}\cite{507}\cite{765}\cite{1033}\cite{1900}}              \\
\scriptsize{}                                       &\scriptsize{}     &\scriptsize{}      &\scriptsize{\textit{Gradient:} $||grad(S)||_1$ \cite{505}\cite{506}\cite{507}\cite{26}}              \\
\hline 
\end{tabular}}
\caption{Loss functions $f_{low}()$, $f_{sparse}()$ and $f_{noise}()$ used for the low-rank, sparse and noise constraints in the different problem formulations. Regularization functions $f_{back}()$, $f_{fore}()$ used to suitably perform DLAM to background/foreground separation.} \centering
\label{LossFunctionRPCA}
\end{table}
\end{landscape}
Depth-weighted group-wise principal component analysis for foreground/background separation

\subsubsection{Definition and Classifications of Outliers}
\begin{figure}
\begin{center}
\includegraphics[width=9cm]{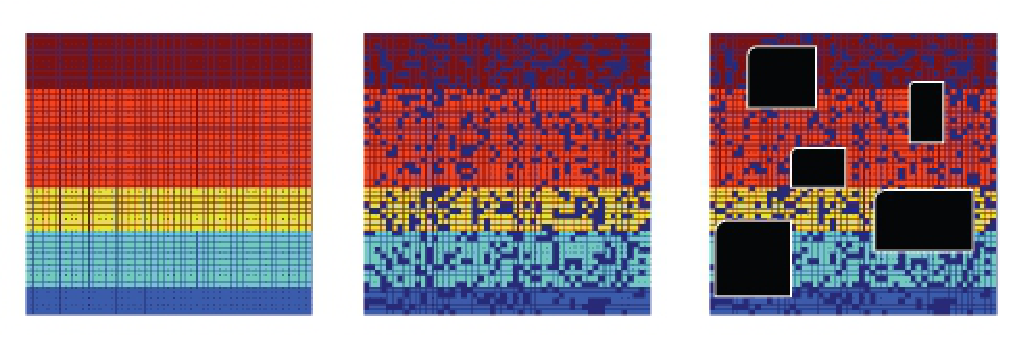}
\caption{Illustration of different types of corruption fo the matrix $A$: a) $A$ without noise b) $A$ with element-wise outliers/noise and c) $A$ with both element-wise outliers/noise and missing data (Illustration from the slides of Ma et al. \cite{5000}}). 
\label{Matrice1}
\end{center}
\end{figure}

The aim of the different problem formulations via decomposition into low-rank plus additive matrices is to be robust to outliers. Thus, the notion of outliers needs to be defined. Outliers can be defined as arbitrarily large valued measurements which do not characterize the true data samples. Practically, outliers can be viewed as observations or values that are considerably different from the majority of the data and usually follow heavy-tailed distribution by assuming that the unobservable noise have a fast-decaying Gaussian probability
distribution. As the observed examples are stacked as rows (or columns) in a data matrix, either some of the entries may be affected by \textit{1)} additive outliers (positive or negative values) or missing data, or \textit{2)} entire rows (or columns) are corrupted. Thus, outliers can be classified by considering their location in the observation matrix $A$ as follows: 
\begin{enumerate}
\item Element-wise outliers/noise. 
\item Row or column wise outliers/noise.
\item Missing data: They can be due to an acquisition problem or it is the presence of moving objects in the application of background initialization \cite{1710}.
\end{enumerate}
Figure \ref{Matrice1} shows each case. Furthermore, outliers can be classified in the subspace where they are the more prominent as developed in Brahma et al.\cite{1930}:
\begin{enumerate}
\item Outliers in the original observation space (OS) (that are commonly addressed) are relatively visible due to their outlying values. They can be effectively handled by PCP.
\item Outliers in the orthogonal complement (OC) subspace (which is the space orthogonal to the primary principal component subspace) for which PCP may fail. OC outliers are observations that have arbitrarily large magnitude when projected onto the OC subspace.
\end{enumerate}
In literature, She et al. \cite{1042} proposed a dedicated method for OC outliers. Only Brahma et al. \cite{1930} addressed both OS and OC outliers. For the application in computer vision, it is important to note that outliers may have different shapes and characteristics like in background/foreground separation where moving objects have different appearances, and spatial and temporal constraints.

\subsubsection{Decomposition into Low-rank plus Additive Matrices}
\label{subsubsec:DLAM} 
From the homogenized overview, we can see that all the decompositions in the different problem formulations can be grouped in a unified view that we called Decomposition into Low-rank plus Additive Matrices (DLAM). 

\subsubsection{Decomposition Problem}
All the decompositions can be written in a general formulation as follows:
\begin{equation}
A= \sum_{k=1}^K M_k = M_1+ M_2+M_3= L+S+E 
\label{EquationDecomposition1}
\end{equation}
with $K \in \left\{1,2,3\right\}$. The matrix $A$ can be corrupted by outliers . The characteristics of the matrices $M_k$ are as follows:
\begin{itemize}
\item The first matrix $M_1=L$ is a low-rank matrix. In some decompositions, $L$ is decomposed as follows: \textbf{1)} a product of two matrices $UV$ obtained by bilateral projection \cite{601} or by matrix factorization \cite{1101} in the RPCA framework, \textbf{(2)} a product of two matrices $UV^T$ \cite{1037}\cite{1043}\cite{2003} obtained by robust matrix factorization in the RMC framework, and \textbf{(3)} a product of two matrices $WH$ \cite{761}\cite{762-1}\cite{762} or three matrices $W\Lambda H$ \cite{765} with the constraints to be positive in the RNMF framework.  \\
\item The second matrix $M_2$ is an unconstrained (residual) matrix in implicit decomposition (LRM, RMC). $M_2$ is a sparse matrix $S$ in explicit decomposition (RPCA, RNMF, RSR, RST). $S$ can be decomposed in two sub-matrices $S_{OS}$ and $S_{OC}$ which contains the outliers in the OS space and the ouliers in the OC space, respectively. In literature, $S$ restricted to $S_{OS}$ is decomposed as follows: \textbf{1)} a sum of two matrices $S_1$ and $S_2$ which correspond to the foreground and the dynamic backgrounds \cite{1900}, and \textbf{(2)} a product of two matrices $RT$ where $R$ is an unconstrained matrix and $T$ is a sparse matrix \cite{24-3}\cite{24-4}\cite{1045} in the framework of RPCA. \\
\item The third matrix $M_3$ is generally the noise matrix $E$. The noise can be modeled by a Gaussian, a mixture of Gaussians (MoG) or a Laplacian distribution. $M_3$ can capture turbulences in the case of background/foreground separation \cite{1031}. \\
\end{itemize}
Thus, the decomposition can be rewritten as follows:
\begin{equation}
A= L + S_{OS} +S_{OC} + E 
\label{EquationDecomposition2}
\end{equation}
\indent Practically, the decomposition is $A \approx L$ when $K=1$. This decomposition is called implicit as the second matrix $S$ is not explicitly used in the problem formulation but it can be obtained by the difference $A-L$. It corresponds to basic formulations (LRM, MC, PCA and NMF). The decomposition is $A=L+S$ when $K=2$. It is called explicit because $S$ is explicitly determined and used in the problem formulation. It corresponds to robust formulations (RLRM, RMC, RPCA, RNMF,RSR and RST). In the case of $K=3$, the decomposition is $A=L+S+E$. This decomposition is called "stable" explicit decomposition as it separates the outliers in $S$ and the noise in $E$. It corresponds to stable robust formulations. \\

\indent Table \ref{DLAMOverview} shows an overview of the matrix $M_k$ in the different problem formulations. Thus, we can see that the decomposition is implicit in LRM and MC because the decomposition is made implicitly, i.e. $A \approx L$. $S$ is the residual matrix in RLRM and RMC whereas $S$ is a sparse matrix in RPCA, RNMF, RSR and RST with constraints on $E$ for the stable versions. \\

\begin{table}[!ht]
\begin{center}
\scalebox{0.79}{
\begin{tabular}{|l|l|l|l|} 
\hline
\scriptsize{Problem Formulations} &\scriptsize{Matrix $M_1=L$} &\scriptsize{Matrix $M_2=S$} &\scriptsize{Matrix $M_3=E$} \\
\hline
\hline
\scriptsize{\textbf{Implicit Decomposition ($K=1$)}}   &\scriptsize{} &\scriptsize{}             &\scriptsize{-}          \\
\hline
\scriptsize{LRM}           &\scriptsize{$L$ Low-rank} &\scriptsize{Not used}                     &\scriptsize{-}          \\
\scriptsize{MC}            &\scriptsize{$L$ Low-rank} &\scriptsize{Not used}                     &\scriptsize{-}          \\
\hline
\scriptsize{\textbf{Explicit Decomposition ($K=2$)}}   &\scriptsize{} &\scriptsize{}             &\scriptsize{-}          \\
\hline
\scriptsize{RLRM}           &\scriptsize{$L$ Low-rank} &\scriptsize{$S$ Residual matrix}             &\scriptsize{-}      \\
\scriptsize{RMC}            &\scriptsize{$L$ Low-rank} &\scriptsize{$S$ Residual matrix}             &\scriptsize{-}      \\
\scriptsize{RPCA}           &\scriptsize{$L$ Low-rank} &\scriptsize{$S$ Sparse (outlier+noise)}   &\scriptsize{-}                 \\
\scriptsize{RNMF}           &\scriptsize{$L=WH^T$Low-rank (positive)} &\scriptsize{$S$ Sparse (outlier+noise)}   &\scriptsize{-}  \\          
\scriptsize{RSR}            &\scriptsize{$L$ Low-rank} &\scriptsize{$S$ Sparse}                      &\scriptsize{-}              \\
\scriptsize{RST}            &\scriptsize{$L$ Low-rank} &\scriptsize{$S$ Sparse}                      &\scriptsize{-}              \\
\hline
\scriptsize{\textbf{Stable Explicit Decomposition ($K=3$)}}   &\scriptsize{} &\scriptsize{}                              &\scriptsize{-}     \\
\hline
\scriptsize{Stable RPCA}              &\scriptsize{$L$ Low-rank}  &\scriptsize{$S$ Sparse (outlier)}                 &\scriptsize{$E$ Noise}  \\
\scriptsize{Stable RNMF}              &\scriptsize{$L=WH^T$Low-rank (positive)} &\scriptsize{$S$ Sparse (outlier)}   &\scriptsize{$E$ Noise}  \\     \scriptsize{Stable RSR}               &\scriptsize{$L$ Low-rank} &\scriptsize{$S$ Sparse}                            &\scriptsize{$E$ Noise}  \\
\scriptsize{Stable RST}               &\scriptsize{$L$ Low-rank} &\scriptsize{$S$ Sparse}                            &\scriptsize{$E$ Noise}  \\
\hline
\end{tabular}}
\end{center}
\caption{Decomposition into Low-rank plus Additive Matrices (DLAM): The different matrices in the different problem formulations.} \centering
\label{DLAMOverview}
\end{table}

\subsubsection{Minimization problem}
\label{SectionMinimizationProblem}
\indent The corresponding minimization problem of Equation \ref{EquationDecomposition1} can be formulated in a general way as follows:
\begin{equation}
\underset{M_i}{\text{min}} ~~   \sum_{i=1}^K \lambda_i f_{i}(M_i)  ~~~ \text{subj} ~~~ C_i
\label{EquationMinimization}
\end{equation}
where the $\lambda_i$ are the regularization parameters. $f_i()$ are the loss functions with $f_1()=f_{low}()$,  $f_2()=f_{sparse}()$ and $f_3()=f_{noise}$. $C_i$ is a constraint on $L$, $S$ and $E$ with $A$ which varies following the value of $K$, that is $C_1$ is $||A-L||_2=0$, $C_2$ is $||A-L-S||_2=0$, $C_3$ is $||A-L-S-E||_2=0$. $C_i$ can be expressed in an inequality form too. \\

\underline{\textbf{Case $K=1$:}} It is the degenerated case for the implicit decompositions (LRM, MC) which are not robust because there are no constraints on the matrix $S=A-L$. In this case, Equation \ref{EquationMinimization}  with $K=1$ can be written as follows:
\begin{equation}
\underset{L}{\text{min}} ~~  \lambda_1 f_{low}(L)  ~~~ \text{subj} ~~~ C_1
\label{EquationMinimization1}
\end{equation}

\underline{\textbf{Case $K=2$:}} Equation \ref{EquationMinimization} with $K=2$ for the explicit decompositions becomes as follows:
\begin{equation}
\underset{L,S}{\text{min}} ~~  \lambda_1 f_{low}(L)  + \lambda_2 f_{sparse}(S) ~~~ \text{subj} ~~~ C_2
\label{EquationMinimization2}
\end{equation}
where $\lambda_1$ and $\lambda_2$ are the regularization parameters. $C_2$ is a constraint on $A$, $L$ and $S$. $f_{low}(L)$ is a loss function which constrains the matrix $L$ to be low-rank as the following ones: $rank(.)$, or the surrogated norm $||.||_{*}$. $f_{sparse}(S)$ is a loss function which constrains the matrix $S$ to be sparse as the following ones: $||.||_{l_{0}}$, or the surrogated norm $||.||_{l_{1}}$. An overview of the different loss functions $f_{low}(.)$ and $f_{sparse}()$ which are used in the literature are shown in Table \ref{LossFunctionRPCA}. \\

\indent This minimization problem can be $NP$-hard, and convex or not following the constraints and the loss functions used. Practically, when the problem is $NP$-hard and/or not convex, the constraints are relaxed by changing the loss functions to obtain a tractable and convex problem. For example, the original formulation in RPCA \cite{3} used the $rank(.)$ and the $l_0$-norm as original loss functions for $L$ and $S$, respectively as shown in Equation \ref{EquationPCP2}. As this problem is $NP$-hard, this formulation is relaxed with the nuclear norm and the $l_1$-norm as shown in Equation \ref{EquationPCP3}. To minimize confusion, the models that minimize rank functions and nuclear norms are named the original model and the relaxed model, respectively. \\

\indent Thus, the corresponding minimization problem of Equation \ref{EquationMinimization2} can be formulated with norms to be convex and solvable as follows:
\begin{equation}
\underset{L,S}{\text{min}} ~~  \lambda_1 ||L||_{norm_1}^{p_1} + \lambda_2 ||S||_{norm_2}^{p_2} ~~~ \text{subj} ~~~ C_2
\label{EquationMinimization21}
\end{equation}
where $\lambda_1$ and $\lambda_2$ are the regularization parameters. $p_1$ and $p_2$ are taken in the set $\left\{1,2\right\}$. $ ||.||_{norm_1}$ and $||.||_{norm_2}$ could be any norm of the following set of norms: $l_\alpha$-norm, $l_\infty$, $l_{\alpha,\beta}$ mixed norm,  $L_\alpha$-seminorm,  Frobenius norm, nuclear norm, dual norm and Schatten norm. $||.||_{norm_1}$ and  $||.||_{norm_2}$ are taken to enforce the low-rank and sparsity constraints of $L$ and $S$, respectively. The constraint $C_1$ is generally based on \textbf{1)} an equality such as $||A-L-S||_{norm_0}^{p_0}=0$ or $rank(L)=r$, or \textbf{(2)} an inequality such as $||A-L-S||_{norm_0}^{p_0} \leq q$ or $rank(L)\leq r$. $||.||_{norm_0}$ is a norm taken in the set of norms previously defined. Moreover, the minimization problem formulated in Equation \ref{EquationMinimization21} can be written in its Lagrangian form as follows:
\begin{equation}
\underset{L,S}{\text{min}} ~~ \frac{\lambda_0}{2}||A-L-S||_{norm_0}^{p_0} +  \lambda_1 ||L||_{norm_1}^{p_1} + \lambda_2 ||S||_{norm_2}^{p_2}   ~~~ \text{subj} ~~~ C_{L2}
\label{EquationLagrangianMinimization}
\end{equation}
where $\lambda_0$ is regularization parameter. $C_{L2}$ is the constraint similar to the constraint $C_2$. \\

\underline{\textbf{Case $K=3$:}} Equation \ref{EquationMinimization} with $K=3$ for the stable explicit decomposition is written as follows:
\begin{equation}
\underset{L,S}{\text{min}} ~~  \lambda_1 f_{low}(L)  + \lambda_2 f_{sparse}(S) + \lambda_3 f_{noise}(E) ~~~ \text{subj} ~~~ C_3
\label{EquationMinimizationSPCP}
\end{equation}
where the $\lambda_i$ are the regularization parameters. $C_3$ is a constraint on $A$, $L$, $S$ and $E$. $f_{noise}(E)$ is a function that contrains $E$. Thus, the corresponding minimization problem of Equation \ref{EquationMinimizationSPCP} can be formulated with norms to be convex and solvable as follows:
\begin{equation}
\underset{L,S}{\text{min}} ~~  \lambda_1 ||L||_{norm_1}^{p_1} + \lambda_2 ||S||_{norm_2}^{p_2} + \lambda_3 ||E||_{norm_3}^{p_3} ~~~ \text{subj} ~~~ C_3
\label{EquationMinimization1SPCP}
\end{equation}
where $p_3$ is taken in the set $\left\{1,2\right\}$. $||.||_{norm_3}$ could be any previous norms. Then, the minimization problem formulated in Equation \ref{EquationMinimization1SPCP} can be written in its Lagrangian form as follows:
\begin{equation}
\underset{L,S}{\text{min}} ~~ \frac{\lambda_0}{2}||A-L-S-E||_{norm_0}^{p_0} +  \lambda_1 ||L||_{norm_1}^{p_1} + \lambda_2 ||S||_{norm_2}^{p_2} \lambda_3 ||E||_{norm_3}^{p_3}   ~~~ \text{subj} ~~~ C_{L3}
\label{EquationLagrangianMinimizationSPCP}
\end{equation}
where $\lambda_0$ is regularization parameter. $C_{L3}$ is the constraint similar to the constraint $C_3$.  Finally, the minimization problem seeks to the following optimization problem: $F(X)= f(x)+ g(x)$ where we have:
\begin{equation} 
f(x)=\frac{\lambda_0}{2}||A-L-S-E||_{norm_0}^{p_0} 
\end{equation}
\begin{equation}
g(x)=\lambda_1 ||L||_{norm_1}^{p_1} + \lambda_2 ||S||_{norm_2}^{p_2} + \lambda_3 ||E||_{norm_3}^{p_3}
\end{equation}
where we have: 
\begin{itemize}
\item $f(x)$ is smooth and convex function which has Lipschitz continuous gradients \cite{5010}. $f(x)$ contains the loss function \cite{1590}.  \\
\item $g(x)$ can be a nonsmooth and nonconvex function \cite{5010}. $g(x)$ contains the low-rank regularizer, the sparse  regularizer and the noise regularizer \cite{1590}. \\
\end{itemize}
In general, solving a nonsmooth and nonconvex objective function is difficult with weak convergence guarantees \cite{5010}.

\subsubsection{Algorithms for solving the optimization problem}
\label{subsubsec:Solvers} 
Algorithms which are called solvers are then used to solve the minimization problem in its original form or in its Lagrangian form. Furthermore, instead of directly solving the original convex optimizations, some authors use their strongly convex approximations in order to design efficient algorithms. Zhang et al. \cite{44} proved that these strongly convex programmings guarantee the exact low-rank matrix recovery as well. Moreover, solvers have different characteristics in terms of complexities: complexity per iteration, complexity to reach an accuracy of $\epsilon$ precision ($\epsilon$-optimal solution), and convergence rate complexity following the number of iterations. The key challenges related to the solvers are the following ones \cite{5001}: 
\begin{enumerate}
\item Choice of the solver to make the iterations as few as possible.
\item Choice of the SVD algorithm to make the iterations as efficient as possible. 
\end{enumerate}
\indent The solvers can be broadly classified into two categories as developed by Chen \cite{1045}: \\
\begin{itemize}
\item \textbf{Regularization based approaches:} The decomposition is formulated as regularized fitting, where the
regularizers are convex surrogates for rank and sparsity. Examples of algorithms in this category include the following solvers: Singular Value Thresholding (SVT) \cite{20}, the Accelerated Proximal Gradient (APG) \cite{19}, and the Augmented Lagrange Multiplier (ALM) \cite{18}.  All the solvers for the different problem formulations are grouped in Table \ref{A-TPCP1Overview}, Table \ref{A-TPCP2Overview}, Table \ref{A-TPCP3Overview}, and Table \ref{A-TPCP4Overview}. \\
\item \textbf{Statistical inference based approaches:} Hierarchical statistical models are used to model the data generation process and prior distributions are selected to capture the low-rank and sparse properties of the respective terms. The joint distribution involving the observations, unknown variables and hyperparameters can be determined from the priors and conditional distributions. Posterior distributions of the unknown variables are approximated using Bayesian inference approaches. Representative algorithms in this category can be found in \cite{7}\cite{24}\cite{94}\cite{1024}. 
\end{itemize}

\indent For the SVD algorithms, approximated SVD solutions exist to avoid full SVD such as partial SVD \cite{18}, linear time SVD \cite{73}, limited memory SVD \cite{2301}, symmetric low-rank product-Gauss-Newton \cite{2302}, and Block Lanczos with Warm Start (BLWS) \cite{57}. 	 

\begin{table}
\scalebox{0.79}{
\begin{tabular}{|l|l|}
\hline
\scriptsize{Solvers for PCP} &\scriptsize{Complexity} \\
\hline
\hline
\scriptsize{\textbf{Basic solvers}} &\scriptsize{} \\
\hline
\scriptsize{Singular Value Threshold (SVT\protect\footnotemark[1])} 	&\scriptsize{$O_{iter}(mnmin(mn))$, $O_{pre}$=unknown, $O_{conv}$=unknown}	   \\ 
\scriptsize{Cai et al. (2008) \cite{20}}					&\scriptsize{}   							    \\ 
\hline
\scriptsize{Iterative Thresholding (IT)}				     																 
&\scriptsize{$O_{iter}(mnmin(mn))$, $O_{pre}(\sqrt{L/\epsilon})$, $O_{conv}=1/T^2$}      \\ 
\scriptsize{Wright et al. (2009) \cite{4}}					 &\scriptsize{}			               \\ 
\hline
\scriptsize{Accelerated Proximal Gradient (APG\protect\footnotemark[1])} 																	
&\scriptsize{$O_{iter}(mnmin(mn))$,  $O_{pre}(\sqrt{1/\epsilon})$, $O_{conv}(1/T^2)$}     \\ 
\scriptsize{Lin et al. (2009) \cite{19}}							  &\scriptsize{Full SVD}		     	  \\ 
\hline
\scriptsize{Dual Method (DM\protect\footnotemark[1])}																							        
&\scriptsize{$O_{iter}(rmn)$, $O_{pre}(\sqrt{1/\epsilon})$, $O_{conv}(1/T^2)$}	          \\	
\scriptsize{Lin et al. (2009)\cite{19}}								 &\scriptsize{Partial SVD}	        \\ 
\hline
\scriptsize{Exacted Augmented Lagrangian Method (EALM)}			  															 
&\scriptsize{$O_{iter}(mnmin(mn))$, $O_{pre}$=unknown, $O_{conv}(1/\mu_T)$}	               \\ 
\scriptsize{(EALM\protect\footnotemark[1]) (2009) Lin et al.\cite{18}}			&\scriptsize{Full SVD} \\ 
\hline
\scriptsize{Inexact Augmented Lagrangian Method (IALM)}		    															 
&\scriptsize{$O_{iter}(rmn)$, $O_{pre}$=unknown, $O_{conv}(1/\mu_T)$}	                     \\ 
\scriptsize{(IALM\protect\footnotemark[1]) (2009) Lin et al. \cite{18}}		        															
&\scriptsize{Partial SVD, Linear Time SVD \cite{73}}	                                     \\ 
\scriptsize{}		    									 						 
&\scriptsize{Limited Memory SVD (LMSVD\protect\footnotemark[2]) \cite{2301}}	             \\ 
\scriptsize{}		        															
&\scriptsize{Symmetric Low-Rank Product-Gauss-Newton \cite{2302}}	                         \\ 
\hline
\scriptsize{Alternating Direction Method (ADM)}		  															
&\scriptsize{$O_{iter}(mnmin(mn))$, $O_{pre}$=unknown, $O_{conv}$=unknown} 	                \\ 
\scriptsize{(LRSD\protect\footnotemark[3]) Yuan and Yang  (2009) \cite{21}}&\scriptsize{}   \\   
\hline
\scriptsize{Symmetric Alternating Direction Method (SADM\protect\footnotemark[4])}	 											 
&\scriptsize{$O_{iter}$=Unknown, $O_{pre}(1/ \epsilon)$, $O_{conv}$=Unknown}                \\
\scriptsize{(SADAL) Ma (2010) \cite{46}, Goldfarb et al. (2010) \cite{47}}	 &\scriptsize{} \\
\hline
\scriptsize{Non Convex Splitting  ADM (NCSADM)}	 											 
&\scriptsize{$O_{iter}$=Unknown, $O_{pre}$=Unknown, $O_{conv}$=Unknown}                   \\
\scriptsize{Chartrand (2012) \cite{65}}	 &\scriptsize{}                                   \\
\hline
\scriptsize{Variant of Douglas-Rachford Splitting Method (VDRSM)}	 											 
&\scriptsize{$O_{iter}$=Unknown, $O_{pre}$=Unknown, $O_{conv}$=Unknown}      \\
\scriptsize{Zhang and Liu (2013) \cite{79}}	 &\scriptsize{}                  \\
\hline
\scriptsize{Proximity Point Algorithm (PPA)}
&\scriptsize{$O_{iter}$=Unknown, $O_{pre}$=Unknown, $O_{conv}$=Unknown}      \\
\scriptsize{Zhu et al. (2014) \cite{1017}}	 &\scriptsize{}                  \\
\hline
\scriptsize{Proximal Iteratively Reweighted Algorithm (PIRA)}	 											 
&\scriptsize{$O_{iter}$=Unknown, $O_{pre}$=Unknown, $O_{conv}$=Unknown}      \\
\scriptsize{Wang et al. (2014) \cite{1022} (5)}	 &\scriptsize{}              \\
\hline
\scriptsize{Alternating Rectified Gradient Method (ARGM)}	 											 
&\scriptsize{$O_{iter}$=Unknown, $O_{pre}$=Unknown, $O_{conv}$=Unknown}      \\
\scriptsize{($l_1$-ARG) Kim et al. (2014) \cite{1016}}	 &\scriptsize{}      \\
\hline
\scriptsize{Parallel Direction Method of Multipliers (PDMM)}	 											 
&\scriptsize{$O_{iter}$=Unknown, $O_{pre}$=Unknown, $O_{conv}(1/T)$}      \\
\scriptsize{Wang et al. (2014)  \cite{1032}}	 &\scriptsize{}             \\
\hline
\scriptsize{Generalized Accelerated Proximal Gradient (GAPG)}	 											 
&\scriptsize{$O_{iter}$=Unknown, $O_{pre}$=Unknown, $O_{conv}$=Unknown}      \\
\scriptsize{He et al. (2013) \cite{87}}	 &\scriptsize{}                      \\
\hline
\scriptsize{Improved alternating direction method (IADM)}	 											 
&\scriptsize{$O_{iter}$=Unknown, $O_{pre}$=Unknown, $O_{conv}$=Unknown}      \\
\scriptsize{Chai et al. (2013) \cite{91}}	 &\scriptsize{}                    \\
\hline
\end{tabular}}
\caption{Solvers for RPCA-PCP: An overview of their complexity per iteration at running time $O_{iter}$, their complexity $O_{pre}$ to reach an accuracy of $\epsilon$ precision and their convergence rate $O_{conv}$ for $T$ iterations. "Unknown" stands for not indicated by the authors.} 
\label{A-TPCP1Overview}
\end{table}

\footnotetext[1]{{http://perception.csl.uiuc.edu/matrix-rank/samplecode.html}}
\footnotetext[2]{{http://www.caam.rice.edu/~yzhang/LMSVD/lmsvd.html}}
\footnotetext[3]{{http://math.nju.edu.cn/~jfyang/LRSD/index.html}}
\footnotetext[4]{Available on request by email to the corresponding author}

\clearpage
\begin{table}
\scalebox{0.70}{
\begin{tabular}{|l|l|}
\hline
\scriptsize{Solvers for PCP} &\scriptsize{Complexity} \\
\hline
\hline
\scriptsize{\textbf{Linearized solvers}} &\scriptsize{} \\
\hline
\scriptsize{Linearized Augmented Lagrangian Method (LALM)}					                
&\scriptsize{$O_{iter}(mnmin(mn))$, $O_{pre}$=Unknown, $O_{conv}$=Unknown}    \\ 
\scriptsize{Yang and Yuan (2011) \cite{22}}				&\scriptsize{}              \\ 
\hline
\scriptsize{Linearized Alternating Direction Method (LADM)}												 
&\scriptsize{$O_{iter}(mnmin(mn))$, $O_{pre}$=Unknown, $O_{conv}$=Unknown}	   \\
\scriptsize{Yang and Yuan (2011) \cite{22}}					&\scriptsize{}					   \\ 
\hline
\scriptsize{LADM with Adaptive Penalty (LADMAP\protect\footnotemark[5])} 
&\scriptsize{$O_{iter}(rmn)$, $O_{pre}(1/ \epsilon)$, $O_{conv}$=Unknown}        \\ 
\scriptsize{Lin et al. (2011) \cite{39}}		&\scriptsize{Accelerated version}    \\ 
\hline
\scriptsize{Linearized Symmetric Alternating Direction Method (LSADM\protect\footnotemark[5])} 						
&\scriptsize{$O_{iter}$=Unknown, $O_{pre}(1/ \epsilon)$, $O_{conv}$=Unknown}  \\
\scriptsize{(ALM) Ma (2010) \cite{46}, Goldfarb et al. (2010) \cite{47}}	 &\scriptsize{}  \\ 
\hline
\scriptsize{Fast Linearized Symmetric Alternating Direction Method (Fast-LSADM\protect\footnotemark[5])} 
&\scriptsize{$O_{iter}$=Unknown, $O_{pre}(\sqrt{1/\epsilon})$, $O_{conv}$=Unknown}  \\
\scriptsize{(FALM) Ma (2010) \cite{46}, Goldfarb et al. (2010) \cite{47}}	&\scriptsize{}          \\ 
\hline
\scriptsize{Linearized Alternating Direction Method (LADM)}						           	
&\scriptsize{$O_{iter}(rmn)$,  $O_{pre}(1/ \epsilon)$, $O_{conv}$=Unknown}                 \\ 
\scriptsize{(LMaFit\protect\footnotemark[6]) Shen et al. (2011) \cite{23}} &\scriptsize{}  \\ 	                  
\hline
\scriptsize{\textbf{Fast solvers}}&\scriptsize{} \\
\hline
\scriptsize{Randomized Projection for ALM (RPALM)}	    				  										
&\scriptsize{$O_{iter}(pmn)$, $O_{pre}$=Unknown, $O_{conv}$=Unknown}      			              \\ 
\scriptsize{Mu et al. (2011) \cite{11}}												   &\scriptsize{}                \\ 
\hline
\scriptsize{$l_1$-filtering (LF\protect\footnotemark[5])}	         																				
&\scriptsize{$O_{iter}(r^2(m+n))$, $O_{pre}$=Unknown, $O_{conv}$=Unknown}                    \\ 
\scriptsize{Liu et al. (2011) \cite{38}\cite{38-1}}							&\scriptsize{}                \\	
\hline
\scriptsize{Block Lanczos with Warm Start}
&\scriptsize{less than $O_{iter}(pmn)$, $O_{pre}$=Unknown, $O_{conv}$=Unknown}               \\ 
\scriptsize{Lin and Wei (2010) \cite{57}}												&\scriptsize{Partial SVD}    \\
\hline
\scriptsize{Exact Fast Robust Principal Component Analysis (EFRPCA)}	         																				
&\scriptsize{$O_{iter}(mk^2)$ with $k \ll n$, $O_{pre}$=Unknown, $O_{conv}$=Unknown}         \\ 
\scriptsize{Abdel-Hakim and El-Saban (2012) \cite{59}}							&\scriptsize{Full SVD}   \\ 
\hline
\scriptsize{Inexact Fast Robust Principal Component Analysis (IFRPCA)}	         																				
&\scriptsize{$O_{iter}(mk^2)$ with $k \ll n$, $O_{pre}$=Unknown, $O_{conv}$=Unknown}        \\ 
\scriptsize{Abdel-Hakim and El-Saban (2012) \cite{59}}						&\scriptsize{Partial SVD} \\
\hline
\scriptsize{Matrix Tri-Factorization (MTF)}	         																			
&\scriptsize{$O_{iter}(n^3+(r^3+r^2n+mn^2+rn^2))$}   \\ 
\scriptsize{Liu et al. (2013) \cite{60}}		&\scriptsize{$O_{pre}$=Unknown, $O_{conv}$=Unknown}     \\
\hline
\scriptsize{Fast Tri-Factorization(FTF)}	         																			
&\scriptsize{$O_{iter}(r^3+r^2(m+n)+r^2m+rmn)$}      \\ 
\scriptsize{Liu et al. (2013) \cite{61}}			&\scriptsize{$O_{pre}$=Unknown, $O_{conv}$=Unknown}     \\
\hline
\scriptsize{PRoximal Iterative SMoothing Algorithm (PRISMA)}	         																			
&\scriptsize{$O_{iter}(nm)$, $O_{pre}(\text{log}(\epsilon)/ \epsilon)$, $O_{conv}$=Unknown}  \\ 
\scriptsize{Orabona et al. (2012) \cite{71}}					&\scriptsize{}                          \\
\hline
\scriptsize{Fast Alterning Minimization (FAM)\protect\footnotemark[7]}
&\scriptsize{$O_{iter}$=Unknown, $O_{pre}$=Unknown, $O_{conv}$=Unknown}                     \\ 
\scriptsize{Rodriguez and Wohlberg (2013) \cite{26-1}}							&\scriptsize{}          \\
\hline
\scriptsize{Fast Alternating Direction Method of Multipliers (FADMM)}
&\scriptsize{$O_{iter}$=Unknown, $O_{pre}$=Unknown, $O_{conv}$=Unknown}                    \\ 
\scriptsize{Yang and Wang (2014) \cite{1028}}					&\scriptsize{}                       \\
\hline
\scriptsize{Fast Alternating Direction Method with Smoothing Technique (FADM-ST)}
&\scriptsize{$O_{iter}$=Unknown, $O_{pre}$=Unknown, $O_{conv}$=Unknown}                     \\ 
\scriptsize{Yang (2014) \cite{1039}}											&\scriptsize{}                    \\
\hline
\scriptsize{\textbf{Online solvers}}& \scriptsize{}\\
\hline
\scriptsize{Online Alternating Direction Method (OADM)}			&\scriptsize{}                  \\ 
\scriptsize{Wang and Banerjee (2013) \cite{1019}}															                 
&\scriptsize{$O_{iter}$=Unknown, $O_{pre}$=Unknown, $O_{conv}(1/T)$}                        \\
\hline
\scriptsize{\textbf{Non convex solvers}} &\scriptsize{}\\
\hline
\scriptsize{Difference of Convex (DC)}					                
&\scriptsize{$O_{iter}$=Unknown, $O_{pre}$=Unknown, $O_{conv}$=Unknown}                      \\ 
\scriptsize{Sun et al. (2013) \cite{82}}						    &\scriptsize{}                       \\
\hline
\scriptsize{Fast Alternating Difference of Convex (FADC)}
&\scriptsize{$O_{iter}$=Unknown, $O_{pre}$=Unknown, $O_{conv}$=Unknown}                      \\ 
\scriptsize{Sun et al. (2013) \cite{82}}								&\scriptsize{}                       \\
\hline
\scriptsize{Non-convex Alternating Projections (AltProj)}
&\scriptsize{$O_{iter}(r^2mn)$, $O_{pre}(\text{log}(1/ \epsilon))$, $O_{conv}$=Unknown}     \\ 
\scriptsize{Netrapalli et al. (2014) \cite{1047}}		&\scriptsize{}                          \\
\hline
\scriptsize{Fixed Rank - Fast Factorization based RPCA (F-FFP)}
&\scriptsize{$O_{iter}(kmn) $ with $k \ll n$, $O_{pre}$=Unknown, $O_{conv}$=Unknown}          \\ 
\scriptsize{Peng et al. (2016) \cite{17110}}		&\scriptsize{}                                \\
\hline
\scriptsize{Unfixed Rank - Fast Factorization based RPCA (U-FFP)}
&\scriptsize{$O_{iter}(kmn) $ with $k \ll n$, $O_{pre}$=Unknown, $O_{conv}$=Unknown}          \\ 
\scriptsize{Peng et al. (2016)  \cite{17110}}		&\scriptsize{}                                \\
\hline
\scriptsize{\textbf{2D solvers}} &\scriptsize{}\\
\hline
\scriptsize{Iterative method for Bi-directional Decomposition (IMBD)}					                
&\scriptsize{$O_{iter}$=Unknown, $O_{pre}$=Unknown, $O_{conv}$=Unknown}                     \\ 
\scriptsize{Sun et al. (2013)\cite{1005}}															       &\scriptsize{} \\
\hline
\end{tabular}}
\caption{Solvers for RPCA-PCP: An overview of their complexity per iteration at running time $O_{iter}$, their complexity $O_{pre}$ to reach an accuracy of $\epsilon$ precision and their convergence rate $O_{conv}$ for $T$ iterations. "Unknown" stands for not indicated by the authors.} 
\label{A-TPCP11Overview}
\end{table}

\footnotetext[5]{Available on request by email to the corresponding author}
\footnotetext[6]{http://lmafit.blogs.rice.edu/}
\footnotetext[7]{{https://sites.google.com/a/istec.net/prodrig/Home/en/pubs}}

\clearpage
\begin{table}
\scalebox{0.70}{
\begin{tabular}{|l|l|}
\hline
\scriptsize{Solvers}&\scriptsize{Complexity} \\
\hline
\hline
\scriptsize{RPCA via SPCP (RPCA-SPCP)} &\scriptsize{Zhou et al. \cite{5}} \\
\hline
\scriptsize{Alternating Splitting Augmented Lagrangian method (ASALM\protect\footnotemark[8])}	  
&\scriptsize{$O_{iter}$=Unknown, $O_{pre}$=Unknown, $O_{conv}$=Unknown}				       \\
\scriptsize{Tao and Yuan \cite{35}}						     &\scriptsize{}				             \\	
\hline
\scriptsize{Variational ASALM (VASALM\protect\footnotemark[8])} &\scriptsize{$O_{iter}$=Unknown, $O_{pre}$=Unknown, $O_{conv}$=Unknown}  	    \\ 
\scriptsize{Tao and Yuan \cite{35}}	&\scriptsize{}  			             \\ 
\hline
\scriptsize{Parallel ASALM (PSALM\protect\footnotemark[8])}	   &\scriptsize{$O_{iter}$=Unknown, $O_{pre}$=Unknown, $O_{conv}$=Unknown}  	    \\ 
\scriptsize{Tao and Yuan \cite{35}}	&\scriptsize{}  			             \\ 
\hline
\scriptsize{Non Smooth Augmented Lagrangian Algorithm (NSA\protect\footnotemark[9])} &\scriptsize{$O_{iter}$=Unknown, $O_{pre}$=Unknown, $O_{conv}$=Unknown} \\
\scriptsize{Aybat et al. \cite{34}}  &\scriptsize{}  			             \\ 
\hline
\scriptsize{First-order Augmented Lagrangian algorithm for Composite norms (FALC\protect\footnotemark[9])}               &\scriptsize{$O_{iter}$=Unknown,  $O_{conv}$=Unknown, $O_{pre}(1/\epsilon)$}  \\
\scriptsize{Aybat et al. \cite{34-1}} &\scriptsize{}  			             \\ 
\hline
\scriptsize{Augmented Lagragian method for Conic Convex  (ALCC\protect\footnotemark[9])}       
&\scriptsize{$O_{iter}$=Unknown,  $O_{conv}$=Unknown, $O_{pre}(\text{log}(1/\epsilon))$} \\
\scriptsize{Aybat et al. \cite{34-2}} &\scriptsize{}  			             \\ 
\hline
\scriptsize{Partially Smooth Proximal Gradient (PSPG\protect\footnotemark[9])}                  
&\scriptsize{$O_{iter}$=Unknown,  $O_{conv}$=Unknown, $O_{pre}(1/\epsilon)$} \\
\scriptsize{Aybat et al. \cite{34-3}} &\scriptsize{}                         \\ 
\hline
\scriptsize{Alternating Direction Method - Increasing Penalty (ADMIP\protect\footnotemark[9])}  &\scriptsize{$O_{iter}$=Unknown,  $O_{conv}$=Unknown, $O_{pre}$=Unknown} \\
\scriptsize{Aybat et al.\cite{34-4}} &\scriptsize{}                           \\ 
\hline
\scriptsize{Inexact Alternating Minimization - Matrix Manifolds (IAM-MM)}            
&\scriptsize{$O_{iter}(rmn)$,$O_{conv}$=Unknown, $O_{pre}$=Unknown} \\
\scriptsize{(R2PCP\protect\footnotemark[10]) Hinterm\"{u}ller and Wu \cite{1036}}	&\scriptsize{} \\
\scriptsize{Partially Parallel Splitting - Multiple Block (PPS-MB)} 
&\scriptsize{$O_{iter}(rmn)$,$O_{conv}=1/t$, $O_{pre}$=Unknown}         \\
\scriptsize{(NEW, NEW-R) Hou et al. (2015) \cite{1394}}		&\scriptsize{} \\
\hline
\scriptsize{RPCA via SPCP (RPCA-SPCP)(2)} &\scriptsize{Zhou and Tao \cite{601}}   \\
\hline
\scriptsize{Greedy Bilateral Smoothing (GreBsmo\protect\footnotemark[11])}	  
&\scriptsize{$O_{iter}(max(\left|\Omega\right|r^2, mnr^3)$, $O_{conv}$=Unknown, $O_{pre}$=Unknown}												         \\
\scriptsize{Zhou and Tao \cite{601}}						       &\scriptsize{}				\\	
\scriptsize{Bilinear Generalized Approximate Message Passing (BiG-AMP\protect\footnotemark[12])}      									&\scriptsize{$O_{iter}(mn+nl+ml)$, $O_{conv}$=Unknown, $O_{pre}$=Unknown} \\ 
\scriptsize{Parker and Schniter \cite{80}}		  		  &\scriptsize{}  		\\ 
\hline
\scriptsize{RPCA via Quantized PCP (RPCA-QPCP)} &\scriptsize{Becker et al. \cite{8}}   \\
\hline
\scriptsize{Templates for First-Order Conic Solvers (TFOCS\protect\footnotemark[13])}	 &\scriptsize{$O_{iter}(m \text{log} n)$, $O_{conv}$=Unknown, $O_{pre}(1/\epsilon)$}  		 \\ 
\scriptsize{Becker et al. \cite{8}}		&\scriptsize{}  			\\	
\hline
\scriptsize{RPCA via Block based PCP (RPCA-BPCP)} &\scriptsize{Tang and Nehorai \cite{41}}  \\
\hline
\scriptsize{Augmented Lagrangian Method (ALM)}	
&\scriptsize{$O_{iter}(mnmin(mn))$, $O_{conv}$=Unknown, $O_{pre}$=Unknown} \\ 
\scriptsize{RPCA-LBD\protect\footnotemark[8]) Tang and Nehorai \cite{41}}		 &\scriptsize{} \\
\hline
\scriptsize{RPCA via Local PCP (RPCA-LPCP)} &\scriptsize{Wohlberg et al. \cite{26}}         \\
\hline
\scriptsize{Split Bregman Algorithm (SBA)}  &\scriptsize{$O_{iter}$=Unknown, $O_{conv}$=Unknown, $O_{pre}$=Unknown}  	   \\ 
\scriptsize{Goldstein and Osher \cite{106}} &\scriptsize{}                                                               \\ 
\hline
\scriptsize{RPCA via Outlier Pursuit (RPCA-OP\protect\footnotemark[14])} &\scriptsize{Xu et al. \cite{56}}               \\
\hline
\scriptsize{Singular Value Threshold (SVT)}	   &\scriptsize{$O_{iter}(mnmin(mn))$, $O_{conv}$=Unknown, $O_{pre}$=Unknown} \\ 
\scriptsize{Cai et al. \cite{20}}		           &\scriptsize{}  		                                                        \\ 
\hline
\scriptsize{RPCA with Sparsity Control (RPCA-SpaCtrl)} &\scriptsize{Mateos and Giannakis \cite{9}\cite{10}}               \\
\hline
\scriptsize{Alternating Minimization (AM)}	&\scriptsize{$O_{iter}$=Unknown, $O_{conv}$=Unknown, $O_{pre}$=Unknown}  		  \\ 
\scriptsize{Zhou et al. \cite{107}}			  	&\scriptsize{}                                                                \\ 
\hline
\scriptsize{RPCA via Sparse Corruptions (RPCA-SpaCorr)} &\scriptsize{Hsu et al. \cite{58}}                                \\
\hline
\scriptsize{-}													 	&\scriptsize{$O_{iter}$=Unknown, $O_{conv}$=Unknown, $O_{pre}$=Unknown}  		    \\ 
\scriptsize{-}				  								  &\scriptsize{}                                                                  \\ 

\hline
\scriptsize{RPCA via Log-sum Heuristic Recovery (RPCA-LHR)} &\scriptsize{Deng et al. \cite{55}}                           \\
\hline
\scriptsize{Majorization-Minimization (MM)}	&\scriptsize{$O_{iter}$=Unknown, $O_{conv}$=Unknown, $O_{pre}$=Unknown}  		  \\ 
\scriptsize{Fazel \cite{108}, Lange et al. \cite{109}}			&\scriptsize{}                                                \\ 
\hline
\scriptsize{Bayesian RPCA (B-RPCA\protect\footnotemark[15])} &\scriptsize{Ding et al. \cite{7}}                            \\
\hline
\scriptsize{Markov chain Monte Carlo (MCMC)}   &\scriptsize{$O_{iter}(r(m+n)+mn)$}  	   																	\\ 
\scriptsize{Robert and Cassela \cite{113}}	   &\scriptsize{$O_{conv}$=Unknown, $O_{pre}$=Unknown}                         \\	
\hline
\scriptsize{Variational Bayesian Inference (VB)}		
&\scriptsize{$O_{iter}$=Unknown, $O_{conv}$=Unknow, $O_{pre}$=Unknow}                                                    \\	
\scriptsize{Beal \cite{102}}			                  &\scriptsize{} 	                                                     \\
\hline
\scriptsize{Variational Bayesian RPCA (VB-RPCA\protect\footnotemark[16])} &\scriptsize{Babacan et al. \cite{24}}         \\
\hline
\scriptsize{Approximate Bayesian Inference (AB)}							 &\scriptsize{$O_{iter}(min(n^3,r^3)+min(m^3,r^3))$}  	   \\
\scriptsize{Beal \cite{102}}	&\scriptsize{$O_{conv}$=Unknown, $O_{pre}$=Unknown}                                        \\
\hline
\scriptsize{Approximated RPCA (A-RPCA)} &\scriptsize{(GoDec\protect\footnotemark[17]) Zhou and Tao \cite{6}}             \\
\hline
\scriptsize{Naive GoDec}				               &\scriptsize{Linear convergence}  \\ 
\scriptsize{Zhou and Tao \cite{6}}			       &\scriptsize{}                    \\
\scriptsize{Fast Godec via Bilateral Random Projection}					&\scriptsize{Linear convergence}  \\
\scriptsize{Zhou and Tao \cite{6}}															&\scriptsize{}                    \\
\hline
\end{tabular}}
\caption{Solvers for RPCA (excepted PCP): An overview of their complexity per iteration at running time $O_{iter}$, their complexity $O_{pre}$ to reach an accuracy of $\epsilon$ precision and their convergence rate $O_{conv}$ for $T$ iterations. "Unknown" stands for not indicated by the authors.}
\label{A-TPCP2Overview}
\end{table}

\footnotetext[8]{Available on request by email to the corresponding author}
\footnotetext[9]{http://www2.ie.psu.edu/aybat/codes.html}
\footnotetext[10]{http://www.uni-graz.at/imawww/ifb/r2pcp/index.html}
\footnotetext[11]{https://sites.google.com/site/godecomposition/GreBsmo.zip}
\footnotetext[12]{http://www2.ece.ohio-state.edu/~schniter/BiGAMP/BiGAMP.html}
\footnotetext[13]{{http://cvxr.com/tfocs/}}
\footnotetext[14]{{http://guppy.mpe.nus.edu.sg/~mpexuh/publication.html}}
\footnotetext[15]{{http://people.ee.duke.edu/~lcarin/BCS.html}} 
\footnotetext[16]{{http://www.dbabacan.info/software.html}}
\footnotetext[17]{{http://sites.google.com/site/godecomposition/code}}

\begin{landscape}
\begin{table}
\scalebox{0.70}{
\begin{tabular}{|l|l|l|}
\hline
\scriptsize{Methods}&\scriptsize{Solvers}&\scriptsize{Complexity} \\
\hline
\hline
\scriptsize{\textbf{Robust Non-negative Matrix Factorization (RNMF)}}	&\scriptsize{}
&\scriptsize{}  \\
\scriptsize{Manhattan Non-negative Matrix Factorization (MahNMF\protect\footnotemark[17])}	&\scriptsize{Rank-one Residual Iteration (RRI)}
&\scriptsize{$O_{iter}(mnr(\text{log}(m)+1))$, $O_{pre}$=Unknown, $O_{conv}$=Unknown}  \\
\scriptsize{Guan et al. \cite{761}}	    &\scriptsize{Guan et al. \cite{761}}                          &\scriptsize{}        \\
\scriptsize{}	&\scriptsize{Nesterov's smoothing method (OGM)}
&\scriptsize{$O_{iter}$=Unknown, $O_{pre}(1/ \epsilon)$,  $O_{conv}$=Unknown}  \\
\scriptsize{}	    &\scriptsize{Nesterov \cite{3003}}                                                  &\scriptsize{}        \\
\hline
\scriptsize{Near-separable Non-negative Matrix Factorization (RobustXray)}	&\scriptsize{Alternating Direction Method of Multipliers (ADMM)}
&\scriptsize{$O_{iter}$=Unknown, $O_{pre}$=Unknown, $O_{conv}$=Unknown}  \\
\scriptsize{Kumar and Sindhwani \cite{762}}	    &\scriptsize{Boyd et al. \cite{111}}                          &\scriptsize{}        \\
\hline
\scriptsize{Robust Asymmetric Non-negative Matrix Factorization (RANMF)}	 &\scriptsize{Soft Regularized Asymmetric Alternating Minimization (SRAM)}
&\scriptsize{$O_{iter}$=Unknown, $O_{pre}$=Unknown, $O_{conv}$=Unknown}  \\
\scriptsize{Woo and Park \cite{765}}	          &\scriptsize{Woo and Park \cite{765}}                          &\scriptsize{}        \\
\hline
\hline
\scriptsize{\textbf{Robust Matrix Completion (RMC)}}	&\scriptsize{}
&\scriptsize{}  \\
\scriptsize{RMC-$l_{\sigma}$ norm loss function}	&\scriptsize{Gradient Descent Iterative Hard Thresholding (IHT)}
&\scriptsize{Linear convergence rate}  \\
\scriptsize{Yang et al. \cite{2001}}	    &\scriptsize{Yang et al. \cite{2001}}              &\scriptsize{}        \\
\scriptsize{}	                                                        &\scriptsize{Gradient Descent Iterative Soft Thresholding (IST)}
&\scriptsize{Linear convergence rate}  \\
\scriptsize{Yang et al. \cite{2001}}	    &\scriptsize{Yang et al. \cite{2001}}              &\scriptsize{}        \\
\hline
\scriptsize{RMC-Robust Bilateral Factorization (RBF)}	&\scriptsize{Alternating direction
Method of Multipliers (ADMM))}
&\scriptsize{$O_{iter}(d^2n + d^2m + mnd)$, $O_{pre}$=unknown, $O_{conv}$=unknown}  \\
\scriptsize{Shang et al. \cite{1043}}	    &\scriptsize{Shang et al. \cite{1043}}              &\scriptsize{$d \ll n <m$}        \\
\hline
\scriptsize{RMC (Convex Formulation}	               &\scriptsize{Convex Alternating Direction Augmented Lagrangian (Convex ADAL)}
&\scriptsize{$O_{iter}(mn^2)$,  $O_{pre}$=unknown, $O_{conv}(1/T)$}  \\
\scriptsize{Shang et al. \cite{2003}}	    &\scriptsize{Shang et al. \cite{2003}}              &\scriptsize{}        \\
\hline
\scriptsize{RMC-Matrix Factorization (MF)}	 &\scriptsize{Non-Convex Alternating Direction Augmented Lagrangian (Non-convex ADAL)}
&\scriptsize{$O_{iter}(d^2m + mnd)$,  $O_{pre}$=unknown, $O_{conv}(1/T)$, $d \ll n <m$}  \\
\scriptsize{Shang et al. \cite{2003}}	    &\scriptsize{Shang et al. \cite{2003}}              &\scriptsize{}        \\
\hline
\scriptsize{Factorized Robust Matrix Completion (FRMC)}	 &\scriptsize{Spectral Projected Gradient Iterations (SPG)}
&\scriptsize{$O_{iter}$=Unknown, $O_{pre}$=Unknown, $O_{conv}$=Unknown}  \\
\scriptsize{Mansour and Vetro \cite{2000}}	    &\scriptsize{Berg and Friedlander \cite{3004}}              &\scriptsize{}        \\
\hline
\scriptsize{Motion-Assisted Matrix Completion (MAMC\protect\footnotemark[19])}	 &\scriptsize{ALM-ADM Framework}
&\scriptsize{$O_{iter}(mnmin(m,n))$, $O_{pre}$=unknown, $O_{conv}(1/\mu_T)$}  \\
\scriptsize{Yang et al. \cite{2002}}	    &\scriptsize{Lin et al. \cite{18}}                                 &\scriptsize{}        \\
\hline
\hline
\scriptsize{\textbf{Robust Subspace Recovery (RSR)}}	&\scriptsize{}
&\scriptsize{}  \\
\scriptsize{Robust Subspace Recovery via Bi-Sparsity (RoSuRe)}	&\scriptsize{Linearized ADMM}
&\scriptsize{$O_{iter}(rmn)$, $O_{pre}(1/ \epsilon)$, $O_{conv}$=Unknown}  \\
\scriptsize{Bian and Krim \cite{1014}}	    &\scriptsize{Lin et a. \cite{39}}                                &\scriptsize{}        \\
\hline
\scriptsize{Robust Orthonomal Subspace Learning (ROSL\protect\footnotemark[20])}	&\scriptsize{inexact ADM/BCD}
&\scriptsize{$O_{iter}(rmn)$, $O_{pre}$=unknown, $O_{conv}$=unknown}  \\
\scriptsize{Xu et al. \cite{759-1}}	    &\scriptsize{Xu et al. \cite{759-1}}                                 &\scriptsize{}        \\
\cline{2-3}
\scriptsize{ROSL+}	&\scriptsize{Random Sampling}
&\scriptsize{$O_{iter}(r^2(m+n))$, $O_{pre}$=unknown, $O_{conv}$=unknown}  \\
\scriptsize{Xu et al. \cite{759-1}}	    &\scriptsize{Xu et al. \cite{759-1}}                                 &\scriptsize{}        \\
\hline
\scriptsize{Robust Orthogonal Complement PCA (ROCPCA)}	&\scriptsize{$M$-estimators}
&\scriptsize{$O_{iter}$=Unknown, $O_{pre}$=Unknown, $O_{conv}$=Unknown}  \\
\scriptsize{She et al. \cite{1042}}	    &\scriptsize{She and Owen \cite{3005}}                                 &\scriptsize{}        \\
\hline
\end{tabular}}
\caption{Solvers for RNMF, RMC, and RSR: An overview of their complexity per iteration at running time $O_{iter}$, their complexity $O_{pre}$ to reach an accuracy of $\epsilon$ precision and their convergence rate $O_{conv}$ for $T$ iterations. "Unknown" stands for not indicated by the authors.}
\label{A-TPCP3Overview}
\end{table}
\footnotetext[18]{https://sites.google.com/site/nmfsolvers/}
\footnotetext[19]{http://cs.tju.edu.cn/faculty/likun/projects/bf-separation/index.htm}
\footnotetext[20]{https://sites.google.com/site/xianbiaoshu/}
\end{landscape}

\clearpage
\begin{landscape}
\begin{table}
\scalebox{0.70}{
\begin{tabular}{|l|l|l|}
\hline
\scriptsize{Methods}&\scriptsize{Solvers}&\scriptsize{Complexity} \\
\hline
\hline
\scriptsize{\textbf{Robust Subspace Tracking (RST)}}	&\scriptsize{} &\scriptsize{}  \\
\scriptsize{GRASTA\protect\footnotemark[21]}	&\scriptsize{Augmented Lagrangian algorithm}
&\scriptsize{$O_{iter}(\left|\Omega\right| d^3+d \left|\Omega\right|+nd^2)$, $O_{pre}$=unknown, $O_{conv}$=unknown}  \\
\scriptsize{}	    &\scriptsize{with the Grassmannian geodesic gradient descent}                                 &\scriptsize{}        \\
\scriptsize{He et al. \cite{27}\cite{28}}	
&\scriptsize{Boyd et al. \cite{111}, Edelman et al. \cite{112}}		                                              &\scriptsize{}        \\
\hline
\scriptsize{pROST\protect\footnotemark[22]}	&\scriptsize{Conjugate Gradient}
&\scriptsize{$O_{iter}$=unknown, $O_{pre}$=unknown, $O_{conv}$=unknown}  \\
\scriptsize{Hage and Kleinstauber \cite{69}}   &\scriptsize{Hage and Kleinstauber \cite{69}}		     &\scriptsize{}        \\
\hline
\scriptsize{GOSUS\protect\footnotemark[23]}	&\scriptsize{ADMM}
&\scriptsize{$O_{iter}$=unknown, $O_{pre}$=unknown, $O_{conv}$=unknown}  \\
\scriptsize{Xu et al. \cite{85}}   &\scriptsize{Xu et al. \cite{85}}		     &\scriptsize{}        \\
\hline
\scriptsize{FARST}	&\scriptsize{ADMM}
&\scriptsize{$O_{iter}(\left|\Omega\right|d^3+d\left|\Omega\right|+nd^2)$, $O_{pre}$=unknown, $O_{conv}$=unknown}  \\
\scriptsize{Ahn \cite{1035}}   &\scriptsize{Ahn \cite{1035}}		     &\scriptsize{}        \\
\hline
\hline
\scriptsize{\textbf{Robust Low Rank Minimization (RLRM)}}	&\scriptsize{} &\scriptsize{}  \\
\scriptsize{LRM with Contiguous Outliers Detection}		&\scriptsize{Alternating Algorithm (SOFT-IMPUTE)}				&\scriptsize{$O_{iter}$=Unknown, $O_{pre}$=Unknown, $O_{conv}$=Unknown}       \\
\scriptsize{(DECOLOR\protect\footnotemark[24]) Zhou et al. \cite{25}}	 
&\scriptsize{Mazumder et al. \cite{103}}														                                                  &\scriptsize{}  		  \\
\hline
\scriptsize{LRM with DRMF}																		
&\scriptsize{Block Coordinate Descent strategy}												        &\scriptsize{$O_{iter}(mn(r+\text{log}(p)))$}           \\
\scriptsize{(DRMF\protect\footnotemark[25])  Xiong et al. \cite{54}}	
&\scriptsize{Xiong et al. \cite{54}}																 	        &\scriptsize{$K$ partial SVD at each iteration}         \\
\hline
\scriptsize{LRM with DRMF-R}	&\scriptsize{Block Coordinate Descent strategy}	&\scriptsize{$O_{iter}(mn(r+\text{log}(p)))$}           \\
\scriptsize{(DRMF-R\protect\footnotemark[25]) Xiong et al. \cite{54}}			  	
&\scriptsize{Xiong et al. \cite{54}}																 	         &\scriptsize{$K$ partial SVD at each iteration}        \\
\hline
\scriptsize{PRMF}	&\scriptsize{Conditional EM Algorithm (CEM)}	&\scriptsize{$O_{iter}$=Unknown, $O_{pre}$=Unknown, $O_{conv}$=Unknown}                         \\
\scriptsize{(PRMF\protect\footnotemark[26]) Wang et al. \cite{5400}}			  	
&\scriptsize{Jebara and Pentland \cite{115}}																 	 &\scriptsize{}                                         \\
\hline
\scriptsize{BRMF}	&\scriptsize{Conditional EM Algorithm (CEM)}	&\scriptsize{$O_{iter}$=Unknown, $O_{pre}$=Unknown, $O_{conv}$=Unknown}                         \\
\scriptsize{(BRMF\protect\footnotemark[27]) Wang and Yeung \cite{5401}}			  	
&\scriptsize{Jebara and Pentland \cite{115}}																 	 &\scriptsize{}                                         \\
\hline
\scriptsize{PLRMF}	&\scriptsize{Inexact ALM with Gauss-Seidel iteration}	&\scriptsize{$O_{iter}=r(max(m,n))^2$, $O_{pre}$=Unknown, $O_{conv}$=Unknown}   \\
\scriptsize{(RegL1-ALM\protect\footnotemark[28]) Zheng et al. \cite{1025}}			  	
&\scriptsize{Zheng et al. \cite{1025}}																 	  &\scriptsize{}            \\
\hline
\scriptsize{LRMF-MOG}	&\scriptsize{EM algorithm}	&\scriptsize{$O_{iter}$=Unknown, $O_{pre}$=Unknown, $O_{conv}$=Unknown}                           \\
\scriptsize{Meng et al. \cite{2-1}}			  	
&\scriptsize{Meng et al. \cite{2-1}}																 	    &\scriptsize{}             \\
\hline
\scriptsize{UNN-BF}	&\scriptsize{ALM}	&\scriptsize{$O_{iter}(mnr + nr^2)$, $O_{pre}$=Unknown, $O_{conv}$=Unknown}                                   \\
\scriptsize{Cabral et al.\cite{2-2}}			 
&\scriptsize{Cabral et al.\cite{2-2}}																 	    &\scriptsize{}            \\
\hline
\scriptsize{RRF}&\scriptsize{Alternative Direction Descent Algorithm (ADDA)}&\scriptsize{$O_{iter}$=Unknown, $O_{pre}$=Unknown, $O_{conv}$=Unknown} \\
\scriptsize{Sheng et al. \cite{1027}}			  	
&\scriptsize{Sheng et al. \cite{1027}}																 	  &\scriptsize{}             \\
\hline
\end{tabular}}
\caption{Solvers for ST and LRM: An overview of their complexity per iteration at running time $O_{iter}$, their complexity $O_{pre}$ to reach an accuracy of $\epsilon$ precision and their convergence rate $O_{conv}$ for $T$ iterations. "Unknown" stands for not indicated by the authors.}
\label{A-TPCP4Overview}
\end{table}
\footnotetext[21]{{http://sites.google.com/site/hejunzz/grasta}}
\footnotetext[22]{{http://www.gol.ei.tum.de/index.php?id=37 L=1}}
\footnotetext[23]{{http://pages.cs.wisc.edu/~jiaxu/projects/gosus/}}
\footnotetext[24]{{http://bioinformatics.ust.hk/decolor/decolor.html}}
\footnotetext[25]{{http://www.autonlab.org/autonweb/downloads/software.html}}
\footnotetext[26]{{http://winsty.net/prmf.html}}
\footnotetext[27]{{http://winsty.net/brmf.html}}
\footnotetext[28]{{https://sites.google.com/site/yinqiangzheng/}}
\end{landscape}

\subsubsection{Adequacy for the background/foreground separation} 
\label{subsubsec:ABFS} 
For each problem formulation, we investigated its adequacy with the application of background/foreground separation in their corresponding section in terms of following criteria: \textbf{(1)} its robustness to noise, \textbf{(2)} its spatial and temporal constraints, \textbf{(3)} the existence of an incremental version, \textbf{(4)} the existence of a real-time implementation, and \textbf{(5)} the ability to deal with the challenges met in video sequences. Table \ref{BFSeparation} show an overview of the existing methods. The following observations can be made:

\begin{enumerate}
\item \textbf{Robustness to noise:} Noise is due to a poor quality image source such as images acquired by a web cam or images after compression. It affects the entries of the matrix $A$. In each problem formulation, assumptions are made to assure the exact recovery of the decomposition. PCP assumed that all entries of the matrix to be recovered are exactly known via the observation and that the distribution of corruption should be sparse and random enough without noise. These assumptions are rarely verified in the case in real applications because only a fraction of entries of the matrix can be observed and the observation can be corrupted by both impulsive and Gaussian noise. The robustness of PCP can be improved by taking into account entry-wise noise like in SPCP, quantization error such like QPCP and the presence of outliers in entire columns like in BPCP. Other methods address sparsity control, recovery guarantees or the entry-wise noise. \\
\item \textbf{Spatial and temporal constraints:} These constraints are very needed to the background/foreground separation task as moving objects present spatial and temporal characteristics. Thus, several approaches tried to take them into account as follows: \\
\begin{enumerate}
\item \textit{Spatial constraints} of the foreground object are addressed by \textbf{(1)} BPCP \cite{41}\cite{501}, LBPCP \cite{26}, BRPCA \cite{7}, IRLS \cite{505}\cite{506}\cite{507} in RPCA framework, \textbf{(2)} RANMF \cite{765} in RNMF, and \textbf{(3)} DECOLOR \cite{25} and MBRMF \cite{5401} in LRM framework. \\
\item \textit{Temporal constraints} are addressed by \textbf{(1)} RPCA with dense optical flow \cite{62}, RPCA with consistent optical flow \cite{1041}, RPCA with smoothness and arbitrariness constraints (RFDSA) \cite{1033}, BRPCA  \cite{7} in RPCA framework, \textbf{(2)} MAMC \cite{2002} and RMAMC \cite{2002} in RMC framework, and \textbf{(3)} DECOLOR \cite{25} and MBRMF \cite{5401} in LRM framework. \\
\item \textit{Spatial and temporal constraints} Less approaches address both the spatial and temporal constraints. For example, RPCA with smoothness and arbitrariness constraints (RFDSA) \cite{1033}, BRPCA \cite{7}, spatio-temporal IRLS \cite{507}, DECOLOR \cite{25} and MBRMF \cite{5401} were the first methods which address both the spatial and temporal constraints. \\
\end {enumerate} 
Practically, the different strategies used to take into account the spatial and/or temporal coherence can be classified as follows: \\
\begin{itemize}
\item For the \textbf{\textit{regularization based approaches}}, the main strategies to take into account the \textbf{\textit{spatial coherence}} consist of using \textbf{(1)} a mixed norm ($||.||_{2,1}$ \cite{41}\cite{501}\cite{505}\cite{506}\cite{507}\cite{1520}\cite{1522}\cite{1523}\cite{1524}\cite{1525}\cite{1527}) on the matrices $L$ and/or $S$, \textbf{(2)} using a structured sparsity norm \cite{1901}, structured group sparsity norm \cite{1542} or dynamic tree structured sparsity nom \cite{1524}\cite{1525}\cite{1527} on the matrix $S$, and \textbf{(3)} adding a term on the matrix $S$ in the minimization problem such as a Total Variation penalty \cite{506}\cite{507}\cite{765}\cite{1033}\cite{1900} or a gradient \cite{505}\cite{506}\cite{507}\cite{26}. For the \textbf{\textit{temporal coherence}}, optical flow (like in TTLO \cite{1031}, two-pass RPCA \cite{62}, block sparse RPCA \cite{1041}, MAMC \cite{2002}\cite{2002-1}, RMAMC \cite{2002}\cite{2002-1}), and saliency motion detection (like in SCM-RPCA \cite{1711} and MODSM \cite{1532}) are used in the RPCA and RMC frameworks. Thus, the motion information can be used in several ways as follows : \textbf{(1)} an adaptive $\lambda$ \cite{62}\cite{1901} which is a function of the motion consistency to ensure that all the changes caused by the foreground motion will be entirely transfered to the matrix $S$, and \textbf{(2)} a weighting matrix $W$ \cite{2002}\cite{2002-1}\cite{1711} which is constructed from the optical flow to suppress slowly-moving objects, to enforce the recovery of the background that appears at only a few frames and to eliminate the influence of light conditions, camouflages, and dynamic backgrounds. To take into account both \textbf{\textit{spatial and temporal coherence}}, strategies can be classified as follows: \textbf{(a)} combination the methods of (1) and (2) like spatiotemporal IRLS \cite{507}, SCM-RPCA \cite{1711} and GSRPCA-LSD \cite{1901}, \textbf{(b)} spatio-temporal coherency clues based approaches like in SMD-RPCA \cite{1537}, \textbf{c)} graph-based approaches which incorporate spectral graph regularization like in RPCAG \cite{1072-1}, FRPCAG \cite{1072-2}, and MAGRPCA \cite{1010-7}, spatiotemporal graph regularization like in SLMC \cite{1010-10} and SRPCA \cite{1010-10} encoding data and feature similarity on low-rank model, or weighted cluster graph regularization like in the piece-wise low-rank model \cite{1505}, and \textbf{d)} depth-based approaches such as DG-PCA \cite{1542} and depth-extended ORPCA (DEOR-PCA) \cite{1010-4}. Thus, the minimization problem expressed in Equation \ref{EquationMinimization} can be extended for background/foreground separation in the following general formulation:
\begin{equation}
\begin{array}{ll}
\underset{L,S,E}{\text{min}} ~~  \underbrace{\lambda_1 f_{low}(T(L)) + \lambda_2 f_{sparse}(\Pi(S)) + \lambda_3  f_{noise}(E)}_{Decomposition} \\
+ \underbrace{\lambda_4 f_{back}(L) + \lambda_5 f_{fore}(S)}_{Application} \\
 ~~~ \text{subj} ~~~ C_3
\end{array}
\label{EquationMinimization22}
\end{equation}
where $\lambda_4$ and $\lambda_5$ are regularization parameters. $f_{back}(L)$ and $f_{fore}(S)$ are regularization functions that allow the minimization to take into account the characteristics of the background and the foreground, respectively. $f_{back}(L)$ can be a mixed norm. $f_{fore}(S)$ can be the gradient, the Total Variation or a static or dynamic tree structured sparsity norm on $S$. The function $T()$ allows to take into account camera jitter like in incPCP-TI \cite{26-6}\cite{26-7}. The function $\Pi()$ allows to add a confidence map on $S$ like in TTLO \cite{1031} and SCM-RPCA \cite{1711}. Thus, the confidence map reinforces the pixels belonging from the moving objects. $C_3$ contains the constraints which can be as follows: \\
\begin{enumerate}
\item only on the recovery such as $A= L + S + E$. \\
\item both on the recovery and the spatial/temporal aspects such as $W \circ A = W\circ (L + S + E)$. $W$ is a weighting matrix based on optical flow in MAMC \cite{2002}\cite{2002-1}, and  based on salient motion detection in SCM-RPCA \cite{1711}. $W$ imposed a shape constraint or region constraint. \\
\item on the recovery and the transformation aspects such as $A \circ \tau=L + S + E$. $\tau$ is a transformation function based on motion vectors like in RASL \cite{2810}, ARPCA \cite{1521}, ARPCA-BS \cite{1520}\cite{1522}\cite{1523}, ARPCA-CSSP \cite{1524}\cite{1525}, DSPSS \cite{1527} and t-GRASTA \cite{2801}\cite{2802}. Practically, $\tau$ models potential global motion that the foreground region undergoes \cite{1521}. Thus, $\tau$ is a set of independent transformations (one per frame), each having a parametric representation, such that $A \circ \tau$ aligns all the observed video frame \cite{26-6}. The main limitation of the
algorithm  to compute the motion model parameter $\tau$ in RASL is its computation cost. To address this problem, Ebadi and Izquierdo \cite{1521} proposed a computationally-cheaper algorithm. \\
\end{enumerate}
Note that \textit{the first part} of Equation \ref{EquationMinimization22} with  $f_{low}(L)$, $f_{sparse}(S)$ and $f_{noise}(E)$ concerns mainly the decomposition into low-rank plus additive matrices, and \textit{the second part} with  $f_{back}(L)$ and $f_{fore}(S)$ concerns mainly the application to background/foreground separation. Thus, the minimization problem can be formulated in a general form as follows:
\begin{equation}
\begin{array}{ll}
\underset{L,S,E}{\text{min}} ~~  \underbrace{ \lambda_1 ||T(L)||_{norm_1}^{p_1} + \lambda_2 ||\Pi(S)||_{norm_2}^{p_2} + \lambda_3 ||E||_{norm_3}^{p_3} }_{Decomposition} \\
+ \underbrace{ \lambda_4 ||L||_{l_{2,1}} + \delta_1 ||grad(S)||_{l_{1}} + \delta_2 TV(S) + \delta_3 \Omega(S)}_{Application} \\
 ~~~ \text{subj} ~~~ W \circ A= W \circ (L + S + E)~~~  or ~~~ \text{subj} ~~~ A \circ \tau=L + S + E
\end{array}
\label{EquationMinimization3}
\end{equation}
where $\delta_1$, $\delta_2$ and $\delta_3$ are regularization parameters. $norm_2$ is usually taken to force spatial homogeneous fitting in the matrix $S$, that is for example the norm $l_{2,1}$ with $p_2=1$ \cite{41}\cite{501}\cite{505}\cite{506}\cite{507}\cite{1520}\cite{1522}\cite{1523}\cite{1524}. The other terms in the second part which concerns mainly the application to background/foreground separation can be described as follows: \\
\begin{enumerate}
\item $||grad(S)||_1$, $TV(S)$ and $\Omega(S)$ are a gradient \cite{505}\cite{506}\cite{507}\cite{26}, a total variation \cite{506}\cite{507}\cite{765}\cite{1033}\cite{1900} and a static or dynamic tree structured sparsity norm \cite{1901}\cite{1542} \cite{1524}\cite{1525}\cite{1527} applied on the matrix $S$, respectively. \\
\item $L$ can be processed with a set of invertible and independent transformations $T()$ like in incPCP-TI \cite{26-6}\cite{26-7} in presence of translational and rotational camera jitter.  \\
\item $S$ can be processed with a linear operator $\Pi()$ that weights its entries according to their confidence of corresponding to a moving object such that the most probable elements are unchanged and the least are set to zero. $\Pi()$ is computed with optical flow in TTLO \cite{1031} and with salient motion detection in SCM-RPCA \cite{1711}. \\
\item The term  $\lambda_4 ||L||_{l_{2,1}}$ ensures the recovered $L$ has exact zero columns corresponding to the outliers.  \\
\item A weighting matrix $W$ \cite{2002}\cite{2002-1}\cite{1711} or a transformation $\tau$ \cite{2810}\cite{1521} \cite{1520}\cite{1522}\cite{1523} \cite{1524}\cite{1525}\cite{1527}\cite{2801}\cite{2802} can be used in the constraints $C_3$: \textbf{(1)} to enforce the recovery of the background that appears at only a few frames and to eliminate the influence of light conditions, camouflages, and dynamic backgrounds, and \textbf{(2)} to model potential global motion that the foreground region undergoes, respectively. \\
\end{enumerate}
\item For \textbf{\textit{the statistical inference based approaches}} Markov Random Fields (MRF) are used to extract temporally and spatially localized moving objects like in BRPCA \cite{7}, DECOLOR \cite{25} and MBRMF \cite{5401}. Statistical total variations can also be used like in the approach based on smoothness and arbitrariness constraints (RFDSA) \cite{1033}.   \\
\end{itemize}
\item \textbf{Incremental algorithms:} Incremental algorithms are needed to update the low-rank and additive matrices when a new data arrives. Several incremental algorithms can be found in the literature as follows: \textbf{(1)} in the RPCA framework (PCP \cite{14}\cite{15}  \cite{16}\cite{17}\cite{17-3}\cite{1007}\cite{26-2}\cite{1038}, SPCP \cite{68}, RPCA-SpaCtrl \cite{9}\cite{10}, Approximated RPCA \cite{1060}), \textbf{(2)} in the subspace tracking framework (GRASTA \cite{27}\cite{28}, t-GRASTA \cite{2801}\cite{2802}, pROST \cite{69}\cite{6901}, GOSUS \cite{85} and FARST \cite{1035}), and \textbf{(3)} in the RNMF framework (COROLA \cite{1077}, ORLRMR \cite{1120}, LSVD-LRR \cite{1330}, ORLRMR \cite{1536}). Thus, the decomposition can be written as follows:
\begin{equation}
A_t=L_t+S_t+ E_t
\label{EquationDecomposition3}
\end{equation}
where $t$ is the indice for the time. $L_t$, $S_t$, $E_t$ are determined from $L_{t-1}$, $S_{t-1}$, $E_{t-1}$ and the current observation. \\
\item \textbf{Real-time implementations:} As background/foreground separation needs to be achieved in real-time, several strategies have been developed and are generaly based on submatrices computation \cite{42} or GPU implementations \cite{12}\cite{13}. Real-time implementations can be found for PCP \cite{12}\cite{13}\cite{42}\cite{1015} and for SPCP \cite{53} \\
\item \textbf{Strategies:} Differents strategies can be used to applied DLAM for background/foreground separation. For example, Gao et al.\cite{62}
developed a two-pass RPCA process for consistent foreground detection. For objects or people which remain immobile for a certain period of time, Tepper et al. \cite{1135} proposed a method which detects foreground objects at different timescales, by exploiting the theoretical and practical properties of RPCA.\\
\item \textbf{Dealing with the challenges met in video sequences:} Several challenges appear in video such as dynamic backgrounds and illumination changes as developed in Section \ref{sec:Challenges} and in Bouwmans \cite{303}. The challenge which is the most addressed (apart dynamic backgrounds and illumination changes) in literature is when the camera is slowly moving like in camera jitter. Thus, the existing approaches can be classified following the challenges that they addressed: \\
\begin{itemize}
\item \textbf{Noisy images:} Javed et al. \cite{1010-5} proposed an input video denoising to cope with noisy videos in presence of rainy or snowy conditions. A real time Active Random Field (ARF) constraints is exploited using probabilistic spatial neighborhood system for image denoising. After that, Online Robust PCA (OR-PCA) is used to separate the low-rank and sparse component from denoised frames. In addition, a color transfer functionis employed between the source and input image for handling global illumination conditions which is a very useful technique for surveillance
agents to handle the night time videos. Experimental results on i-LIDS and Change Detection (CDnet) 2014 datasets show that OR-PCA with ARF outperforms by integrating the original MOG with ARF, PBAS with ARF, and Codebook model with ARF. In an other work, Chen et al. \cite{24-3} developed a variational Bayesian Sparse Estimator (VBSE) which achieved background/foreground separation in blurred and noisy video sequences. Furthermore, VBSE is free of input parameters and is hence suitable to automated deployment. \\
\item \textbf{Bootstrapping:} Javed et al. \cite{1010-7} developed a Motion-Aware Graphs Regularized RPCA, named MAGRPCA which is robust in clutter
scenes, where background is always occluded by heavy foreground objects. MAGRPCA outperforms both of RMAMC \cite{2002}\cite{2002-1} and GoDec \cite{6}. \\
\item \textbf{Camera motion:} Several strategies are used in literature to deal with camera motion: \textbf{(1)} transformations based methods in which a transformation $\tau()$ is applied on the data matrix $A$ (like in RASL \cite{2810}, ARPCA \cite{1521}, ARPCA-BS \cite{1520}\cite{1522}\cite{1523}, ARPCA-CSSP \cite{1524}, DSPSS \cite{1525}\cite{1527}, OR-SGD \cite{1534}, t-GRASTA \cite{2801}\cite{2802}) or on the low-rank matrix $S$ (like incPCP-TI \cite{26-6}\cite{26-7}), \textbf{(2)} compensation based methods in which the motion due the camera is compensated in pre-processing step like in DG-PCA \cite{1542}, FRMC \cite{2000}, FRMC-MVP \cite{2000-1} and MAGRPCA \cite{1010-7}, and \textbf{(3)} endogenous convolutional based methods in which convolutional sparse representations to model the effects of non-linear transformations such as translation and rotation, thereby simplifying or eliminating the alignment pre-processing task like in ECSR \cite{26-75}. First, Peng et al. \cite{2810} proposed a robust alignment RPCA model for linearly correlated images. In an other way, Rodriguez and Wohlberg \cite{26-6} developed a translational and rotational incremental principal component pursuit when camera jitter appears. A real-time implementation of this method was proposed by Silva and Rodriguez \cite{26-7}. Ebadi and Izquierdo \cite{1521} proposed an approximated RPCA (ARPCA) for decomposing in batch way unaligned and corrupted images as the sum of a low-rank and a sparse corruption matrix, while simultaneously aligning the images according to the optimal image transformations. This decomposition is called $\tau$-decomposition and includes parameters modeling global motion of background regions and also entails a Gaussian additive noise part to be robust to camera movement and dynamic backgrounds. In further work, Ebadi et al. \cite{1520}\cite{1522}\cite{1523} improved ARPCA by imposing block-sparsity on the pixels of each video frame with the $l_{2,1}$-norm rather than a whole column in the matrix $S$ as made in RPCA-LBD \cite{41}\cite{501}. Therefore, this algorithm called ARPCA-BS gives more  robustness  than  RPCA-PCP \cite{3} and RPCA-LBD \cite{41}\cite{501} in presence of varying foreground object sizes, illumination changes and dynamicbackgrounds. Furthermore, Ebadi et al. \cite{1520}\cite{1522}\cite{1523} used a SVD-free algorithm for the case of rank-$1$ background. Thus, ARPCA-BS outperforms RPCA \cite{3}, RPCA-LBD \cite{41}\cite{501} and GoDec \cite{6} in computation cost/time as well as performance. In an other work, Ebadi et al. \cite{1524} proposed a dynamic tree-structured sparse matrix, and solved ARPCA extended to handle camera motion. The dynamicity of group structures is controlled via a patch-based group selection algorithm that preserves the natural shape of objects in the scene. The size and structure of these patches are dynamically refined in an iterative process. Moreover, to reduce the problem of dimensionality and scale, a low-rank background modeling solved as Column Subset Selection Problem (CSSP) reduces the order of complexity, decreases computation time, and eliminates the huge storage need for large videos. Experimental results \cite{1524} show that ARPCA-CSSP outperforms SemiSoftGoDec \cite{6}, GSRPCA-LSD \cite{1901} and SPGFL \cite{1712}. In a further work, Ebadi and Izquierdo \cite{1525}\cite{1527} improved ARPCA-CSS by using a superpixel approach to impose spatial coherence on the regions, and to obtain crisp and meaningful foreground regions. This algorithm called Dynamic SuperPixel Structured-Sparse (DSPSS) gives better scores on the ChangeDetection.net dataset than SemiSoftGoDec \cite{6}, GSRPCA-LSD \cite{1901} and SPGFL \cite{1712}. In an other way, Song et al. \cite{1534} proposed an image alignment method for an online RPCA solved via a stochastic gradient descent algorithm called ORPCA-SGD. Instead of computing the warp update using noisy input samples like RASL, ORPCA-SGD directly linearizes the object function by performing warp update on the recovered samples. In an other way, Han et al. \cite{1541}\cite{1911} improved the OR-PCA algorithm to be robust against camera jitter. An other work proposed by Tian et al. \cite{1542} used a depth-enhanced homography model for global motion compensation before the a Depth-weighted group-wise PCA (DG-PCA) method is executed. For RMC, Mansour et al. \cite{2000} developed factorized robust matrix completion (FRMC) algorithm, and used the motion vectors extracted from the coded video bitstream to compensate for the change in the camera perspective. In a further work, Kao et al. \cite{2000-1} proposed to improve FRMC-MVP with a label propagation scheme based on motion vanishing point (MVP) analysis to address the case of moving cameras. This method is called FRMC-MVP. For background initialization, MAGRPCA developed by Javed et al. \cite{1010-7}  is robust in presence of camera jitter by learning the locality and similarity information within a video. Thus, inter-frame and intra-frame graphs are constructed to preserve the notion of geometric information in low-rank component. For RST, He et al. \cite{2801}\cite{2802} developed an iterative Grassmannian optimization called t-GRASTA which is robust to camera jitter. t-GRASTA is an extension of GRASTA combined with RASL. \\
\item \textbf{Illumination Changes:} To be robust to illumination changes, Javed et al. \cite{1010-7} incorporated spectral graph regularization in the RPCA framework while Newson et al. \cite{1505} used a weighted cluster graph. \\
\item \textbf{Dynamic Backgrounds:} Javed et al. \cite{1010-2}\cite{1010-6} used Markov Random Field (MRF) in OR-PCA to deal with dynamic backgrounds. In RPCA based on Salient Motion Detection (SMD-RPCA), Chen et al. \cite{1537} defined a saliency clue over the sparse matrix $S$ to filter out the dynamic backgrounds globally. The idea is based from the following observations: \textbf{(1)} the sparsity degree of stable background region varies less frequently around their mean value than the dynamic background regions, \textbf{(2)} the sparsity degree of dynamic background regions become relatively weak when moving objects are passing through, and \textbf{(3)} the sparsity degree of moving objects can be either frequently changing or not, but both its amplitude and its duration are larger than those of dynamic background regions. Thus, a short-term thresholding separates the stable region from the dynamic backgrounds by performing statistics on the variation of sparse residual. In SRPCA, Javed et al. \cite{1010-10} employed spatio-temporal graph regularization. Experimental results on the CD.net 2014 dataset show that SRPCA outperforms GoDec \cite{6}, GRASTA \cite{27}, DECOLOR \cite{25}, and RMAMC \cite{2002}. \\
\item \textbf{Intermittent Motion of Foreground Objects:} In MAGRPCA, Javed et al. \cite{1010-7} used an optical flow algorithm between consecutive frames to generate the binary mask of motion. This motion mask allows to remove the motionless video frames and create a matrix comprising only dynamic video clips. Thus, MAGRPCA incorporates the motion message and encodes the manifold constraints. MAGRPCA is more effective than to RMAMC \cite{2002}\cite{2002-1} because motionless frames are removed in order to handle large outliers in the background model. In SMD-RPCA, Chen et al. \cite{1537} leveraged the previously detected salient motion to guide the update of the current low-rank prior. First, the background maintenance is suspended for those regions where the detected moving objects come to standstill by making the static object keep a high saliency value. Second, for the newly-exposed background areas that are previously covered by the "current static object", their updating are boosted. The idea is that the newly-exposed backgrounds present strong similarity with respect to its non-salient surroundings in RGB feature space, while the currently-stopped object should keep high contrasts. Thus, the updating strength of the low-rank information respect a saliency metric which allow to obtain a saliency clue mask to guide the updating of the low-rank prior. In an other work, Newson et al. \cite{1505} used a weighted cluster graph. \\
\item \textbf{Ghost Suppression:} For the ghost problem, Rodriguez and Wohlberg \cite{26-71} proposed an algorithm called gs-incPCP which can suppress the ghost by using two simultaneous background estimates based on observations over the previous $N_1$ and $N_2$ frames with $N_1 \ll N_2$ in order to identify and diminish the ghosting effect. In DSPSS, Ebadi et al. \cite{1525}\cite{1527} proposed a tandem algorithm which involves an initialization step before the optimization takes place. It is different from algorithms that require a two-pass optimization \cite{62}, where the optimization is twice performed to refine results. Introducing a prior knowledge of the spatial distribution of the outliers to the model, Ebadi et al. \cite{1525}\cite{1527} obtained a faster convergence. \\
\end{itemize} 
\end{enumerate} 
All these key issues need to be addressed in the different problem formulations based on the decomposition into low-rank plus sparse matrices to be suitably applied to background modeling and foreground detection in video taken by a static camera. The algorithms which address the largest number of challenges/requirements for the background initialization are MAGRPCA \cite{1010-7}, SLMC \cite{1010-10} and FRMC \cite{2000}. For background/foreground separation, it is SMD-RPCA \cite{1537}, SRPCA \cite{1010-10} and incPCP-TI \cite{26-7}. 

\begin{landscape}
\begin{table*}
\scalebox{0.50}{
\begin{tabular}{|l|l|l|l|l|l|} 
\hline
\scriptsize{Background/Foreground Separation} &\scriptsize{RPCA} &\scriptsize{RNMF} &\scriptsize{RST} &\scriptsize{RMC} &\scriptsize{RLRM} \\
\hline
\hline
\scriptsize{Spatial Coherence} &\scriptsize{RPCA-LBD (Tang and Nehorai \cite{41}) (Guyon et al. \cite{501})}  &\scriptsize{RANMF (Woo and Park\cite{765})}  &\scriptsize{-} &\scriptsize{-} &\scriptsize{LSVD-LRR (Dou et al. \cite{1330})}  \\
\scriptsize{} &\scriptsize{LPCP (Wohlberg et al. \cite{26})}             		          &\scriptsize{}  &\scriptsize{} &\scriptsize{} &\scriptsize{} \\
\scriptsize{} &\scriptsize{IRLS (Guyon et al. \cite{505}\cite{506}\cite{507})}        &\scriptsize{}  &\scriptsize{} &\scriptsize{} &\scriptsize{} \\
\scriptsize{} &\scriptsize{ARPCA-BS (Ebadi et al.\cite{1520}\cite{1522}\cite{1523})}  &\scriptsize{}  &\scriptsize{} &\scriptsize{} &\scriptsize{} \\
\scriptsize{} &\scriptsize{ARPCA-CSSP (Ebadi et al.\cite{1524})} &\scriptsize{} &\scriptsize{} &\scriptsize{} &\scriptsize{} \\
\scriptsize{} &\scriptsize{DSPSS (Ebadi et al.\cite{1525}\cite{1527})} &\scriptsize{} &\scriptsize{} &\scriptsize{} &\scriptsize{} \\
\scriptsize{} &\scriptsize{SPGFL (Javed et al.\cite{1712})}   &\scriptsize{} &\scriptsize{} &\scriptsize{} &\scriptsize{} \\
\hline
\scriptsize{Temporal Coherence} &\scriptsize{Dense optical flow (Gao et al. \cite{62})}         &\scriptsize{-}  &\scriptsize{}-  &\scriptsize{MAMC (Yang et al. \cite{2002})} &\scriptsize{-} \\
\scriptsize{}                   &\scriptsize{Consistent optical flow (Huang et al. \cite{1041})} &\scriptsize{}  &\scriptsize{} &\scriptsize{RMAMC (Yang et al. \cite{2002})} &\scriptsize{} \\
\scriptsize{}                   &\scriptsize{TTLO (Oreifej et al. \cite{1031})} &\scriptsize{}  &\scriptsize{} &\scriptsize{RMAMC (Yang et al. \cite{2002})} &\scriptsize{} \\
\hline
\scriptsize{Spatio-Temporal Coherence} &\scriptsize{Smoothness and arbitrariness constraints (RFDSA) (Guo et al. \cite{1033})}  &\scriptsize{-} &\scriptsize{-} &\scriptsize{SLMC (Javed et al. \cite{1010-10})}  &\scriptsize{DECOLOR (Zhou et al. \cite{25})} \\
\scriptsize{}   &\scriptsize{Total Variation (TV) Regularizer (Gao et al. \cite{1900})}  &\scriptsize{}  &\scriptsize{} &\scriptsize{}  &\scriptsize{} \\
\scriptsize{}   &\scriptsize{Piece-wise Low-rank Model (Newson et al. \cite{1505})}     &\scriptsize{}  &\scriptsize{} &\scriptsize{}  &\scriptsize{} \\
\scriptsize{}   &\scriptsize{RPCAG (Shahid et al. \cite{1072-1})}                       &\scriptsize{}  &\scriptsize{} &\scriptsize{}  &\scriptsize{} \\
\scriptsize{}   &\scriptsize{FRPCAG (Shahid et al. \cite{1072-2})}                      &\scriptsize{}  &\scriptsize{} &\scriptsize{}  &\scriptsize{} \\
\scriptsize{}   &\scriptsize{SCM-RPCA (Sobral et al. \cite{1711})}                  &\scriptsize{}  &\scriptsize{} &\scriptsize{}  &\scriptsize{MBRMF (Wang et al. \cite{5401})} \\
\scriptsize{}   &\scriptsize{GSRPCA-LSD (Liu et al.\cite{1901})}   &\scriptsize{} &\scriptsize{} &\scriptsize{} &\scriptsize{} \\
\scriptsize{}   &\scriptsize{MODSM (Pang et al. \cite{1532})}                      &\scriptsize{}  &\scriptsize{} &\scriptsize{}  &\scriptsize{} \\
\scriptsize{}   &\scriptsize{SMD-RPCA (Chen et al. \cite{1537})}   &\scriptsize{}  &\scriptsize{} &\scriptsize{} &\scriptsize{} \\   
\scriptsize{}   &\scriptsize{DG-PCA (Tian et al. \cite{1542})}                      &\scriptsize{}  &\scriptsize{} &\scriptsize{}  &\scriptsize{} \\
\scriptsize{}   &\scriptsize{MAGRPCA (Javed et al. \cite{1010-7})}                  &\scriptsize{}  &\scriptsize{} &\scriptsize{}  &\scriptsize{} \\
\scriptsize{}   &\scriptsize{BRPCA (Ding et al.\cite{7})}                           &\scriptsize{} &\scriptsize{} &\scriptsize{}   &\scriptsize{} \\
\scriptsize{}   &\scriptsize{Spatio-temporal IRLS (Guyon et al. \cite{507})}        &\scriptsize{}  &\scriptsize{} &\scriptsize{}  &\scriptsize{} \\
\scriptsize{}   &\scriptsize{DEOR-PCA (Javed et al. \cite{1010-4})}                 &\scriptsize{}  &\scriptsize{} &\scriptsize{}  &\scriptsize{} \\
\scriptsize{}   &\scriptsize{SRPCA (Javed et al. \cite{1010-10})}                   &\scriptsize{}  &\scriptsize{} &\scriptsize{}  &\scriptsize{} \\
\hline
\scriptsize{Incremental Algorithms} &\scriptsize{ReProCS (Qiu and Vaswani \cite{16})}         &\scriptsize{-}           &\scriptsize{GRASTA (He et al. \cite{28})} &\scriptsize{-} &\scriptsize{COROLA (Shakeri et al. \cite{1077})} \\
\scriptsize{}  &\scriptsize{Support-Predicted Modified-CS RR-PCP (Qiu and Vaswani \cite{15})} &\scriptsize{}           &\scriptsize{t-GRASTA (He et al. \cite{2801}\cite{2802})} &\scriptsize{} &\scriptsize{ORLRMR (Guo \cite{1120})} \\
\scriptsize{}  &\scriptsize{Support-Predicted Modified-CS (Qiu and Vaswani \cite{16})}        &\scriptsize{}           &\scriptsize{GAS21 (He and Zhang \cite{2803})} &\scriptsize{} &\scriptsize{LSVD-LRR (Dou et al. \cite{1330})} \\
\scriptsize{}  &\scriptsize{Automated ReProCS (Qiu and Vaswani \cite{17})}        &\scriptsize{}           &\scriptsize{pROST (Hage et al. \cite{69})} &\scriptsize{} &\scriptsize{ORLRMR (Guo \cite{1536})} \\
\scriptsize{}  &\scriptsize{Prac-ReProCS (Guo et al. \cite{17-3})}                    &\scriptsize{}       &\scriptsize{GOSUS (Xu et al. \cite{85})} &\scriptsize{} &\scriptsize{} \\
\scriptsize{}                       &\scriptsize{iLR (Wei et al. \cite{1007})}        &\scriptsize{}        &\scriptsize{FARST (Ahn \cite{1035})} &\scriptsize{} &\scriptsize{} \\
\scriptsize{}                       &\scriptsize{incPCP (Rodriguez and Wohlberg \cite{26-2})}    &\scriptsize{}    &\scriptsize{ROSETA (Mansour et al. \cite{1075}} &\scriptsize{} &\scriptsize{} \\
\scriptsize{}                       &\scriptsize{incPCP-TI (Rodriguez and Wohlberg \cite{26-6}}  &\scriptsize{}    &\scriptsize{} &\scriptsize{} &\scriptsize{} \\
\scriptsize{}                       &\scriptsize{ORPCA (Xu \cite{1038})}                    &\scriptsize{}  &\scriptsize{} &\scriptsize{} &\scriptsize{} \\
\scriptsize{}                       &\scriptsize{ORPCA-SGD (Song et al.\cite{1534})}        &\scriptsize{}  &\scriptsize{} &\scriptsize{} &\scriptsize{} \\
\scriptsize{}                       &\scriptsize{Projection RPCA (Lee and Lee \cite{1535})} &\scriptsize{}  &\scriptsize{} &\scriptsize{} &\scriptsize{} \\
\scriptsize{}                       &\scriptsize{OTNNR (Hong et al. \cite{1540})}           &\scriptsize{}  &\scriptsize{} &\scriptsize{} &\scriptsize{} \\
\hline\scriptsize{Real-time Algorithms} &\scriptsize{CAQR (Anderson et al. \cite{12})}      &\scriptsize{-} &\scriptsize{Real Time pROST (Hage et al. \cite{6901})} &\scriptsize{-} &\scriptsize{-} \\
\scriptsize{} &\scriptsize{Real time PCP (Pope et al. \cite{42}) }                                 &\scriptsize{} &\scriptsize{} &\scriptsize{} &\scriptsize{} \\
\scriptsize{} &\scriptsize{LR Submatrix Recovery/Reconstruction (LRSRR) (Guo et al. \cite{1015})}  &\scriptsize{} &\scriptsize{} &\scriptsize{} &\scriptsize{} \\
\scriptsize{} &\scriptsize{Real time inPCP (Rodriguez \cite{26-4}) }     &\scriptsize{} &\scriptsize{} &\scriptsize{} &\scriptsize{} \\
\scriptsize{} &\scriptsize{Real time incPCP-TI (Silva and Rodriguez \cite{26-7})} &\scriptsize{} &\scriptsize{} &\scriptsize{} &\scriptsize{} \\
\hline
\scriptsize{Dealing with the challenges} &\scriptsize{\textbf{Noisy videos:}} &\scriptsize{\textbf{Noisy videos:}} &\scriptsize{-} &\scriptsize{\textbf{Noisy videos:}} &\scriptsize{-} \\
\scriptsize{}                            &\scriptsize{OR-PCA with ARF (Javed et al.\cite{1010-5})}  &\scriptsize{RobustXray (Kumar et al. \cite{762})} &\scriptsize{} &\scriptsize{RMAMC (Yang et al. \cite{2002})} &\scriptsize{} \\
\scriptsize{}                            &\scriptsize{VBSE (Chen et al. \cite{24-3})}  &\scriptsize{} &\scriptsize{} &\scriptsize{} &\scriptsize{} \\
\scriptsize{}  &\scriptsize{\textbf{Bootstrapping:} MAGRPCA (Javed et al. \cite{1010-7})} &\scriptsize{} &\scriptsize{} &\scriptsize{} 
&\scriptsize{} \\
\scriptsize{}  &\scriptsize{\textbf{Camera jitter:}} &\scriptsize{} &\scriptsize{\textbf{Camera jitter:}} &\scriptsize{\textbf{Camera jitter:}} &\scriptsize{} \\
\scriptsize{}                            &\scriptsize{RASL (Peng et al. \cite{2810})}      &\scriptsize{} &\scriptsize{t-GRASTA (He et al. \cite{2801}\cite{2802})} &\scriptsize{FRMC (Mansour et al. \cite{2000})} &\scriptsize{} \\
\scriptsize{}                            &\scriptsize{incPCP-TI (Rodriguez and Wohlberg \cite{26-6})}  &\scriptsize{} &\scriptsize{} 
&\scriptsize{FRMC-MVP (Kao et al. \cite{2000-1})} &\scriptsize{} \\
\scriptsize{}                            &\scriptsize{ARPCA (Ebadi et al. \cite{1521})}     &\scriptsize{} &\scriptsize{} &\scriptsize{} &\scriptsize{} \\
\scriptsize{}                            &\scriptsize{ARPCA-BS (Ebadi et al.\cite{1520}\cite{1522}\cite{1523})} &\scriptsize{} &\scriptsize{} &\scriptsize{} &\scriptsize{} \\
\scriptsize{}                            &\scriptsize{ARPCA-CSSP (Ebadi et al.\cite{1524})}       &\scriptsize{} &\scriptsize{} &\scriptsize{} &\scriptsize{} \\
\scriptsize{}                            &\scriptsize{DSPSS (Ebadi et al.\cite{1525}\cite{1527})} &\scriptsize{} &\scriptsize{} &\scriptsize{} &\scriptsize{} \\
\scriptsize{}                            &\scriptsize{ORPCA-SGD (Song et al. \cite{1534})}        &\scriptsize{} &\scriptsize{} &\scriptsize{} &\scriptsize{} \\
\scriptsize{}                            &\scriptsize{modified OR-PCA (Han et al. \cite{1541}\cite{1911})} &\scriptsize{} &\scriptsize{} &\scriptsize{} &\scriptsize{} \\
\scriptsize{}                            &\scriptsize{DG-PCA (Tian et al. \cite{1542})}  &\scriptsize{} &\scriptsize{} &\scriptsize{} &\scriptsize{} \\
\scriptsize{}                            &\scriptsize{MAGRPCA (Javed et al. \cite{1010-7})} &\scriptsize{} &\scriptsize{} &\scriptsize{} &\scriptsize{} \\
\scriptsize{}                            &\scriptsize{ECSR (Wohlberg  \cite{26-75})}  &\scriptsize{} &\scriptsize{} &\scriptsize{} 
&\scriptsize{} \\
\scriptsize{}                            &\scriptsize{\textbf{Illumination Changes:} MAGRPCA (Javed et al. \cite{1010-7})} &\scriptsize{} &\scriptsize{} &\scriptsize{} &\scriptsize{} \\
\scriptsize{}                            &\scriptsize{Piece-wise Low-rank Model (Newson et al. \cite{1505})} &\scriptsize{} &\scriptsize{} &\scriptsize{} &\scriptsize{} \\
\scriptsize{}                            &\scriptsize{\textbf{Dynamic Backgrounds:} OR-PCA with MRF (Javed et al. \cite{1010-2}\cite{1010-6})} &\scriptsize{} &\scriptsize{} &\scriptsize{} &\scriptsize{} \\
\scriptsize{}                            &\scriptsize{SMD-RPCA (Chen et al. \cite{1537})} &\scriptsize{} &\scriptsize{} &\scriptsize{} &\scriptsize{} \\
\scriptsize{}                            &\scriptsize{SRPCA (Javed et al. \cite{1010-10})} &\scriptsize{} &\scriptsize{} &\scriptsize{} &\scriptsize{} \\
\scriptsize{}                            &\scriptsize{\textbf{Intermittent Motion of Foreground Objects:} } &\scriptsize{} &\scriptsize{} &\scriptsize{} &\scriptsize{} \\
\scriptsize{}                            &\scriptsize{MAGRPCA (Javed et al. \cite{1010-7})} &\scriptsize{} &\scriptsize{} &\scriptsize{} &\scriptsize{} \\
\scriptsize{}                            &\scriptsize{SMD-RPCA (Chen et al. \cite{1537})} &\scriptsize{} &\scriptsize{} &\scriptsize{} &\scriptsize{} \\
\scriptsize{}                            &\scriptsize{Piece-wise Low-rank Model (Newson et al. \cite{1505})} &\scriptsize{} &\scriptsize{} &\scriptsize{} &\scriptsize{} \\
\scriptsize{}                            &\scriptsize{\textbf{Ghost Suppression:} gs-incPCP (Rodriguez and Wohlberg \cite{26-71})} &\scriptsize{} &\scriptsize{} &\scriptsize{} &\scriptsize{} \\
\scriptsize{}                            &\scriptsize{DSPSS (Ebadi et al.  \cite{1525}\cite{1527})} &\scriptsize{} &\scriptsize{} &\scriptsize{} &\scriptsize{} \\
\hline
\end{tabular}}
\caption{Existing Methods which address the requirements of background/foreground separation. "-" corresponds to requirements which need to be investigated for the corresponding problem formulation. MAGRPCA \cite{1010-7}, and incPCP-TI \cite{26-7} are the algorithms which address the largest number of challenges/requirements of background initialization and background/foreground separation, respectively.} \centering
\label{BFSeparation}
\end{table*}
\end{landscape}

\subsubsection{Sparse decompositions}
\label{subsubsec:OD} 
Sparse decompositions are similar to low-rank decompositions except that the first matrix is considered to be sparse instead of low-rank.  Sparse decompositions are achieved in the different following problem formulations:
\begin{itemize}
\item \textbf{Sparse Dictionary learning:} Sparse dictionary learning (DL) builds data representation by decomposing each datum into a linear combination of a few components selected from a dictionary of basic elements, called atoms \cite{LD-2}. Sparse dictionary learning is also called sparse coding in the literature \cite{DL-27-1}\cite{DL-27-2}. Thus, the observation matrix is decomposed as follows:
\begin{equation}
A= X + N = D \alpha + N
\label{EquationDL1}
\end{equation}
where $A$ is the matrix which contains the observations, $X$ is a sparse noiseless matrix and $N$ is the noise matrix. $X$ is the product between $D$ which is a dictionary, and $\alpha$ which is a sparse vector. Thus, the assumption of the sparse decomposition is that the observed image is an approximated linear combination over a dictionary $D$ and a vector coefficients $\alpha$. In order to recover the noiseless image (background), the decomposition problem seeks to the following minimization problems:
\begin{equation}
\underset{D,\alpha}{\text{min}} ~~ ||A-X||_{l_2}^{2} +  ||D \alpha-X||_{l_2}^{2} + ||\alpha||_0
\label{EquationDL2}
\end{equation}
The first term minimizes the error between the recovered image and the observed version. The second term ensures that the denoised image is an approximated linear combination over the dictionary $D$ and coefficients $\alpha$. Finally, the third term determines the degree of sparsity of the coefficients, in fact $0$ counts the null coefficients. Thus, the recovered image is represented with the smallest possible number of vectors from the dictionary. The minimization problem is solved iteratively in three steps. First, a solver such as a matching pursuit type algorithm is used to estimate the coefficients of the linear decomposition of the denoised image over the dictionary  Second, the dictionary is updated. And finally, the last step updates the denoised image $X$. Applied to background/foreground separation, a dictionary learning method considers that \textbf{(1)} The background has a sparse linear representation over a learned dictionary, and \textbf{(2)} the foreground is sparse in the sense that majority pixels of the frame belong to the background. Learning the dictionary is a key step to the success of background modeling. The different approaches developed in the literature differs from the case 1 the algorithm used to learn the dictionary (K-SVD \cite{LD-1}\cite{DL-20}\cite{DL-21}\cite{DL-22}\cite{DL-24}\cite{DL-25}\cite{DL-29}, RDL \cite{DL-23}, MOD \cite{DL-28}), BPFA \cite{DL-28}), 2) the decomposition (Two terms \cite{DL-20}\cite{DL-21}\cite{DL-23}, three terms \cite{DL-22}), 3) the minimization problem with a different norm on the sparse error ($l-1$ norm \cite{DL-23}\cite{DL-24}\cite{DL-26}) or with a different norm on the degree of sparsity ($l_1$-norm \cite{DL-23}\cite{DL-27}\cite{DL-27-1}, Frobenius norm \cite{DL-24}, $l_{1,2}$-norm \cite{SA-70}\cite{SA-71}), and 4) the solvers. Examples of solvers include matching pursuit type algorithm such as Matching Pursuit \cite{DL-20}, Orthogonal Matching Pursuit  \cite{SO-11}\cite{DL-21}\cite{DL-22}\cite{DL-28}, Lasso \cite{SO-20}\cite{DL-24}, Group Lasso \cite{DL-24}, IRLS \cite{DL-26} and Least Angle Regression (LARS) \cite{SO-70}\cite{DL-30}. Furthermore, online dictionary learning algorithms are developed like in Lu et al. \cite{DL-26}, and Zhang et al. \cite{DL-31} with Symmetric Positive Definite (SPD) matrices. \\
\item \textbf{Sparse Linear Approximation/Regression:} This formulation problem is similar to sparse dictionnary learning and leads to the same decomposition. First, Dikmen et al. \cite{SEE-40}\cite{SEE-41}\cite{SEE-42} refer to linear approximation of the sparse error estimation, and basis selection (i.e. the dictionary). This method viewed foreground objects as sparse corruption signals and estimated them by the sparse recovering method. Second, other authors \cite{SEE-50}\cite{SEE-51}\cite{SEE-60} refer to sparse outlier estimation in a linear regression model regarding foreground objects as outliers and consider that the observation error is composed of foreground outlier and background noise. Thus, the foreground detection task has been converted into a outlier estimation problem. \\
\item \textbf{Compressive sensing:} The CS theory states that a signal can be reconstructed from a small number of measurements with high probability, provided that the signal is sparse in the spatial domain or some transform domains \cite{LD-2}. Assume that a signal $X$ can be represented as $X= \Psi \Theta$ , where $\Psi$ denotes a basis and $\Theta$ is the coefficients corresponding to the basis. The signal is said to be $k$-sparse if all other elements in $\Theta$ vanish except for nonzero coefficients. According to CS, for a sparse signal, compressive measurements can be collected by the following random projections:
\begin{equation}
A= \Phi X + N
\label{EquationCS1}
\end{equation}
where $\Phi \in \mathbf{R}^{m\times n}$ is the measurement matrix with $m \leq n$, $A$ contains $m$ measurements, and $N$ is the measurement noise. Specifically, a high dimensional $X$ vector is converted into a much lower dimensional measurement vector $A$. Moreover, the compressive measurements in contain almost all the information of the sparse vector $X$. This means that CS works with data of significantly lower dimension so as to achieve computation efficiency as well as accuracy. In order to recover the noiseless signal, the decomposition problem seeks to the following minimization problems:
\begin{equation}
\underset{D,\alpha}{\text{min}} ~~ ||\Theta||_{l_0} +  \frac{1}{2} ||A - \Phi X||^{2} 
\label{EquationCS2}
\end{equation}
Because Equation \ref{EquationCS2} is an NP hard problem, the sparse solution can be obtained by replacing the nom $l_0$ by the norm $l_1$ as done in \cite{CS-1} \cite{CS-2}\cite{CS-20}\cite{CS-40}\cite{CS-50}\cite{CS-90}. Thus, the background/foreground separation problem can be viewed as a sparse approximation problem where convex optimization and greedy methods can be applied. It is not necessary to learn the background itself to detect the changes and the foreground objects which can be directly detected on the compressive samples. Hence, no foreground
reconstruction is done until a detection is made to save computation. The different approaches developed in the literature differs mainly by the minimization problem ($l_1$-$l_1$ minimization \cite{CS-3}\cite{CS-4}) and the solvers. Examples of solvers include Basis Pursuit (BP) \cite{SO-50}\cite{CS-70}, Basis Pursuit  Denoising (BPDN) \cite{SO-60}\cite{CS-1}, Orthogonal Matching Pursuit (OMP) \cite{CS-70}, Stagewise OMP  (StOMP)\cite{SO-12}\cite{CS-60}, Lattice Matching Pursuit (LaMP) \cite{CS-2}, Convex Lattice Matching Pursuit (CoLaMP) \cite{CS-100},  Compressive sampling matching pursuit (CoSaMP) \cite{CS-10}\cite{CS-20}, and Gradient Projection for Sparse Reconstruction (GPSR) \cite{CS-50}. Furthermore, structured sparsity \cite{SS-1}\cite{SS-2}\cite{SS-3}\cite{CS-2} can be used to exploit a priori spatial information on coefficient structure in addition to signal sparsity as the foreground objects are usually not only sparse but also clustered in a distinct way. Dynamic Group Sparsity (DGS) \cite{DGS-10}\cite{DGS-11}\cite{DGS-15}\cite{CS-90} can also be used to exploit both temporal and spatial information. In a similar way, Liu et al.  \cite{GS-1} used Spatio-Temporal Group Sparsity (STGS) for background subtraction. In an other way, an adaptive algorithm called Adaptive Rate Compressive Sensing (ARCS) \cite{CS-30} \cite{CS-30} allows to choose the number of measurements so as to limit the data rate of the sensor while simultaneously maintaining enough information such that to be able to robustly detect the foreground objects.
\end{itemize}
Like we proposed the unified view DLAM, it is possible to define a unified view of the sparse decompositions that we called Decomposition in Sparse plus Additive Matrices (DSAM) but it is out of the scope of this paper. For more information, the reader can refer to the DSAM Website (https://sites.google.com/site/dsamwebsite/).

\subsubsection{Mixed decompositions}
Mixed decompositions stand at the intersection of the previous problem formulations. There are two main approaches in the literature:
\begin{itemize}
\item{\textbf{RPCA-CS:}} Waters et al. \cite{29}\cite{30} proposed to recover the entries of a matrix $A$ in terms of a low-rank matrix $L$ and sparse matrix $S$ from a small set of compressive measurements $y =\mathbf{A}(L+S)$ where $\mathbf{A}$ is an underdetermined linear operator. The optimization problem that unites the above two problem classes above is:
\begin{equation}
\underset{}{\text{min}} ~~ ||y-\mathbf{A}(L+S)||_{l_2} ~~ \textbf{subj} ~~ rank(L)<r, ~~ ||vec(S)||_{l_0} <K
\label{EquationMD1}
\end{equation}
Waters et al. \cite{29}\cite{30} developed an algorithm for solving Equation \ref{EquationMD1}, called SPArse  and  low  Rank  decomposition  via  Compressive  Sensing (SpaRCS). It combines CoSaMP \cite{CS-10} for sparse vector recovery and ADMiRA \cite{SO-100} for low-rank matrix recovery. 
To accelerate the convergence speed of SpaRCS, Kyrillidis and Cevher \cite{1008} proposed an algorithm called Matrix ALPS and based on acceleration techniques from convex analysis and exploited well-known memory-based acceleration technique. As  incorporating priori knowledge into the basic compressive sensing results in significant improvement of its performance,  Zoonobi and  Kassim \cite{88} extended SpaRCS with partial known support. Jiang et al. \cite{45} reformulated the problem \ref{EquationMD1}) into an equivalent problem by introducing some splitting variables, and  applied the ADM framework. Furthermore, Jiang et al. \cite{45} extented this model to deal with the joint reconstruction of multiple color components. Jiang et al. \cite{45-1} improved this model by adding low latency. In an other way, Yang et al. \cite{72} developed an online algorithm in which the background is learned adaptively as the compressive measurements are processed.  \\
\item{\textbf{Sparse Dictionary Learning-CS:}} Huang et al. \cite{DL-25} proposed an algorithm of moving object detection via the sparse representation and learned dictionary. First, compress image in order to reduce data redundancy and bandwidth. Then, data dictionary with CS
measurement values and sparse basis is initialized, trained and updated through the K-SVD. Finaly, moving object detection is achieved via PCP. In an other approach, Jiang et. \cite{SA-72} used  a spatial-temporal image patch (bricks) as atomic unit for sparse dictionary representation. Furthermore, Random Projection emerged from Compressive Sensing theory is used to reduce the dimension of the bricks so as to speed up the algorithm. \\
\end{itemize}
Like low-rank decompositions, sparse and mixed decompositions can be applied to background/foreground separation too but their study is out of the scope of this paper.

\subsection{Motivations and Contributions}
Since the works of Candes et al. \cite{3} and Chandrasekaran et al. \cite{3-1}, this research field witnessed very significant publications on problem formulations based on the decomposition into low-rank plus additive matrices, and applications in computer vision generate new developments as developed in the handbook \cite{7500}. Furthermore, the different robust problem formulations based on the decomposition into low-rank plus additive matrices often outperform state-of-the-art methods in several computer vision applications \cite{7101}\cite{16010}. Indeed, as this decomposition is nonparametric and does not make many assumptions, it is widely applicable to a large scale of problems ranging from: \\
\begin{itemize}
\item \textbf{Latent variable model selection:} Chandrasekaran et al. \cite{7100} proposed to discover the number of latent components, and to learn a statistical model over the entire collection of variables by only observing samples of a subset of a collection of random variables.
The geometric properties of the decompostion of low-rank plus sparse matrices play an important role in this approach \cite{7100}\cite{7110}.  \\
\item \textbf{Image processing:} Sometimes, it is needed to separate information from noise or outliers in image processing. RPCA framework was applied with success in image analysis \cite{1003} such as image denoising \cite{1012}, image composition \cite{1150}, image decomposition \cite{17020}, image  Mosaicking \cite{17030}, image colorization \cite{1160}, image alignment and rectification \cite{2810}, multi-focus image \cite{1170} and face recognition \cite{1190}. \\
\item \textbf{Video processing:} This application of DLAM is the most investigated one. Indeed, numerous authors used the RPCA and RLRM problem formulations in applications such as action recognition \cite{1200}, motion estimation \cite{1320}, motion saliency detection \cite{49}\cite{1531}\cite{1592}, video coding \cite{81}\cite{83}\cite{97}\cite{1060}, key frame extraction \cite{1030}, hyperspectral video processing \cite{96}, video restoration \cite{1390}, video stabilization \cite{1521}, change detection \cite{17010}, moving target detection \cite{17040}, video object segmentation \cite{1593} and in background and foreground separation \cite{1018}\cite{1031}\cite{1048}.   \\
\item \textbf{3D Computer Vision:} DLAM can be used in structure from motion \cite{38-2}\cite{1370}\cite{1370-1} and 3D motion recovery \cite{17100}. Structure from Motion (SfM) refers to the process of automatically generate a 3D structure of an object by its tracked 2D image frames. Practically, the goal is to recover both 3D structure, namely 3D coordinates of scene points, and motion parameters, namely attitude (rotation) and position of the cameras, starting from image point correspondences. Then, finding the full 3D reconstruction of this object can be posed as a low-rank matrix recovery problem \cite{38-2}\cite{1370}\cite{1370-1}. In an other work, Wang et \cite{17100} developed a 3D motion recovery based on low-rank matrix analysis to correct invalid or corrupted motions.\\
\end{itemize}

\indent In this context, the aim of this survey is then to provide a first complete overview of all the decomposition into low-rank plus additive matrices for \textbf{(1)} novices who could be students or engineers beginning in the field of computer vision, \textbf{(2)} experts as we put forward the recent advances that need to be improved, and \textbf{(3)} reviewers to evaluate papers in journals, conferences, and workshop such as RSL-CV 2015\protect\footnotemark[29]. So, this survey is intended to be a reference for researchers and developers in industries, as well as graduate students, interested in robust decomposition applied to computer vision. \\

\indent Pratically, the publications on decomposition into low-rank plus additive matrices can be classified in two types: \textbf{(1)} publications from researchers of the mathematical community which are more devoted to the fundamental aspect with proofs and with experimentations little investigated in different computer applications, and \textbf{(2)} publications from researchers of the computer vision community which focus on how adapt the decomposition into low-rank plus additive matrices to a specific application by taking into account the specific constraints of this application. Here, we decide to focus on the application of background/foreground separation due to the following reasons:
\begin{enumerate}
\item This application witnessed very numerous papers (more than $390$) since 2009. 
\item Background/foreground separation is the most representative and demanding application as it needs to take into account both spatial and temporal constraints with incremental and real-time constraints \cite{510}\cite{7501}.
\end{enumerate}
Thus, first type of publications are mostly reviewed in Section \ref{sec:RPCA} to Section \ref{sec:LRM} while the second type of publications are mostly grouped in Section \ref{subsubsec:ABFS} and Table \ref{BFSeparation}). \\

\indent Although there is this large number of publications, no algorithm today seems to emerge and to be able to simultaneously address all the key challenges that accompany real-world videos. This is due, in part, to the absence of a rigorous quantitative evaluation with large-scale datasets with accurate ground truth providing a balanced coverage of the range of challenges present in the real world. Indeed, in the first publications, the authors usually compared qualitatively their method to RSL \cite{1} or PCP \cite{3}. Recent quantitative evaluations in foreground detection using the performance metrics have been made but they are limited to one algorithm \cite{40}\cite{66}\cite{502}\cite{501}. In a more recent work, Guyon et al. \cite{500} compared five algorithms RSL \cite{1}, RPCA-PCP solved via EALM \cite{18}, RPCA-PCP solved via IALM \cite{18}, QPCP \cite{8} and BRPCA \cite{7} with the Wallflower dataset \cite{200}, the I2R dataset \cite{203} and Shah dataset \cite{204}. Experimental results show that BRPCA which addresses both spatial and temporal constraints outperforms the other methods. However, this evaluation is limited to five methods and it is not made on large datasets that present a coverage of the range of challenges. A similar study made by Rueda et al. \cite{63} compared RPCA-PCP solved via EALM \cite{18}, BRPCA \cite{7} and GoDec \cite{6}. The authors also concluded that the BRPCA offers the best results in dynamic and static scenes by exploiting the existing correlation between frames of the video sequence using Markov dependencies. In a more complete survey, Bouwmans and Zahzah \cite{510} evaluated ten RPCA-PCP algorithms on the BMC dataset but this evaluation is limited to the framework of RPCA solved via PCP. \\

\footnotetext[29]{{http://rsl-cv2015.univ-lr.fr/workshop/}}

\begin{figure}
\begin{center}
\includegraphics[width=10cm]{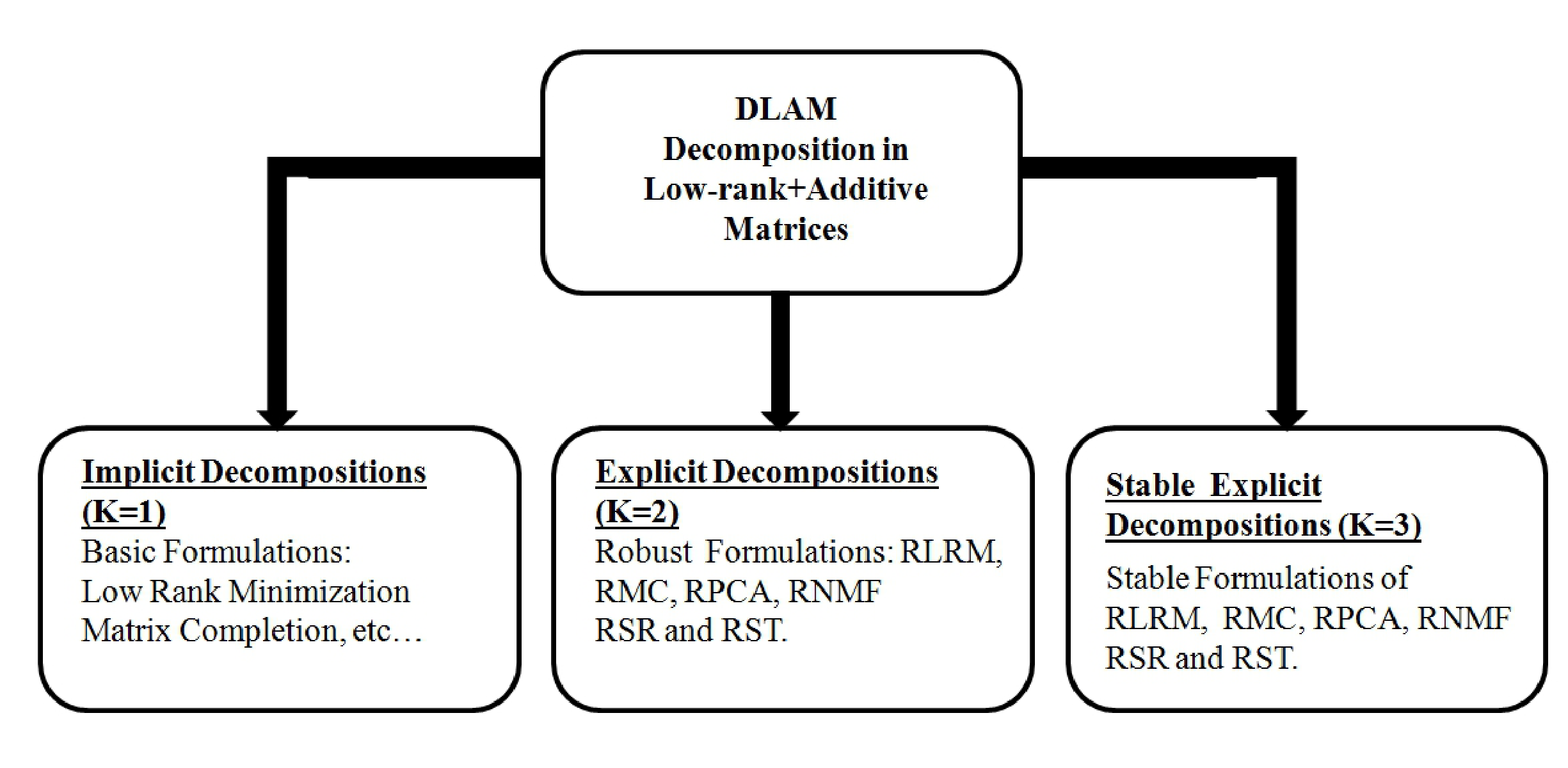}
\caption{Unified View of the Different Decomposition into Low-rank plus Additive Matrices (DLAM).} 
\label{UnifiedViewDLAM}
\end{center}
\end{figure}

\indent Moreover, we believe that we are living in a key transition in the field of background subtraction as we are progressively migrating from the conventional statistical models as MOG \cite{700}\cite{300}, KDE \cite{710}\cite{302} and naive subspace learning models \cite{301} to models based on robust decomposition into low-rank plus additive matrices (RPCA, RNMF, RMC, RSR, RST, LRM) which can achieve at least the same performance in terms of precision than the conventional statistical models \cite{510}. Thus, the aim of this survey is to review and evaluate the robust decomposition into low-rank plus additive matrices for the application of background/foreground separation. For this, it reviews all the models since the first works of Candes et al. \cite{3} and Chandrasekaran et al. \cite{3-1} to the recent ones. By reviewing both existing and new ideas, this survey gives a complete overview of the decompositions, solvers, and applications related to background/foreground separation. Moreover, an accompanying website called the DLAM Website\protect\footnotemark[30] is provided. It allows the reader to have a quick access to the main resources, and codes in the field. Finally, with this survey, we aim to bring a one-stop solution, i.e., access to a number of different decompositions, solvers, implementations and benchmarking techniques in a single paper. Considering all of this, we present a comprehensive review of different methods based on decomposition into low-rank plus additive matrices for testing and ranking existing algorithms for foreground detection.  Contributions of this paper can be summarized as follows: 

\begin{itemize}
\item \textbf{A unified view of the decomposition into low-rank plus additive matrices:} After a preliminary overview on the different robust problem formulations in Section \ref{subsec:Overview}, we provided in Section \ref{subsec:UnifiedFramework} a unified view of the different decompositions into low-rank plus additive matrices. Figure \ref{UnifiedViewDLAM} shows an overview of this unified view. \\
\item \textbf{A review regarding different decomposition methods in low-rank plus additive matrices:} RPCA models are reviewed in Section \ref{sec:RPCA}. For each method, we investigate how they are solved, and if incremental and real-time versions are available for foreground detection. Furthermore, their advantages and drawbacks are discussed in the case of outliers due to dynamic backgrounds or illumination changes. In the same manner, we review the RNMF models in Section \ref{sec:RNMF}, RMC models in Section \ref{sec:RMC}, RSR models in Section \ref{sec:RSR}, and the robust subspace tracking models in Section \ref{sec:ST}. Finally, robust low-rank minimization models are reviewed in Section \ref{sec:LRM}.  \\
\item \textbf{A systematic evaluation and comparative analysis:} We compare and evaluate different decomposition methods in low-rank and additive matrices on a large-scale dataset in Section \ref{sec:ER}. This dataset is the Background Models Challenge (BMC 2012) dataset\protect\footnotemark[31] \cite{205} and we used the provided quantitative evaluation framework which allows us to do a fair and complete comparison. 
\end{itemize}

\indent The rest of this paper is organized as follows. Firstly, we review each original method in its section (Section \ref{sec:RPCA} to Section \ref{sec:LRM}). For each method, we investigate how they are solved, and if incremental and real-time versions are available for background/foreground separation. Then, the performance evaluation using quantitative metrics over the BMC dataset is given in Section \ref{sec:ER}. Finally, we conclude with promising research directions in Section \ref{sec:Conclusion}.

\footnotetext[30]{{https://sites.google.com/site/robustdlam/}}
\footnotetext[31]{{http://bmc.iut-auvergne.com/}}

\section{Robust Principal Component Analysis}
\label{sec:RPCA}

\subsection{RPCA via Principal Component Pursuit}
\label{subsec:RPCA-PCP}
RPCA via PCP proposed by Candes et al. \cite{3} in 2009 is currently the most investigated method. In the following sub-sections, we reviewed this method and all these modifications in terms of decomposition, solvers, incremental algorithms and real time implementations. Table \ref{PCPOverview1}, Table \ref{PCPOverview2} and Table \ref{PCPOverview3} show an overview of the Principal Component Pursuit methods and their key characteristics.

\begin{table*}
\scalebox{0.79}{
\begin{tabular}{|l|l|} 
\hline
\scriptsize{Categories} &\scriptsize{Authors - Dates}                                                \\
\hline
\hline
\scriptsize{\textbf{Decomposition}}                        &\scriptsize{}                             \\
\scriptsize{1) original PCP}                              &\scriptsize{Candes et al. (2009) \cite{3}} \\
\scriptsize{2) modified-PCP (Fixed Rank)}                 &\scriptsize{Leow et al. (2013) \cite{78}}  \\
\scriptsize{3) modified-PCP (Nuclear Norm Free)}          &\scriptsize{Yuan et al. (2013) \cite{75}}  \\
\scriptsize{4) modified-PCP (Capped Norms)}               &\scriptsize{Sun  et al. (2013) \cite{82}}  \\
\scriptsize{5) modified-PCP (Inductive)}                  &\scriptsize{Bao et al. (2012)  \cite{67}}  \\
\scriptsize{6) modified-PCP (Partial Subspace Knowledge)} &\scriptsize{Zhan and Vaswani (2014) \cite{17-4}}                    \\
\scriptsize{7) $p$,$q$-PCP (Schatten-$p$ norm, $l_q$ norm)}               &\scriptsize{Wang et al. (2014)  \cite{1022}}        \\
\scriptsize{8) modified $p$,$q$-PCP (Schatten-$p$ norm, $L_q$ seminorm)}   &\scriptsize{Shao et al. (2014)  \cite{1023}}       \\
\scriptsize{9) modified PCP (2D-PCA)}                                       &\scriptsize{Sun et al. (2013)  \cite{1005}}       \\
\scriptsize{10) modified PCP (Rank-N Soft Constraint)}                      &\scriptsize{Oh  (2012)  \cite{1011}}              \\
\scriptsize{11) Joint Video Frame Set Division  RPCA (JVFSD-RPCA)}          &\scriptsize{Wen (2014) \cite{1026}}               \\
\scriptsize{12) Nuclear norm and Spectral norm Minimization Problem (NSMP)} &\scriptsize{Wang and Feng (2014) \cite{1034}}     \\
\scriptsize{13) Weighted function NSMP (WNSMP)}                             &\scriptsize{Wang and Feng (2014) \cite{1034}}     \\
\scriptsize{14) Implicit Regularizers (IR)}                                 &\scriptsize{He et al. (2013)  \cite{87}}          \\
\scriptsize{15) Random Learning (RL)}                                       &\scriptsize{Rahmani and Atia (2015)  \cite{1063}} \\
\scriptsize{16) Shape Constraint (SC)}                                      &\scriptsize{Yang et al. (2015)  \cite{1902}}      \\
\scriptsize{17) Generalized Fused Lasso regularization (GFL)}               &\scriptsize{Xin et al. (2015)  \cite{1070}}       \\
\scriptsize{18) Double Nuclear Norm-Based Matrix Decomposition (DNMD)}      &\scriptsize{Zhang et al. (2015)  \cite{1100}}     \\
\scriptsize{19) Double Nuclear Norm-Based RPCA (DNRPCA)}      							&\scriptsize{Zhou and Jin (2016)  \cite{1577}}     \\
\scriptsize{20) Self-paced Matrix Factorization (SPMF)}                     &\scriptsize{Zhao et al. (2015)  \cite{1101}}      \\
\scriptsize{21) K-Sparsity Prior (K-SP)}                                    &\scriptsize{Karl and Osendorfer (2015)  \cite{1391}}   \\
\scriptsize{22) Multi-scale Low Rank Matrix Decomposition (MLR)}            &\scriptsize{Ong and Lustig (2015)  \cite{1510}}        \\
\scriptsize{23) Non Convex-RPCA (NC-RPCA)}                                  &\scriptsize{Kang et al. (2015)  \cite{1543}}           \\
\scriptsize{24) Schatten p-Norm low-rank + dual}                            &\scriptsize{Wang et al. (2015)  \cite{1545}}           \\
\scriptsize{25) Weighted Schatten p-Norm Minimization (WSNM)}               &\scriptsize{Xie et al. (2015)  \cite{1546}}            \\
\scriptsize{26) Multi-stage Convex Relaxation (MCR)}                        &\scriptsize{Han and Zhang (2016)  \cite{1573}}         \\
\scriptsize{27) Weighted Low-rank Decomposition (WELD)}  										&\scriptsize{Li et al. (2016)  \cite{1576}}             \\
\scriptsize{28) Regularized Incomplete RPCA (RIRPCA)}  										  &\scriptsize{Shi et al. (2016)  \cite{1591}}            \\
\scriptsize{29) RPCA with PCP with Features (PCPF)}  												&\scriptsize{Chiang  et al. (2016)  \cite{16020}}       \\
\scriptsize{30) Fixed Rank RPCA with Coupled Dictionaries(FRPCA-CD)}  		  &\scriptsize{Lai et al. (2016)  \cite{16060}}           \\
\scriptsize{31) Modified RPCA with Hessian Matrix (RPCA-HM)}                &\scriptsize{Kiruba et al. (2014)  \cite{16070}}        \\
\scriptsize{32) Switched Randomized RPCA (SR-RPCA}                          &\scriptsize{Kaloorazi and Lamare (2016)  \cite{17000}} \\
\scriptsize{33) Approximated RPCA (ARPCA}                                   &\scriptsize{Ebadi et Izquierdo (2015)  \cite{1521}}    \\
\scriptsize{34) Truncated Nuclear Norm (LRSD-TNN)}												  &\scriptsize{Cao et al.  (2016) \cite{17140}}           \\
\scriptsize{35) Truncated Nuclear Norm Minimization (TNNM)}                 &\scriptsize{Zhang et al.  (2016) \cite{17120}}         \\
\scriptsize{36) Online Truncated Nuclear Norm Regularization (OTNNR)}       &\scriptsize{Hong et al.  (2015) \cite{1540}}           \\
\scriptsize{37) Polar n-(Bi)complex PCP}                                    &\scriptsize{Chan and Yang (2016) \cite{17130}}         \\
\scriptsize{38) Optimized Polar n-(Bi)complex PCP}                          &\scriptsize{Chan and Yang (2016) \cite{17130}}         \\
\scriptsize{39) Quaternionic PCP}                                           &\scriptsize{Chan and Yang (2016) \cite{17130}}         \\
\hline
\end{tabular}}
\caption{Principal Component Pursuit: A Complete Overview (Part 1). The first column indicates the concerned category and the second column the name of each method. Their corresponding acronym is indicated in the first parenthesis. The third column gives the name of the authors and the date of the related publication.} \centering
\label{PCPOverview1}
\end{table*}

\begin{table*}
\scalebox{0.79}{
\begin{tabular}{|l|l|} 
\hline
\scriptsize{Categories} &\scriptsize{Authors - Dates}                                                \\
\hline
\hline
\scriptsize{\textbf{Solvers}}                      &\scriptsize{}                                                   \\
\scriptsize{\textbf{1) Basic solvers}}             &\scriptsize{}                                                   \\
\scriptsize{Singular Values Decomposition (SVT)}   &\scriptsize{Cai et al. (2008) \cite{20}}                        \\
\scriptsize{Iterative Thresholding (IT)}           &\scriptsize{Wright et al. (2009) \cite{4}}                      \\
\scriptsize{Accelerated Proximal Gradient (APG)}   &\scriptsize{Lin et al.(2009) \cite{19}}                         \\
\scriptsize{Dual Method (DM)}                      &\scriptsize{Lin et al.(2009) \cite{19}}                         \\
\scriptsize{Exacted Augmented Lagrangian Method (EALM)}            &\scriptsize{Lin et al. (2009) \cite{18}}        \\
\scriptsize{Inexact Augmented Lagrangian Method (IALM)}            &\scriptsize{Lin et al. (2009) \cite{18}}        \\
\scriptsize{Alternating Direction Method (ADM)}                    &\scriptsize{Yuan and Yang (2009) \cite{21}}     \\
\scriptsize{ADM with Gaussian back substitution (ADM-G)}           &\scriptsize{He and Yuan (2012) \cite{51}}       \\
\scriptsize{Symmetric Alternating Direction Method (SADM)}         &\scriptsize{Goldfarb et al. (2010) \cite{47}}   \\
\scriptsize{Non Convex Splitting ADM (NCSADM)}                     &\scriptsize{Chartrand (2012) \cite{65}}         \\
\scriptsize{Douglas-Rachford Splitting Method (DRSM)}              &\scriptsize{Gandy and Yamada (2010] \cite{74}}  \\
\scriptsize{Variant of Douglas-Rachford Splitting Method (VDRSM)}  &\scriptsize{Zhang and Liu (2013) \cite{79}}     \\
\scriptsize{Proximity Point Algorithm (PPA)}                       &\scriptsize{Zhu et al. (2014) \cite{1017}}      \\
\scriptsize{Proximal Iteratively Reweighted Algorithm (PIRA)}      &\scriptsize{Wang et al. (2014)  \cite{1022}}    \\
\scriptsize{Alternating Rectified Gradient Method (ARGM)}          &\scriptsize{Kim et al. (2014)  \cite{1016}}     \\
\scriptsize{Parallel Direction Method of Multipliers (PDMM)}       &\scriptsize{Wang et al. (2014)  \cite{1032}}    \\
\scriptsize{Generalized Singular Value Thresholding (GSVT)}        &\scriptsize{Lu et al. (2014)  \cite{1061}}      \\
\scriptsize{Generalized Accelerated Proximal Gradient (GAPG)}      &\scriptsize{He et al. (2013)  \cite{87}}        \\
\scriptsize{Improved alternating direction method (IADM)}          &\scriptsize{Chai et al. (2013) \cite{91}}       \\ 
\scriptsize{Iterative Thresholding with Primal-Dual Method (IT-PDM)} &\scriptsize{Fan et al. (2014) \cite{1040}}    \\
\scriptsize{Optimal Singular Values Shrinkage (OptShrink)}         &\scriptsize{Moore et al. (2014) \cite{1046}}    \\  
\scriptsize{Alterning Minimization (AM)}                           &\scriptsize{Gu et al. (2016) \cite{1574}}       \\  
\scriptsize{Bi-Factored Gradient Descent (BFGD)}                   &\scriptsize{Park  et al. (2016) \cite{16040}}   \\ 
\cline{1-2}
\scriptsize{\textbf{2) Linearized solvers}}                         &\scriptsize{}                                  \\
\scriptsize{Linearized ADM (LADM)}                                  &\scriptsize{Yang and Yuan (2011) \cite{22}}    \\
\scriptsize{Linearized ADM with Adaptive Penalty (LADMAP)}          &\scriptsize{Lin et al. (2011) \cite{39}}       \\
\scriptsize{Linearized Symmetric ADM (LSADM)}                       &\scriptsize{Goldfarb et al. (2010) \cite{47}}  \\
\scriptsize{Fast Linearized Symmetric ADM (Fast-LSADM)}             &\scriptsize{Goldfarb et al. (2010) \cite{47}}  \\
\scriptsize{Linearized IAD Contraction Methods (LIADCM)}            &\scriptsize{Gu et al. (2013) \cite{1020}}      \\
\cline{1-2}
\scriptsize{\textbf{3) Fast solvers}}                                   &\scriptsize{}    \\
\scriptsize{Randomized Projection for ALM (RPALM)}	    					      &\scriptsize{Mu et al. (2011) \cite{11}}                  \\
\scriptsize{$l_1$-filtering (LF)}                                       &\scriptsize{Liu et al. (2011) \cite{38}}                 \\ 
\scriptsize{Block Lanczos with Warm Start}                              &\scriptsize{Lin and Wei (2010) \cite{57}}                \\ 
\scriptsize{Exact Fast Robust Principal Component Analysis (EFRPCA)}    &\scriptsize{Abdel-Hakim and El-Saban (2012) \cite{59}}   \\
\scriptsize{Inexact Fast Robust Principal Component Analysis (IFRPCA)}  &\scriptsize{Abdel-Hakim and El-Saban (2012) \cite{59}}   \\
\scriptsize{Matrix Tri-Factorization (MTF)}	         										&\scriptsize{Liu et al. (2013) \cite{60}}			           	\\								\scriptsize{Fast Tri-Factorization(FTF)}	         											&\scriptsize{Liu et al. (2013) \cite{61}}	                \\
\scriptsize{PRoximal Iterative SMoothing Algorithm (PRISMA)}            &\scriptsize{Orabona et al. (2012) \cite{71}}             \\
\scriptsize{Fast Alterning Minimization (FAM)}                          &\scriptsize{Rodriguez and Wohlberg (2013) \cite{26-1}}   \\
\scriptsize{Fast Alternating Direction Method of Multipliers (FADMM)}    &\scriptsize{Yang and Wang (2014) \cite{1028}}           \\
\scriptsize{Fast Alternating Direction Method with Smoothing Technique (FADM-ST)} &\scriptsize{Yang (2014)\cite{1039}}            \\ 
\scriptsize{Fast Randomized Singular Value Thresholding (FRSVT)}        &\scriptsize{Oh et al. (2015)\cite{1071}}                 \\ 
\scriptsize{Coherence Pursuit (CoP)}                                    &\scriptsize{Rahmani and Atia\cite{1571-2}}               \\ 
\cline{1-2}
\scriptsize{\textbf{4) Online solvers}}                             &\scriptsize{}                                              \\
\scriptsize{Online Alternating Direction Method (OADM)}             &\scriptsize{Wang and Banerjee (2013) \cite{1019}}          \\
\cline{1-2}
\scriptsize{\textbf{5) Non convex solvers}}                             &\scriptsize{}                                           \\
\scriptsize{Difference of Convex (DC)}                                  &\scriptsize{Sun et al. (2013) \cite{82}}                \\
\scriptsize{Fast Alternating Difference of Convex (FADC)}               &\scriptsize{Sun et al. (2013) \cite{82}}                \\
\scriptsize{Non-convex Alternating Projections(AltProj)}                &\scriptsize{Netrapalli et al. (2014) \cite{1047}}       \\
\scriptsize{Iterative Shrinkage-Thresholding/Reweighted Algorithm (ISTRA)}  &\scriptsize{Zhong et al. (2015) \cite{1064}}        \\
\scriptsize{Gauss-Newton ADMM (GN-ADMM)}                                    &\scriptsize{Tran-Dinh and Zhang(2016) \cite{16030}} \\
\scriptsize{Fast NonConvex Low-rank (FaNCL)}                                &\scriptsize{Yao et al. (2015) \cite{1590}}          \\
\scriptsize{Fast RPCA via Gradient Descent (GD)} 														&\scriptsize{Yi et al. (2016) \cite{1595}}           \\
\scriptsize{Fixed Rank - Fast Factorization based RPCA (F-FFP)}             &\scriptsize{Peng et al. (2016) \cite{17110}}        \\
\scriptsize{Unfixed Rank - Fast Factorization based RPCA (U-FFP)}             &\scriptsize{Peng et al. (2016) \cite{17110}}      \\
\cline{1-2}
\scriptsize{\textbf{6) 2D solvers}}                                              &\scriptsize{}                                  \\
\scriptsize{Iterative method for Bi-directional Decomposition (IMBD)}            &\scriptsize{Sun et al. (2013) \cite{1005}}     \\    
\cline{1-2}
\scriptsize{\textbf{7) Free SVD solvers}}                                       &\scriptsize{}                                  \\
\scriptsize{Free SVD algorithm}            																			&\scriptsize{Ebadi and Izquierdo (2015) \cite{1522}} \\
\hline
\end{tabular}}
\caption{Principal Component Pursuit: A Complete Overview (Part 2). The first column indicates the concerned category and the second column the name of each method. Their corresponding acronym is indicated in the first parenthesis. The third column gives the name of the authors and the date of the related publication.} \centering
\label{PCPOverview2}
\end{table*}

\begin{table*}
\scalebox{0.65}{
\begin{tabular}{|l|l|l|} 
\hline
\scriptsize{Categories}             &\scriptsize{Methods} &\scriptsize{Authors - Dates}   \\
\hline
\hline
\scriptsize{Incremental Algorithms}  &\scriptsize{Recursive Robust PCP (RR-PCP)}   &\scriptsize{Qiu and Vaswani (2010) \cite{14}}            \\
\scriptsize{}        &\scriptsize{Recursive Project Compressive Sensing (ReProCS)}  &\scriptsize{Qiu and Vaswani (2011)  \cite{16}}          \\
\scriptsize{}        &\scriptsize{Support-Predicted Modified-CS RR-PCP}            &\scriptsize{Qiu and Vaswani (2011) \cite{15}}            \\
\scriptsize{}        &\scriptsize{Support-Predicted Modified-CS}                   &\scriptsize{Qiu and Vaswani (2011) \cite{16}}            \\
\scriptsize{}        &\scriptsize{Automated ReProCS}                               &\scriptsize{Qiu and Vaswani (2012) \cite{17}}            \\
\scriptsize{}        &\scriptsize{Practical ReProCS (Prac-ReProCS)}                &\scriptsize{Guo et al. (2013) \cite{17-3}}               \\
\scriptsize{}        &\scriptsize{Incremental Low-Rank (iLR) Algorithm}            &\scriptsize{Wei et al. (2014) \cite{1007}}               \\
\scriptsize{}        &\scriptsize{Incremental PCP (incPCP)}                        &\scriptsize{Rodriguez and Wohlberg (2014) \cite{26-2}}   \\ 
\scriptsize{}        &\scriptsize{Incremental PCP TI (incPCP-TI)}                  &\scriptsize{Rodriguez and Wohlberg (2015) \cite{26-6}}   \\ 
\scriptsize{}        &\scriptsize{Online RPCA (ORPCA)}                             &\scriptsize{Xu (2014) \cite{1038}}                       \\ 
\scriptsize{}        &\scriptsize{Online RPCA via Stochastic Gradient Descent (ORPCA-SGD)} &\scriptsize{Song et al. (2015) \cite{1534}}      \\ 
\scriptsize{}        &\scriptsize{Projection based RPCA (ProjectionRPCA)}          &\scriptsize{Lee and Lee (2015) \cite{1535}}              \\ 
\scriptsize{}        &\scriptsize{Online RPCA with Truncated Nuclear Norm Regularization (OTNNR)}   &\scriptsize{Hong et al. (2015) \cite{1540}}  \\ 
\hline
\hline
\scriptsize{Real time Implementations}      &\scriptsize{CAQR}             &\scriptsize{Anderson et al. (2010) \cite{12}}        \\
\scriptsize{}                               &\scriptsize{Real-Time PCP}    &\scriptsize{Pope et al. (2011) \cite{42}}            \\
\scriptsize{} &\scriptsize{LR Submatrix Recovery/Reconstruction (LRSRR)}   &\scriptsize{Guo et al. (2014) \cite{1015}}           \\
\scriptsize{} &\scriptsize{Real time inPCP on TK1}                         &\scriptsize{Rodriguez (2015) \cite{26-4}}            \\
\scriptsize{} &\scriptsize{Real time inPCP-TI on TK1}                      &\scriptsize{Silva and Rodriguez (2015) \cite{26-7}}  \\
\hline
\scriptsize{Multi-Features Algorithms}      &\scriptsize{Multi-Features Algorithm (MFA)}     &\scriptsize{Gan et al. (2013) \cite{95}}      \\
\scriptsize{}                               &\scriptsize{Multi-Task RPCA (MTRPCA)}           &\scriptsize{Wang and Wan (2014) \cite{1051}}  \\
\hline
\scriptsize{Spatial-Temporal Algorithms}    &\scriptsize{Dense Optical Flow}           &\scriptsize{Gao et al. (2012) \cite{62}}           \\
\scriptsize{}           &\scriptsize{Consistent Optical Flow}                          &\scriptsize{Huang et al. (2013) \cite{1041}}       \\
\scriptsize{}           &\scriptsize{Smoothness and Arbitrariness Constraints (RFDSA)} &\scriptsize{Guo et al. (2014) \cite{1033}}         \\
\scriptsize{}           &\scriptsize{Total Variation (TV) Regularizer (TVR)}         &\scriptsize{Gao et al. (2015) \cite{1900}}           \\
\scriptsize{}           &\scriptsize{Piece-wise Low-rank Model (PLM)}                &\scriptsize{Newson et al. (2015) \cite{1505}}        \\
\scriptsize{}           &\scriptsize{Graphs Model (RPCAG)}                           &\scriptsize{Shahid et al. (2015) \cite{1072-1}}      \\
\scriptsize{}           &\scriptsize{Fast Graphs Model (FRPCAG)}                     &\scriptsize{Shahid et al. (2015) \cite{1072-2}}      \\
\scriptsize{}           &\scriptsize{Shape and Confidence Map-based (SCM-RPCA)}      &\scriptsize{Sobral et al. (2015) \cite{1711}}        \\
\scriptsize{}           &\scriptsize{Connectivity and Saliency Map (MODSM)}          &\scriptsize{Pang et al. (2015) \cite{1532}}          \\
\scriptsize{}           &\scriptsize{Salient Motion Detection (SMD-RPCA}             &\scriptsize{Chen et al. (2015) \cite{1537}}          \\
\scriptsize{}           &\scriptsize{Depth-weighted Group-wise PCA (DG-PCA)}         &\scriptsize{Tian et al. (2015) \cite{1542}}          \\
\scriptsize{}           &\scriptsize{Structured Sparsity RPCA (GSRPCA-LSD)}          &\scriptsize{Liu et al. (2015) \cite{1901}}           \\
\scriptsize{}           &\scriptsize{Superpixel Structured Foreground (SPGFL)}       &\scriptsize{Javed  et al. (2015) \cite{1712}}        \\
\scriptsize{}           &\scriptsize{Block Structured ARPCA (ARPCA-BS)}              &\scriptsize{Ebadi and Izquierdo (2015) \cite{1522}}  \\
\scriptsize{}           &\scriptsize{Dynamic Tree-Structured ARPCA (ARPCA-CSSP)}     &\scriptsize{Ebadi et al. (2016) \cite{1524}}         \\
\scriptsize{}           &\scriptsize{Dynamic SuperPixel Structured-Sparse (DSPSS)}   &\scriptsize{Ebadi and Izquierdo (2016) \cite{1525}}  \\
\scriptsize{}           &\scriptsize{Motion-Aware Graphs Regularized RPCA (MAGRPCA)} &\scriptsize{Javed et al. (2016) \cite{1010-7}}       \\
\scriptsize{} 					&\scriptsize{Spatiotemporal RPCA (SRPCA)}										 &\scriptsize{Javed et al. (2016) \cite{1010-10}}      \\
\hline
\scriptsize{Compressive Sensing Algorithms} &\scriptsize{Sparsity Reconstruction for Compressive Sensing (SpaRCS)}                                   &\scriptsize{Waters et al. (2011) \cite{29}} \\
\scriptsize{}   &\scriptsize{SpaRCS with Partial Support Knowledge (SpaRCS-PSK)}  &\scriptsize{Zonoobi and Kassim (2013) \cite{88}}   \\
\scriptsize{}   &\scriptsize{Adaptive Reconstruction Compressive Sensing (ARCS)}  &\scriptsize{Yang et al. (2013) \cite{72}}          \\
\scriptsize{}   &\scriptsize{LRSD for Compressive Sensing (LRSDCS)}               &\scriptsize{Jiang et al. (2014) \cite{1008}}       \\
\scriptsize{}   &\scriptsize{Recursive Low-rank and Sparse decomposition (rLSDR)} &\scriptsize{Li and Qi (2014) \cite{1049}}          \\
\scriptsize{}   &\scriptsize{Compressive PCA on Graphs (CPCA)}                    &\scriptsize{Shahid et al. (2016) \cite{1072-3}}    \\
\hline
\scriptsize{Optimal PCP Solutions}   &\scriptsize{Minimum Description Length (MDL)}               &\scriptsize{Ramirez and Shapiro (2012) \cite{37}} \\
\scriptsize{}                       &\scriptsize{Saliency Measure}                                &\scriptsize{Gao et al. (2012) \cite{62}} \\
\hline
\scriptsize{SVD Algorithms} &\scriptsize{Full SVD}            &\scriptsize{-} \\
\scriptsize{}               &\scriptsize{Partial SVD}         &\scriptsize{-} \\
\scriptsize{}               &\scriptsize{Linear Time SVD}     &\scriptsize{Yang and  An (2013) \cite{73}}    \\
\scriptsize{}               &\scriptsize{smaller-scale SVD}   &\scriptsize{Zhang and  Tian (2013) \cite{90}} \\
\scriptsize{}               &\scriptsize{block-SVD}                                           &\scriptsize{Chai et al. (2013) \cite{91}}       \\
\scriptsize{}               &\scriptsize{Limited Memory SVD (LMSVD)}                          &\scriptsize{Liu et al. (2013)  \cite{2301}}     \\
\scriptsize{}               &\scriptsize{Symmetric Low-Rank Product-Gauss-Newton (SLRPGN)}    &\scriptsize{Liu et al. (2014) \cite{2302}}      \\
\scriptsize{}               &\scriptsize{randomized SVD (rSVD)}                               &\scriptsize{Erichson et al. (2016) \cite{2027}} \\
\hline
\end{tabular}}
\caption{Principal Component Pursuit: A Complete Overview (Part 3). The first column indicates the concerned category and the second column the name of each method. Their corresponding acronym is indicated in the first parenthesis. The third column gives the name of the authors and the date of the related publication.} \centering
\label{PCPOverview3}
\end{table*}

\subsubsection{Principal Component Pursuit}
\label{sec:PCP1}
Candes et al. \cite{3}\cite{4} proposed a convex optimization to address the robust PCA problem. The observation matrix $A$ is assumed represented as: 
\begin{equation}
A = L + S
\label{EquationPCP1}
\end{equation}
where $L$ is a low-rank matrix and $S$ must be sparse matrix with a small fraction of nonzero entries. The straightforward formulation is to use $l_0$-norm to minimize the energy function:
\begin{equation}
\underset{L,S}{\text{min}} ~~ rank(L) + \lambda ||S||_{l_{0}}  ~~~ \text{subj} ~~~ A-L-S=0
\label{EquationPCP2}
\end{equation}
where $\lambda>0$ is an arbitrary balanced parameter.
But this problem is $NP$-hard, typical solution might involve a search with combinatorial complexity. This research seeks to solve for $L$ with the following optimization problem:
\begin{equation}
\underset{L,S}{\text{min}} ~~ ||L||_* + \lambda ||S||_{l_{1}}  ~~~ \text{subj} ~~~ A-L-S=0
\label{EquationPCP3}
\end{equation}
where $||.||_*$  and $||.||_{l_{1}}$ are the nuclear norm (which is the $l_1$-norm of singular value) and $l_1$-norm, respectively, and
$\lambda>0$ is an arbitrary balanced parameter. Usually, $\lambda=\frac{1}{\sqrt{max(m,n)}}$. Under these minimal assumptions, this approach called Principal Component Pursuit (PCP) solution perfectly recovers the low-rank and the sparse matrices. \\
\indent Candes et al. \cite{5} showed results on face images and background modeling that demonstrated encouraging performance.  The low-rank minimization concerning $L$ offers a suitable framework for background modeling due to the correlation between frames. So, minimizing $L$ and $S$ implies that the background is approximated by a low-rank subspace that can gradually change over time, while
the moving foreground objects constitute the correlated sparse outliers which are contained in $S$. To obtain the
foreground mask, $S$ needs to be thresholded. The threshold is determined experimentally. $rank(L)$ influences the number of modes of the background that can be represented by $L$: If $rank(L)$ is to high, the model will incorporate the moving objects in its representation; if the $rank(L)$ is to low, the model tends to be uni-modal and then the multi-modality which appears in dynamic backgrounds will be not captured. The quality of the background/foreground separation is directly related to the assumption of the low-rank and sparsity of the background and foreground, respectively.
In this case, the best separation is then obtained only when the optimization algorithm has converged. \\
\indent Essentially, the nuclear-norm term corresponds to low-frequency components along the temporal while the $l_1$ norm describes
the high-frequency components. However, the low-frequency components can leak into extracted background images for areas that are dominated by moving objects. The leakage as ghost artifacts which appear in extracted background cannot be well handled by adjusting the weights between the two
regularization parameters. Practically, RPCA-PCP present other several limitations developed in the Section \ref{subsubsec:RPCAFramework} and an overview of their solutions is given in the following sections.

\subsubsection{Algorithms for solving PCP}
Several algorithms called solvers have been proposed for solving PCP. An overview of these solvers as well as their complexity (when they are available) are grouped in Table \ref{A-TPCP1Overview} and Table \ref{A-TPCP11Overview}. For an $m\times n$ input matrix $A$ with estimated rank $r$, the complexity per iteration at running time is formulated as $O_{iter}(f_{iter}(m,n,r,..))$ where $f_{iter}()$ is a function. The complexity to reach an accuracy of $\epsilon$ precision ($\epsilon$-optimal solution) is formulated as $O_{pre}(f_{pre}(\epsilon))$ where $f_{pre}()$ is a function.  The convergence rate is formulated as $O_{conv}(f_{conv}(T))$, where $f_{conv}()$ is a function of $T$ which is the number of iterations.
All these algorithms require solving the following type of subproblem in each iteration:
\begin{equation}
\underset{L,S}{\text{min}} ~~ \eta ||L||_{norm_1}^{p_1} + \lambda ||S||_{norm_2}^{p_2}
\label{EquationPCP4}
\end{equation}
The above problem can have a closed form solution or not following the application. So, several solvers can be found in the the literature: \\
\begin{itemize}
\item \textbf{Basic solvers:} When the problem is supposed to have a closed form solution, PCP can be reformulated as a semidefinite program and then be solved by standard interior point methods \cite{100}. However, interior point methods have difficulty in handling large matrices because the complexity of computing the step direction is $O((mnmin(m,n))^2)$, where $m \times n$ is the size of the data matrix. If $m=n$, then the complexity is $O(n^6)$. So the generic interior point solvers are too limited for many real applications where the number of data are very large. To overcome the scalability issue, only the first-order information can be used. Cai et al. \cite{20} showed that this technique, called Singular Value Thresholding (SVT), can be used to minimize the nuclear norm for matrix completion. As the matrix recovery problem in Equation (\ref{EquationPCP3}) needs minimizing a combination of both the $l_1$-norm and the nuclear norm, Wright et al. \cite{4} adopted a iterative thresholding technique (IT) to solve it and obtained similar convergence and scalability properties than interior point methods. However, the iterative thresholding scheme converges extremely slowly with $O_{pre}=\sqrt{L/\epsilon}$ where $L$ is the Lipschitz constant of the gradient of the objective function. To alleviate this slow convergence, Lin et al. \cite{19} proposed two algorithms: the accelerated proximal gradient (APG) algorithm and the gradient-ascent algorithm applied to the dual of the problem in Equation (\ref{EquationPCP3}). However, these algorithms are all the same too slow for real application. More recently, Lin et al. \cite{18} proposed two algorithms based on augmented Lagrange multipliers (ALM). The first algorithm is called Exact ALM (EALM) method that has a Q-linear convergence speed, while the APG is in theory only sub-linear. The second algorithm is an improvement of the EALM that is called Inexact ALM (IALM) method, which converges practically as fast as the EALM, but the required number of partial SVDs is significantly less. The IALM is at least five times faster than APG, and its precision is also higher \cite{18}. However, the direct application of ALM treats Equation (\ref{EquationPCP3}) as a generic minimization problem and ignores its separable structure emerging in both the objective function and the constraint \cite{18}. Hence, the variables $S$ and $L$ are minimized simultaneously. Yuan and Yang \cite{21} proposed to alleviate this ignorance by the Alternating Direction Method of Multipliers (ADMM) which minimizes the variables $L$ and $S$ serially. The convergence of ADMM for convex objective functions has been proved \cite{1130}\cite{1131}\cite{1076}\cite{1600}\cite{1602}\cite{1603}. The iteration complexity was  analyzed for Multi-Block ADMM \cite{1601}. ADMM achieves it with less computation cost than ALM. Recently, Chartrand \cite{65} proposed a non convex splitting version of the ADMM \cite{21} called NCSADM. This non convex generalization of \cite{18} produces a sparser model that is better able to separate moving objects and stationary objects. Furthermore, this splitting algorithm maintains the background model while removing substantial noise, more so than the convex regularization does. The ALM neglects the separable structure in both the objective function and the constraint. Thus, Zhang and Liu \cite{79} proposed a variant of the Douglas-Rachford splitting method (VDRSM) for accomplishing recovery in the case of illumination changes and dynamic backgrounds. In a similar way, Zhu et al. \cite{1017} proposed a Proximity Point Algorithm (PPA) based on the Douglas-Rachford splitting method. The convex optimization problem is solved by canceling the constraint of the variables, and the proximity operators of the objective function are computed alternately. The new algorithm can exactly recover the low-rank and sparse simultaneously, and it is proved to be convergent.
An other approach developed by Chai et al. \cite{91} is an improved alternating direction method (IADM) algorithm with a block based SVD approach. Experimental results \cite{91} on the I2R dataset \cite{203} show that IADM outperforms SVT \cite{20}, APG \cite{19}, IALM \cite{18} and ADM \cite{21} with less computation time. \\
\item \textbf{Linearized solvers:} When the resulting subproblems do not have closed-form solutions, Yang and Yuan \cite{22} proposed to linearize these subproblems such that closed-form solutions of these linearized subproblems can be easily derived. Global convergence of these Linearized ALM (LALM) and ADM (LADM) algorithms are established under standard assumptions. Recently, Lin et al. \cite{39} improved the convergence for the Linearized Alternating Direction Method with an Adaptive Penalty (LADMAP). They proved the global convergence of LADM and applied it to solve Low-Rank Representation (LRR). Furthermore, the fast version of LADMAP reduces the complexity $O(m \times n \times min(m,n))$ of the original LADM based method to $O(r\times m\times n)$, where r is the rank of the matrix to recover, which is supposed to be smaller than $m$ and $n$. In a similar way, Ma \cite{46} and Goldfarb et al. \cite{47} proposed a Linearized Symmetric Alternating Direction Method (LSADM) for minimizing the sum of two convex functions. This method requires at most $O(1/ \epsilon)$ iterations to obtain an $\epsilon$-optimal solution, and its fast version called Fast-LSADM requires at most $O(1/ \sqrt{\epsilon})$ with little change in the computational effort required at each iteration. \\
\item \textbf{Fast solvers:} All the previous solvers require computing SVDs for some matrices, resulting in  $O(m\times n\times min(m,n))$ complexity. Although partial SVDs are used to reduce the complexity to $O(r\times m\times n)$ such a complexity is still high for large data sets. Therefore, recent researches focus on the reduction of the complexity by avoiding computation of SVD. Shen et al. \cite{23} presented a method where the low-rank matrix is decomposed in a product of two low-rank matrices and then minimized over the two matrices alternatively. Although, they do not require nuclear norm minimization and so the computation of SVD, the convergence of the algorithm is not guaranteed as the problem is non-convex. Furthermore, both the matrix-matrix multiplication and QR decomposition based rank estimation technique require $O(r\times m\times n)$ complexity. So, this method does not essentially reduce the complexity. In an other way, Mu et al. \cite{11} reduced the problem scale by random projections (linear or bilinear projection) but different random projections may lead to radically different results. Furthermore, additional constraints to the problem slow down the convergence. The complexity of this method is $O(p\times m\times n)$ where $p \times m$ is the size of the random projection matrix with $p \ll m$, $p \ll n$ and $p>r$. So, this method is still nor linear complexity with respect to the matrix size. Its convergence needs more iterations than IALM but it requires less computation time. In an other way, Liu et al. \cite{38}\cite{38-1} proposed an algorithm, called $l_1$-filtering for exactly solving PCP with an $O(r^2(m+n))$ complexity. This method is a truly linear cost method to solve PCP problem when the data size is very large while the target rank is small. Moreover, $l_1$-filtering is highly parallelizable. It is the first algorithm that can exactly solve a nuclear norm minimization problem in linear time. Numerical experiments \cite{38}\cite{38-1} show the great performance of $l_1$-filtering in speed compared to the previous algorithms for solving PCP. In an other way,  Orabona et al. \cite{71} proposed an optimization algorithm  called PRoximal Iterative SMoothing Algorithm (PRISMA) which is decomposed into three parts: a smooth part, a simple non-smooth Lipschitz part, and a simple non-smooth non-Lipschitz part. Furthermore, a time variant smoothing strategy is used to obtain a guarantee that does not depend on knowing in advance the total number of iterations nor a bound on the domain. Numerical experiments \cite{71} show that PRISMA required less iterations than Fast-LSADM \cite{47}. An other approach developed by Rodriguez and Wohlberg  \cite{26-1} is able to compute a sparse approximation even after the first outer loop, (taking approximately 12 seconds for a $640\times480\times400$ color test video) which is approximately an order of magnitude faster than IALM \cite{18} with the same accuracy. Yang and Wang \cite{1028} proposed a Fast Alternating Direction Method of Multipliers (FADMM) algorithms which outperforms slightly IALM \cite{18} and ADM \cite{21} in term of computation times. Yang \cite{1039} improved FADMM by using a smoothing technique which is used to smooth the non-smooth terms in the objective function. \\
\item \textbf{Online solvers:} The previous solvers are mainly bacth ones but online algorithms are better adapted for real-time application. So, Wang and Banerjee \cite{1019} proposed an efficient online learning algorithm named online ADM (OADM) which can solve online convex optimization under linear constraints where the objective could be nonsmooth. \\
\item \textbf{Non convex solvers:} Sun et al. \cite{82} developed for the non-convex RPCA formulation of RPCA with capped norms two algorithms called Difference of Convex (DC) and Fast Alternating Difference of Convex (FADC), respectively. DC programming treats a non-convex function as the difference
of two convex functions, and then iteratively solves it on the basis of the combination of the first convex part and the linear approximation of the second convex part. Numerical measurements  \cite{82} demonstrate that DC approach performs better than both IALM \cite{18} and NSA \cite{34} in terms of the low-rank and sparsity. In an other way, Netrapalli et al. \cite{1047} proposed a Non-convex Alternating Projections algorithm (AltProj) to solve a non-convex formulation of RPCA. The overall complexity of AltProj is $O(r^2mn \text{log}(1/\epsilon)$). This is drastically lower than the best known bound of $O(m^2n / \epsilon)$ on the number of iterations required by convex methods, and just a factor $r$ away from the complexity of naive PCA. AltProj is around 19 times faster than IALM \cite{18}. Moreover, visually, the background extraction seems to be of better accuracy. \\
\item \textbf{2D solvers:} Sun et al. \cite{1005}\cite{1005-1} developed an iterative algorithm for robust 2D-PCA via alternating optimization which learns the projection matrices by bi-directional decomposition. To further speed up the iteration, Sun et al. \cite{1005}\cite{1005-1} proposed an alternating greedy approach or $l_0$-norm regularization, minimizing over the low-dimensional feature matrix and the sparse error matrix. \\
\end{itemize}

\subsubsection{Algorithms for incremental PCP}
PCP is an offline method which treats each image frame as a column vector of the matrix $A$. In real-time application such as foreground detection, it would be more useful to estimate the sparse matrix in an incremental way quickly as new frame comes rather than in batch way. Furthermore, the sparsity structure may change slowly or in a correlated way, which may result in a low-rank sparse matrix. In this case, PCP assumption will not get satisfied and $S$ can't be separated from $L$. Moreover, the principal directions can change over time. So, the rank of the matrix $L$ will keep increasing over time thus making PCP infeasible to do after time. This last issue can be solved  by not using all frames of the sequences. So, several algorithms for incremental algorithms are available in literature.

\begin{enumerate}
\item \textbf{Recursive Projected Compressive Sensing (ReProCS):}  To address the two first issues, Qiu and Vaswani \cite{14} proposed an online approach called Recursive Robust PCP (RR-PCP) in \cite{14}, and Recursive Projected Compressive Sensing (ReProCS) in \cite{16}\cite{17-1}\cite{17-21}\cite{17-22}. The aim of ReProCS is to causally keep updating the sparse matrix $S_t$ at each time, and keep updating the principal directions. The $t$-th column of $A$, $A_t$, is the data acquired at time $t$ and can be decomposed as follows:
\begin{equation}
A_t=L_t+S_t=\left[U I\right]\left[x_t S_t\right]^t
\label{EquationRRPCP1}
\end{equation}
where $x_t=U^TL_t$ and the matrix $U$ is an unknown $m \times m$ orthonormal matrix. The support of $S_t$ changes slowly over time. Let $N_t$ denote the support of $x_t$ which is assumed piecewise constant over time and given an initial estimate of $P_t=(U)_{N_t}=\hat{P}_t$, Qiu and Vaswani \cite{14} solved for the sparse component $S_t$ by finding the orthogonal complement matrix $\hat{P}_{t,\perp}$, and then using the projection $M_t$ onto $\hat{P}_{t,\perp}$, denoted by $y_t$:
\begin{equation}
y_t=\hat{P}_{t,\perp}^TM_t=\hat{P}_{t,\perp}^T L_t+\hat{P}_{t,\perp}^TS_t
\label{EquationRRPCP2}
\end{equation}
to solve $S_t$. The low-rank component is close to zero if $\hat{P_t}\approx P_t$, otherwise new directions are added. Furthermore, recent estimates of $L_t=A_t-S_t$ are stored and used to update $P_t$. Confirming the first results obtained in \cite{17-5}\cite{17-51}, a correctness result for ReProCS is given by Lois and Vawani \cite{17-6}. However, ReProCS requires the support $x_t$ to be fixed and quite small for a given support size $S_t$, but this does often not hold. So, ReProCS could not handle large outliers support sizes. \\
\item \textbf{Support-Predicted Modified-CS :} Qiu and Vaswani \cite{15} address the large outliers support sizes by using time-correlation of the outliers. This method called Support-Predicted Modified-CS RR-PCP \cite{15} and Support-Predicted Modified-CS \cite{16} is also an incremental algorithm and outperforms the ReProCS. However, this algorithm is only adapted for specific situation where there are only one or two moving objects that remain in scene. But, this is not applicable to real videos where multiple and time-varying number of objects can enter of leave the scene. Moreover, it requires knowledge of foreground motion. \\
\item \textbf{Automated Recursive Projected CS (A-ReProCS):} To address the limitation of the Support-Predicted Modified-CS, Qiu and Vaswani \cite{17} proposed a method called automated Recursive Projected CS (A-ReProCS) that ensures robustness when there are many nonzero foreground pixels, that is, there are many moving objects or large moving objects. Furthermore, A-ReProCS outperforms the previous incremental algorithms when foreground pixels are correlated spatially or temporally and when the foreground intensity is quite similar to the background one. \\
\item \textbf{ReProCS with cluster-PCA (ReProCS-cPCA):} To address the fact that the structure that we require is that $L_t$ is in a low dimensional subspace and the eigenvalues of its covariance matrix are "clustered", Qiu and Vaswani \cite{17-2}\cite{17-21} introduced a Recursive Projected Compressive Sensing with cluster-PCA (ReProCS-cPCA). Under mild assumptions,  ReProCS-cPCA with high probability can exactly recover the support set of $S_t$ at all times. Furthermore, the reconstruction errors of both $S_t$ and $L_t$ are upper bounded by a time-invariant and small value. \\
\item \textbf{Practical ReProCS (Prac-ReProCS) :} Guo et al. \cite{17-3}\cite{17-31}\cite{17-32}\cite{17-33}  designed a practically usable modification of the theoretical ReProCS algorithm. This practical ReProCS (Prac-ReProCS) requires much fewer parameters which can be set without any model knowledge and it exploits practically valid assumptions such as denseness for $L_t$, slow subspace change for $L_t$, and correlated support change of $S_t$. \\
\item \textbf{Incremental Low-Rank (iLR) Algorithm:} Wei et al. \cite{1007} proposed an incremental low-rank matrix
decomposition algorithm that maintains a clean background matrix adaptive to dynamic changes with both effectiveness and efficiency guarantees. Instead of a batch RPCA which requires a large number of video frames (usually 200 frames) for each time period, 15 frames only are required with iLR. iLR algorithm is about 9 times faster than a batch RPCA. \\
\item \textbf{Incremental PCP (incPCP):} Rodriguez and Wohlberg \cite{26-2}\cite{26-5} proposed an incremental PCP which processes one frame at a time. Obtaining similar results to batch PCP algorithms, it has an extremely low memory footprint and a computational complexity that allows real-time processing. Furthermore, incPCP is also able to quickly adapt to changes in the background. A Matlab-only implementation of this algorithm \cite{26-3} running on a standard laptop (Intel i7- 2670QM quad-core, 6GB RAM, 2.2 GHz) can process color videos of size $640\time480$ and $1920\time1088$ at a rate of $8$ and $1.5$ frames per second respectively. On the same hardware, an ANSI-C implementation \cite{26-3} can deliver a rate of $49.6$ and
$7.2$ frames per second for grayscale videos of size $640\time480$ and $1920\time1088$ respectively. This algorithm has real-time performance on GPU \cite{26-4}. Furthermore, Rodriguez and Wohlberg \cite{26-6} developed a translational and rotational jitter invariant incPCP which reach real-time performanc on GPU \cite{26-7}. This method was applied in automatic vehicle counting \cite{26-72}. \\
\end{enumerate}

\subsubsection{Methods for real time implementation of PCP}
Despite the efforts to reduce the time complexity, the corresponding algorithms have prohibitive computational time for real application such as foreground detection.  The main computation in PCP is the singular value decomposition of the large matrix $A$. Instead of computing a large SVD on the CPU, Anderson et al. \cite{12}\cite{13} proposed an implementation of the communication-avoiding QR (CAQR) factorization that runs entirely on the GPU. This implementation achieved $30\times$ speedup over the implementation on CPU using Intel Math Kernel Library (Intel MKL).
\\
In an other way, Pope et al. \cite{42} proposed a variety of methods that significantly reduce the computational time of the ALM algorithm. These methods can be classified as follows:
\begin{itemize}
\item \textbf{Reducing the computation time of SVD:} The computation of the SVD is reduced using the Power method \cite{42} that enables to compute the singular values in a sequential manner, and to stop the procedure when a singular value is smaller than a threshold. The use of the Power method by itself results in $4.32 \times$ lower runtime. Furthermore, the gain is improved by a factor of $2.02\times$ speedup if  the procedure is stopped when the singular value is smaller than the threshold. If the rank of $L$ is fixed and the Power SVD is stopped when the number of singular value is equal to $rank(L)$, the additional speedup is $17.35$.
\item \textbf{Seeding the PCP algorithm:} PCP operates on matrices consisting of blocks of contiguous frames acquired with a fixed camera. So, the low-rank matrix does not change significantly from one block to the next. Thus, Pope et al. \cite{42} use the low-rank component obtained by the ALM algorithm from the previous block as a starting point for the next block. This method allows an additional speedup of $7.73$.
\item \textbf{Partioning into subproblems:} Pope et al. \cite{42} proposed to partition the matrix $A$ into $P$ smaller submatrices. The idea is to combine the solutions of the $P$ corresponding PCP subproblems to recover the solution of the full matrix $A$ at lower computational complexity. 
\end{itemize}
\indent In this way, Pope et al. \cite{42} demonstrated that the PCP algorithm can be in fact suitable for real-time foreground/background separation for video-surveillance application using the corresponding hardware. \\
\indent In a similar manner,  Guo et al. \cite{1015} proposed a low-rank matrix recovery scheme, which splits the original RPCA into two small ones: a low-rank submatrix recovery and a low-rank submatrix reconstruction problems. This method showed a speedup of the ALM algorithm by more than $365$ times compared to a C implementation with less requirement of both time and space. In addition, this method significantly cuts the computational load for decomposing the remaining frames.

\subsubsection{Methods for finding the optimal PCP solution}
PCP recovers the true underlying low-rank matrix when a large portion of the measured matrix is either missing or arbitrarily corrupted. However, in the absence of a true underlying signal $L$ and the deviation
$S$, it is not clear how to choose a value of $\lambda$ that produces a good approximation of the given data $A$ for a given application.A typical approach would involve some cross-validation step to select $\lambda$ to maximize the final results of the application. The key problem with cross-validation in this case is that the best model is selected indirectly in terms of the final results, which can depend in unexpected ways on later stages in the data processing chain of the application. Instead, Ramirez and Sapiro \cite{36}\cite{37} addressed this issue via the Minimum Description Length (MDL) principle \cite{114} and so proposed a MDL-based low-rank model selection. The principle is to select the best low-rank approximation by using a direct measure on the intrinsic ability of the resulting model to capture the desired regularity from the data. To obtain the family of models $\mathbf{M}$ corresponding to all possible low-rank approximation of $A$, Ramirez and Sapiro \cite{36}\cite{37} applied the RPCA decomposition for a decreasing sequence of values of $\lambda$, $\left\{\lambda_t:t=1,2,3,...\right\}$ obtaining a corresponding sequence of decomposition $\left\{(L_t,S_t),t=1,2,3,...\right\}$. This sequence is obtained via a simple modification of the ALM algorithm \cite{18} to allow warm restarts, that is, where the initial ALM iterate for computing $(L_t,S_t)$ is $(L_{t-1},S_{t-1})$. Finally, Ramirez and Sapiro \cite{36}\cite{37} select the pair $(L_{\hat{t}},S_{\hat{t}})$,$\hat{t}=\underset{t}{\text{arg min}}\left\{MDL(L_t)+MDL(S_t)\right\}$ where $MDL(L_t)+MDL(S_t)=MDL(A|M)$ denoted the description length in bits of $A$ under the description provided by a given model $M \in \mathbf{M}$. Experimental results show that the best $\lambda$ is not the one determined by the theory in Candes et al. \cite{3}. \\
\indent An other approach developed by Gao et al. \cite{62} consist of two-pass RPCA process. The first-pass RPCA done on block resolution detect region with salient motion. Then, a saliency measure in each area is computed and permits to adapt the value of  $\lambda$ following the motion in the second-pass RPCA. Experimental results show that this block-sparse RPCA outperforms the original PCP \cite{3} and the ReProCS \cite{16}. In a similar way using a block-based RPCA, Biao and Lin \cite{1506} determined $\lambda$ with the affiliation of block to the class of "moving objects". Experimental results show that this approach gives better robustness on the I2R dataset \cite{203} than the single Gaussian, the MOG and the KDE.

\subsubsection{Modified-PCP}
\label{subsubsec:Modified-PCP}
In the literature, there are several modifications which concern the improvements of the original PCP and they can be classified as follows: \\

\begin{enumerate}
\item \textbf{Fixed rank: } Leow et al. \cite{78} proposed a fixed rank algorithm for solving background recovering problems because low-rank matrices have known ranks. The decomposition involves the same model than PCP in Equation \ref{EquationPCP1} but the minimization problem differs by the constraint as follows:
\begin{equation}
\underset{L,S}{\text{min}} ~~ ||L||_* + \lambda ||S||_{l_{1}}  ~~~ \text{subj} ~~~ rank(L)= r 
\label{EquationPCP11}
\end{equation}
with $r$ is the rank of the matrix $L$ and $r$ is known. Lai et al.  \cite{1507} developed an incremental fixed rank algorithm.\\
\item \textbf{Nuclear norm free: } A nother variant of PCP was formulated by Yuan et al. \cite{75} who proposed a nuclear-norm-free algorithm to avoid the SVD computation. The low-rank matrix is thus represented as $u1^T$ where $u \in \mathbf{R}^m$ and $1$ denotes the vector $\mathbf{R}^n$. Accordingly, a noiseless decomposition is formulated as follows:
\begin{equation}
A = u1^T+S
\label{EquationPCP12}
\end{equation}
Then, the corresponding minimization problem is the following one:
\begin{equation}
\underset{u}{\text{min}} ~~ ||A-u1^T||_{l_{1}}  
\label{EquationPCP13}
\end{equation}
Note the closed-form solution of Equation \ref{EquationPCP13} is given by the median of the entries of the matrix $A$. In other words, the background is extracted as the median at each pixel location of all frames of a surveillance video. As no iteration is required at all
to obtain the solution of \ref{EquationPCP13}, its computation for solving should be significantly cheaper than any iterative schemes for solving Equation \ref{EquationPCP3} numerically. Furthermore, this model extracts more accurately the background than the original PCP. Moreover, Yuan et al. \cite{75} developed a noise and a blur and noise nuclear-norm-free models SPCP which are detailed in Section \ref{subsubsec:Modified-SPCP}. In a similar way, Yang et al. \cite{1393} proposed a nonconvex model for background/foreground separation, that can incorporate both the nuclear-norm-free model and the use of nonconvex regularizers. \\
\item \textbf{Capped norms:} In an other way, Sun et al. \cite{82} presented a nonconvex formulation using the capped norms for matrices and vectors,
which are the surrogates of the rank function and the $l_0$-norm, and called capped nuclear norm and the capped $l_1$-norm, respectively. The minimization problem is formulated as follows:
\begin{equation}
\underset{L,S}{\text{min}} ~~ rank(L) + \lambda||S||_{l_{0}} ~~~ \text{subj} ~~~ ||A-L-S||_F^2\leq \sigma^2 
\label{EquationPCP14}
\end{equation}
where $\sigma^2$ is the level of Gaussian noise. 
The capped nuclear norm is then:
\begin{equation}
\frac{1}{\theta_1} \left[||L||_*+\sum_{i=1}^p max(\sigma_i(L)-\theta_1),0)\right]
\label{EquationPCP15}
\end{equation}
and the capped $l_1$-norm is formulated as follows:
\begin{equation}
\frac{1}{\theta_2} \left[||S||_{l_{1}}+\sum_{i=1}^p max(S_{ij})-\theta_2),0)\right] 
\label{EquationPCP16}
\end{equation}
for some small parameters $\theta_1$ and $\theta_2$. If all the singular values of $L$ are greater than $\theta_1$ and all the absolute values of elements in $S$ are greater than $\theta_2$, then the approximation will become equality. The smaller $\theta_1$ and $\theta_2$ are, the more accurate the capped norm approximation would be. The recovery precision is controled via $\theta_1$ and $\theta_2$. By carefully choosing
$\theta_1$ and $\theta_2$, $L$ and $S$ are more accurately determined than with the nuclear norm and the $l_1$-norm approximation. This capped formulation can be solved via two algorithms. One is based on the Difference of Convex functions (DC) framework and the other tries to solve the sub-problems via a greedy approach. Experimental results \cite{82} show better performance for the capped formulation of PCP than the original PCP \cite{3} and SPCP \cite{5} on the I2R dataset \cite{203}. \\
\item \textbf{Inductive approach: } Bao et al. \cite{67} proposed the following decomposition: 
\begin{equation}
A=PA+S
\label{EquationPCP17}
\end{equation}
where $P \in \mathbf{R}^{n \times n}$ is the low-rank projection matrix. The related optimization problem is formulated as follows:
\begin{equation}
\underset{P,S} {\text{min}} ~~ ||P||_* + \lambda ||S||_{l_{1}}  ~~~ \text{subj} ~~~ A-PA-S=0
\label{EquationPCP18}
\end{equation}
This is solved by IALM \cite{18}. Furthermore, Bao et al. \cite{67} developed an inductive version which requires less computational cost in processing new samples. \\
\item \textbf{Partial Subspace Knowledge:} Zhan and Vaswani \cite{17-4} proposed a modified-PCP with partial subspace knowledge. They supposed that a partial estimate of the column subspace of the low-rank matrix $L$ is available. This information is used to improve the PCP solution, i.e. allow recovery under weaker assumptions. So, the modified-PCP requires significantly weaker incoherence assumptions than PCP, when the available subspace knowledge is accurate. The corresponding optimization problem is written as follows: 
\begin{equation}
\underset{L,S} {\text{min}} ~~ ||L||_* + \lambda ||S||_{l_{1}}  ~~~ \text{subj} ~~~ L+P_{\Gamma^\perp}S=P_{\Gamma^\perp}A
\label{EquationPCP19}
\end{equation}
where $P_{\Gamma^\perp}$ is a projection matrix, $\Gamma$ is a linear space of matrices with column span equal to that of the columns of $S$, and $\Gamma^\perp$ is the orthogonal complement. Zhan and Vaswani \cite{17-4} applied with success their modified-PCP to the background-foreground separation problem, in which the subspace spanned by the background images is not fixed but changes over time and the changes are gradual. \\
\item  \textbf{Schatten-$p$,$l_q$-PCP ($p$,$q$-PCP):} The introduced norms by Candes et al. \cite{3} are not tight approximations, which may deviate the solution from the authentic one. Thus, Wang et al. \cite{1022} considered a non-convex relaxation which consists of a Schatten-$p$ norm and a $l_q$-norm with $0<p,q\leq1$ that strengthen the low-rank and sparsity, respectively. The Schatten-$p$ norm ($||.||_{S_p}$) is a popular non-convex surrogate of the rank function. Thus, the miminization problem is the following one:
\begin{equation}
\underset{L,S}{\text{min}} ~~ ||L||_{S_p}^p + \lambda ||S||_{l_q} ~~~ \text{subj} ~~~ A-L-S=0
\label{EquationPCP20}
\end{equation}
By replacing the Schatten-$p$ norm and a $l_q$-norm  by their expression, the miminization problem can be written as follows:
\begin{equation}
\underset{L,S}{\text{min}} ~~ \lambda_1 \sum_{i=1}^{min(m,n)} (\sigma_i(L))^p +  \lambda_2 \sum_{i=1}^{m}\sum_{j=1}^{n} |S_{ij}|^q
\label{EquationPCP21}
\end{equation}
where $\sigma_i(L)$ denotes the $i^{th}$ singular values of $L$. When $p=q=1$, $p$,$q$-PCP degenerates into the original convex PCP. Smaller values
of $p$ and $q$ help $p$,$q$-PCP to well approximate the original formulation of RPCA. The solver used is a Proximal Iteratively Reweighted Algorithm (PIRA) based on alternating direction method of multipliers, where in each iteration the underlying objective function is linearized to have a closed form solution. Experimental results \cite{1022} on the I2R dataset \cite{203} show better performance for $p$,$q$-PCP (in its stable formulation) in comparison to the original SPCP \cite{5} and SPCP solved via NSA \cite{34}. \\
\item  \textbf{Modified Schatten-$p$,$l_q$-PCP: } Shao et al. \cite{1023} proposed a similar approach than $p$,$q$-PCP \cite{1022} but they used the $L_q$-seminorm as a surrogate to the $l_1$-norm instead of the $l_q$-norm. Thus, the miminization problem is the following one:
\begin{equation}
\underset{L,S}{\text{min}} ~~ ||L||_{S_p}^p + \lambda ||S||_{L_q}^q ~~~ \text{subj} ~~~ A-L-S=0
\label{EquationPCP22}
\end{equation}
Furthermore, Shao et al. \cite{1023} used two different solvers based on the ALM and the APG methods as well as efficient root-finder strategies.  \\
\item  \textbf{Robust 2D-PCA:} To take into account the two-dimensional spatial information, Sun et al. \cite{1005} extracted a distinguished feature matrix for image representation, instead of matrix-to-vector conversion. Thus, the miminization problem is the following one:
\begin{equation}
\underset{U, V, S}{\text{min}} ~~  \lambda ||S||_{l_0} +\frac{1}{2}  ||A-U\Sigma V^T-S||_{F}^2 ~~~ \text{subj} ~~~ UU^T=I , VV^T=I
\label{EquationPCP23}
\end{equation}
where $U\Sigma V^T=L$ Different from $l_1$-norm relaxation,  Sun et al. \cite{1005} developed an iterative method to solve Equation (\ref{EquationPCP22}) efficiently via alternating optimization, by specific greedy algorithm for the $l_0$-norm regularization. So, a robust 2D-PCA model by sparse regularization is then solved via an alternating optimization algorithms. Results on dynamic backgrounds from the I2R dataset \cite{203} show the effectiveness of the Robust 2D-PCA (R2DPCA), compared with the conventional 2D-PCA \cite{3002} and PCP solved via IALM \cite{18}. \\
\item  \textbf{Rank-$N$ Soft Constraint:} Oh \cite{1011} proposed a RPCA with Rank-$N$ Soft Constraint (RNSC) based on the observation that the matrix $A$ should be rank $N$ without corruption and noise. Hence, the decomposition is formulated as estimating sparse error matrix and minimizing rank of low-rank matrix consisting of $N$ principal components associated to the $N$ largest singular largest values. Thus, the miminization problem with rank-$N$ soft constraint is the following one:
\begin{equation}
\underset{L,S}{\text{min}} ~~ \sum_{i=N+1}^{min(m,n)} ||\sigma_i(L)|| + \lambda ||S||_{l_1} ~~~ \text{subj} ~~~ A-L-S=0
\label{EquationPCP24}
\end{equation}
where $\sigma_i(L)$ represents the $i^{th}$ singular value of the low-rank matrix $L$, and $N$ is a constraint parameter for rank-$N$. Minimizing partial sum of singular values can minimize the rank of the matrix $L$ and satisfies the rank-$N$ constraint. Then, Oh \cite{1011} applied the RPCA with Rank-$1$ Soft Constraint on the edge images for moving objects detection under global illumination changes.In the case of moving camera, Ebadi and Izquierdo \cite{1522}\cite{1523} proposed a SVD-free algorithm to solve Rank-$1$ RPCA that achieved more than
double the amount of speed-up in computation time for the same performance target compared to its counterpart with SVD. This approach \cite{1520} can handle camera movement, various foreground object sizes, and slow-moving foreground pixels as well as sudden and gradual illumination changes.\\
\item  \textbf{JVFSD-RPCA:}  Wen et al. \cite{1026} reconstructed the input video data and aimed to make the foreground pixels not only sparse in space
but also sparse in "time" by using a Joint Video Frame Set Division and RPCA-based (JVFSD-RPCA) method. In addition, they used the motion as a priori knowledge. The proposed method consists of two phases. In the first phase, Lower Boundbased Within-Class Maximum Division (LBWCMD) method divided the video frame set into several subsets. In this way, the successive frames are assigned to different subsets in which
the foregrounds are located at the scene randomly. In the second phase, each subset with the frames are increased with a small quantity of motion. This method shows robustness in the case of dynamic backgrounds. \\
\item  \textbf{NSMP/WNSMP:} Wang and Feng \cite{1034} improved the RPCA method to find a new model to separate the background and foreground, and to reflect the correlation between them as well. For this, they proposed a "low-rank + dual" model and they used the reweighted dual function norm instead of the normal norms so as to get a better and faster model. So, the original minimization problem is improved by a nuclear norm and spectral norm minimization problem (NSMP). Thus, the minimization problem with the dual norm is the following one:
\begin{equation}
\underset{L,S}{\text{min}} ~~ \lambda ||L||_* + \mu||S||_2 ~~~ \text{subj} ~~~ A-L-S=0
\label{EquationPCP25}
\end{equation}
where the spectral norm $||.||_2$ is the dual norm of the nuclear norm, and it corresponds to the largest singular value of the matrix \cite{1034}. 
As the nuclear norm regularized is not a perfect approximation of the rank function, Wang and Feng \cite{1034} proposed a weighted function nuclear norm and spectral norm minimization problem (WNSMP) with the corresponding minimization problem:
\begin{equation}
\underset{L,S}{\text{min}} ~~ \lambda ||\omega(L)||_* + \mu||\omega^{-1}(S)||_2 ~~~ \text{subj} ~~~ A-L-S=0
\label{EquationPCP26}
\end{equation}
where $\omega()$ denotes the weighted function which directly adds the weights onto the singular values of the matrix, and, for any matrix $X$, weighted function norm is defined as follows: $||\omega(X)||_*=\sum_{i=1}^{min(m,n)} \omega_i \sigma_i(X)$ and  $||\omega^{-1}(X)||_2= \underset{i}{\text{max}} \frac{1}{\omega_i} \sigma_i(X)$w where $\sigma_i(X)$ represents the $i^{th}$ singular value of the matrix $X$. Although this minimization problem with the weighted function nuclear norm is nonconvex, fortunately it has a closed form solution due to the special choice of the value of weights, and it is also a better approximation to the rank function. NSMP and WNSMP show more robustness on the I2R dataset \cite{203} than RPCA solved IALM \cite{3} and GoDec \cite{6}. \\
\item  \textbf{Implicit Regularizers:} He et al. \cite{87} proposed a robust framework for low-rank matrix recovery via implicit regularizers of robust $M$-estimators (Huber, Welsch, $l_1$-$l_2$) and their minimizer functions. Based on the additive form of halfquadratic optimization, proximity operators of implicit regularizers are developed such that both low-rank structure and corrupted errors can be alternately recovered. Thus, the minimization problem with implicit regularizers is formulated as follows:
\begin{equation}
\underset{L,S}{\text{min}} ~~ \lambda ||L||_* + \varphi(S) ~~~ \text{subj} ~~~ A-L-S=0
\label{EquationPCP27}
\end{equation}
where the implicit regularizer $\varphi(y)$ is defined as the dual potential function of a robust loss function $\phi(x)$ where $\phi(x)=\underset{y}{\text{min}}\frac{1}{2} ||x-y||_2^2 + \varphi(x)$. If $\phi(x)$ is Huber $M$-estimator, the
implicit regularizer $\varphi(y)$ becomes $\mu\lambda ||(L)||_{l_{1}}$. When the $M$-estimator $\phi(x)$  is Welsch $M$-estimator, the minimization
problem becomes the sample based maximum correntropy problem. Compared with the mean square error, the model in Equation (\ref{EquationPCP25}) is more robust to outliers due to $M$-estimation. Experimental results \cite{87} on the I2R dataset \cite{203} show that the Welsch $M$-estimator outperforms the Huber-estimator and the $l_1$-$l_2$-estimator. \\
\end{enumerate}

\subsection{RPCA via Stable Principal Component Pursuit}
\label{subsec:SPCP}
PCP is limited to the low-rank component being exactly low-rank and the sparse component being exactly sparse but the observations in real applications are often corrupted by noise affecting every entry of the data matrix. Therefore, Zhou et al. \cite{5} proposed a stable PCP (SPCP) that guarantee stable and accurate recovery in the presence of entry-wise noise. In the following sub-sections, we reviewed this method and all these modifications in terms of decomposition, solvers, incremental algorithms and real time implementations. Table \ref{SPCPOverview} shows an overview of the Stable Principal Component Pursuit methods and their key characteristics.  

\begin{table*}
\scalebox{0.70}{
\begin{tabular}{|l|l|l|} 
\hline
\scriptsize{Stable PCP} &\scriptsize{Categories} &\scriptsize{Authors - Dates} \\
\hline
\hline
\scriptsize{Decompositions} &\scriptsize{1) Original SPCP}                       					&\scriptsize{Zhou et al. (2010) \cite{5}}               \\
\scriptsize{} &\scriptsize{2) Modified-SPCP (Bilateral Projection)}             					&\scriptsize{Zhou and Tao (2013) \cite{601}}            \\
\scriptsize{} &\scriptsize{3) Modified-SPCP (Nuclear-Norm Free)}                 				  &\scriptsize{Yuan et al. (2013) \cite{75}}              \\
\scriptsize{} &\scriptsize{4) Modified-SPCP (Nuclear-Norm Free for blur in noisy video)}  &\scriptsize{Yuan et al. (2013) \cite{75}}              \\
\scriptsize{} &\scriptsize{5) Modified-SPCP (Undercomplete Dictionary)}                   &\scriptsize{Sprechman et al. (2012) \cite{68}}         \\
\scriptsize{} &\scriptsize{6) Variational SPCP (Huber Penalty)}                           &\scriptsize{Aravkin et al. (2014) \cite{1029}}         \\
\scriptsize{} &\scriptsize{7) Three Term Low-rank Optimization (TTLO)}                    &\scriptsize{Oreifej et al. (2012)  \cite{1031}}        \\
\scriptsize{} &\scriptsize{8) Inequality-Constrained (RPCA)}                              &\scriptsize{Li et al. (2015)  \cite{1360}}             \\
\scriptsize{} &\scriptsize{9) Double-noise-dual-problem (DNDP)}                           &\scriptsize{Cheng et al. (2015)  \cite{1550}}          \\

\hline
\scriptsize{Solvers} &\scriptsize{Alternating Splitting Augmented Lagrangian method (ASALM)}         &\scriptsize{Tao and Yuan (2011) \cite{35}} \\
\scriptsize{} &\scriptsize{Variational ASALM (VASALM)}                                 &\scriptsize{Tao and Yuan (2011) \cite{35}}       \\
\scriptsize{} &\scriptsize{Parallel Splitting ALM (PSALM)}                                     &\scriptsize{Tao and Yuan (2011) \cite{35}}       \\
\scriptsize{} &\scriptsize{Non Smooth Augmented Lagrangian Algorithm (NSA)}            &\scriptsize{Aybat et al. (2011) \cite{34}}       \\
\scriptsize{} &\scriptsize{First-order Augmented Lagrangian algorithm for Composite norms (FALC)} &\scriptsize{Aybat et al. (2012) \cite{34-1}}     \\
\scriptsize{} &\scriptsize{Augmented Lagragian method for Conic Convext (ALCC)}                   &\scriptsize{Aybat et al. (2012) \cite{34-2}}     \\
\scriptsize{} &\scriptsize{Partially Smooth Proximal Gradient (PSPG)}                  &\scriptsize{Aybat et al. (2012) \cite{34-3}}             \\
\scriptsize{} &\scriptsize{Alternating Direction Method - Increasing Penalty (ADMIP)}  &\scriptsize{Aybat et al. (2012) \cite{34-4}}             \\
\scriptsize{} &\scriptsize{Greedy Bilateral Smoothing (GreBsmo)}                       &\scriptsize{Zhou and Tao (2013) \cite{601}}              \\
\scriptsize{} &\scriptsize{Bilinear Generalized Approximate Message Passing (BiG-AMP)} &\scriptsize{Parker and Schniter (2012) \cite{80}}        \\
\scriptsize{} &\scriptsize{Inexact Alternating Minimization - Matrix Manifolds (IAM-MM)}&\scriptsize{Hinterm\"{u}ller and Wu (2014) \cite{1036}} \\
\scriptsize{} &\scriptsize{Customized Proximal Point Algorithm (CPPA)}                  &\scriptsize{Huai et al. (2015) \cite{1073}}             \\
\scriptsize{} &\scriptsize{multi-block Bregman (BADMM)}                                 &\scriptsize{Wang et al. (2015) \cite{1076}}             \\
\scriptsize{} &\scriptsize{Partially Parallel Splitting - Multiple Block (PPS-MB)}      &\scriptsize{Hou et al. (2015) \cite{1394}}              \\
\scriptsize{} &\scriptsize{Local Convex Relaxation (LCR)}                               &\scriptsize{Mao and Zhang (2016) \cite{1560}}           \\  
\scriptsize{} &\scriptsize{Distributed Douglas-Rachford splitting method (DDRSM)}       &\scriptsize{He and Han (2016) \cite{1561}}              \\ 
\scriptsize{} &\scriptsize{Twisted ADMM (TADMM)}  																	    &\scriptsize{Wang and Song (2016) \cite{1562}}           \\ 
\scriptsize{} &\scriptsize{Dual Smoothing (DS)}																          &\scriptsize{Aravkin and Becker (2016) \cite{1563}}      \\ 
\hline
\scriptsize{Compressive Sensing Algorithms} &\scriptsize{Frank-Wolfe-Thresholding}  &\scriptsize{Mu et al. (2014) \cite{34-5}} \\
\hline
\scriptsize{Incremental Algorithms} &\scriptsize{Fast Trainable Encoders}  &\scriptsize{Sprechman et al. (2012) \cite{68}} \\
\hline
\scriptsize{Real time Implementations} &\scriptsize{DFC-PROJ}           &\scriptsize{Mackey et al. (2011) \cite{53}} \\
\scriptsize{}          &\scriptsize{DFC-PROJ-ENS}       &\scriptsize{Mackey et al. (2011) \cite{53}} \\
\scriptsize{}          &\scriptsize{DFC-NYS}            &\scriptsize{Mackey et al. (2011) \cite{53}} \\
\scriptsize{}          &\scriptsize{DFC-NYS-ENS}        &\scriptsize{Mackey et al. (2011) \cite{53}} \\
\hline
\end{tabular}}
\caption{Stable Principal Component Pursuit: A Complete Overview. The first column indicates the concerned category and the second column the name of each method. Their corresponding acronym is indicated in the first parenthesis. The third column gives the name of the authors and the date of the related publication.} \centering
\label{SPCPOverview}
\end{table*}

\subsubsection{Stable Principal Component Pursuit}
\label{subsubsec:SPCP}
Zhou et al. \cite{5} proposed a stable PCP (SPCP) which assumes that the observation matrix $A$ is represented as follows: 
\begin{equation}
A = L + S + E
\label{EquationSPCP1}
\end{equation}
where $E$ is a noise term (say i.i.d. noise on each entry of the matrix) and $||E||_F<\delta$ for some $\delta>0$.
To recover $L$ and $S$, Zhou et al. \cite{5} proposed to solve the following optimization problem, as a relaxed version to PCP:
\begin{equation}
\underset{L,S}{\text{min}} ~~ ||L||_* + \lambda ||S||_{l_{1}}  ~~~ \text{subj} ~~~||A-L-S||_F<\delta
\label{EquationSPCP2}
\end{equation}
where  $||.||_F$ is the Frobenius norm and $\lambda = \frac{1}{\sqrt{n}}$. 

\subsubsection{Algorithms for solving SPCP}
\label{subsubsec:Algorithms-SPCP}
Like in Equation (\ref{EquationPCP2}) for PCP, Tao and Yuan \cite{35} showed that an easy reformulation of the constrained convex programming for Equation (\ref{EquationSPCP2}) falls perfectly in the applicable scope of the classical ALM. Moreover, the favorable separable structure emerging in both the objective function and the constraints entails the idea of splitting the corresponding augmented Lagrangian function to derive efficient numerical algorithms. So, Tao and Yuan \cite{35} developed the alternating splitting augmented Lagrangian method (ASALM) and its variant (VASALM), and the parallel splitting augmented Lagrangian method (PSALM) for solving Equation (\ref{EquationSPCP2}). ASALM and its variants converge to an optimal solution. However, ASALM iterations are too slow for real time application and its complexity is not known. To address this problem, Aybat et al. \cite{34} studied how fast first-order methods can be applied to SPCP with low complexity iterations and showed that the subproblems that arise when applying optimal gradient methods of Nesterov, alternating linearization methods and alternating direction augmented Lagrangian methods to the SPCP problem either have closed-form solutions or have solutions that can be obtained with very modest effort. Furthermore, Aybat et al. \cite{34} developed a new first order algorithm called Non Smooth augmented Lagrangian Algorithm (NSA), based on partial variable splitting. All but one of the methods analyzed require at least one of the non-smooth terms in the objective function to be smoothed and obtain an $\epsilon$-optimal solution to the SPCP problem in $O(\sqrt{\epsilon})$ iterations. NSA, which works directly with the fully non-smooth objective function, is proved to be convergent under mild conditions on the sequence of parameters it uses. NSA, although its complexity is not known, is the fastest among the optimal gradient methods, alternating linearization methods and alternating direction augmented Lagrangian methods algorithms and substantially outperforms ASALM. In a similar way, Aybat et al. \cite{34-3} proposed a proximal gradient algorithm called Partially Smooth Proximal Gradient (PSPG). Experimental results show that both the number of partial SVDs and the CPU time of PSPG are significantly less than those for NSA and ASALM. An overview of these algorithms as well as their complexity can be seen in Table \ref{A-TPCP2Overview}.

\subsubsection{Methods for real time implementation of SPCP}
\label{subsubsec:RTI-SPCP}
Mackey et al. \cite{53} proposed a real time implementation framework, entitled Divide-Factor-Combine (DFC). DFC randomly divides the original matrix factorization task into cheaper subproblems, solves those subproblems in parallel using any base matrix factorization (MF) algorithm, and combines the solutions to the subproblem using an efficient technique from randomized matrix approximation. The inherent parallelism of DFC allows for near-linear to superlinear speedups in practice, while the theory provides high-probability recovery guarantees for DFC comparable to those provided by its base algorithm. So, Mackey et al. \cite{53} proposed two algorithms called DFC-PROJ and DFC-NYS, that differ from the method used to divide the original matrix. DFC-PROJ randomly partitions the orthogonal projection of the matrix $A$ onto the $t$ $l$-column submatrices $C_1,...,C_t$ by using a column sampling method, while DFC-NYS selects an $l$-column submatrix $C$ and an $d$-row submatrix $R$ using the generalized Nystr\"{o}m method. DFC significantly reduces the per-iteration complexity to $O(mlr_{C_1})$ where $r_{C_1}$ is the rank of the matrix $C_1$ for the DFC-PROJ. The cost of combining the submatrix estimates is even smaller, since the outputs of standard MF algorithms are returned in factored form. Indeed, the column projection step of DFC-PROJ requires only $O(mr^2 + lr^2)$ time for $r=max_i k_{C_i}$, $O(mr^2 + lr^2)$ time for the pseudoinversion of $C_i$ and $O(mr^2 + lr^2)$ time for matrix multiplication with each $C_i$ in parallel. For the DFC-NYS, the per-iteration complexity $O(mlr_{C})$ where $r_{C}$ is the rank of the matrix $C$ and $O(mlr_{R})$ where $r_{R}$ is the rank of the matrix $R$. The cost of combining requires $O(l\bar{r}^2+d\bar{r}^2+min(m,n)\bar{r}^2)$ time where $\bar{r}=max(r_C,r_R)$. Mackey et al. \cite{53} improved these real time implementations by using ensemble methods that improve performance of matrix approximation algorithms, while straightforwardly leveraging the parallelism of modern many-core and distributed architectures \cite{110}. As such, an set variants of the DFC algorithms have been developed reducing recovery error while introducing a negligible cost to the parallel running time. For DFC-PROJ-ENS, rather than projecting only onto the column space of $C_1$, the  projection of $C_1,...,C_t$ is done onto the column space of each $C_i$ in parallel and then average the $t$ resulting low-rank approximations. For DFC-NYS-ENS, a random $d$-row submatrix is chosen like in DFC-NYS and independently partition the columns of the matrix in $l$ like in DFC-PROJ. After running the base MF algorithm on each submatrix, the generalized Nystr\"{o}m method is applied to each pair of matrices in parallel and then the $t$ resulting low-rank approximations is obtained by average.

\subsubsection{Modified-SPCP}
\label{subsubsec:Modified-SPCP}
In the literature, there are several modifications which concern the original SPCP and they can be classified as follows: \\
\begin{enumerate}
\item \textbf{Bilateral factorization: } Zhou and Tao \cite{601} proposed a noisy robust PCA by replacing $L$ with its bilateral factorization $L=UV$ and regularizing the $l_1 norm$ of $S$'s entries. The corresponding minimization problem is then formulated as follows:
\begin{equation}
\underset{U, V, S}{\text{min}} ~~ \lambda||S||_{l_{1}} +  ||A-UV-S||_F^2 \nonumber
\end{equation}
\begin{equation}
 ~~~ \text{subj} ~~~rank(U) = rank(V) \leq r
\label{EquationSPCP11}
\end{equation}
The $l_1$ regularization permits soft-thresholding in updating $S$. Zhou and Tao \cite{601} solved this minimization problem using a Greedy Bilateral Smoothing algorithm (GreBsmo). GreBsmo considerably speed up the decomposition and performs 30-100
times faster than most existing algorithms such as IALM \cite{18}. \\
\item \textbf{Nuclear norm free: } Other variants of SPCP were given by Yuan et al. \cite{75} who developed a nuclear-norm-free algorithm to avoid the SVD computation in SPCP. The low-rank matrix is thus represented as $u1^T$ where $u \in \mathbf{R}^m$ and $1$ denotes the vector $\mathbf{R}^n$. Accordingly, a noiseless decomposition is formulated as follows:
\begin{equation}
A = u1^T+S+E
\label{EquationSPCP12}
\end{equation}
Then, the corresponding minimization problem is the following one:
\begin{equation}
\underset{S \in \mathbf{R}^{m\times n}, u \in \mathbf{R}^m}{\text{min}} ~~ ||S||_{l_{1}} + \frac{\mu}{2}  ||A-u1^T-S||_F^2 \nonumber
\end{equation}
\begin{equation}
~~~ \text{subj} ~~~rank(u1^T) = 1
\label{EquationSPCP13}
\end{equation}
where $\mu$ is a penalty parameter. This model has no closed-form solution and need to be solved iteratively. This model extracts more accurately the background than the original SPCP. An other advantage of this model against the original SPCP is that it has only one parameter
in the minimization function to be tuned in the implementation. Considering, that there might be a blur in a noisy video surveillance video, Yuan et al. \cite{75} extended the model developed in Equation \ref{EquationSPCP13} to:
\begin{equation}
\underset{S \in \mathbf{R}^{m\times n}, u \in \mathbf{R}^m}{\text{min}}{\text{min}} ~~ ||S||_{l_{1}} + \frac{\mu}{2}  ||A+H(U1^T+S)||_F^2  \nonumber
\end{equation}
\begin{equation}
~~~ \text{subj} ~~~ rank(u1^T) = 1
\label{EquationSPCP14}
\end{equation}
where $H$ is the matrix representation of a regular blurring operator. The blur is assumed to appear in a frame-wise way. In this blur configuration, this nuclear-norm-free model extracts more robustly the background than the original SPCP with the blur option. \\
\item \textbf{Under-complete dictionary: } Considering a dictionary approach, Sprechman et al.\cite{68} proposed the following decomposition:
\begin{equation}
A=UT+S+E
\end{equation}
where $U \in \mathbf{R}^{m \times r}$, $T \in \mathbf{R}^{r \times n}$, and $S \in \mathbf{R}^{m \times n}$. Then, the corresponding minimization problem is as follows:
\begin{equation}
\underset{U, T, S}{\text{min}} ~~  \lambda ||S||_{l_{1}} + \frac{\lambda_1}{2} (||U||_F^2 + ||T||_F^2) + \frac{1}{2} ||A-UT-S)||_F^2 \nonumber
\end{equation} 
\begin{equation}
 ~~~ \text{subj} ~~~ rank(UT) \leq r
\label{EquationSPCP15}
\end{equation} 
The low-rank component can be considered as an under-complete dictionary $U$, with $r$ atoms, multiplied by a matrix $T$
containing in its columns the corresponding coefficients for each data vector in $A$. This interpretation brings the SPCP problem close to that of dictionary learning in the sparse modeling domain. This is solved via an alternating minimization problem. Furthermore, Sprechman et al.\cite{68} developed an online version of their SPCP via fast trainable encoders. \\
\item \textbf{Variational formulation:} Aravkin et al. \cite{1029} proposed a convex variational framework which is accelerated with quasi-Newton methods. The corresponding minimization problem is then formulated as follows:
\begin{equation}
\text{min}  ~~ \Phi(L,S)  \nonumber  ~~~ \text{subj} ~~~ \rho(L+S-A) \leq \epsilon
\label{EquationSPCP16}
\end{equation}
where in classical formulation $\Phi(L,S)= ||L||_* + \lambda ||S||_{l_{1}}$ and $\rho$ is assumed to be the Frobenius norm. As this restriction is not necessary, Aravkin et al. \cite{1029} considered $\rho$ to be smooth and convex. Then, $\rho$ is taken to be the robust Huber penalty. This approach offers advantages over the original SPCP formulation in terms of scalability and practical parameter selection. \\
\item  \textbf{Three Term Low-rank Optimization:} Oreifej et al. \cite{1031} proposed a three term decomposition for video stabilization and moving object detection in turbulence as follows:
\begin{equation}
\underset{A,L,S}{\text{min}} ~~ \lambda ||L||_* + \lambda_1 ||f_{\pi}(S)|| + \lambda_2 ||E||_F^2 ~~~ \text{subj} ~~~ A=L+S+E
\label{EquationSPCP17} 
\end{equation}  \\
where the frames of the sequence are stacked in the matrix $A$. Thus, low-rank matrix $L$ corresponds to the background, the sparse matrix $E$ corresponds to the moving object and the dense error matrix $E$ corresponds to the turbulence. The turbulence causes dense and Gaussian noise, and therefore can be captured by Frobenius norm. Therefore, Oreifej et al. \cite{1031} enforced an additional constraint on the objects with $f_{\pi}(.)$ which is the object confidence map, which is a linear operator that weights the entries of $S$ according to their confidence of corresponding to a moving object such that the most probable elements are unchanged and the least are set to zero. Oreifej et al. \cite{1031} used the solver IALM \cite{18} to solve Equation \ref{EquationSPCP17}. Experimental results \cite{1031} show that this decomposition outperforms  Mixture of Gaussians (MoG) \cite{700}, Kernel Density Estimation \cite{710} and PCA \cite{310}. Furthermore, the code called ThreeWayDec\protect\footnotemark[32] is provided.
\end{enumerate}

\footnotetext[32]{{http://www.vision.eecs.ucf.edu/projects/Turbulence/}}

\subsection{RPCA via Quantization based Principal Component Pursuit}
\label{subsec:TFOCS}
Becker et al. \cite{8} proposed a inequality constrained version of RPCA proposed by Candes et al. \cite{3} to take into account the quantization error of the pixel's value. Indeed, each pixel has a value between $0,1,2,\ldots,255$. This value is the quantized version of the real value which is between $[0,255]$. So, the idea is to apply RPCA to the real observations instead of applying it to the quantized ones. Indeed, it is unlikely that the quantized observation can be split nicely into a low-rank and sparse component. So, Becker et al. \cite{8} supposed that $L+S$ is not exactly equal to $A$, but rather that $L+S$ agrees with $A$ up to the precision of the quantization. The quantization can induce at most an error of 0.5 in the pixel value. This measurement model assumes that the observation matrix $A$ is represented as follows: 
\begin{equation}
A=L+S+Q
\label{EquationTFOCS1}
\end{equation}
where $Q$ is the error of the quantization. Then, the  objective function is the same than the equality version in Equation (\ref{EquationPCP3}), but instead of the constraints $L+S=A$, the constraints are $\|A-L-S\|_{l_{\infty}} \le 0.5$. So, the quantization based PCP (QPCP) is formulated as follows:
\begin{equation}
\underset{L,S}{\text{min}} ~~ ||L||_* + \lambda ||S||_{l_{1}}  ~~~ \text{subj} ~~~ \|A-L-S\|_{l_{\infty}} \le 0.5
\label{EquationTFOCS2}
\end{equation}
The $l_\infty$-norm allows to capture the quantization error of the observed value of the pixel. 
\newline
\newline
\noindent \textbf{Algorithms for solving QPCP:}
Becker et al. \cite{8} used a general framework for solving this convex cone problem called Templates for First-Order Conic Solvers  (TFOCS). First, this approach determines a conic formulation of the problem and then its dual. Then, Becker et al. \cite{8} applied smoothing and solved the problem using an optimal first-order method. This approach allows to solve the problem in compressed sensing.

\subsection{RPCA via Block based Principal Component Pursuit}
\label{sec:BPCP}
Tang and Nehorai \cite{41} proposed a block based PCP (BPCP) that enforces the low-rankness of one part and the block sparsity of the other part. This decomposition involves the same model than PCP in Equation (\ref{EquationPCP1}), that is $A=L+S$,  where $L$ is the low-rank component but $S$ is a block-sparse component. The low-rank matrix $L$ and the block-sparsity matrix $S$ can be recovered by the following optimization problem \cite{6}:
\begin{equation}
\underset{L,S}{\text{min}} ~~ ||L||_* + \kappa (1-\lambda) ||L||_{l_{2,1}} + \kappa \lambda||S||_{l_{2,1}}
\nonumber
\end{equation}
\begin{equation}
~~~ \text{subj} ~~~ A-L-S=0
\label{EquationBPCP1}
\end{equation}
where $||.||_*$  and $||.||_{l_{2,1}}$ are the nuclear norm and the $l_{2,1}$-norm, respectively. The $l_{2,1}$-norm corresponds to the $l^1$-norm of the vector formed by taking the $l^2$-norms of the columns of the underlying matrix. 
The term $\kappa(1-\lambda)||L||_{l_{2,1}}$ ensures that the recovered matrix $L$ has exact zero columns corresponding to the outliers. In order to eliminate ambiguity, the columns of the low-rank matrix $L$ corresponding to the outlier columns are assumed to be zeros. \\

\indent This approach also called RPCA-LBD was evaluated in background/foreground separation in Guyon et al. \cite{501} and gives better results than the original RPCA-PCP \cite{3}. But RCPA-LBD made the assumption that the matrix $S$ contains mostly zero columns, with several non-zero ones
corresponding toforeground  elements. This assumption cannot be made in background/foreground separation because the columns of the matrix $S$ are assumed to correspond to foreground objects in the frames. Moreover, assuming that most columns are zero
contradicts the definition of sparse matrix. Indeed, when a whole column in the sparse matrix is zero, it means the information in that column is
assigned to the low-rank subspace. Furthermore, if the video sequence contains foreground objects in all the frames then this assumption
does not help. Instead, it is more suitable if the  block-sparsity was imposed on the pixels of each video frame like in RPCA-BS \cite{1520}\cite{1522}\cite{1523} rather than a whole column in the matrix $S$. \\

\noindent \textbf{Algorithm for solving BPCP:}
Tang and Nehorai \cite{41} designed an efficient algorithm to solve the convex problem in Equation (\ref{EquationBPCP1}) based on the ALM method. This algorithm decomposed the matrix $A$ in a low-rank and block-sparse matrices in respect to the $||.||_{l_{2,1}}$ and the extra term $\kappa (1-\lambda) ||L||_{l_{2,1}}$. 

\subsection{RPCA via Local Principal Component Pursuit}
\label{sec:LPCP}
PCP is highly effective but the underlying model is not appropriate when the data are not modeled well by a single low-dimensional subspace. Wohlberg et al. \cite{26} proposed a decomposition corresponding to a more general underlying model consisting of a union of low-dimensional subspaces.
\begin{equation}
A=AU+S
\label{EquationLPCP1}
\end{equation}
This idea can be implemented as the following problem:
\begin{equation}
\underset{U,S}{\text{min}} ~~ |U||_{l_{1}} + \alpha ||U||_{l_{2,1}} + \beta ||S||_{l_{1}}  ~~~ \text{subj} ~~~ A-AU-S=0
\label{EquationLPCP2}
\end{equation}
The explicit notion of low-rank, and its nuclear norm proxy, is replaced by representability of a matrix as a sparse representation on itself. The $l_{2,1}$-norm encourages rows of $U$ to be zero, but does not discourage nonzero values among the entries of a nonzero row. The $l_1$-norm encourages zero values within each nonzero row of $S$.
\\To better handle noisy data, Wohlberg et al. \cite{26} modified Equation (\ref{EquationLPCP2}) with a penalized form and added a Total Variation penalty on the sparse deviations for contiguous regions as follows:
\begin{equation}
\underset{U,S}{\text{min}} ~~ \frac{1}{2} ||A-DU-S||_{l_2}^2 + \alpha||U||_{l_{1}}
\nonumber
\end{equation}
\begin{equation}
+ \beta ||U||_{l_{2,1}} + \gamma ||S||_{l_{1}} + \delta ||grad(S)||_{l^{1}}  
\label{EquationLPCP3}
\end{equation}
where the dictionary $D$ is derived from the data $A$ by mean subtraction and scaling, and  $grad(S)$ is a vector valued discretization of the 3D gradient of $S$. An appropriate sparse $U$ can be viewed as generating a locally low-dimensional approximation $DU$ of $A-S$. When the dictionary is simply the data (i.e., $D=A$), the sparse deviations (or outliers) $S$ are also the deviations of the dictionary
$D$, so constructing the locally low-dimensional approximation as $(D-S)U$, implying an adaptive dictionary $D-S$, should allow $U$ to be even sparser.
\newline 
\newline
\noindent \textbf{Algorithm for solving LPCP:} Wohlberg et al. \cite{26} proposed to solve Equation (\ref{EquationLPCP2}) using the Split Bregman Algorithm (SBA) \cite{106}. Adding terms relaxing the equality constraints of each quantity and its auxiliary variable, Wohlberg et al. \cite{26} introduced Bregman variables in Equation (\ref{EquationLPCP2}). So, the problem is split into an alternating minimization of five subproblems. Two subproblems are $l_2$ problems that are solved by techniques for solving linear systems such as conjugate gradient. The other three subproblems are solved very cheaply using shrinkage, i.e. generalized shrinkage and soft shrinkage.

\subsection{RPCA via Outlier Pursuit}
\label{sec:OP}
Xu et al. \cite{56} proposed a robust PCA  via Outlier Pursuit (OP) to obtain a robust decomposition when the outliers corrupted entire columns, that is every entry is corrupted in some columns. This method involves the nuclear norm minimization and recover the correct column space of the uncorrupted matrix, rather than the exact matrix itself. The decomposition involves the same model than PCP in Equation (\ref{EquationPCP1}), that is $A=L+S$. A straightforward formulation to minimize the energy function can be written as follows:
\begin{equation}
\underset{L,S}{\text{min}} ~~ rank(L) + \lambda ||S||_{0,c}  ~~~ \text{subj} ~~~ A-L-S=0
\label{EquationOP1}
\end{equation}
where $||S||_{0,c}$ stands for the number of nonzero columns of a matrix, and it is equivalent to $||S||_{l_{2,0}}$ which corresponds to the number of non-zero columns too \cite{1050}. $\lambda>0$ is an arbitrary balanced parameter. But this problem is $NP$-hard, typical solution might involve a search with combinatorial complexity. This research seeks to solve for $L$ with the following optimization problem:
\begin{equation}
\underset{L,S}{\text{min}} ~~ ||L||_* + \lambda ||S||_{l_{1,2}}  ~~~ \text{subj} ~~~ A-L-S=0
\label{EquationOP2}
\end{equation}
where $||.||_*$  and $||.||_{l_{1,2}}$ are the nuclear norm and the $l_{1,2}$-norm, respectively. The $l_{1,2}$-norm corresponds to the $l_2$-norm of the vector formed by taking the $l_1$-norms of the columns of the underlying matrix. $\lambda>0$ is an arbitrary balanced parameter. Adapting the OP algorithm to the noisy case,that is $A=L+S+E$, Xu et al. \cite{56}  proposed a robust PCA via Stable Outlier Pursuit (SOP):
\begin{equation}
\underset{L,S}{\text{min}} ~~ ||L||_* + \lambda ||S||_{l_{1,2}}  ~~~ \text{subj} ~~~||A-L-S||_F<\delta
\label{EquationOP3}
\end{equation}
where $S$ is supported on at most $\gamma n$ columns and $\lambda = \frac{3}{7\sqrt{\gamma n}}$.
\newline
\newline
\textbf{Algorithm for solving OP and SOP:} Xu et al. \cite{56} used the Singular Value Threshold (SVT) algorithm to solve these two minimization problems.

\subsection{RPCA with Sparsity Control}
\label{sec:SpaCtrl}
Mateos and Giannakis \cite{9}\cite{10} proposed a robust PCA using a bilinear decomposition with Sparsity Control (RPCA-SpaCtrl). The decomposition involves the following model:
\begin{equation}
A=M+PU^T+S+E
\label{EquationSpaCtrl1}
\end{equation}
where $M$ is the mean matrix, the matrix $U$ has orthogonal columns, $P$ are the principal components matrix, $S$ is the outliers matrix and $E$ is a zero-mean matrix. The percentage of outliers determines the degree of sparsity in $S$. The criterion for controlling outlier sparsity is seek to the relaxed estimation:
\begin{equation}
\underset{U,S}{\text{min}} ~~ ||X+1_N M^{T} -P U^{T}-S||_F^2 + \lambda ||S||_{l_{2(r)}} ~~~ \text{subj} ~~~ U U^{T}=I_q
\label{EquationSpaCtrl2}
\end{equation}
where $||S||_{l_{2(r)}}= \sum_{i=1}^{m\times n} ||S_i||_{l^2}$ is the row-wise $l^2$-norm. The non-differentiable $l^2$-norm regularization term controls rows-wise sparsity on the estimator of $S$. The sparsity is then also controlled by the parameter $\lambda$. To optimize Equation (\ref{EquationSpaCtrl2}), Mateos and Giannakis \cite{9}\cite{10} used an alternating minimization algorithm \cite{107}.
\newline
\newline
\textbf{Algorithm for incremental SpaCtrl:}
An incremental version of Equation (\ref{EquationSpaCtrl2}) is obtained using the Exponentially Weighted Least Squares (EWLS) estimator as follows:
\begin{equation}
\underset{U,S}{\text{min}} ~~ \sum_{i=1}^{m\times n} \beta^{(m\times n)-i} \left[||X_n+m-U^TT_i-S_i||_{l_{2}}^2 + \lambda ||S_i||_{l_{2}} \right]
\label{EquationSpaCtrl3}
\end{equation}
where $\beta$ is a learning rate between $0$ and $1$. So, the entire history of data is incorporated in the online estimation process.
Whenever $\beta<1$, past data are exponentially discarded thus enabling operation in nonstationary backgrounds. Towards deriving a real-time, computationally efficient, and recursive solver of Equation (\ref{EquationSpaCtrl3}), an AM scheme is adopted in which iteration $k$ coincides with the time scale $i=1,2,...$ of data acquisition. Experimental results \cite{10} show that RPCA-SpaCtrl with $\lambda=9.69\times 10^{-4}$ presents better performance than the naive PCA \cite{310} and RSL\cite{1} with less time computation.

\subsection{RPCA with Sparse Corruption}
\label{sec:SpaCorr}
Even if the matrix $A$ is exactly the sum of a sparse matrix $S$ and a low-rank matrix $L$, it may be impossible to identify these components from the sum. For example, the sparse matrix $S$ may be low-rank, or the low-rank matrix $L$ may be sparse. So, Hsu et al. \cite{58} imposed conditions on the sparse and low-rank components in order to guarantee their identifiability from $A$ .
This method requires that $S$ not be too dense in any single row or column, and that the singular vectors of $L$ not be
too sparse. The level of denseness and sparseness are considered jointly in the conditions in order to obtain the
weakest possible conditions. This decomposition RPCA with Sparse Corruption (RPCA-SpaCorr) involves the same model than PCP in Equation (\ref{EquationPCP1}), that is $A=L+S$. Then, Hsu et al. \cite{58} proposed two convex formulations. The first is the following constrained formulation:
\begin{equation}
\underset{L,S}{\text{min}} ~~ ||L||_* + \lambda ||S||_{l_{1}}  ~~~\text{subj}~~~ ||A-L-S||_{l_{1}} \leq \epsilon_1 
\nonumber
\end{equation}
\begin{equation}
 ~~~ \text{and} ~~~ ||A-L-S||_* \leq \epsilon_*
\label{EquationSpaCor1}
\end{equation}
where $\lambda>0$, $\epsilon_1 \geq 0$ and $\epsilon_* \geq 0$.
The second is the regularized formulation:
\begin{equation}
\underset{L,S}{\text{min}} ~~ ||L||_* + \lambda ||S||_{l_{1}} + \frac{1}{2 \mu} ||A-L-S||_F^2 
\nonumber
\end{equation}
\begin{equation}
~~~ \text{subj} ~~ ||A-L-S||_{l_{1}} \leq \epsilon_1 ~~~ \text{and} ~~~ ||A-L-S||_* \leq \epsilon_*
\label{EquationSpaCor2}
\end{equation}
where  $\mu>0$ is the regularization parameter. Hsu et al. \cite{58} added a constraint to control the entry-wise
$\infty$-norm of $L$, that is $||L||_{{l_\infty}}$. That is $||L||_{l_{\infty}} \leq b$ is added in Equation(\ref{EquationSpaCor1}) and  $||A-S||_{l_{\infty}} \leq b$ is added in Equation (\ref{EquationSpaCor2}). The parameter $b$ is a natural bound for $L$ and is typically $510$ for image processing. Hsu et al. \cite{58} determined two identifiability conditions that guarantee the recovery. The first one measures the maximum number of non-zero entries in any row or column of $S$. The second one measures the sparseness of the singular vectors $L$. Hence, a mild strengthening of these measures is achieved for the recovery guarantees.

\subsection{RPCA via Log-sum Heuristic Recovery}
\label{sec:LHR}
When the matrix has high intrinsic rank structure or the corrupted errors become dense, the convex approaches may not achieve good performances. Then, Deng et al. \cite{55} used the Log-sum Heuristic Recovery (LHR) to learn the low-rank structure. The decomposition involves the same model than PCP in Equation (\ref{EquationPCP1}), that is $A=L+S$. Although the objective in Equation (\ref{EquationPCP3}) involves the nuclear norm and the $l_1$-norm, it is based on the $l_1$ heuristic since nuclear norm can be regarded as a specific case of $l_1$-norm \cite{55}. Replacing the nuclear norm by its $l_1$-norm formulation, the problem can be solved as follows:
\begin{equation}
\underset{\hat{X} \in \hat{D}}{\text{min}} ~~ \frac{1}{2}(||diag(Y)||_{l_{1}} + ||diag(Z)||_{l_{1}}) +\lambda ||E||_{l_{1}}
\label{EquationLSR1}
\end{equation}
where $\hat{X}=\left\{Y,Z,L,S\right\}$ and \\
\\
$\hat{D}=\left\{\left(Y,Z,L,S\right):\begin{pmatrix} Y & L \\ L^T & Z \end{pmatrix} \geq 0, (L,S) \in C \right\}$ \\
\\
$(L,S) \in C$ stands for convex constraint. $Y$ and $Z$ are both symmetric and positive definite. $\geq$ represents semi-positive definite. The convex problem with two norms in Equation (\ref{EquationPCP3}) has been converted to an optimization only
with $l_1$-norm and therefore it is called $l_1$-heuristic. Next, Deng et al. \cite{55} used the logsum term to represent the sparsity of signals and obtained the Log-sum Heuristic Recovery (LHR) model:
\begin{equation}
\underset{\hat{X} \in \hat{D}}{\text{min}} ~~ \frac{1}{2}(||diag(Y)||_L + ||diag(Z)||_L)  +\lambda ||E||_L
\label{EquationLSR2}
\end{equation}
where $||X||_L=\sum_{ij}log~(\left|X_ij\right|+\delta)$ with $\delta>0$ is a small regularization constant. This model is non convex but the convex upper bound can be easily defined. LHR can remove much denser errors from the corrupted matrix rather compared to PCP.
\newline
\newline
\noindent \textbf{Algorithm for solving LHR:}
Deng et al. \cite{55} used the majorization-minimization (MM) \cite{108}\cite{109} algorithm that replaces the hard problem by a sequence of easier ones. It proceeds in an Expectation Maximization (EM)-like fashion by repeating two steps of majorization and minimization in an iterative way. During the majorization step, it constructs the convex upper bound of the non-convex objective. In the minimization step, it minimizes the upper bound.

\subsection{RPCA via Iteratively Reweighted Least Squares Minimization}
Guyon et al. \cite{505} proposed the decomposition solved via IRLS with the following model:
\begin{equation}
A = L+S = UV + S
\label{EquationIRLS1}
\end{equation}
\noindent where $U$ is a low-rank matrix corresponding to the background model plus noise, and $V$ reconstructs $L$ by linear combination. $S$ corresponds to the moving objects. The model involves the error reconstruction determined by the following constraints:
\begin{equation}
\underset{U \in \mathbb{R}^{n \times p}, V \in \mathbb{R}^{p \times m}}{min} ~~ \mu ||UV||_* + ||(A-UV)\circ W_1||_{l_{\alpha,\beta}}
\label{EquationIRLS2}
\end{equation}
where $||.||_*$ denote the nuclear norm and $||.||_{l_{\alpha,\beta}}$ is a mixed norm. $W_1$ which is a weighted matrix is iteratively computed and aims to enforce the fit exclusively on guessed background region. A function $\Phi(.)$ smoothes the error like spatial median filtering and transforms the error for obtain a suitable weighted mask for regression:
\begin{equation}
W=\Phi(A-UV) , \Phi(x)=e^{-\gamma TV(A-UV)}
\label{EquationIRLS3}
\end{equation}
By including local penalty as a constraint in RPCA, it explicitly increases local coherence of the sparse component as filled/plain shapes for moving objects. Furthermore, the decomposition is split into two parts. The first part tracks 1-Rank decomposition since the first eigen-vector is strongly dominant in video surveillance. For the mixed norm, Guyon et al. \cite{505} used $||.||_{l_{2,1}}$ instead of the usual $||.||_{l_{1,1}}$ because it forces spatial homogeneous fitting. Thus, the SVD algorithm can be seen as an iterative regression and then IRLS algorithm is then used. So, Guyon et al. \cite{505} increased local coherence of the error for moving objects by including local penalty as a constraint in the decomposition. Using the same approach, Guyon et al. \cite{506} added spatial constraint in the minimization based on the gradient and Guyon et al. \cite{507} proposed a spatio-temporal version. An other variant of RPCA via IRLS have been developed by Lu et al \cite{1006}.

\subsection{RPCA via Stochastic Optimization}
Feng et al. \cite{1010} proposed an Online Robust PCA (OR-PCA) algorithm. The main idea is to develop a stochastic optimization algorithm to minimize the empirical cost function, which processes one sample per time instance in an online manner. The coefficients which correspond to noise and the basis
are optimized in an alternative manner. The low dimensional subspace called low-rank matric basis are first initialized randomly and then updated after every frame per time instance. Moreover, OR-PCA decomposes the nuclear norm of the objective function of the traditional PCP algorithms into an explicit product of two low-rank matrices, i.e. basis and coefficients. The main function in OR-PCA is formulated as:
\begin{equation}
\underset{L \in \mathbb{R}^{n \times p}, R \in \mathbb{R}^{n \times r}}{\text{min}} ~~ \frac{1}{2} ||A-LR^T-S||_F^2+ \frac{\lambda_1}{2} (||L||_F^2+ ||R||_F^2) + \lambda_2 ||S||_{l_{1}}\\
\label{EquationSO}
\end{equation}
where $R$ is a coefficient matrix. $\lambda_1$ controls the basis and coefficients for low-rank matrix, whereas $\lambda_2$ controls the sparsity pattern, which can be tunned according to video analysis. In addition, basis and coefficient depend on the value of rank. In case of video background modeling, no visual results \cite{1010} have been found using this technique. Therefore, Javed et. al \cite{1010-1} modified OR-PCA via stochastic optimization method for background subtraction applications. An initialization scheme is adopted which converges the algorithm very fastly as compared to original OR-PCA. \\

\indent In order to perform OR-PCA, a number of video frames are first initialized as a low dimensional basis then stochastic optimization is performed on each input frame to separate the low-rank and sparse component. As compare to conventional RPCA via PCP based schemes, no batch optimizations are needed therefore OR-PCA is applicable for real time processing. In addition, a global pre-processing steps such as Laplacian and Gaussian images are introduced in modified OR-PCA which increase the detection rate. Using these modifications in original scheme, both memory cost and computational time is decreased, since the idea is based on to process one single frame per time instance, but the method shows some weak performance when large variations in the background scenes occurs such as waving trees and water surface. \\

\indent Therefore, Javed et al. \cite{1010-2}\cite{1010-6} further improved the foreground segmentation using the continuous constraints with Markov Random Field (MRF). OR-PCA via image decomposition using initialization scheme including continuous MRF with tuned parameters shows a drastic improvements in the experimental results specially in case highly dynamic backgrounds. In their work a good parameters range is provided according to different background scenarios. A huge amount of experimental results are provided which shows a very nice potential for its real time applicability. This scheme was improved with dynamic feature selection \cite{1010-3}\cite{1010-6}, a depth-extended version caled DEOR-PCA with spatiotemporal constraints \cite{1010-4}, and for noisy videos with Active Random Field (ARF) \cite{1010-5}. In an other way, Han et al. \cite{1541}\cite{1911} improved OR-PCA to be robust against camera jitter. \\

\indent In an other approach, Chen and Li \cite{16050} proposed an online algorithm based on Incremental Nonnegative Matrix Factorization (INMF) \cite{16051} to solve the OR-PCA problem. Firstly, nonnegative constraints are added on the coefficient vector and background subspace, and multiplicative update rules are used to compute them in order to avoid unreasonable negative pixels appearing in the estimated background images.
Then the structural relationship of the foreground pixels is taken into account by using structured sparsity-inducing norm. Weights to update
the background subspace are chosen appropriately as follows: 1) When new frames contain rather slowly moving objects, small weights
are assigned for new frames to avoid ghosts in the computed background images, and 2) if background changes quickly,
greater weights are used for adapting to the background changes. So a tradeoff is considered between the two situations when setting weights. Experimental results \cite{16050} show that this method outperforms the original OR-PCA \cite{1010}, the original RPCA-PCP \cite{3} and DECOLOR \cite{25}.

\subsection{Bayesian Robust Principal Component Analysis}
\label{sec:BRPCA}
Bayesian Robust Principal Component Analysis approaches have also been investigated for RPCA and used a Bayesian framework in the decomposition into low-rank plus sparse matrices. Ding et al. \cite{7} modeled the singular values of $L$ and the entries of $S$ with beta-Bernoulli priors, and used a Markov chain Monte Carlo (MCMC) sampling scheme to perform inference. This method called Bayesian RPCA (BRPCA) needs many sampling iterations, always hampering its practical use. In a similar approach, Babacan et al. \cite{24} adopted the automatic relevance determination (ARD) approach to model both $L$ and $S$, and utilized the variational Bayes (VB) method to do inference. This method called Variational Bayesian RPCA (VBRPCA) is more computationally efficient. 
However, these three methods assume a certain noise prior (a sparse noise plus a
dense noise), which cannot always effectively model the diverse types of noises occurring in practice. To address this problem, Zhao et al. \cite{1024} proposed a generative RPCA model under the Bayesian framework by modeling data noise as a mixture of Gaussians (MoG). Table \ref{BRPCAOverview} shows an overview of the Bayesian Robust Principal Component Analysis methods. 

\begin{table*}
\scalebox{0.90}{
\begin{tabular}{|l|l|l|} 
\hline
\scriptsize{Bayesian RPCA} &\scriptsize{Categories} &\scriptsize{Authors - Dates} \\
\hline
\hline
\scriptsize{Decompositions} &\scriptsize{1) Original BRPCA}                 &\scriptsize{Ding et al. (2011) \cite{7}}      \\
\scriptsize{} &\scriptsize{2) Variational BRPCA (VBRPCA)}             		  &\scriptsize{Babacan et al. (2011) \cite{24}}  \\
\scriptsize{} &\scriptsize{3) Factorized Variational BRPCA  (FVBRPCA)}      &\scriptsize{Aicher (2013) \cite{94}}          \\
\scriptsize{} &\scriptsize{4) MOG-RPCA}                                     &\scriptsize{Zhao et al. (2014) \cite{1024}}   \\
\scriptsize{}	&\scriptsize{5) Bayesian-Ising-Signal (BIS)}                  &\scriptsize{Huan et al. (2016) \cite{1572}}   \\
\hline
\scriptsize{Solvers} &\scriptsize{Markov chain Monte Carlo (MCMC)}                     &\scriptsize{Robert and Cassela (2004) \cite{113}} \\
\scriptsize{} &\scriptsize{Variational Bayesian Inference (VB)}                        &\scriptsize{Beal (2003) \cite{102}}       \\
\scriptsize{} &\scriptsize{Approximate Bayesian Inference (AB)}                        &\scriptsize{Beal (2003) \cite{102}}       \\
\hline
\scriptsize{Spatio-Temporal Algorithms} &\scriptsize{Markov Random Field }             &\scriptsize{Ding et al. (2011) \cite{7}} \\
\hline
\end{tabular}}
\caption{Bayesian Robust Principal Component Analysis: A Complete Overview. The first column indicates the concerned category and the second column the name of each method. Their corresponding acronym is indicated in the first parenthesis. The third column gives the name of the authors and the date of the related publication.} \centering
\label{BRPCAOverview}
\end{table*}

\subsubsection{Bayesian Robust PCA}
Ding et al.\cite{7} proposed a Bayesian Robust PCA (BRPCA). Assuming that the observed data matrix $A$ can be decomposed in three matrix like in SPCP \cite{5}, the Bayesian model is then as follows:
\begin{equation}
A = D(ZG)W_2 + BX + E
\label{BRCA}
\end{equation}
where $D \in \mathbf{R}^{n \times r}$, $W \in \mathbf{R}^{r \times m}$ and $G \in \mathbf{R}^{r \times r}$ are diagonal matrices and $X \in \mathbf{R}^{n \times m}$. The diagonal matrix $Z$ has binaries entries along the diagonal, and the binary matrix $B \in \left\{0,1^{n \times m}\right\}$ is sparse. $r$ defines the largest possible rank that may inferred for $L$, and $r$ is set to a large value. The low-rank, sparse and noise component are obtained as follows.

\begin{itemize}
\item \textbf{Low-rank Component:}
The low-rank component is modeled as $L=D(ZG)W_2$.  This is similar to SVD excepted for the extra diagonal matrix $Z$ with diagonal elements $z_{k,k}\in{0,1}$ for $k=1,...,r$. The product $ZG$ is a diagonal matrix too. The use of $Z$ decouples the rank learning and the singular value learning. $r$ is chosen large and then the diagonal entries of $ZG$ are sparse. The binary diagonal matrix $Z$ is modeled as follows:
\begin{equation}
z_{k,k} \sim Bernoulli(p_k)
\label{EquationLRC1}
\end{equation}
\begin{equation}
p_k \sim Beta (\alpha_0, \beta_0), k=1,...,r.
\label{EquationLRC2}
\end{equation}
with $\alpha_0>0$ and $\beta_0>0$. The parameters $\alpha_0$ and $\beta_0$ imposed the sparseness of the diagonal of $Z$. $\alpha_0$ and $\beta_0$ are set respectively to $1/K$ and $(K-1)/K$.
Each diagonal entry in $G$, denoted as $g_{k,k}$ for $k=1,...,r$, is obtained from a normal-gamma distribution:
\begin{equation}
g_{k,k} \sim \mathcal{N}(0,\tau^{-1}) k=1,...,r.
\label{EquationLRC3}
\end{equation}
\begin{equation}
\tau \sim Gamma (a_0,b_0) 
\label{EquationLRC4}
\end{equation}
with $a_0>0$ and $b_0>0$. $a_0$ and $b_0$ are set to $10^{-7}$.
The columns of matrices $D$ and $W_2$ are obtained from normal distribution:
\begin{equation}
d_k \sim \mathcal{N}(0,(1/N)I_N) k=1,...,K.
\label{EquationLRC5}
\end{equation}
\begin{equation}
w_{2,m} \sim \mathcal{N}(0,(1/K)I_K) m=1,...,M.
\label{EquationLRC6}
\end{equation}
with $I_N$ is the $N \times N$ identity matrix. The decomposition can be rewritten as follows:
\begin{equation} 
l_m=D(ZG)w_{2,m}=\sum_{k=1}^{K}z_{k,k}g_{k,k}w_{2,(k,m)} d_k, m=1,...,M
\label{EquationLRC7}
\end{equation}
So, each column of $L$ is the weighted sum of the dictionary elements in $D$, and $K$ is the size of the dictionary. The weights ${z_{k,k}g_{k,k}}_{k=1:K}$ determined the dictionary elements that are active to construct $L$. The weights ${w_{2,(k,m)}}_{k=1:K}$ determined the importance of the selected dictionary elements for the representation of the $m^{th}$ column of $L$.

\item \textbf{Sparse Component:}
The sparse component is modeled as $S=BX$, where $B$ is a binary matrix. This decomposition separates the learning sparseness from the learning of values. Each column of $B$ is modeled as follows:
\begin{equation}
b_m \sim \prod_{n=1}^{N}Bernoulli(\pi_n), m=1,...,M 
\label{EquationSC1}
\end{equation}
\begin{equation}
\pi_n \sim Beta (\alpha_1, \beta_1), n=1,...,N.
\label{EquationSC2}
\end{equation}
The sparseness prior is made with the parameters $\alpha_1$ and $\beta_1$.  $\alpha_1$ and $\beta_1$ are set respectively to $1/N$ and $(N-1)/N$. The columns of X are obtained from a normal-gamma distribution:
\begin{equation}
x_m \sim \mathcal{N}(0,v^{-1}I_N), m=1,...,M
\label{EquationSC3}
\end{equation}
\begin{equation}
v \sim Gamma (c_0, d_0).
\label{EquationSC4}
\end{equation}
with $c_0>0$ and $d_0>0$. $c_0$ and $d_0$ are set to $10^{-6}$.
Ding et al.\cite{7} addressed the dependency of the sparse component in time and space with a Markov structure. If the parent node of $I_t(i,j)$ noted $I_{t-1}(i,j)$ is non-zero, its child node is also non-zero with a high probability. To introduce spatial dependence, Ding et al. \cite{7} defined the state of $F_t(i,j)$ as follows:
\begin{equation}
S(F_t(i,j))=active   ~~ if   ~~ ||N(F_t(i,j)||_0 \geq \rho
\end{equation}
\begin{equation}
S(F_t(i,j))=inactive   ~~ otherwise
\end{equation}
where $\rho=5$ which imposes that a node is active if the sparse component contains at least $5$ non-zero members in its neighborhood defined by  $N(F_t(i,j))=\left\{ F_{(k,l)} : \left|k-i\right| \leq 1, \left|l-j\right| \leq 1 \right\}$. Then, a child node depends on its parent node in time and on its neighbors in space. Markov dependency is then imposed by modifying Equation (\ref{EquationSC1}) and Equation (\ref{EquationSC2}) as follows.
\begin{equation}
b_t \sim \prod_{n=1}^{N}Bernoulli(\pi_{nt}), t=1,...,M 
\label{EquationSC5}
\end{equation}
\begin{equation}
\pi_{nt} \sim Beta (\alpha_H, \beta_H) ~~ if ~~ S(b_{n,t-1})=active  
\nonumber
\end{equation}
\begin{equation}
~~ \text{with} ~~ n=1,...N, t=2,...,M.
\label{EquationSC6}
\end{equation}
\begin{equation}
\pi_{nt} \sim Beta (\alpha_L, \beta_L) ~~ if ~~ S(b_{n,t-1})=inactive 
\label{EquationSC7}
\end{equation}
\begin{equation}
~~ \text{with} ~~ n=1,...N, t=2,...,M.
\nonumber
\end{equation}
where $H$ and $L$ indicate the high and low states in the Markov model and $\alpha_H$,$\alpha_H$, $\beta_L$ and $\beta_H$ are set to assume that the sparseness will be propagated along time with high probability. For $t=1$, Equation (\ref{EquationSC1}) and Equation (\ref{EquationSC2}) are used since there are no parent nodes for the first frame.

\item \textbf{Noise Component:}
The noise is modeled by a Gaussian distribution as follows:
\begin{equation}
e_{n,m} \sim \mathcal{N}(0,\gamma_m^{-1}), ~~ \text{with} ~~  n=1,...,N  
\label{EquationNC1}
\end{equation}
\begin{equation}
\gamma_m \sim Gamma(e_0,f_0) ~~ \text{for} ~~ m=1,...,M,
\label{EquationNC2}
\end{equation}
with $e_{n,m}$ is the entry at row $n$ and column $m$ of $E$. $c_0$ and $d_0$ are set to $10^{-6}$.
\end{itemize}

Then, the posterior density function of the BRPCA is as follows:
\begin{equation} 
- log~(p(\Theta | A, H)) = \frac{\tau}{2} ||G||_F^2 - log~(\left[f_{BB}(Z;H))\right] 
\nonumber
\end{equation}
\begin{equation} 
+ \frac{N}{2} \sum_{k=1}^r ||d_k||_{l^2}^2 + \frac{1}{2} \sum_{m=1}^M ||w_m||_{l^2}^2 +  \frac{v}{2} ||X||_F^2
\nonumber
\end{equation}
\begin{equation} 
- log~(\left[f_{BB}(B;H))\right] + \frac{1}{2} ||Y-L-S||_F^2 
\nonumber
\end{equation}
\begin{equation} 
- log \left[Gamma(\tau | H) Gamma(v | H) Gamma (\gamma | H) \right] 
\nonumber
\end{equation}
\begin{equation} 
+ constant
\label{EquationPDFBRPCA}
\end{equation}
where $\Theta$ represents all model parameters, $f_{BB}(.|H)$ represents the beta-Bernoulli prior, and $H=\left\{\alpha_0, \alpha_1, \beta_0,\beta_1,a_0,b_0,c_0,d_0,e_0,f_0 \right\} $ are model hyper parameters. \\

\noindent \textbf{Algorithms for solving BRPCA:} Ding et al. \cite{7} proposed to approximate the posterior density function in Equation (\ref{EquationPDFBRPCA}) with two algorithms:
\begin{itemize}
\item  \textbf{Markov chain Monte Carlo (MCMC) analysis implemented with Gibbs sampler \cite{113}:} The posterior distribution is approximated by a set of samples, collected by iteratively drawing each random variable from its conditional posterior distribution given the most recent values of all the other parameters.
\item \textbf{Variational Bayesian inference (VB) \cite{102}:} A set of distributions $q(\Theta)$ allow to approximate the true posterior distributions $p(\Theta|A)$, and uses a lower bound to approximate the true log-likelihood of the model $log~(p(A|\Theta)$. The algorithm iteratively updates $q(\Theta)$ so that the lower bound approaches to $log~(p(A|\Theta)$.
\end{itemize}
The computational complexity of MCMC and VB iteration is approximatively the same. The VB solution may find a local-optimal solution which may be not be the global-optimal best solution. Ding et al. \cite{7} found that MCMC work quite effectively in practice. \\

\noindent \textbf{Relation to PCP and SPCP:} For the low-rank component, Ding et al. \cite{7} employed a Gaussian prior to obtain a constraint on Frobenius norm $||G||_F^2$ with a beta-Bernoulli distribution to address the sparseness of singular value and to obtain a small number of non-zero singular values, while PCP employs the rank function that is relaxed to the nuclear norm when solving the problem in a convex way. For the sparse component, the constraint on Frobenius matrix norm $||X||_F^2$ and the beta-Bernoulli distribution are used to impose sparseness while PCP uses the $l_0$-norm that is relaxed to the $l_1$-norm. The error term $(2 \mu)^{-1} ||A-L-S||_F^2$ in SPCP \cite{4} corresponds to the Gaussian prior placed on the measurement noise in Equation (\ref{EquationNC1}). For solving the problem, the main difference is that BRPCA uses numerical methods to estimate a distribution for the unknown parameters, whereas optimization based methods effectively search a single solution that minimizes a analogous functional to $-log~(p(\Theta|A,H))$.

\subsubsection{Variational Bayesian Robust Principal Component Analysis}
\label{sec:VBRPCA}
Babacan et al. \cite{24} proposed a Variational Bayesian Robust PCA (VBRPCA). Assuming that the observed data matrix $A$ can be decomposed in three matrix like in SPCP \cite{5}, the variational Bayesian model is then as follows:
\begin{equation}
A = DB^T + S + E
\label{EquationVBRPCA1}
\end{equation}
where $DB^T$ is the low-rank component with $D \in \mathbf{R}^{m \times r}$ and $B \in  \mathbf{R}^{r \times n}$, $S$ is the sparse component with arbitrarily large coefficients and $E$ is the dense error matrix with relatively smaller coefficients. The low-rank, sparse and noise component are obtained as follows. 
The low-rank component $L$ is then given by $DB^T$. So, $L$ is the sum of outer-products of the columns of $D$ and $B$, that is,
\begin{equation}
L=\sum_{i=1}^k d_{.i}b_{.i}^T
\label{EquationVBRPCA2}
\end{equation}
where $k \geq r$. $d_{.i}$ and $d_{i.}$ denote the $i^{th}$ column and row of $D$, respectively. To impose column sparsity in $D$ and $B$, such that most columns in $D$ and $B$ are set equal to zero, the columns are defined with Gaussians priors as follows:
\begin{equation}
p(D|\gamma)=\prod_{i=1}^k \mathcal{N}(d_{.i}|0,\sigma_i I)
\label{EquationVBRPCA3}
\end{equation}
\begin{equation}
p(B|\gamma)=\prod_{i=1}^k \mathcal{N}(b_{.i}|0,\sigma_i I)
\label{EquationVBRPCA4}
\end{equation}
where $\sigma_i$ is the variance. Most of the variances are very small values during inference to reduce the rank of the estimate. Then, the following conditional distribution for the observations are obtained:
\begin{equation}
p(A|D,B,S,\beta)=\mathcal{N}(A|DB^T+S,\gamma^{-1}I)
\label{nonumber}
\end{equation}
\begin{equation}
=exp[\frac{\beta}{2}||A-DB^T-S||_F^2
\label{EquationVBRPCA5}
\end{equation}
where $\beta$ is a uniform hyperprior. The modeling of the sparse component $S$ is done by using independent Gaussian priors on its coefficients $S_{ij}$ as follows:
\begin{equation}
p(S|\alpha)=\prod_{i=1}^m\prod_{j=1}^n \mathcal{N}(S_{ij}|0,\alpha_{ij}^{-1})
\label{EquationVBRPCA6}
\end{equation}
where $\alpha=\left\{\alpha_{ij}\right\}$, $\alpha_{ij}$ is the precision of the Gaussian on the $(i,j)^{th}$ coefficient and $p(\alpha_{ij})$=const $\forall i,j$.
Finally, the joint distribution is expressed as follows:
\begin{equation}
p(A,D,B,S,\gamma,\alpha,\beta)
\nonumber
\end{equation}
\begin{equation}
=p(A|D,B,S,\beta)p(D|\gamma)p(B|\gamma)p(S|\alpha)p(\gamma)p(\alpha)p(\beta)
\label{EquationVBRPCA7}
\end{equation}

where $p(\gamma_i)=\frac{1}{\gamma_i}^{a+1} exp(\frac{-b}{\gamma_i})$ and $p(\beta)$ is a constant assuming that the noise precision have a uniform prior.\\

\noindent \textbf{Algorithm for solving VBRPCA:} The exact full-Bayesian inference using joint distributions in Equation (\ref{EquationVBRPCA7}) is intractable because $p(y)$ can't be computed by marginalizing all variables. Therefore, Babacan et al. \cite{24} used an inference procedure based on mean field variational Bayes. The aim is to compute posterior distribution approximations by minimizing the Kullback-Leibler divergence in an alternating way for each variable. Let $z=(D,D,S,\gamma,\alpha,\beta)$, the posterior approximation $q(z_k)$ of each variable $z_k \in z$ is then determined as follows:
\begin{equation}
log~(q(z_k)) = \left\langle log~(p(A,z))\right\rangle_{\frac{z}{z_k}}+const
\end{equation}
where $\frac{z}{z_k}$ is the set $z$ without $z_k$. The distribution $p(A,z)$ is the joint probability distribution given in Equation (\ref{EquationVBRPCA7}). The posterior factorization $q(z)=\prod q(z_k)$ is used such that the posterior distribution of each unknown is estimated by holding the others fixed using their most recent distributions. Thus, the expectations of all parameters in the joint distribution are taken with respect to their most recent distributions, and the result is normalized to find the approximate posterior distribution. Since all distributions are in the conjugate exponential family, the form of each posterior approximation is easily determined.

\subsubsection{Factorized Variational Bayesian RPCA (FVBRPCA)}
Aicher \cite{94} proposed a Factorized Variational Bayesian RPCA. This model is slightly different from BRPCA \cite{7} and VBRPCA \cite{24} in how sparse noise is modeled and incorporated as well as the use of variational Bayes instead of MCMC.
\begin{equation}
A = UV^T+ Z^*\circ B + E
\label{Equation-FVBRPCA-1}
\end{equation}
where $\circ$ denotes the Hadamard element-wise multiplication. The low-rank matrix is $L=UV^T$ and $U$ is restrited to be
an $n \times r$ matrix and $V$ to be an $r \times m$ matrix so that the rank of $L$ less than or equal to $r$. The sparse matrix is $S = Z^*\circ B$ and $B$ is set to be a sparse binary matrix and $Z^*$ is without constraint. For numerical reasons, $Z^*$ is treated as a very diffuse Gaussian matrix. To induce sparsity in $S$, a prior on $B$ is selected such that it is sparse. $E$ is a small Gaussian noise term and the prior on its variance
small compared with the variance of $Z^*$. Instead of solving Equation \ref{Equation-FVBRPCA-1}, it is more numerical convenient to solve the following problem:
\begin{equation}
A = UV^T+ Z^*\circ B + E \circ (1-B)
\label{Equation-FVBRPCA-2}
\end{equation}
To infer $U$,$V$,$B$,$Z$ and $E$, Aicher \cite{94} approximated the posterior distribution with a factorizable distribution. This a
variational approach selects the distribution $q$ closest to the posterior  in the sense of Kullback-Leibler (KL) divergence. By parameterizing $q$, Aicher \cite{94} converted the inference scheme back into an objective maximization problem. After selecting a distribution to approximate
the posterior, the expectations of $U$,$V$,$B$,$Z$ and $E$ are taken to estimate them. Experimental results \cite{94} show that FVBRPCA performs slightly better than RPCA solved via IALM \cite{18}, VBRPCA \cite{24} and GoDec \cite{6}.
 
\subsubsection{Bayesian RPCA with MoG noise(MoG-BRPCA)}  
Zhao et al. \cite{1024} developed a generative RPCA model under the Bayesian framework by modeling data noise as a mixture of Gaussians (MoG). The MoG is a universal estimator to continuous distributions and thus MoG-BRPCA is able to fit a wide range of noises such as Laplacian, Gaussian, sparse
noises and any combinations of them. 

\subsection{Approximated RPCA}
\label{sec:A-RPCA}

\subsubsection{"Go Decomposition" (GoDec)}
Zhou and Tao \cite{6} proposed a randomized low-rank and sparse matrix decomposition called "Go Decomposition" (GoDec). GoDec estimates the low-rank part $L$ and the sparse part $S$ by using the same decomposition than SPCP \cite{5}:
\begin{equation}
A = L + S + E
\label{EquationGD1}
\end{equation}
To solve the problem in Equation (\ref{EquationGD1}), GoDec alternatively assigns the low-rank approximation to $A-S$ to $L$ and the sparse approximation to $A-L$ to $S$. This approximated decomposition problem seeks to solve the minimization of the following decomposition error:
\begin{equation}
\underset{L,S}{\text{min}} ~~ ||A-L-S||_F^2   ~~~ \text{subj} ~~~ rank(L) \le r, card(S) \le k.
\label{EquationGD2}
\end{equation}
\\
\noindent \textbf{Algorithm for solving GoDec:} The optimization problem in Equation (\ref{EquationGD2}) is solved by alternatively solving the two following subproblems:
\begin{equation}
L_t=arg \underset{rank(L) \le e}{\text{min}} ~~ ||A-L-S_{t-1}||_F^2
\label{EquationGD3}
\end{equation}
\begin{equation}
S_t=arg \underset{card(S) \le e}{\text{min}} ~~ ||A-L_t-S||_F^2
\label{EquationGD4}
\end{equation}
Although both subproblems have nonconvex constraints, their global solutions $L_t$ and $S_t$ exist. Indeed, these subproblems can be solved by updating $L_t$ via singular value hard thresholding of $A-S_{t-1}$ and updating $S_t$ via entry-wise hard thresholding of $A-L_t$, respectively as follows:
\begin{equation}
L_t=\sum_{i=1}^{r} \lambda_i U_i V_i^T ~~ with ~~ SVD(A-S_{t-1})=UGV^T
\label{EquationGD5}
\end{equation}
\begin{equation}
S_t=P_{\Omega}(A-L_t) ~~ \text{with} ~~ \Omega: \left|(A-L_t)_{i,j \in \Omega} \right| \ne 0
\nonumber
\end{equation}
\begin{equation}
~~ \text{and} ~~ \ge \left|(A-L_t)_{i,j \in \bar{\Omega}} \right|, \left| \Omega \right| \ge k
\label{EquationGD6}
\end{equation}
where $P_{\Omega}(.)$ is defined as the projection of the matrix on the observed entries following the sampling set $\Omega$. The main computation time is due to the computation of the SVD for $A-S_{t-1}$ in the updating $L_t$ sequence. To significantly reduce the time cost, Zhou and Tao \cite{6} replaced the SVD by a Bilateral Random Projection(BRP) based low-rank approximation.

\subsubsection{Semi-Soft GoDec}
Zhou and Tao \cite{6} proposed a Semi-Soft GoDec which adopts soft thresholding to the entries of $S$, instead of GoDec which imposes hard thresholding to both the singular values of the low-rank part $L$ and the entries of the sparse part $S$. This improvement has two  two main advantages: 1) the parameter $k$ in constraint $card(S)\leq k$ is automatically determined by a soft-threshold $\tau$, thus avoids the situation when $k$ is chosen too large and some part of noise $E$ is leaked into $S$; 2) the time cost is substantially smaller than the ordinary GoDec. For example, the background modeling experiments can be accomplished with a speed 4 times faster then ordinary GoDec, while the error is kept the same or even smaller. The approximated decomposition problem seeks to solve the minimization of the following decomposition error:
\begin{equation}
\underset{L,S}{\text{min}} ~~ ||A-L-S||_F^2   ~~~ \text{subj} ~~~ rank(L) \le r, card(S) \le \tau.
\label{EquationGD}
\end{equation}
where $\tau$ is the soft threshold. Chen et al. \cite{1060} proposed to use Semi-Soft GoDec for video coding in the existing standard codecs H.264/AVC and HEVC via background/foreground separation. For this, Chen et al. \cite{1060} developed an extension of the Semi-Soft GoDec that is
able to perform LRSD on new matrix columns with a given low-rank structure, which is called incremental low-rank and sparse decomposition (ILRSD).

\subsection{Sparse Additive Matrix Factorization (SAMF)}
Nakajima et al. \cite{24-1}\cite{24-2} extented the original robust PCA \cite{3} by proposing a unified view called Sparse Additive Matrix Factorization (SAMF). Instead of RPCA which only copes with element-wise sparsity (spiky noise) and low-rank sparsity (low-dimensional matrix), SAMF 
handles various types of sparse noise such as row-wise and column-wise sparsity. Thus, the decomposition is written as follows:
\begin{equation}
A=\sum_{k=0}^K S + E
\label{EquationSAMF-1}
\end{equation}
where $K$ is the number of sparse matrices. $K=2$ in the original RPCA \cite{3} in which the element-wise sparse term is added to the low-rank term.
For background/foreground separation, the low-rank term and the element-wise sparse term capture the static background and the moving
foreground, respectively. Nakajima et al. \cite{24-1}\cite{24-2} relied on the natural assumption that a pixel segment which has similar intensity values in an image tends to belong to the same object. Thus, Nakajima et al. \cite{24-1}\cite{24-2} adopted a segment-wise sparse term, where the matrix is constructed using a precomputed over-segmented. Experimental results \cite{24-1}\cite{24-2} on the CAVIAR dataset \cite{206} show that SAMF based on image segmentation (sSAMF) outperforms PCP via IALM \cite{3} which correponds to 'LE'-SAMF in \cite{24-1}\cite{24-2}. \\

\noindent \textbf{Algorithm for solving SAMF:} First, Nakajima et al. \cite{24-1}\cite{24-2} reduced  the partial SAMF problem
to the standard MF problem, which can be solved analytically. Then, Nakajima et al. \cite{24-1}\cite{24-2} derived an iterative algorithm called the mean update (MU) for the variational Bayesian approximation to SAMF, which gives the global optimal solution for a large subset of parameters in each step. 

\subsection{Variational Bayesian Sparse Estimator (VBSE)}
Chen et al. \cite{24-3}\cite{24-4}\cite{1045} proposed a generalization of the original RPCA \cite{3}, where a linear transformation through the
use of a known measurement matrix, is applied to the outlier corrupted data. The aim is to estimate the outlier amplitudes given the transformed observation. This approach called variational Bayesian Sparse Estimator (VBSE) can achieved background/foreground separation in blurred and noisy video sequences. Thus, the decomposition is written as follows:
\begin{equation}
A=L+RS+E
\label{EquationVBSE-1}
\end{equation}
where $R$ models the linear transformation performed on the data. The aim is to obtain accurate estimates for the sparse term $S$ and the low-rank term $L$, given the noise corrupted observation $A$. Although $S$ is sparse, the multiplication with a wide matrix $R$ has an effect of compression, and
hence the product $RE$ is not necessarily sparse.  Then, Chen et al. \cite{24-3}\cite{24-4} modeled the lowk-rank part as follows:
\begin{equation}
||L||_*=\underset{U,V}{\text{min}} ~~ \frac{1}{2} ||U||_F^2 + ||V||_F^2  ~~~ \text{subj} ~~~ L=UV^T
\label{EquationVBSE-2}
\end{equation}
With these relaxation and parametrization, Chen et al. \cite{24-3}\cite{24-4} obtained the following optimization problem:
\begin{equation}
\underset{U,V,S}{\text{min}} ~~ \frac{1}{2} ||A-UV^T-RS||_F^2 + \lambda_* (||U||_F^2 + ||V||_F^2) + \lambda_1  ||E||_{l_{1}}  ~~~ \text{subj} ~~~ L=UV^T
\label{EquationVBSE-3}
\end{equation}
where $\lambda_*$ and $\lambda_1$  are regularization parameters. To enforce column sparsity in $U$ and $V$, the columns of $U$ and $V$ are modeled with Gaussian priors of precision. Then, Chen et al. \cite{24-3}\cite{24-4} incorporated conjugate Gamma hyperprior on the precisions. The sparse part $S$ is modeled by setting the entries be independent of each other, and their amplitudes are modeled by zero-mean Gaussian distributions with independent precisions. For the noise part $E$, Gaussian priors with zero mean are used to model the dense observation noise. By combining these different stages in a hierarchical Bayesian model, a joint distribution of the observation and all the unknown variables is expressed as follows:
\begin{equation}
\rho(A,U,V,S,\gamma,\alpha,\beta)
\label{EquationVBSE-4}
\end{equation}
where $\gamma$ and $\alpha$ are hyperparameters and $\beta$ is the noise precision. To solve VBSE, Chen et al. \cite{24-3}\cite{24-4} used an an approximate Bayesian inference. Experimental results  \cite{24-3}\cite{24-4} on the CAVIAR dataset \cite{206} show that VBSE outperforms PCP solved via APG \cite{19} and PCA solved via IALM \cite{3}.

\section{Robust Non-negative Matrix Factorization}
\label{sec:RNMF}
Non-negative matrix factorization (NMF) approximates a non-negative matrix $A$ by a product of two non-negative low-rank factor matrices $W$ and $H$. A complete review of the algorithms for nonnegative matrix factorization (NMF) and nonnegative tensor factorization (NTF) based on the block coordinate descent (BCD) framework is available in Kim et al; \cite{771}. Table \ref{TPCP2Overview} shows an overview of the different robust NMF decompositions. Their corresponding solvers as well as their complexity can be seen in Table \ref{A-TPCP3Overview}. 

\subsection{Manhattan Non-negative Matrix Factorization (MahNMF)} 
Guan et al. \cite{761} proposed a robust non-negative matrix factorization when the noise distribution is heavy tailed. The method called Manhattan NMF (MahNMF) minimizes the Manhattan distance between $A$ and $W^TH$ for modeling the heavy tailed Laplacian noise. Thus, Guan et al. \cite{761} minimized the Manhattan distance between an $m \times n$-dimensional non-negative matrix $A$ and $W^TH$ as follows:

\begin{equation}
\underset{W\geq0,H\geq0}{\text{min}} ~~ f(W,H)=||A-W^TH||_M   
\label{EquationMahNMF}
\end{equation}

where $||.||_M $ is the Manhattan distance which is equal to the summation of the $l_1$ norm, and the reduced dimensionality $r$ satisfies that $r \ll min(m,n)$. Since $W$ and $H$ are low-rank matrices, MahNMF actually estimates the nonnegative low-rank part, i.e., $W^TH$, and the sparse part, i.e., $A-W^TH$, of a non-negative matrix $A$. MahNMF performs effectively and robustly when data are contaminated by outliers because it benefits from both the modeling ability of Laplace distribution to the heavy tailed behavior of noise and the robust recovery capability of the sparse and low-rank decomposition. Experimental results \cite{761} on the I2R dataset \cite{203} show that MahNMF gives similar visual results than PCP via IALM \cite{3} and GoDec \cite{6}.\\

\noindent \textbf{Algorithms for solving MahNMF:} Two fast optimization algorithms for MahNMF were developed by Guan et al. \cite{761}. They are called the rank-one residual iteration (RRI) method, and Nesterov's smoothing method, respectively. Each variable of $W$ and $H$ are iteratively updated in a closed form solution in the RRI method by approximating the residual matrix with the outer product of one row of $W$ and one row of $H$. The RRI method is neither scalable to large scale matrices nor flexible enough to optimize all MahNMF extensions. As the objective functions of MahNMF are neither convex nor smooth, Guan et al. \cite{761} proposed a Nesterov's smoothing method to recursively optimize one factor matrix with an other matrix fixed. Thus, the smoothing parameter are set inversely proportional to the iteration number so improving the approximation accuracy iteratively.

\subsection{Near-separable Non-negative Matrix Factorization (NS-NMF)}
Promising robust NMF approaches have emerged based under the assumption that the data matrix satisfies a separability condition which enables the NMF problem to be solved efficiently and exactly. Under this assumption, the data matrix $A$ is said to be $r$-separable if all columns of $X$ are contained in the conical hull generated by a subset of $r$ columns of $A$. In other words, if $A$ has a factorization $WH$ then the separability assumption states that the columns of $W$ are present in $A$ at positions given by an unknown index set $B$ of size $r$. Equivalently, the corresponding columns of the right factor matrix $H$ constitute the $r \times r$ identity matrix, i.e., $H_B=I$. These columns indexed by
$B$ are called anchor columns. \\
In this framework, Kumar et al. \cite{762-1} proposed a family of conical hull finding procedures called Xray for near-separable NMF (NS-NMF) problems with Frobenius norm loss. The minimization problem can be formulated as follows:

\begin{equation}
\underset{A_B\geq 0, H\geq 0}{\text{min}} ~~ ||A-A_BH||_F^2  ~~~ \text{subj} ~~~ A_B\geq 0, H \geq 0
\label{EquationXray}
\end{equation}

Geometrically, Xray finds anchor columns one after the other, incrementally expanding the cone and using exterior columns to locate the next anchor. Xray present several advantages for background/foreground separation: (1) it requires no more than
$r$ iterations each of which is parallelizable, (2) it empirically robust to noise, (3) it admits efficient model selection, and (4) it does not require normalizations or preprocessing needed in other methods. However, the use of Frobenius norm approximations is not very suitable in the presence of outliers or different noise characteristics \cite{762}. \\
In this context, Kumar and Sindhwani \cite{762} improved Xray to provide robust factorizations with respect to $1_1$ loss, and approximations with respect to the family of Bregman divergences. In the case of background/foreground separation, it is natural to seek a low-rank background matrix $L$ that minimizes $||A-L||$ where $A$ is the frame-by-pixel video matrix, and the $l_1$ loss imposes a sparsity prior on the residual foreground. Instead of low-rank approximations in Frobenius or spectral norms, there is not a SVD-like tractable solution. For this, Kumar et al. \cite{762} imposed the separable NMF assumption on the background matrix. This constraint implies that the variability of pixels across the frames can be considered as observed variability in a small set of pixels. Under a more restrictive setting, it is equivalent to median filtering on the video frames, while a full near-separable NMF model conveys more degrees of freedom to model the background. The minimization problem can be formulated as follows:

\begin{equation}
\underset{A_B\geq 0, H\geq 0}{\text{min}} ~~ ||A-A_BH||_{l_{1}}   ~~~ \text{subj} ~~~ A_B \geq 0, H \geq 0
\label{EquationRobustXray-1}
\end{equation}

where ${A_B}$ are the columns of $A$ indexed by set $B \subset {1,2,...,n}$. Experimental results \cite{762} on the I2R dataset \cite{203} show that RobustXray outperforms the robust NMF (local search) which minimizes $\underset{W\geq 0, H\geq 0}{\text{min}} ~~ ||A-WH||_{l_{1}}$ \cite{762}, the robust Low-rank (local-search) which minimizes $\underset{W, H}{\text{min}} ~~ ||A-WH||_{l_{1}}$ \cite{762},  XRay-$l_2$ \cite{762-1} and PCP via IALM \cite{3}. \\

\noindent \textbf{Algorithms for solving XRay-$l_2$ and RobustXray:} Algorithms of Kumar et al. \cite{762-1} are not suitable for noise distributions
other than Gaussian. The algorithm for RobustXray proceeds by identifying one anchor column in each iteration and adding it to the current set of anchors, thus expanding the cone generated by anchors. Each iteration consists of two steps: (1) anchor selection step that finds the column of $A$
to be added as an anchor, and (2) a projection step where all data points (columns of $A$) are projected to the current cone in terms of minimizing the $l_1$ norm. 

\subsection{Robust Asymmetric Non-negative Matrix Factorization (RANMF)}
Woo and Park  \cite{765} proposed a formulation called $l_\infty$-norm based robust asymmetric nonnegative matrix factorization (RANMF) for the grouped outliers and low nonnegative rank separation problems. The main advantage of RANMF is that the denseness of the low nonnegative rank factor matrices can be controlled. To control distinguishability of the column vectors in the low nonnegative rank factor matrices for stable basis, Woo and Park \cite{765} imposed asymmetric constraints, i.e., denseness condition on the coefficient factor matrix only. As a by product, a well-conditioned basis factor matrix is obtained. Compared to the nuclear norm based low-rank enforcing models, RANMF is not sensitive to the nonnegative rank constraint parameter due to the soft regularization method. Thus, the decomposition is made as follows:
\begin{equation}
A=L+S= W\Lambda H +S
\label{EquationRANMF-1}
\end{equation}
where $L=W\Lambda H$ is low nonnegative rank matrix and $S$ contains the grouped outliers. $\Lambda$ is a diagonal matrix with $\Lambda_{ii} = \lambda_i$ which is be considered as an asymmetric singular value of $L$. Since  $\Lambda$ is subsumed into $W$ or $H$, Equation \ref{EquationRANMF-1} is a typical nonnegative matrix factorization (NMF) formulation and its corresponding minimization problem can be written as follows:
\begin{equation}
\underset{L,S, \Phi}{\text{min}} ~~ \frac{\alpha}{2} ||A-L-S||_F^2 + \Phi(S) +  \beta \Psi(S, \Phi) + \gamma TV(Q(\Phi))   ~~~ \text{subj} ~~~ Rank(L) \leq \tau, 0<L<b_L 
\label{EquationRANMF-2}
\end{equation}
where $\Phi(.)$ is a sparsity enforcing function, such as $l_p$-norm ($0<p<1$) or log-function. $Q$ is the reshaping operator from 2D to 3D, $TV(.)$ is the 3D Total Variation and $b_L=255$ for image data. $Rank(L)$ is a low nonnegative rank enforcing function such as nonnegative nuclear norm, and $\tau\geq0$ is a rank nonnegative constraint parameter. As grouped outliers is not foreground mask and can be very noisy, the TV appears as an additional denoising/segmentation process to detect foreground mask. To solve Equation \ref{EquationRANMF-2}, Woo and Park  \cite{765} developed a  Soft Regularized Asymmetric alternating Minimization (SRAM) algorithm.\\
\indent Experimental results \cite{765} on the I2R dataset \cite{203} show that RANMF outperforms PCP via IALM \cite{3} and DECOLOR \cite{25}.

\section{Robust Matrix Completion}
\label{sec:RMC}
Robust matrix completion RMC, also called RPCA plus matrix completion problem can also be used for background/foreground separation. Althought RPCA via principal component pursuit \cite{3} can be considered as RMC using $l_1$-norm loss function, the main difference lies in that in
RMC problems the support of missing entries is given, whereas in RPCA corrupted entries are never known \cite{4}\cite{2001}. From a statistical learning viewpoint, RPCA is a typical unsupervised learning problem while the RMC can be interpreted as a supervised learning problem \cite{2001}.
Table \ref{TPCP2Overview} shows an overview of the different RMC decompositions. Their corresponding solvers as well as their complexity can be seen in Table \ref{A-TPCP3Overview}.

\subsection{$l_{\sigma}$-norm loss function (RMC-$l_{\sigma}$)}
RPCA via principal component pursuit can be considered as RMC using $l_1$-norm loss function. Following this idea, Yang et al. \cite{2001} proposed  a nonconvex relaxation approach to the matrix completion problems when the entries are contaminated by non-Gaussian noise or outliers. Based on a nonconvex $l_{\sigma}$ loss function, Yang et al. \cite{2001} developed a rank constrained as well as a nuclear norm regularized model. The nuclear norm heuristic model is formulated in the following form:
\begin{equation}
\underset{S \in \mathbf{R}^{m \times n}}{\text{min}} ~~ \lambda ||L||_* +  l_{\sigma}(L)
\label{EquationIHT-1}
\end{equation}  
where $\lambda$ is a regularization parameter and the data fitting risk $l_{\sigma}(L)$ is given by:

\begin{equation}
l_{\sigma}(L)=\frac{\sigma^2}{2} \sum_{(i,j) \in \Omega} (1-exp(-(L_{ij}-A_{ij})^2 / \sigma^2))
\label{EquationIHT-2}
\end{equation} 
Experimental results \cite{2001} show that RMC-$l_{\sigma}$ performs slightly better than PCP solved via IALM \cite{3} because details of the background images are recovered well, whereas PCP solved via IALM \cite{3} does not seem to perform as well as  RMC-$l_{\sigma}$ where some details
of the background are added to the foreground. It can be also observed that none of the two methods can recover the missing entries in the foreground. Furthermore, RMC-$l_{\sigma}$ is more than 3 times faster than PCP solved via IALM \cite{3}.\\ 

\noindent \textbf{Algorithms for solving RMC-$l_{\sigma}$-IHT:} $l_{\sigma}$-IHT leads to computational difficulty due to its nonconvexity.
To solve this problem, Yang et al. \cite{2001} developed two algorithms based on iterative soft thresholding (IST) and iterative hard thresholding (IHT). These two algorithms are called $l_{\sigma}$-IST and $l_{\sigma}$-IHT. By verifying the Lipschitz continuity of the gradient of the datafitting risk, $l_{\sigma}$-IST and $l_{\sigma}$-IHT converge. Under proper conditions, the recoverability as well as the linear convergence rate results are obtained. Only RMC-$l_{\sigma}$-IHT was tested on the I2R dataset \cite{203}.

\subsection{Robust Bilateral Factorization (RMC-RBF)}  
For RMC, Shang et al. \cite{1037}\cite{1043} proposed a scalable and provable structured low-rank matrix factorization method to recover the low-rank plus sparse matrices from missing and grossly corrupted data. Thus, a Robust Bilinear Factorization (RBF) method recovered the low-rank plus sparse matrices from incomplete and/or corrupted data, or a small set of linear measurements. The decomposition is the following one:
\begin{equation}
A=L+S=UV^T+S
\label{EquationRBF-1}
\end{equation}
where $U$ and $V$ are two matrices of compatible dimensions, where $U$ has orthogonal columns. Then, the corresponding minimization problem is formulated as follows:
\begin{equation}
\underset{U,V,S}{\text{min}} ~~ \lambda ||V||_* + ||P_{\Omega}(S)||_{l_{1}}   ~~~ \text{subj} ~~~ P_{\Omega}(A)=P_{\Omega}(UV^T+S),U^TU=I
\label{EquationRBF-2}
\end{equation}      
where $\lambda \geq 0$ is the regularization parameter, $||V||_*$ is the nuclear norm of the low-rank matrix $V \in  \mathbf{R}^{m \times n}$, $S \in  \mathbf{R}^{m \times n}$  is the sparse error matrix. $\Omega$ is the index set of observed entries and $P_{\Omega}(.)$ is the projection operator onto that subspace. RBF not only takes into account the fact that the observation is contaminated by additive outliers or missing data, but can also identify both low-rank and sparse noisy components from incomplete and grossly corrupted measurements. So, Shang et al. \cite{1037} developed two small-scale matrix nuclear norm regularized bilinear structured factorization models for RMC problems, in which repetitively calculating SVD of a large-scale matrix is replaced by updating two much smaller factor matrices. Then, Shang et al. \cite{1037}\cite{1043} applied the alternating direction method of multipliers (ADMM) to efficiently solve the RMC problems. Experimental results show that RBF gives similar visual results than RPCA solved via IALM \cite{3} and GRASTA \cite{27} but RBF is more
than 3 times faster than GRASTA \cite{27} and more than 2 times faster than RPCA solved via IALM \cite{3}.

\subsection{Matrix Factorization (RMC-MF)} 
The general RMC problem aims to simultaneously recover both low-rank and sparse components from incomplete and grossly corrupted observations via the
following convex optimization problem:

\begin{equation}
\underset{L,S}{\text{min}} ~~ ||L||_*+ \lambda ||S||_{l_{1}}   ~~~ \text{subj} ~~~ P_{\Omega}(L+S)=P_{\Omega}(A)
\label{EquationMF-1}
\end{equation}   

where $P_{\Omega}(A)$ is defined as the projection of the matrix $A$ on the observed entries $\Omega$: $P_{\Omega}(A_{ij})=A_{ij}$ if $(i,j) \in \Omega$ and $P_{\Omega}(A_{ij})=0$ otherwise. From Equation \ref{EquationMF-1}, Shang et al. \cite{2003} find that the optimal solution $E_{\Omega^C}=0$ where $\Omega^C$ is the complement of $\Omega$ and corresponds to the index set of the unobserved entries. So, the RMC problem is equivalent to the following convex optimization problem:

\begin{equation}
\underset{L,S}{\text{min}} ~~ ||L||_*+ \lambda ||P_{\Omega}(S)||_{l_{1}}   ~~~ \text{subj} ~~~ P_{\Omega}(L+S)=P_{\Omega}(A), E_{\Omega^C}=0
\label{EquationMF-2}
\end{equation}  

To efficiently solve the RMC problem and avoid introducing some auxiliary variables, Shang et al. \cite{2003} assumed that the constraint with a linear projection operator $P_{\Omega}$ can be simplified into $A=L+S$. To further improve the efficiency of this convex model and the scalability of handling large data sets, Shang et al. \cite{2003} proposed a scalable non-convex model in which the desired low-rank matrix $L$ is factorized into two much smaller matrices $U \in \mathbf{R}^{m \times d}$ and $V \in \mathbf{R}^{d \times n}$ where $d$ is
an upper bound for the rank of the matrix $L$, i.e., $d \geq r = \text{rank}(L)$. Thus, the decomposition problem is formulated as follows:
\begin{equation}
A=UV^T+S 
\label{EquationMF-3}
\end{equation}  
Finally, the RMC with Matrix Factorization (RMC-MF) problem is equivalent to the following convex optimization problem:
\begin{equation}
\underset{U,V,S}{\text{min}} ~~ ||V||_* + \lambda ||P_{\Omega}(S)||_{l_{1}} ~~~ \text{subj} ~~~ A=UV^T+S, U^TU=I
\label{EquationMF-4}
\end{equation}  
Experimental results show  that RMC (convex formulation) and RMC-MF (non-convex formulation) are slightly better than that of GRASTA \cite{27} and UNN-BF \cite{2-2}. The theoretical reason for the unsatisfactory performance of the $l_1$-penalty is that the irrepresentable condition is not met. Hence, RMC-MF incorporating with matrix factorization is more accurate in recovering the low-rank matrix than RMC (convex formulation). Furthermore, RMC-MF is more than 7 times faster than RMC (convex formulation), more than 4 times faster than GRASTA \cite{27}, and more than 2 times faster than UNN-BF \cite{2-2}.\\

\noindent \textbf{Algorithms for solving RMC and RMC-MF:} Shang et al. \cite{2003} developed two efficient alternating direction augmented Lagrangian (ADAL) solvers for solving the convex model and the non-convex model, respectively. For the convex problem, the running time of the corresponding algorithm is dominated by that of performing SVD on the matrix of size $m\times n$. For the non-convex problem, the corresponding algorithm performs SVD on much smaller matrices of sizes $m\times d$ and $d\times n$, and some matrix multiplications. Hence, the total time complexity of the algorithm for the convex RMC and the algorithm for the non-convex are $O(tmn^2)$ and $O(t(d^2m + mnd))$ with $(d \ll n<m)$, respectively, where $t$ is the number of iterations.

\subsection{Factorized Robust Matrix Completion (FRMC)}  
Mansour and Vetro \cite{2000} developed a factorized robust matrix completion (FRMC) algorithm with global motion compensation to solve the  background/foreground separation problem for videos with moving background. Since the main drawbacks in Equation \ref{EquationPCP3} is that it requires the computation of full or partial singular value decompositions of $L$ in every iteration of the algorithm, Mansour and Vetro \cite{2000} adopted a surrogate for the nuclear norm of a rank-$r$ matrix $L$ defined by the following factorization: \\

\begin{equation}
||L||_*=\underset{L_{L} \in \mathbf{R}^{m,r}, L_{R} \in \mathbf{R}^{n,r}} {\text{min}} ~~\frac{1}{2} (||L_{L}||_F^2+ ||L_{R}||_F^2)   ~~~ \text{subj} ~~~ L_{L}L_{R}^T=L
\label{EquationFRMC}
\end{equation}

This nuclear norm surrogate can be used in standard nuclear norm minimization algorithms that scale to very large matrix
completion problems. Moreover, it was shown that when the factors $L_{L}$ and $L_{R}$ have a rank greater than or equal
to the true rank of $L$.  Hence, each subproblem in the FRMC algorithm is a Lasso problem that Mansour and Vetro \cite{2000} solved using spectral projected gradient iterations. FRMC was developed in batch mode and online mode. The FRMC algorithm in online mode completes the recovery $7$ to $9$ times faster than GRASTA \cite{27} and results in a comparable separation quality. To apply FMRC in the case of moving cameras, Kao et al. \cite{2000-1} proposed a label propagation scheme, which combines the advantages of FRMC and spectral clustering.

\subsection{Motion-Assisted Matrix Completion (MAMC)} 
Yang et al. \cite{2002}\cite{2002-1} proposed a motion-assisted matrix completion (MAMC) model for foreground-background separation. Thus, a dense motion field is estimated for each frame, and mapped into a weighting matrix  $W_3$ which indicates the likelihood that each pixel belongs to the background as follows:

\begin{equation}
\underset{L,S}{\text{min}} ~~ ||L||_*+ \lambda ||S||_{l_{1}}   ~~~ \text{subj} ~~~ W_3 \circ A = W_3 \circ (L+S)	
\label{EquationMAMC-1}
\end{equation}  
where $\circ$ denotes the element-wise multiplication of two matrices and $W_3$ is constructed from motion information. By incorporating
this information, areas dominated by slowly-moving objects are suppressed while the background that appears at only a few frames has more chances to be recovered in the foreground detection results. The influence of light conditions, camouflages, and dynamic backgrounds can also be decreased. \\

\indent In addition, Yang et al. \cite{2002}\cite{2002-1} extended MAMC to a robust MAMC model (RMAMC) which is robust to noise for practical applications as follows:
\begin{equation}
\underset{L,S,E}{\text{min}} ~~ ||L||_*+ \lambda ||S||_{l_{1}} + \gamma ||E||_F^2   ~~~ \text{subj} ~~~ W_3 \circ A = W_3 \circ (L+S+E)	
\label{EquationMAMC-2}
\end{equation} 
where $\gamma$ is a positive constant and $E$ is the matrix which contains the noise. Yang et al. \cite{2002} adapted the ALM algorithm \cite{18} to solve MAMC and RMAMC. Experimental results \cite{2002}\cite{2002-1} on several datasets show that RMAMC outperforms RPCA solved via IALM \cite{3}.

\section{Robust Subspace Recovery}
\label{sec:RSR}
In this category are the robust decompositions other than RPCA and RNMF decompositions. Table \ref{TPCP3Overview} shows an overview of the different RSR decompositions. Their corresponding solvers as well as their complexity can be seen in Table \ref{A-TPCP3Overview}.

\subsection{Robust Subspace Recovery via Bi-Sparsity (RoSuRe)} 
High dimensional data is distributed in a union of low dimensional subspaces in sparse models but the underlying structure may be affected by sparse errors and/or outliers. To solve this problem, Bian and Krim \cite{1014} proposed a bi-sparsity model and provided an algorithm to recover the
union of subspaces in presence of sparse corruption. Thus, the proposed decomposition is the following one:
\begin{equation}
A = L+S = LW+S
\label{EquationRoSure-1}
\end{equation} 
where $W_4$ is sparse matrix. Then, the corresponding minimization problem is formulated as follows:
\begin{equation}
\underset{W,S}{\text{min}} ~~ ||W_4||_{l_{1}}+ \lambda ||S||_{l_{1}}   ~~~ \text{subj} ~~~ A=L+S, L=LW_4, W_{4,(ii)}=0, \forall i
\label{EquationRoSure-2}
\end{equation}  
Experimental results \cite{1014}\cite{1014-1}\cite{1014-2} on static and moving camera cases show the ability of RoSuRe to separate foreground from background in both cases. More interestingly, the sparse coefficient matrix $W_4$ gives information about the relations among data points, which potentially may be used to cluster data into individual clusters. Indeed, for each column of the coefficient matrix $W_4$, the nonzero entries appear periodically. In the case of the periodic motion of the camera, every frame is mainly represented by the frames when the camera is in a similar position, i.e. a similar background, with the foreground moving objects as outliers. After permuting the rows and columns of $W_4$ according to the position of cameras, a block-diagonal structure can be extracted. Thus, images with similar backgrounds are grouped as one subspace. \\

\noindent \textbf{Algorithms for solving RoSuRe:}  Bian and Krim  \cite{1014} developed an algorithm via Bi-Sparsity Pursuit based on linearized ADMM \cite{39}. Practically, Bian and Krim \cite{1014} pursued the sparsity of $S$ and $W_4$ alternatively until convergence. Besides the effectiveness of ADMM on $l_1$ minimization problems, the augmented Lagrange multiplier (ALM) method can address the non-convexity of Equation \ref{EquationRoSure-2}. It hence follows that with a sufficiently large augmented Lagrange multiplier, the global optimizer is approximated by solving the dual problem.

\subsection{Robust Orthonomal Subspace Learning (ROSL)}    
Shu et al. \cite{759}\cite{759-1} presented a computationally efficient low-rank recovery method, called as Robust Orthonormal Subspace
Learning (ROSL). ROSL speeds the rank-minimization of a matrix $L$ by imposing the group sparsity of its coefficients $\alpha$ under orthonormal subspace spanned by orthonormal bases $D$. Its underlying idea is that, given the subspace representation $L=D\alpha$, the rank
of $L$ is upper bounded by the number of non-zero rows of $\alpha$, that is $||\alpha||_{row-0}$. ROSL can be considered as a non-convex relaxation of RPCA by replacing nuclear norm with this rank heuristic \cite{759}. So, ROSL involved the following decomposition:
\begin{equation}
A=D\alpha+S
\label{EquationROSL-1}
\end{equation}  
Thus, ROSL recovers the low-rank matrix $L$ from $A$ by minimizing the number of non-zero rows of $\alpha$, and the sparsity of $S$ as follows:
\begin{equation}
\underset{S,D, \alpha}{\text{min}} ~~ ||\alpha||_{row-0}+ \lambda ||S||_{l_{0}}   ~~~ \text{subj} ~~~ D \alpha+S=A, D^TD=I_k, \forall i
\label{EquationROSL-2}
\end{equation}  
where the subspace bases $D$ = $U$, the coefficients $\alpha = SV^T$ and $L = USV^T$ obtained by SVD. As the sparsity-inducing $l_1$-norm is an acceptable surrogate for the sparsity measure with $l_0$-norm, Shu et al. \cite{759-1} reformulated ROSL as the following non-convex optimization problem:
\begin{equation}
\underset{S,D, \alpha}{\text{min}} ~~ ||\alpha||_{row-1}+ \lambda ||S||_{l_{1}}   ~~~ \text{subj} ~~~ D \alpha+S=A, D^TD=I_k, \forall i
\label{EquationROSL-3}
\end{equation}  
where the $row-1$-norm is defined as $||\alpha||_{row-1}=\sum_i^k ||\alpha_i||_{l^2}$

Experimental results \cite{759}\cite{759-1} on the I2R dataset \cite{203} show that the recovery accuracy and efficiency of ROSL is slightly better than PCP solved via EALM \cite{18}, PCP solved via IALM \cite{18}, PCP solved by Random Projection \cite{11} and PCP solved by LMaFit \cite{23}. Furthermore, ROSL is more than 10 times faster than RPCA solved via IALM \cite{18}. \\
 
\noindent \textbf{Algorithms for solving ROSL:} Shu et al. \cite{759-1} presented an efficient sparse coding algorithm to minimize this rank measure and recoverthe low-rank matrix at quadratic complexity of the matrix size. This, ROSL is solved using inexact ADM (Alternating Direction Method) at the higher scale and inexact BCD (Block Coordinate Descent) at the lower scale. This solver is called inexact ADM/BCD.  Finally, Shu et al. \cite{759-1} developed a random sampling algorithm to further speed up ROSL such that its accelerated version (ROSL+) has linear complexity with respect to the matrix size. ROSL+ is more than 92 times faster than RPCA solved via IALM \cite{18}.

\subsection{Robust Orthogonal Complement Principal Component Analysis (ROC-PCA)}
She et al. \cite{1042} proposed a robust orthogonal complement principal component analysis (ROC-PCA). The aim is to deal with orthogonal outliers that are not necessarily apparent in the original observation space but could affect the principal subspace estimation. For this, She et al. \cite{1042} introduced a projected mean-shift decomposition as follows:
\begin{equation}
AV_{\perp} = L+S+E
\label{EquationROC-PCA-1}
\end{equation} 
where $V_{\perp}$ is $n \times m$ matrix verifying $V_{\perp}^TV_{\perp}=I$ and characterizes the subspace orthorgonal to the rank-$r$ principal component subspace. $AV_{\perp}$ gives the coordinates after projecting the data onto the orthogonal components subspace and it decomposed into three parts: mean $L$, outlier $S$ and noise $E$. The corresponding minimization problem of Equation \ref{EquationROC-PCA-1} is formulated as follows:

\begin{equation}
\underset{V_{\perp},\mu,S}{\text{min}} ~~ \frac{1}{2}||AV_{\perp}-L-S||_F^2+ \sum_{ij} P(||s_{ij}||_{l^{2}}; \lambda_{i})   ~~~ \text{subj} ~~~ V_{\perp}^TV_{\perp}=I
\label{EquationROC-PCA-2}
\end{equation} 
where $I$ is the identity matrix and $s_{ij}$ is the $i^{th}$ row vector of $S$. $P(S; \lambda)= \sum_{ij} \lambda_{ij}||s_{ij}||_{l^2}$ where $||.||_{l^2}$ allows the minimization to address outliers in a row-wise manner. She et al. \cite{1042} used generalized $M$-estimator to solve this minimization. The computation is related to the orthogonality constraint, in addition to the non-smooth and possibly non-convex $P$, and She et al. \cite{1042} developed a fast alternating optimization algorithm on the basis of Stiefel manifold optimization and iterative nonlinear thresholdings.

\section{Robust Subspace Tracking}
\label{sec:ST}
Subspace tracking addresses the problem when new observations come progressively like in online streaming application. The algorithm cannot store all the input data in memory. Thus, the new incoming observations need to be processed and then discarded.  The involved subspaces can have low-rank and/or sparse structures like in the previous decomposition frameworks. Table \ref{TPCP3Overview} shows an overview of the different RMC decompositions. Their corresponding solvers as well as their complexity can be seen in Table \ref{A-TPCP4Overview}.

\subsection{Grassmannian Subspace Tracking (GRASTA)}
He et al. \cite{27}\cite{28} proposed a Grassmannian robust adaptive subspace tracking algorithm (GRASTA). This algorithm uses a robust $l_1$-norm cost function in order to estimate and track non-stationary subspaces when the streaming data vectors are corrupted with outliers. This problem is solved via an efficient Grassmannian augmented Lagrangian Alternating Direction Method. 
\\
Let denote the evolving subspace of $\mathbf{R}^{n \times m}$ as $S_t$ at time $t$ with its dimension $d$ that is supposed to be much smaller than $m$ and $n$. Let the columns of an $mn \times d$ matrix $U_t$ be orthonormal and span $S_t$. Tracking the subspace $S_t$ is equivalent to estimating $U_t$ at each time step $t$. The observed vector data $A_t$ is assumed to be generated at each time step $t$ as follows:
\begin{equation}
A_t=U_t w_t+S_t+E_t
\label{EquationG1}
\end{equation}
where $U_t w_t=L_t$ has a low-rank structure, and $w_t$ is a $d\times 1$ weight vector. The orthonormal columns of $U-t$ span the low-rank subspace of the images. The set of all subspaces of  $\mathbf{R}^n$ dimension $d$ is called the Grassmannian, which is a compact Riemannian manifold and is denoted by  $G(d,n)$. $S_t$ is the $n\times 1$ sparse outlier vector whose nonzero entries may be arbitrarily large,  and  $S_t$ models foreground pixels in the background/foreground separation. $E_t$ is the $n\times 1$ zero-mean Gaussian white noise with small variance. Then, He et al. \cite{27}\cite{28} subsampled $A_t$ on the index set $\Omega_t\subset \left\{1,...,n\right\}$. So, only a small subset of entries of $A_t$ are kept. $U_{\Omega_t}$ is the submatrix of $U_t$ consisting of the rows indexed by $\Omega_t$. For a vector $A_t \in \mathbf{R}^n$ , $A_{\Omega_t}$ is the vector in $\mathbf{R}^{\left|\Omega_t\right|}$ whose entries are those of $A_t$ indexed by $\Omega_t$. To quantify the subspace error when the data are incomplete and corrupted, GRASTA uses the $l_1$-norm to measure the subspace error from the subspace spanned by the column of $U_t$ to the observed vector $A_{\Omega_t}$:
\begin{equation}
F(S;t)=\underset{w}{\text{min}} ~~ ||U_{\Omega_t}w-A_{\Omega_t}||_{l_{1}}
\label{EquationG2}
\end{equation}
If $U_{\Omega_t}$ is known or can be estimated, this $l_1$-minimization problem can be solved by Alternating Direction Method of Multipliers (ADMM) \cite{111}. According to ADM, Equation (\ref{EquationG2}) is equivalent to the following problem by introducing a sparse outlier vector $S \in \mathbf{R}^{\left|\Omega_t\right|}$:
\begin{equation}
\underset{U,w,S}{\text{min}} ~~ ||S||_{l_{1}}  ~~~ \text{subj} ~~~ A_{\Omega_t}=U_{\Omega_t}w+S, ~~,~~U \in G(d,n)
\label{EquationG4}
\end{equation}
This problem is not convex but it offers much more computationally efficient solutions. Experimental results \cite{27} on the I2R dataset \cite{203} show that GRASTA show more robustness than RPCA solved via IALM \cite{3} and ReProCS \cite{16} with less time requirement. \\

\noindent \textbf{Algorithm for solving GRASTA:} The problem in Equation (\ref{EquationG4}) is solved via the augmented Lagrangian function:
\begin{equation}
L(S,w,y)=||S||_1 +y^T(U_{\Omega_t}w+S-A_{\Omega_t})
\nonumber
\end{equation}
\begin{equation}
+\frac{\rho}{2}||U_{\Omega_t}w+S-A_{\Omega_t}||_{l_{2}}^2
\label{EquationG5}
\end{equation}
where $y$ is the dual vector. The unknown variables are $S$, $w$, $y$ and $U$. If $U$ is fixed, the triple ($S$, $w$, $y$) is solved by ADMM \cite{111} and if the triple $(S,w,y)$ is fixed, $U$ is estimate by Grassmannian geodesic gradient descent \cite{112}. GRASTA is then composed by this alternating approach. The total computational cost of GRASTA is $O(\left|\Omega\right| d^3+Td \left|\Omega\right|+nd^2)$ where $\left|\Omega\right|$ is the number of samples per vector used, $d$ is the dimension of the subspace, $n$ is the ambient dimension, and $T$ is the number of ADMM iterations.\\

\subsection{Transformed Grassmannian Subspace Tracking (t-GRASTA)}
\label{subsec:t-GRASTA}
He et al. \cite{2801}\cite{2802} proposed t-GRASTA (transformed-GRASTA) which iteratively performs incremental gradient descent constrained to the Grassmannian manifold of subspaces in order to simultaneously estimate a decomposition of a collection of images into a low-rank subspace, a sparse part of occlusions and foreground objects, and a transformation such as rotation or translation of the image. Based on RASL (Robust Alignment by Sparse and Low-rank decomposition) \cite{2810} which poses the robust image alignment problem as a transformed version of RPCA, He et al. \cite{2801}\cite{2802} adapted Equation \ref{EquationG4} as follows:
\begin{equation}
\underset{U,w,S,\tau}{\text{min}} ~~ ||S||_{l_{1}}  ~~~ \text{subj} ~~~  A_{\Omega_t} \circ \tau=U_{\Omega_t}w+S ~~,~~U \in G(d,n)
\label{EquationTGRASTA1}
\end{equation}
where $\tau$ are the transformations. He et al. \cite{2801}\cite{2802} developed batch mode and online mode algorithms. For batch mode, $U$ is the iteratively learned aligned subspace in each iteration; while for online mode, $U$ is a collection of subspaces which are used for approximating the
nonlinear transform, and they are updated iteratively for each video frame. To solve t-GRASTA,  He et al. \cite{2801}\cite{2802} used a ADDM solver suitable for the locally linearized problem.Experimental results \cite{2801}\cite{2802} on sequences with simulating camera jitters show that t-GRASTA outperforms RASL \cite{2810} and GRASTA \cite{27}. Furthermore,  t-GRASTA is four faster than state-of-the-art algorithms and has half the memory requirement.

\subsection{Grassmannian Adaptive Stochastic Gradient with $l_{2,1}$-norm (GASG21)}
In the presence of column outliers corruption, He and Zhang \cite{2803} formulated the Grassmannian Adaptive Stochastic Gradient for $l_{2,1}$-norm minimization (GASG21). Moreover, the  classical matrix $l_{2,1}$-norm  minimization problem is formulated in its stochastic programming counterpart. 
\begin{equation}
\underset{w}{\text{min}} ~~ ||Uw-A||_{l_{2,1}}- \sum_{j=1}^m||U_jw_j-A_j||_{l_2} ~~~ \text{subj} ~~~ U \in G(d,n)
\label{EquationGASG21}
\end{equation}
The $l_{2,1}$-norm minimization is well suitable for column outliers corruption. For inliers which can be well represented by the subspace, the residues are small. For outliers which can not be fitted into the subspace, the residues are large. Then Equation \ref{EquationGASG21} means that He and Zhang \cite{2803} are optimizing $U$ which can best fit inliers to reduce the sum of $l_2$ fit residues. To solve GAS21, He and Zhang \cite{2803} solved the $l_{2,1}$- norm minimization by stochastic gradient descent (SGD). Experimental results \cite{2803} show that GAS21 outperfoms slightly OP \cite{56} with less time computation.

\subsection{$L_p$-norm Robust Online Subspace Tracking (pROST)}
GRASTA performs a gradient descent on the Grassmannian and aims at optimizing an $l_1$-cost function to mitigate the
effects of heavy outliers in the subspace tracking stage. He et al. \cite{27} overcame the nondifferentiability of the $l_1$-norm by formulating an augmented Lagrangian optimization problem at the cost of doubling the number of unknown parameters.
Although the $l_1$-norm leads to favorably conditioned optimization problems it is well-known that penalizing with non-convex $l_0$-surrogates allows reconstruction even in the case when $l_1$-based methods fail. Therefore, Hage and Kleinsteuber \cite{69}\cite{6901} proposed a method which used the combination of the Grassmannian optimization and the non-convex sparsity measures. This method called pROST firstly focuses on reconstructing and tracking the underlying subspace. pROST can be applied on both fully and incompletely observed data sets. The involved decomposition is the same than in GRASTA (See Equation \ref{EquationG1}) but the minimization problem is formulated as follows:
\begin{equation}
\underset{Rank(L)\leq k}{\text{min}} ~~ ||UW-A||_{L_{p}}
\label{EquationpROST1}
\end{equation}
where $UW=L$. $||.||_{L_{p}}$ is a smoothed and weighted $L_p$-quasi-norm cost function to achieve robustness against outliers, and it is defined as follows:
\begin{equation}
L_p(X)= \sum_{i=1}^m ~ (x_i^2+ \mu)^{\frac{1}{p}} ~~ 0<p<1
\label{EquationpROST2}
\end{equation}
Corresponding pixels in consecutive frames present a big probability to have the same label due to their spatial and temporal proximity. Thus, Hage and Kleinsteuber \cite{69}\cite{6901} used this knowledge to increase the robustness of the residual cost. The contribution of
labeled foreground pixels to the overall penalty is reduced by introducing additional pixel weights $w_i$. If the pixel was previously labeled a foreground pixel and is then likely to remain an outlier in the current
frame, the weight is small to avoid foreground objects compromising the background. If the pixel is labeled a background pixel, the weight
is equal to one for an adaptive model maintenance. Thus, the weighted smoothed $L_p$-quasi-norm cost function is defined as follows:
\begin{equation}
L_p(X)= \sum_{i=1}^m ~ w_i(x_i^2+ \mu)^{\frac{1}{p}} ~~ 0<p<1
\label{EquationpROST3}
\end{equation}
Then, an alternating online optimization framework for estimating the subspace makes the algorithm suitable for online subspace tracking. In contrast to GRASTA, the method presented directly optimizes the cost function and thus operates with less than half the number of unknowns. \\
\indent pROST can be applied in real-time background/foreground separation and makes use of the spatio-temporal dependencies between pixel labels. This leads to robustness in presence of bootstrapping, large foreground objects (which often arise in RPCA-based methods) and jittery cameras. Experimental results \cite{6901} on the ChangeDetection.net dataset \cite{202} confirm that the proposed method can cope with more outliers and with an underlying matrix of higher rank than GRASTA. Particularly, pROST outperforms GRASTA in the case of multi-modal backgrounds. \\

\noindent \textbf{Algorithm for solving pROST:}
Hage and Kleinsteuber \cite{69} \cite{6901} used a Conjugate Gradient (CG) type algorithm on the Grassmannian for solving the individual minimization tasks. Like all optimization methods on the Grassmannian, the algorithm allows to upper-bound the dimension of the underlying subspace and easily extends to the problem of robustly tracking this subspace. 

\subsection{Grassmannian Online Subspace Updates with Structured-sparsity (GOSUS)}
Xu et al. \cite{85} studied the problem of online subspace learning when sequential observations involves structured perturbations. As the observations are an unknown mixture of two components presented to the model sequentially, if no additional constraints is imposed on the residual, it often corresponds to noise terms in the signal which were unaccounted for by the main effect. To address this problem, Xu et al. \cite{85} imposed structural contiguity, which has the effect of leveraging the secondary terms as a covariate that helps the estimation of the subspace itself, instead of merely serving as a noise residual. 
\begin{equation}
\underset{U^TU=I_d,W,S}{\text{min}} ~~ \sum_{i=1}^{l} \mu_i||D^iS||_{l_{2}}+\frac{\lambda}{2}||UW+S-A||_{l_{2}}^2
\label{EquationGOSUS1}
\end{equation}
where $UW=L$ possess a low-rank structure, $S$ possess a sparse structure, and $D^i$ is diagonal matrix which denotes a "group" $i$. Each diagonal element of $D^i$ corresponds to the presence/absence of a pixel in the $i^th$ group. So, $D_{ij}$ is equal to one if pixel $j$ is in group $i$, and it is equal to zero otherwise. Thus, the term  $\sum_{i=1}^{l} \mu_i||D^iS||_{l_{2}}$ is penalty function where $\mu_i$ is the weight for group $i$ and $l$ is the number of such groups. This group sparsity function has a mixed norm structure. The inner norm is either $l_2$ forcing pixels in the corresponding group to be similarly weighted, and the outer norm is $l_1$ which encourages sparsity, that is.only few groups are selected.  The corresponding online estimation procedure for Equation \ref{EquationGOSUS1} is written as an approximate optimization process on a Grassmannian, called Grassmannian Online Subspace Updates with Structured-sparsity (GOSUS). GOSUS is solvable via an alternating direction method of multipliers (ADMM) \cite{111} applied in a block-wise manner. Experimental results \cite{85} on the Wallflower dataset and the I2R dataset \cite{203} show that GOSUS outperforms slightly RPCA solved via IALM \cite{18}, GRASTA\cite{27} and BRMF \cite{5401}.

\subsection{Fast Adaptive Robust Subspace Tracking (FARST)}
In spite of these good properties, the global convergence of GRASTA is not proved as developed in \cite{1035}. Empirically, it is slowly adapted to background change or not adapted to dynamic background. On the contrary, FARST \cite{1035} has an optimal global convergence while sharing some favorable properties with GRASTA. FARST shares the procedure of separating frames into background and foreground with
GRASTA, but it uses a recursive least square algorithm for subspace tracking, which makes it fast adapted to background change and dynamics.
Every time a video frame streams in, two alternating procedures are repeatedly done. First, basis images are updated by a recursive least
square algorithm. Secondly, foreground images are extracted by solving the $l_1$-minimization problem. Furthermore, FARST is an online algorithm fast adapted to background change. Results \cite{1035} show that FARST outperforms GRASTA \cite{27} and PRMF \cite{5400} in the presence of dynamic backgrounds. FARST is solvable via an alternating direction method of multipliers (ADMM) \cite{111}.

\section{Robust Low Rank Minimization}
\label{sec:LRM}
Low-rank minimization is a minimization problem involving a cost function which measures the fit between a given data matrix $A$ and an approximating low-rank matrix $L$. Table \ref{LRMOverview} shows an overview of the different Robust Low Rank Minimization (RLRM) methods and the corresponding solvers. Furthermore, the complexity of the solvers can be seen in Table \ref{A-TPCP4Overview}.

\begin{table*}
\scalebox{0.75}{
\begin{tabular}{|l|l|} 
\hline
\scriptsize{Categories} &\scriptsize{Authors - Dates} \\
\hline
\hline
\scriptsize{\textbf{Decompositions}}  &\scriptsize{} \\
\hline
\scriptsize{Contiguous Outlier Detection (DECOLOR)}         &\scriptsize{Zhou et al. (2011) \cite{25}}     \\
\scriptsize{Direct Robust Matrix Factorization (DRMF)}             			 &\scriptsize{Xiong et al. (2011) \cite{54}}    \\
\scriptsize{Direct Robust Matrix Factorization-Rows (DRMF-R)}            &\scriptsize{Xiong et al. (2011) \cite{54}}    \\
\scriptsize{Probabilistic Robust Matrix Factorization (PRMF)}  					 &\scriptsize{Wang et al. (2012) \cite{5400}}   \\
\scriptsize{Bayesian Robust Matrix Factorization (BRMF)}         			   &\scriptsize{Wang et al. (2013) \cite{5401}}   \\
\scriptsize{Markov Bayesian Robust Matrix Factorization (MBRMF)}         &\scriptsize{Wang et al. (2013) \cite{5401}}   \\
\scriptsize{Practical Low-Rank Matrix Factorization (PLRMF)}         		 &\scriptsize{Zheng et al. (2012) \cite{1025}}  \\
\scriptsize{Low Rank Matrix Factorization with MoG noise (LRMF-MOG)}     &\scriptsize{Meng et al. (2013) \cite{2-1}}    \\
\scriptsize{Unifying Nuclear Norm and Bilinear Factorization (UNN-BF)}   &\scriptsize{Cabral et al. (2013) \cite{2-2}}  \\
\scriptsize{Low Rank Matrix Factorization with General Mixture noise (LRMF-GM)}   &\scriptsize{Cao et al. (2013) \cite{2-3}}  \\
\scriptsize{Robust Rank Factorization (RRF)}         	                	 &\scriptsize{Sheng et al. (2014) \cite{1027}}  \\
\scriptsize{Variational Bayesian Method (VBMF-$l_1$)}                    &\scriptsize{Zhao et al. (2015) \cite{1024-1}} \\
\scriptsize{Robust Orthogonal Matrix Factorization (ROMF)}               &\scriptsize{Kim and Oh (2015) \cite{1074}}    \\
\scriptsize{Contiguous Outliers Representation via Online Low-Rank Approximation (COROLA)}  &\scriptsize{Shakeri and Zhang (2015) \cite{1077}}  \\
\scriptsize{Online Low Rank Matrix Completion (ORLRMR)}                    &\scriptsize{Guo (2015) \cite{1120}}  \\ 
\scriptsize{Matrix Factorization - Elastic-net Regularization (FactEN)}    &\scriptsize{Kim et al. (2015) \cite{1310}}    \\
\scriptsize{Incremental Learning Low Rank Representation - Spatial Constraint (LSVD-LRR)}  &\scriptsize{Dou et al. (2015) \cite{1330}}  \\
\scriptsize{Online Robust Low Rank Matrix Recovery (ORLRMR)}                               &\scriptsize{Guo (2015) \cite{1536}}  \\
\hline
\scriptsize{\textbf{Solvers}} &\scriptsize{} \\
\hline
\scriptsize{Alternating Algorithm (AA)}                          &\scriptsize{Zhou et al. (2011) \cite{25}}          \\
\scriptsize{Block Coordinate Descent Strategy (BCDS)}            &\scriptsize{Xiong et al. (2001) \cite{54}}         \\
\scriptsize{Conditional EM Algorithm (CEM)}                      &\scriptsize{Jebara and Pentland (1999) \cite{115}} \\
\scriptsize{Augmented Lagrangian Multiplier (ALM)}               &\scriptsize{Zheng et al. (2012) \cite{1025}}       \\
\scriptsize{Alternative Direction Descent Algorithm (ADDA)}      &\scriptsize{Sheng et al. (2014) \cite{1027}}       \\
\hline
\scriptsize{\textbf{Spatial and Temporal Algorithms }} &\scriptsize{} \\
\scriptsize{Markov Random Field}                                  &\scriptsize{Zhou et al. (2011) \cite{25}}     \\
\scriptsize{Markov Random Field}                                  &\scriptsize{Wang et al. (2013) \cite{5401}}   \\
\hline
\end{tabular}}
\caption{Robust Low Rank Minimization: A Complete Overview. The first column indicates the concerned category and the second column the name of each method. Their corresponding acronym is indicated in the first parenthesis. The third column gives the name of the authors and the date of the related publication.} 
\centering
\label{LRMOverview}
\end{table*}

\subsection{LRM with contiguous outliers detection (DECOLOR)}
Zhou et al. \cite{25} proposed a formulation of outlier detection in the low-rank representation, in which the outlier support and the low-rank matrix are estimated. This method is called Detecting Contiguous Outlier detection in the Low-rank Representation (DECOLOR). 
The decomposition involves the same model than SPCP in Equation (\ref{EquationSPCP1}), that is $A=L+S+E$. So, the following formulation is proposed to achieve the decomposition:
\begin{equation}
\underset{L,S}{\text{min}} ~~ \alpha rank(L) + \beta ||S||_{l_{0}} + \frac{1}{2}||A-L-S||_F^2 
\nonumber
\end{equation}
\begin{equation}
 ~~~ \text{subj} ~~~ rank(L) \leq r
\label{EquationDC1}
\end{equation}
where $\alpha$ and $\beta$ are regularization parameters. Then, Zhou et al. \cite{25} defined the foreground support matrix of $S$, denoted $F \in \left\{0,1\right\}^{m \times n}$ as follows: \\

\begin{center}
$F_{ij}=0$ if the pixel $ij$ is background \\
$F_{ij}=1$ if the pixel $ij$ is foreground \\
\end{center}

Suppose that $(L^*,S^*)$ is a minimizer of Equation (\ref{EquationDC1}). As long as $S_{ij}^* \neq 0$, $S_{ij}^*=A_{ij}-L{ij}^*$ minimizes Equation (\ref{EquationDC1}). That is:\\

\begin{center}
$A_{ij}-L{ij}^*-S_{ij}^*=A_{ij}-L{ij}^* ~~~ \text{if} ~~~ S_{ij}^*=0 ~~(F_{ij}=0)$ \\
$A_{ij}-L{ij}^*-S_{ij}^*=0 ~~~ \text{if} ~~~ S_{ij}^* \neq 0 ~~ (F_{ij}=1)$ \\
\end{center}

Thus, the following equation is obtained:
\begin{equation}
\underset{L,S}{\text{min}} ~~ \alpha rank(L) + \beta ||S||_{l_{0}} + \frac{1}{2} \sum_{(i,j):F_{ij}=0} (A_{ij}-L_{ij})^2 
\nonumber
\end{equation}
\begin{equation}
 ~~~ \text{subj} ~~~ rank(L) \leq r
\label{EquationDC2}
\end{equation}
Let $P_F(X)$ represents the orthogonal projection of the matrix $X$ onto the linear space of matrices supported by $F$:
\begin{equation}
P_{F}(X)(i,j)=0 ~~~ \text{if} ~~~ F_{ij}=0 
\label{EquationDC3}
\end{equation}
\begin{equation}
P_{F}(X)(i,j)=X_{ij} ~~~ \text{if} ~~~ F_{ij}=1
\label{EquationDC4}
\end{equation}
and $P_{\bar{F}}(X)$ is its complementary projection, i.e.  $P_F(X)+P_{\bar{F}}(X)=X$.
Thus, Equation (\ref{EquationDC5}) is obtained:
\begin{equation}
\underset{L,S}{\text{min}} ~~ \alpha rank(L) + \beta ||S||_{l_{0}} + \frac{1}{2} |P_{\bar{F}}(L-S)||_F^2 
\nonumber
\end{equation}
\begin{equation}
~~~ \text{subj} ~~~ rank(L) \leq r
\label{EquationDC5}
\end{equation}
The binary states of entries in foreground support $F$ are modeled by a Markov Random Field because the foreground objects are contiguous pieces with relatively small size. Based on the first order MRFs, the following regularizer on $F$ is used:
\begin{equation}
||C vec(F)||_{l^1} = \sum_{(ij,kl) \in N} \left|F_{ij}-F_{kl}\right|
\label{EquationDC6}
\end{equation}
where $C$ is the node-edge incidence matrix of a graph $G$ with $m \times n$ nodes, and $N$ is the set of all pairs of adjacent nodes in $G$. $ij$ and $kl$ denote the nodes corresponding to $S_{ij}$ and $S_{kl}$, respectively.
Then, the following energy function is obtained by relaxing $rank(L)$ with the nuclear norm and adding the continuity constraint on $F$:
\begin{equation}
\underset{L,S}{\text{min}} ~~  \alpha ||L||_* +  \beta ||F||_{l_{1}} + \frac{1}{2}||P_{\bar{F}}(A-L)||_F^2 + \gamma ||C vec(F)||_{l^{1}}
\label{EquationDC7}
\end{equation}
where $\alpha>0$ is a parameter which controls the complexity of the background model. $S$ is recovered by $S=P_F(L-S)$. \\
Experimental results \cite{25} on the I2R dataset \cite{203} show that DECOLOR outperforms PCP solved via IALM \cite{3} and two conventional models which are the mean model and the MOG model \cite{700} but the main drawback of DECOLOR is its prohibitive computation time. However, the code called DECOLOR\protect\footnotemark[33] is provided. Zhou and Jin \cite{1940} improved DECOLOR by using two-dimension principal component analysis (2DPCA) rather than traditional principal component analysis (PCA) to obtain the principal components of background. \\

\footnotetext[33]{{http://bioinformatics.ust.hk/decolor/decolor.html}}

\noindent \textbf{Algorithm for solving DECOLOR:} The objective function is non-convex and it includes both continuous and discrete variable. Zhou et al. \cite{35} adopted an alternating algorithm that separates the energy minimization over $B$ and $F$ into two steps. $B$-step is a convex optimization problem and $F$-step is a combinatorial optimization problem. The optimal $B$ is computed efficiently by the SOFT-IMPUTE \cite{103} algorithm and the outlier support $S$ is estimated in polynomial time using graph-cut \cite{104}\cite{105}.

\subsection{LRM with Direct Robust Matrix Factorization (DRMF)}
\label{sec:DRMF}
Xiong et al. \cite{54} proposed a direct robust matrix factorization (DRMF) assuming that a small portion of the
matrix $A$ has been corrupted by some arbitrary outliers. The aim is to get a reliable estimation of the true low-rank structure
of this matrix and to identify the outliers. To achieve this, the outliers are excluded from the model estimation. The decomposition involves the same model than PCP in Equation (\ref{EquationPCP1}), that is $A=L+S$. The direct formulation of DRMF is written as follows:
\begin{equation}
\underset{L,S}{\text{min}} ~~ ||A-S-L||_F ~~~ \text{subj} ~~~ rank(L) \leq r, ||S||_{l_{0}} \leq p
\label{EquationDNMF1}
\end{equation}
where $L$ is the low-rank approximation, $r$ is the rank, $S$ is the matrix of outliers, and $p$ is the maximal number of entries that can be ignored as outliers. Comparing DRMF to the conventional LRM, the difference is that the outliers $S$ can be excluded from the low-rank approximation, as long as the number of outliers is not too large, that is, $S$ is sufficiently sparse. By excluding the outliers from the low-rank approximation, Xiong et al. \cite{54} ensured the reliability of the estimated low-rank structure. PCP \cite{3} is the convex relaxation of DRMF. \\
Experimental results \cite{54} on the I2R dataset \cite{203} show that DRMF outperforms SVD \cite{3000}, PCP solved via IALM \cite{3} and SPCP \cite{5}. Furthermore, the code called DRMF\protect\footnotemark[34] is provided. \\

\footnotetext[34]{{http://www.cs.cmu.edu/~lxiong/}}

\noindent \textbf{Algorithm for solving DRMF:}
Optimization problems involving the rank or the $l_0$-norm that is  set cardinality are difficult to solve. Nevertheless, the DRMF problem admits a simple solution due to its decomposable structure in $L$ and $S$. This problem can be solved by a block coordinate descent strategy. First, the current outliers $S$ are fixed. Secondly, they are excluded from $A$ to get the "clean" data $C$. Then, Xiong et al. \cite{54} fit $L$ based on $C$. Then, the outliers $S$ are updated based on the current errors $E=A-L$. The algorithm solved the factorization problem as follows:
\begin{equation}
\underset{L,S}{\text{arg min}_L} ~~ ||C-L||_F ~~~ \text{subj} ~~~ rank(L) \leq r
\label{EquationDNMF2}
\end{equation}
where $C=A-S$.Then, the outlier detection problem is solved as follows: 
\begin{equation}
\underset{L,S}{\text{arg min}_S} ~~ ||E-L||_F ~~~ \text{subj} ~~~  ||S||_{l_{0}} \leq p
\label{EquationDNMF3}
\end{equation}
where $E=A-L$. The solution to the low-rank approximation problem (\ref{EquationDNMF2}) is directly given by SVD. Since only
the first $r$ singular vectors are required, the computation is accelerated using a partial SVD algorithm.

\subsection{LRM with Direct Robust Matrix Factorization for Rows (DRMF-R)}
\label{sec:DRMF-R}
Xiong et al. \cite{54} proposed an extension of DRMF to deal with the presence of outliers in entire columns. This method is called DRMF-Row (DRMF-R). Instead of counting the number of outlier entries, the number of of outliers patterns is counted using the structured $l_{2,0}$-norm. The direct formulation of DRMF-R is written as follows:
\begin{equation}
\underset{L,S}{\text{min}} ~~ ||A-S-L||_F ~~~ \text{subj} ~~~ rank(L) \leq r, ||S||_{l_{2,0}} \leq p
\label{DNMF}
\end{equation}
where $p$ is the maximal number of outlier rows allowed. OP \cite{56} is the convex relaxation of DRMF-R.  No experimental results in Xiong et al. \cite{54} are provided for DRMF-R on background/foreground separation.

\subsection{Probabilistic Robust Matrix Factorization (PRMF)}
Wang et al. \cite{5400} proposed a  probabilistic method for robust matrix factorization (PRMF) based on the $l_1$-norm loss and $l_2$-regularizer, which bear duality with the Laplace error and Gaussian prior, respectively. For model learning, Wang et al. \cite{5400} used an efficient expectation-
maximization (EM) algorithm by exploiting a hierarchical representation of the Laplace distribution as a scaled mixture of Gaussians. So, Wang et al. \cite{5400} considered the following probabilistic model:

\begin{equation}
\begin{array}{lll}
A=UV^\prime+S   ~~~ \text{subj} ~~~ \\
U_{ij}| \lambda_U \sim N(U_{ij} | 0, \lambda_U^{-1}), \\
V_{ij}| \lambda_V \sim N(V_{ij} | 0, \lambda_V^{-1})
\end{array}
\label{EquationPRMF-1}
\end{equation}
where $UB^{-1}BV^\prime=UV^\prime$ holds for any $r \times r$ non singular matrix $B$. By exploiting $U$ and $V$ as model parameters with $\lambda_U$, $\lambda_V$ and $\lambda$ as hyperparameters with fixed values, MAP estimation is used to find $U$ and $V$. From the rule of Bayes, the following equivalence can be written:
\begin{equation}
p(U,V|A,\lambda,\lambda_U,\lambda_V)\propto p(A|U,V,\lambda) p(U,\lambda_U) p(V,\lambda_V)
\label{EquationPRMF-2} 
\end{equation}
Thus, 
\begin{equation}
log~(p(U,V|A,\lambda,\lambda_U,\lambda_V))=-\lambda ||A-UV^\prime||_{l_{1}} - \frac{\lambda_U}{2} ||U||_{l_{2}}^2 - \frac{\lambda_V}{2} ||V||_{l_{2}}^2 + C
\label{EquationPRMF-3} 
\end{equation}
where $C$ is a constant term independent of $U$ and $V$. Maximizing $log~(p(U,V|A,\lambda,\lambda_U,\lambda_V))$ w.r.t. $U$ and $V$ is equivalent to the following minimization problem:
\begin{equation}
\underset{U,V}{\text{min}} ~~ ||A-UV^\prime||_{l_{1}} +  \frac{\lambda_U^ \prime}{2}||U||_{l_{2}}^2 +  \frac{\lambda_V^\prime}{2} ||V||_{l_{2}}^2
\label{EquationPRMF-4}
\end{equation}
where $\lambda_U^\prime=\frac{\lambda_U}{\lambda}$ and $\lambda_V^\prime=\frac{\lambda_V}{\lambda}$. Experimental results \cite{5400} on the I2R dataset \cite{203} show that PRMF gives similar visual results than PCP solved via IALM \cite{3}, GoDec \cite{6} and BRPCA \cite{7} with less computation time. The corresponding code for PRMF\protect\footnotemark[35] are provided in batch mode and online mode.\\

\footnotetext[35]{{http://winsty.net/prmf.html}}

\noindent \textbf{Algorithm for solving PRMF:}
While the model formulation given in Equation \ref{EquationPRMF-3} is rather straightforward, solving the optimization problem directly would be computationally challenging due to the non-smooth nature of the Laplace distribution. To address this computational issue,
Wang et al. \cite{5400} reformulated the model by exploiting a two-level hierarchical representation of the Laplace distribution and EM algorithm is then used to solve this hierarchical model.

\subsection{Bayesian Robust Matrix Factorization (BRMF)}         			   
PRMF present the following drawbacks: 1) it assumed that the basis and coefficient matrices are generated from zero-mean fixed-variance Gaussian distributions. This  assumption is too restrictive, limiting the model flexibility needed for many real-world applications, 2) PRMF treats each pixel independently with no clustering effect but the moving objects in the foreground usually form groups with high within-group spatial or temporal proximity, 3) the loss function is defined based on the $l_1$ norm and it results to be not robust enough when the number of outliers is large. To address these three limitations, Wang et al. \cite{5401} proposed a full Bayesian formulation called Bayesian Robust Matrix Factorization (BRMF). The framework is similar to PRMF but the differences are the following ones:
\begin{enumerate}
\item For the generative proces, BRMF assumes that the mean vectors and precision matrices of the rows of $U$ and $V$ have conjugate priors. Learning the mean vectors offers more flexibility for the generation of $A$ and learning the precision matrices captures the correlation between different features.
\item BRMF used a Laplace mixture with the generalized inverse Gaussian distribution as the noise model to further enhance model robustness. 
\item BRMF contained a Markov extension (MBRMF) which assumes that the outliers exhibit spatial or temporal proximity. 
\end{enumerate}

Experimental results \cite{5401} on the SABS dataset \cite{201} show that MBRMF outperforms PCP solved via IALM \cite{3}, BRPCA \cite{7}, VBRPCA \cite{24}, DECOLOR \cite{25} and PRMF \cite{5400}. The corresponding codes for BRMF\protect\footnotemark[36] and MBRMF\protect\footnotemark[36] are provided.

\footnotetext[36]{{http://winsty.net/brmf.html}}

\subsection{Practical Low-Rank Matrix Factorization (PLRMF)}         		
Several LRM methods usually fail to minimize the $l_1$-based nonconvex objective function sufficiently. Zheng et al. \cite{1025} proposed to add
a convex nuclear-norm regularization term to improve convergence, without introducing too much heterogenous information.  Thus, the robust  $l_1$-norm is choosen as the measurement. Then,  Zheng et al. \cite{1025} enforced $U$ to be column orthogonal to shrink the solution space, and added a nuclear- norm regularization term so as to improve convergence. This method called Practical Low-Rank Matrix Factorization (PLRMF) is based on the following problem:

\begin{equation}
\underset{U,V}{\text{min}} ~~ ||W_5 \odot (A-UV) ||_{l_{1}} 
\label{EquationPLRMF-1}
\end{equation}
where $A$ is the data matrix with a priori rank of $r$. The unknown variables are $U \in \mathbf{R}^{m \times r}$ and $V \in \mathbf{R}^{r \times n}$, which account for the rank constraint implicitly. $W_5$ is an indicator matrix of the same size as $A$ where the entry value of $1$ means that the component at the same position in $A$ is observed, and $0$ otherwise. The operator  $\odot$ denotes the Hadamard element-wise matrix multiplication. $U$ is constrained to be an orthogonal matrix and the following minimization problem is then obtained:
\begin{equation}
\underset{U,V}{\text{min}} ~~ ||W_5 \odot (A-UV) ||_{l_{1}} ~~~ \text{subj} ~~~ U^TU=I_{r}
\label{EquationPLRMF-2}
\end{equation}
Therefore, Zheng et al. \cite{1025} used a nuclear-norm regularizer $||B||_*=||UV||_*$ where  $||UV||_*=||V||_*=$ due to $U^TU=I_{r}$. The regularized minimization problem is then formulated as follows:
\begin{equation}
\underset{U,V}{\text{min}} ~~ ||W_5 \odot (A-UV) ||_{l_{1}} + \lambda ||V||_*~~~ \text{subj} ~~~ U^TU=I_{r}
\label{EquationPLRMF-3}
\end{equation}
where $\lambda$ is is a weighting factor and should be small enough to keep the regularized objective in Equation \ref{EquationPLRMF-3}.  Experimental results \cite{1025} are presented on Structure from Motion (SFM) but not on background/foreground separation. \\

\noindent \textbf{Algorithm for solving PLRMF (RegL1-ALM):}  Zheng et al. \cite{1025} developed a scalable first-order optimization algorithm to solve the regularized formulation on the basis of the augmented Lagrange multiplier (ALM) method with Gauss-Seidel iteration. The corresponding code called  RegL1-ALM\protect\footnotemark[37] is provided. \\ 
 
\footnotetext[37]{{https://sites.google.com/site/yinqiangzheng/}}

\subsection{Low Rank Matrix Factorization with MoG noise (LRMF-MOG)}     
The previous low-rank factorization used loss functions such as the $l_2$-norm and $l_1$-norm losses. $l_2$-norm is optimal for Gaussian noise, while $l_1$-norm  is for Laplacian distributed noise. However, real data in video are often corrupted by an unknown noise distribution, which is unlikely to be purely Gaussian or Laplacian. To address this problem, Meng et al. \cite{2-1} proposed a low-rank matrix factorization problem with a Mixture of Gaussians (MoG) noise model. Since the MoG model is a universal estimator for any continuous distribution, it is able to represent a wider range of noise distributions. The parameters of the MoG model are estimated with a maximum likelihood method, while the subspace is computed with standard approaches. Thus, the decomposition is made as follows: $A=UV^T+S$. Given the likelihood related to the MoG distribution, the aim is to maximize the loglikelihood function w.r.t the MoG parameters ($\Pi$,$\Sigma$) and the LRMF parameters ($U$, $V$) as follows:
\begin{equation}
\underset{U,V,\Pi,\Sigma}{\text{max}} ~~ \sum_{i,j \in \Omega} \sum_{k=1}^K \pi_k \mathbf{N}(x_{ij}|(u^i)^Tv^j, \sigma_k^2)
\label{EquationLRMF-MOG-1}
\end{equation}
with the MoG distribution constraints on $S$. Experimental results \cite{2-1} on the I2R dataset \cite{203} show that LRMF-MOG outperforms SVD \cite{3000}, RSL \cite{1}, PCP solved via IALM \cite{3}, CWM \cite{3001}, PCA-$l_1$ \cite{3001}.

\subsection{Unifying Nuclear Norm and Bilinear Factorization}     
Cabral et al. \cite{2-2} proposed a low-rank matrix decomposition which can be achieved with both bilinear factorization and nuclear norm regularization models. By analyzing the conditions under which these two decompositions are equivalent, Cabral et al. \cite{2-2} proposed a unified model that inherits the benefits of both which is formulated as follows:

\begin{equation}
\underset{U,V}{\text{min}} ~~ f(A-UV^T) + \frac{\lambda}{2}(||U||_F^2 + \lambda ||V||_F^2)
\label{EquationUNNBF-1}
\end{equation}
where $L=UV^T$ is  the low-rank matrix with known rank $r$. Cabral et al. \cite{2-2} showed that the existence of local minima in Equation \ref{EquationUNNBF-1} depends only on the dimension $r$ imposed on matrices $U$ and $V$. Equation \ref{EquationUNNBF-1} can be written as follows:
\begin{equation}
\underset{L,U,V}{\text{min}} ~~ ||W_5 \odot (A-L) ||_{l_{1}} + \frac{\lambda}{2}(||U||_F^2 + \lambda ||V||_F^2)  ~~~ \text{subj} ~~~ L=UV^T
\label{EquationUNNBF-2}
\end{equation}
where $W_5 \in \mathbf{R}^{m \times n}$  is a weight matrix that is used to denote missing data (i.e., $w_{5,(ij)} = 0$) and observed data (i.e., $w_{5,(ij)}=1$). Furthermore, Cabral et al. \cite{2-2} developed an ALM (Augmented Lagrange Multiplier) to solve Equation \ref{EquationUNNBF-1}. \\
\indent Experimental results \cite{2-2} on the I2R dataset \cite{203} show that UNNBF outperforms PCP solved via IALM \cite{3}, GRASTA \cite{27} and PRMF \cite{5400} with less computation times.

\subsection{Robust Rank Factorization (RRF)}      
Sheng et al. \cite{1027} proposed a $l_1$-regularized Outlier Isolation and REgression (LOIRE) model. The measurement process is written in the following decomposition form:
\begin{equation}
A=L+S+E=BX+S+E
\label{EquationRRF-1}
\end{equation}
where $A$ the observation through $B$, $S$ denotes the outlier vector and $E$ denotes a dense Gaussian noise. By adding a penalty term  $||S||_1$ to the least mean squares on $E$, Sheng et al. \cite{1027} derived a minimization problem for $X$ as follows:
\begin{equation}
\underset{S,E}{\text{min}} ~~ ||E||_{l_{2}} + \mu ||S||_{l_{1}}   ~~~ \text{subj} ~~~ \mu>0, A=BX+S+E
\label{EquationRRF-2}
\end{equation}
where $\mu$ is a regularization term. This formulation is then extended to realize robust rank factorization which can be applied to recover low-rank structures from massive contaminations.
\begin{equation}
\underset{S,X}{\text{min}} ~~ ||A-BX-S||_{l_{2}}^2 + \frac{\lambda}{2}||S||_{l_{1}}   ~~~ \text{subj} ~~~ \mu>0, A=BX+S+E
\label{EquationRRF-3}
\end{equation}
where the matrix $B$ is generally unknown, then a simple way to find a most appropriate matrix $B$ that fits the problem is to search one that minimizes the above optimization problem:
\begin{equation}
\underset{B}{\text{min}} ~~ \underset{S,X}{\text{min}} ~~ ||A-BX-S||_{l_{2}}^2 + \frac{\lambda}{2}||S||_{l_{1}}   ~~~ \text{subj} ~~~ \lambda>0, A=BX+S+E
\label{EquationRRF-4}
\end{equation}
To ensure a unique solution for matrix $B$ and $X$, each column of $B$ should have a unit length. To solve LOIRE, Sheng et al. \cite{1027} developed an 
Alternative Matrix Descent Algorithm (AMDA). Experimental results \cite{1027} on the I2R dataset \cite{203} show that LOIRE outperforms the
PCP solved via IALM \cite{3} and SemiSoft GoDec \cite{6} in terms of computation efficiency with a similar visual accuracy.

\section{Experimental Evaluation}
\label{sec:ER}
First, we remind the challenges met in video-surveillance, a brief description of the Background Models Challenges (BMC) dataset \cite{205} and the measures used for the performance evaluation. Then, we provide the evaluation and the comparison on 32 algorithms of decomposition into low-rank plus additive matrices. We have chosen the BMC dataset as a recent large-scale dataset in order to compare these algorithms and to allow comparison with the experimentations conducted in the survey on the PCA framework \cite{510}. Qualitative and quantitative results are provided and discussed among this dataset. Several codes are available in the LRSLibrary \cite{550} and some software packages for partial SVD computation can be found in \cite{1001}. Furthermore, we have grouped different results found in literature of algorithms for which the code is not available and the experimentation were conducted by their authors on the I2R dataset \cite{203} and the ChangeDetection.net dataset \cite{202}.

\subsection{Challenges in Video Surveillance}
\label{sec:Challenges}
Three main conditions assure a good functioning of the background subtraction in video surveillance: the camera is fixed, the illumination is constant and the background is static. In practice, several challenges appear and perturb this process. They are the following ones:
\begin{itemize}
\item \textbf{Noise image:} It is due to a poor quality image source such as images acquired by a web cam or images after compression.
\item \textbf{Camera jitter:} In some conditions, the wind may cause the camera to sway back and so it cause nominal motion in the sequence. Foreground mask show false detections due to the motion without a robust maintenance mechanism.
\item \textbf{Camera automatic adjustments:} Many modern cameras have auto focus, automatic gain control, automatic white balance and
auto brightness control. These adjustments modify the dynamic in the color levels between different frames in the sequence.
\item \textbf{Illumination changes:} They can be gradual such as ones in a day in an outdoor scene or sudden such as a light switch in an indoor scene. \item \textbf{Bootstrapping:} During the training period, the background is not available in some environments. Then, it is impossible to compute a representative background image.
\item \textbf{Camouflage:} A foreground object's pixel characteristics may be subsumed by the modeled background. Then, the foreground and the background can be distinguished. 
\item \textbf{Foreground aperture:} When a moved object has uniform colored regions, changes inside these regions may not be detected. Thus, the entire object may not appear as foreground. Foreground masks contain false negative detections.
\item \textbf{Moved background objects:} Background objects can be moved. These objects should not be considered part of the foreground.
Generally, both the initial and the new position of the object are detected without a robust maintenance mechanism.
\item \textbf{Inserted background objects:} A new background object can be inserted. These objects should not be considered part of the
foreground. Generally, the inserted background object is detected without a robust maintenance mechanism.
\item \textbf{Dynamic backgrounds:} Backgrounds can vacillate and this requires models which can represent disjoint sets of
pixel values.
\item \textbf{Beginning moving object:} When an object initially in the background moves, both it and the newly revealed parts of the background called "ghost" are detected. 
\item \textbf{Sleeping foreground object:} Foreground object that becomes motionless cannot be distinguished from a background
object and then it will be incorporated in the background. How to manage this situation depends on the context.
Indeed, in some applications, motionless foreground objects must be incorporated and in others it is not the case.
\item \textbf{Shadows:} Shadows can be detected as foreground and can come from background objects or moving objects \cite{1000}. 
\end{itemize}

\subsection{Background Models Challenge Dataset}
\label{sec:BMC}
The BMC (Background Models Challenge\protect\footnotemark[38]) dataset consists of both synthetic and real videos to permit a rigorous comparison of background subtraction techniques for the corresponding workshop organized within Asian Conference in Computer Vision (ACCV). This dataset \cite{205} consists of the following sequences:
\begin{itemize}
\item \textbf{Synthetic Sequences:} A set of 20 urban video sequences rendered with the SiVIC simulator. With this tool, the associate ground truth was rendered frame by frame for each video at 25 fps. Several complex scenarios are simulated such as fog, sun and acquisition noise for two environments (a rotary and a street). A first part of 10 synthetic videos are devoted to the learning phase, while the 10 others are used for the evaluation. \\
\item \textbf{Real Sequences:} The BMC dataset also contains 9 real videos acquired from static cameras like video-surveillance contexts for evaluation. This real dataset has been built in order test the algorithms reliability during time and in difficult situations such as outdoor scenes. So, real long videos about one hour and up to four hours are available, and they may present long time change in luminosity with small density of objects in time compared to previous synthetic ones. Moreover, this real dataset allows to test the influence of some difficulties encountered during the object extraction phase, as the presence of vegetation, cast shadows or sudden light changes in the scene.  \\
\end{itemize}

\footnotetext[38]{{http://bmc.iut-auvergne.com/}}

\subsection{Performance Evaluation Metrics}
\label{sec:Performance}
We used ground truth based metrics computed from the true positives (TP), true negatives (TN), false positives (FP) and false negatives (FN). FP and FN refer to pixels misclassified as foreground (FP) or background (FN) while TP and TN account for accurately classified pixels respectively as foreground and background. Then, we computed the metrics used in the BMC dataset \cite{204} such as the detection rate, the specificity, the false positive rate, the false negative rate, the percentage of wrong classifications, the precision and the F-Measure. Detection rate gives the percentage of corrected pixels classified as background when compared with the total number of background pixels in the ground truth: \\
\begin{equation}
DR=\frac{TP}{TP+FN}
\end{equation}
The specificity is computed as follows: \\
\begin{equation}
Specificity=\frac{TN}{TN+FP}
\end{equation}
The false positive rate and the false negative rate are defined as follows: \\
\begin{equation}
FPR=\frac{FP}{FP+TN}
\end{equation}
\begin{equation}
FNR=\frac{FN}{TP+FN}
\end{equation}
The percentage of wrong classifications is defined as follows: \\
\begin{equation}
PWC=\frac{100 (FN+FP)}{TP+FN+FP+TN}
\end{equation}
Precision gives the percentage of corrected pixels classified as background as compared at the total pixels classified as background by the method: \\
\begin{equation}
Precision=\frac{TP}{TP+FP}
\end{equation}
A good performance is obtained when the \textit{detection rate} is high without altering the \textit{precision}. A \textit{precision} score of 1.0 means that every pixel labeled as belonging to the class foreground in the mask does indeed belong to the corresponding class in the ground-truth but pixels classified as background in the mask can be labeled incorrectly whereas a recall of 1.0 means that every pixel from the class foreground was labeled as belonging to the class foreground but says nothing about how many other pixels were incorrectly also labeled as belonging to foreground. We also computed the F-Measure (or effectiveness measure) as follows: \\
\begin{equation}
F=\frac{2*DR*Precision}{DR+Precision}
\end{equation}
The F-Measure characterizes the performance of classification in Precision-Detection Rate space. The aim is to maximize F close to one. \\

\subsection{Experimental Results}
We made the experimental evaluation by using the quality metrics which are computable thanks to a free software named BMC Wizard. The results of the first workshop BMC 2012 are available at the related website. We evaluated the performance of the following 32 algorithms grouped by category: \\
\begin{itemize}
\item \textbf{Basic (2):} PCA \cite{310} and RSL  \cite{1}.
\item \textbf{Robust Principal Components Analysis (20):} 
\begin{enumerate}
\item RPCA-PCP: EALM \cite{18}, IALM \cite{18}, ADM \cite{21}, LADMAP \cite{39}, LSADM \cite{47}, LADM \cite{23}, BLWS \cite{57}, FAM \cite{26-1}. \item PCA-SPCP: NSA \cite{34}, PSPG \cite{34-1}, R2PCP \cite{1036} and Lag-SPCP-QN \cite{1029}.
\item RPCA-QPCP \cite{8}. 
\item RPCA-BPCP \cite{41}.
\item RPCA-SO: OR-PCA \cite{1010-1} and OR-PCA with MRF \cite{1010-2}.
\item Bayesian RPCA: BRPCA \cite{7}, VBRPCA \cite{24} and MOG-RPCA \cite{1024}.
\item Approximated RPCA: GoDec \cite{6} and SemisoftGoDec \cite{6}.
\end{enumerate}
\item \textbf{Robust Non-negative Matrix Factorization (1):} MahNMF \cite{761}.
\item \textbf{Robust Subspace Recovery (1):} ROSL \cite{759-1}.
\item \textbf{Robust Subspace Tracking (3):} GRASTA \cite{27}, pROST \cite{69} and GOSUS \cite{85}.
\item \textbf{Robust Low-rank Minimization (4):} DECOLOR \cite{25}, DRMF \cite{54}, PRMF \cite{5400} and PLRMF (RegL1-ALM) \cite{1025}. 
\item \textbf{MOG (1):} Adaptive MOG \cite{701}. \\
\end{itemize}

Table \ref{BMC1} and Table \ref{BMC11} show the evaluation results using the synthetic videos. Table \ref{BMC2} and Table \ref{BMC21} shows the evaluation results using the real videos for evaluation phase. First, we provide a short qualitative analysis in presence of illumination changes and dynamic backgrounds. Then, we give a full quantitative evaluation. For the experimental setup, we used the parameters set in each original paper of the corresponding algorithm. To reduce the computation time in the experiments, we initialized batch algorihms with 200 frames instead of 300 frames (taken in the survey restricted on the RPCA framework \cite{510}). Thus, the performance in terms of F-measure are lower for the shared algorithms between this paper and the survey on the PCA framework \cite{510}. But, the rank of the algorithms is the same showing the stability of the algorithms and the evaluation. Thus, the conclusions made in \cite{510} are preserved. Because visual results of each algorithm seem very similar (like in \cite{510}), we have done a full quantitative evaluation.

\subsection{Analysis of the Experimental Results}

\subsubsection{Synthetic Videos}
From Table \ref{BMC1} and Table \ref{BMC11}, we can see that the algorithms that gives the best recall (more than 0.9) are the following ones (in italic in the tables): FAM \cite{26-1}, MahNMF \cite{761}, ROSL \cite{759-1} and DECOLOR \cite{25}. For the precision, the algorithms R2PCP \cite{1036}, OR-PCA with MRF \cite{1010-2}, GRASTA \cite{27}, pROST \cite{69} and DRMF present the higher precision. For the F-Measure, OR-PCA with MRF \cite{1010-2} gives the best performance followed by Lag-SCP-QN \cite{1029}, OR-PCA without MRF \cite{1010-1}, DECOLOR \cite{25} and  R2PCP \cite{1036}. We have indicated in the tables for the fifth best score the rank of the algorithm between parenthesis. Figure \ref{F-Measure} shows a visual overview of the F-Measure for the 32 algorithms and the adaptive MOG \cite{701}. The algorithm OR-PCA with MRF gives the highest F-Measure. For the stable version of PCA, Lag-SCP-QN \cite{1029} shows the best performance. DECOLOR \cite{25} shows the best robustness for the low-rank methods but it is very time consuming instead of DRMF \cite{54} and PRMF \cite{5400}.

\subsubsection{Real Videos}
\begin{figure}
\begin{center}
\includegraphics[width=12cm]{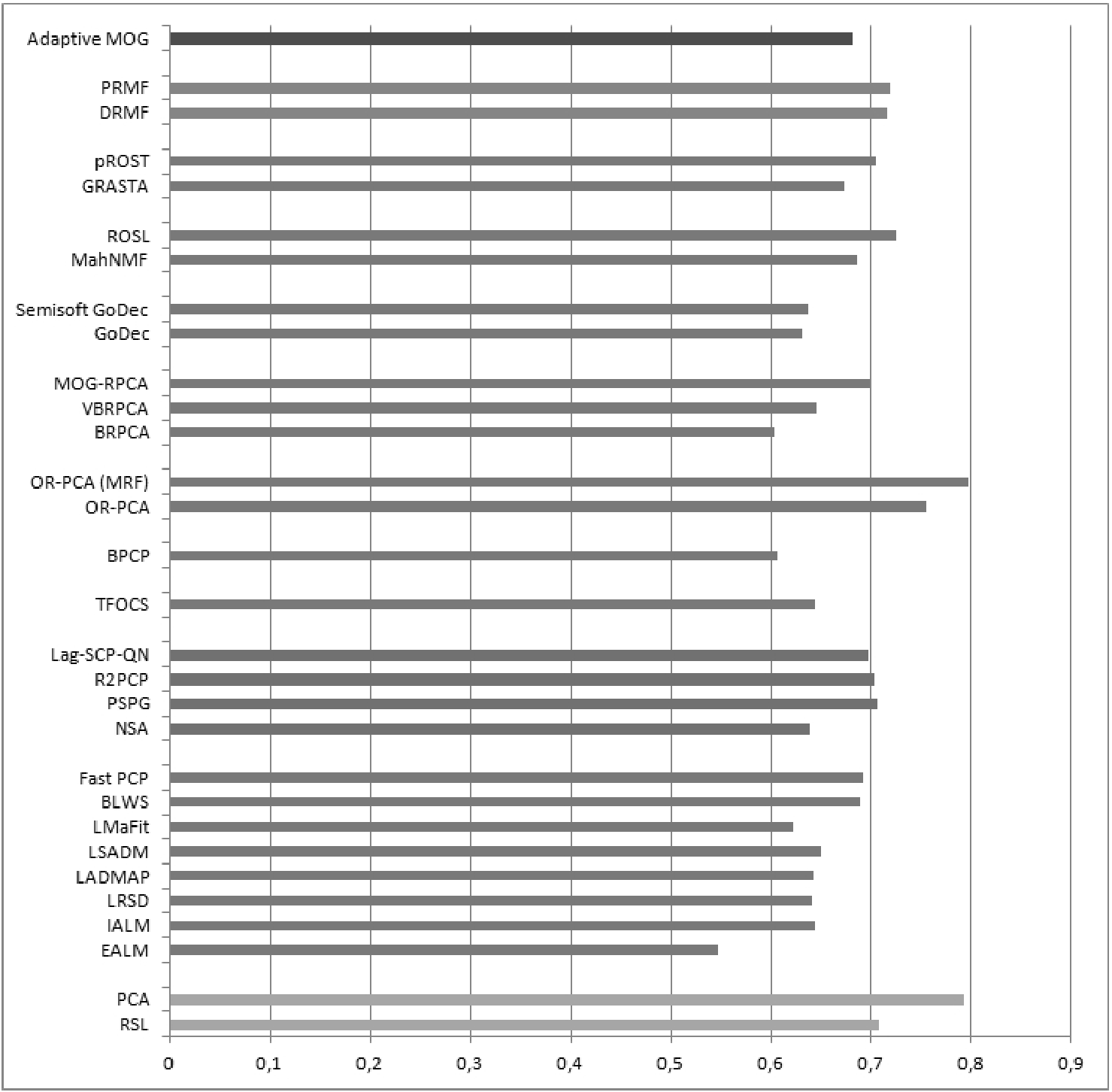} \\
\caption{F-Measure of the 32 Algorithms for the Evaluation Results using the Real Videos for Evaluation Phase.} 
\label{F-Measure-1}
\end{center}
\end{figure}

From Table \ref{BMC2} and Table \ref{BMC21}, we can see that for the F-Measure, OR-PCA with MRF \cite{1010-2} gives the best performance followed by OR-PCA without MRF \cite{1010-1}, ROSL \cite{759-1},  PRMF \cite{5400} and DRMF \cite{54}. Thus, the algorithm OR-PCA with and without MRF has stable performance both on synthetic and real videos by preserving their rank. Otherwise, ROSL, PRMF and DRMF are more robust on real video than on synthetic videos. We have not conducted experiments on real videos for DECOLOR and RegL1-ALM due to very expensive computation time, and for GOSUS due to an implementation problem in the original code when very long sequences appear. Figure \ref{F-Measure-1} shows a visual overview of the F-Measure for the 32 algorithms and the adaptive MOG \cite{701}.\\

\begin{figure}
\begin{center}
\includegraphics[width=12cm]{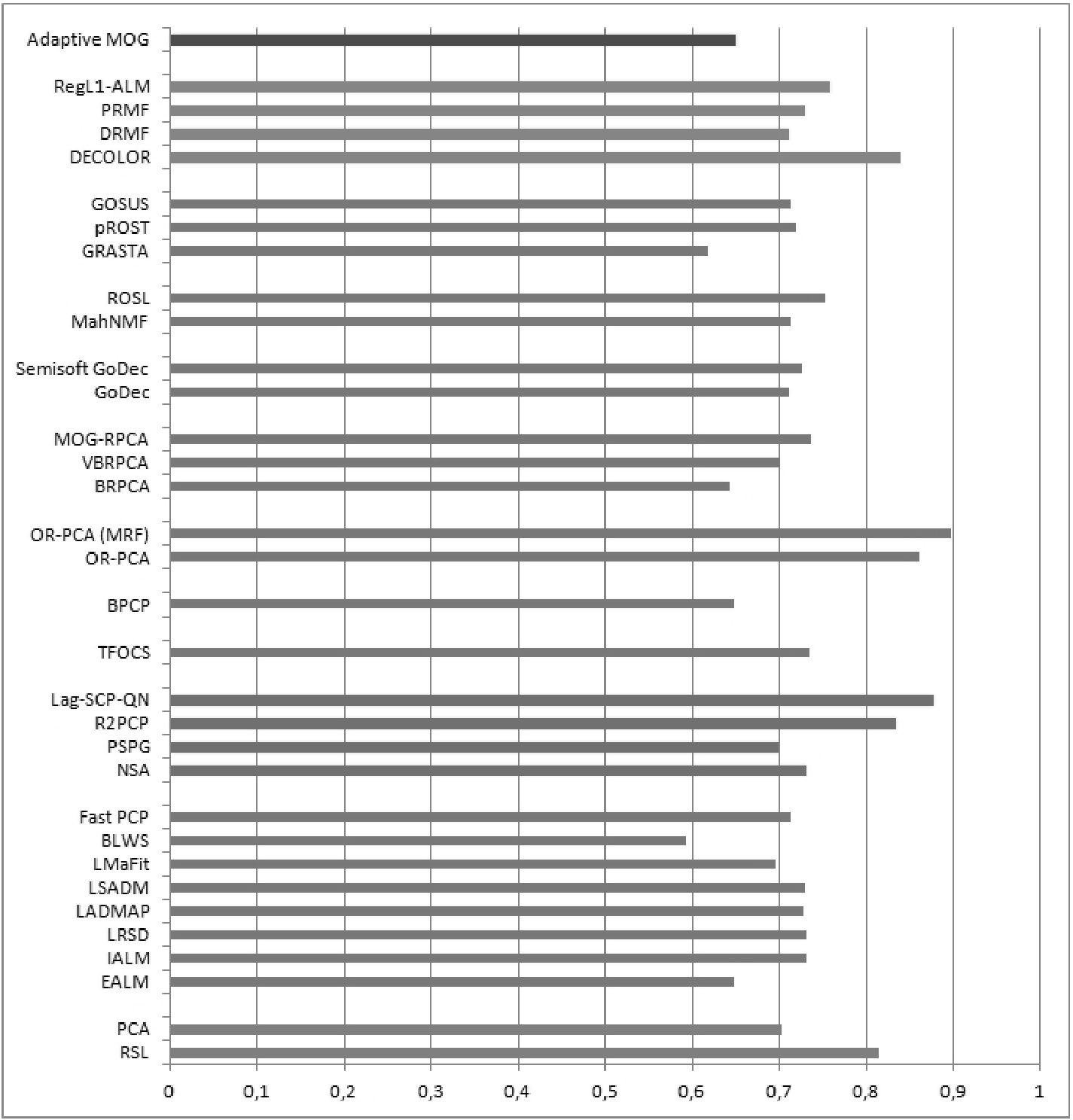} \\
\caption{F-Measure of the 32 Algorithms for the Evaluation Results using the Synthetic Videos for Evaluation Phase.} 
\label{F-Measure}
\end{center}
\end{figure}

\indent An other main conclusion is that most of the RPCA algorithms outperform the adaptive MOG for the synthetic videos as for real videos too. Figure \ref{F-Measure} and  Figure \ref{F-Measure-1} show the F-Measure of the evaluated algorithms for synthetic and real videos, respectively.

\begin{table*}
\begin{center}
\scalebox{0.55}{
\begin{tabular}{|l|l||c|c|c|c|r| |c|c|c|c|r| |c|} 
\hline
Algorithms & Measure & \multicolumn{5}{c|}{Street}     & \multicolumn{5}{c|}{Rotary}  & \multicolumn{1}{c|}{Average}     \\
\cline{3-12} 
           &         & 112 & 212 & 312 & 412 & 512     & 122 & 222 & 322 & 422 & 522  &                                  \\
\hline
\hline
RSL                           & Recall     & 0.877	& 0.874	& 0.804	& 0.821	& 0.871	
																					 & 0.872	& 0.870	& 0.867	& 0.773	& 0.741	& -	    \\	
De La Torre et al. \cite{1}   & Precision  & 0.646	 & 0.642 & 0.616 & 0.551 & 0.526	
																					 & 0.659	& 0.656	& 0.649	& 0.618	& 0.607	& -     \\
                              & F-measure  & 0.746	& 0.743	& 0.699	& 0.672	& 0.661	
																					 & 0.752	& 0.750	& 0.744	& 0.688	& 0.668	& 0.712 \\
\hline
\hline
PCA                           & Recall      & 0.683 & 0.723 & 0.700 & 0.726 & 0.726 
																						& 0.716 & 0.635 & 0.558 & 0.584 & 0.613   &- \\
Oliver et al.\cite{310}       & Precision   & 0.710 & 0.794 & 0.816 & 0.785 & 0.748 
																						& 0.755 & 0.760 & 0.796 & 0.625 & 0.719   &- \\
                               & F-measure  & 0.742 & 0.757 & 0.655 & 0.659 & 0.637 
                                            & 0.730 & 0.747 & 0.721 & 0.653 & 0.716   &0.701 \\
\hline
\hline
\textbf{RPCA-PCP}    & Candes et al. \cite{3}             & & & & & & & & & &        &   \\
\hline
EALM                  & Recall      & 0.607 & 0.599 & 0.533 & 0.516 & 0.509
    																& 0.656 & 0.651 & 0.569 & 0.468 & 0.596 &- \\
Lin et al. \cite{18}  & Precision   & 0.831 & 0.821 & 0.822 & 0.800 & 0.606
                                    & 0.756 & 0.753 & 0.760 & 0.762 & 0.639 &- \\
                      & F-measure   & 0.705 & 0.696 & 0.652 & 0.633 & 0.554    
                        					  & 0.703 & 0.699 & 0.653 & 0.587 & 0.617 &0.649 \\
\hline 
 IALM                  & Recall     & 0.774 & 0.689 & 0.741 & 0.738 & 0.677
																		& 0.743 & 0.750 & 0.741 & 0.705 & 0.705 &- \\
Lin et al. \cite {18}  & Precision  & 0.662 & 0.811 & 0.719 & 0.743 & 0.664
     																& 0.779 & 0.769 & 0.740 & 0.773 & 0.747 &- \\
                       & F-measure  & 0.715 & 0.746 & 0.730 & 0.741 & 0.670
                                    &0.761  & 0.759 & 0.740 & 0.737 & 0.725  &0.732 \\   
\hline
ADM                            & Recall      & 0.699 & 0.691 & 0.713 & 0.737 & 0.680
     																         & 0.727 & 0.728 & 0.711 & 0.658 & 0.718 &- \\
(LRSD)Yuan and Yang \cite{21}  & Precision   & 0.795 & 0.815 & 0.750 & 0.749 & 0.672
     																				 & 0.787 & 0.788 & 0.744 & 0.791 & 0.736 &- \\
                               & F-measure   & 0.744 & 0.749 & 0.731 & 0.742 & 0.676
    																			   & 0.756 & 0.757 & 0.727 & 0.719 & 0.727 &0.732 \\ 
\hline 
LADMAP                & Recall      & 0.699 & 0.691 & 0.724 & 0.738 & 0.681
                                    & 0.727 & 0.730 & 0.713 & 0.655 & 0.718 &- \\
Lin et al. \cite{39}  & Precision   & 0.795 & 0.815 & 0.741 & 0.748 & 0.673
    																& 0.787 & 0.787 & 0.745 & 0.793 & 0.737 &- \\
                      & F-measure   & 0.744 & 0.749 & 0.733 & 0.743 & 0.677
                                    & 0.756 & 0.757 & 0.729 & 0.718 & 0.727 &0.728 \\        
\hline 
LSADM                       & Recall      & 0.724 & 0.707 & 0.756 & 0.712 & 0.690
                                          & 0.728 & 0.730 & 0.742 & 0.632 & 0.680 &- \\
Goldfarb et al. \cite{47}   & Precision   & 0.803 & 0.815 & 0.787 & 0.785 & 0.592
    																			& 0.790 & 0.788 & 0.742 & 0.806 & 0.719 &- \\
                            & F-measure   & 0.762 & 0.758 & 0.736 & 0.747 & 0.638
                                          & 0.758 & 0.758 & 0.742 & 0.710 & 0.699 &0.730 \\                      
\hline 
LADM                  & Recall      & 0.679 & 0.653 & 0.627 & 0.570 & 0.590
                        						& 0.725 & 0.712 & 0.600 & 0.597 & 0.642 &- \\
(LMaFit) Shen et al. \cite{23}      & Precision   & 0.829 & 0.834 & 0.813 & 0.816 & 0.574
     																              & 0.794 & 0.803 & 0.752 & 0.760 & 0.682 &- \\
                                    & F-measure   & 0.748 & 0.734 & 0.710 & 0.675 & 0.582
                                                  & 0.758 & 0.756 & 0.669 & 0.671 & 0.662 &0.696 \\  
\hline
BLWS                   & Recall      & 0.550 &0.548 &0.514 &0.483 &0.470
																	   & 0.636 &0.635 &0.513 &0.453 &0.643 &- \\
Lin and Wei \cite{57}  & Precision   & 0.831 &0.821 &0.816 &0.410 &0.625
																		 & 0.697 &0.698 &0.771 &0.388 &0.561 &- \\
                       & F-measure   & 0.667 &0.662 &0.637 &0.621 &0.539
                        						 & 0.666 &0.665 &0.621 &0.419 &0.600 &0.610\\  
\hline 
FAM (Fast PCP)                         & Recall   & \textit{0.979}	& \textit{0.980}	& \textit{0.948}	& \textit{0.929}	& \textit{0.949}
																									& \textit{0.956}	& \textit{0.978}	& \textit{0.958}	& \textit{0.906}	& \textit{0.945}	& -  \\ 
Rodriguez and Wohlberg \cite{26-1}     & Precision  & 0.707	& 0.691	& 0.531	& 0.522	& 0.531	
																									  & 0.671	& 0.680	& 0.567	& 0.532	& 0.499	& -  \\ 
                                       & F-measure  & 0.821	& 0.811	& 0.681	& 0.668	& 0.681
																									  & 0.789	& 0.802	& 0.712	& 0.670	& 0.497	& 0.713 \\
\hline
\hline
\textbf{RPCA-SPCP}    & Zhou et al. \cite{5}             & & & & & & & & & &       &     \\
\hline
NSA                     & Recall      & 0.707 & 0.706 & 0.722 & 0.700 & 0.688 
     																	& 0.717 & 0.722 & 0.742 & 0.636 & 0.687  &- \\
Aybat et al. \cite{34} & Precision    & 0.820 & 0.817 & 0.749 & 0.798 & 0.712 
   																		& 0.799 & 0.795 & 0.742 & 0.797 & 0.598 &- \\
                        & F-measure   & 0.760 & 0.758 & 0.735 & 0.746 & 0.64
                                      & 0.756 & 0.757 & 0.742 & 0.710 & 0.699 &0.730 \\                                       
\hline 
PSPG                                   & Recall      & 0.980	& 0.968	& 0.934	& 0.926	& 0.943	
	                                                   & 0.975	& 0.968	& 0.938	& 0.917	& 0.964	&- \\
Aybat et al. \cite{34-3}               & Precision   & 0.615	& 0.546	& 0.520	& 0.518	& 0.523	
	                                                   & 0.593	& 0.570	& 0.533	& 0.531	& 0.562	&- \\
                                       & F-measure   & 0.756	& 0.699	& 0.668	& 0.664	& 0.673	
																										& 0.737	& 0.717	& 0.680	& 0.672	& 0.710	& 0.698 \\
\hline
IAM-MM (R2PCP)                         & Recall      & 0.828	& 0.837	& 0.793	& 0.727	& 0.819	
                                                     & 0.819	& 0.818	& 0.763	& 0.689	& 0.814	&-      \\
Hinterm\"{u}ller and Wu \cite{1036}    & Precision   & 0.901	& 0.905	& 0.909	& 0.861	& 0.828	
                                                     & 0.897	& 0.894	& 0.898	& 0.825	& 0.870	&-      \\
                                       & F-measure   & 0.863	& 0.870	& 0.847	& 0.788	& 0.823	
                                                     & 0.856	& 0.855	& 0.825	& 0.751	& 0.841	& 0.832  (5) \\
\hline 
Variational SPCP  (Lag-SPCP-QN))       & Recall      & 0.976	& 0.952	& 0.937	& 0.932	& 0.946	
	                                                   & 0.972	& 0.930	& 0.909	& 0.880	& 0.925	& -     \\
Arakvin and Becker \cite{1029}         & Precision   & 0.776	& 0.818	& 0.821	& 0.837	& 0.819	
                                                     & 0.803	& 0.836	& 0.838	& 0.857	& 0.839	& -     \\
                                       & F-measure   & 0.865	& 0.880	& 0.875	& \textbf{0.882}	& \textbf{0.878}	
																										 & 0.879	& 0.880	& 0.872	& 0.868	& \textbf{0.880}	& 0.876 (2) \\                  
\hline
\hline
\textbf{RPCA-QPCP}     & Becker et al. \cite{8}            & & & & & & & & & &       &    \\
\hline
TFOCS                  & Recall      & 0.760 & 0.693 & 0.717 & 0.753 & 0.684
                                     & 0.748 & 0.742 & 0.742 & 0.681 & 0.727 &- \\
Becker et al. \cite{8} & Precision   & 0.680 & 0.815 & 0.749 & 0.740 & 0.674 
 																		 & 0.774 & 0.782 & 0.737 & 0.790 & 0.729  &- \\
                       & F-measure   & 0.718 & 0.750 & 0.733 & 0.746 & 0.679 
                         	           & 0.761 & 0.762 & 0.739 & 0.732 & 0.728 & 0.735 \\
\hline
\hline
\textbf{RPCA-BPCP}    & Tang and Nehorai \cite{41}             & & & & & & & & & &     &      \\
\hline
ALM                        & Recall      & 0.607 & 0.599 & 0.533 & 0.516 & 0.509 
																				 & 0.656 & 0.651 & 0.569 & 0.469 & 0.596 &- \\
Tang and Nehorai \cite{41} & Precision   & 0.831 & 0.821 & 0.822 & 0.800 & 0.606 
																				 & 0.755 & 0.753 & 0.760 & 0.762 & 0.639  &- \\ 																																											 & F-Measure   & 0.705 & 0.696 & 0.652 & 0.633 & 0.544
																				 & 0.703 & 0.699 & 0.653 & 0.587 & 0.617 &0.649 \\
\hline
\hline
\textbf{RPCA via Stochastic Optimization}    &              & & & & & & & & & &     &       \\
\hline
OR-PCA without MRF          & Recall      &0.851	&0.853	&0.892	&0.860	&0.820
    																		  &0.885	&0.889	&0.831	&0.857	&0.741 &- \\
Javed et al. \cite{1010-1}  & Precision   &0.911	&0.911	&0.848	&0.836	&0.873
                                          &0.908	&0.907	&0.878	&0.816	&0.905 &- \\
                            & F-measure   &0.880	&0.881	&0.870	  &0.848	&0.846   
                        									&0.896	&0.898	&0.854	&0.836	&0.815 &0.862 (3) \\
\hline 
OR-PCA with MRF           & Recall       &0.871	&0.870 &0.894	&0.850	&0.860 
       																	 &0.937	&0.940 &0.923	&0.917	&0.841 &- \\
Javed et al. \cite{1010-2}& Precision    &0.956	&0.952 &0.882	&0.873	&0.894
						                             &0.924	&0.924 &0.901	&0.846  &0.925 &- \\
                          & F-measure    &\textbf{0.911}	&\textbf{0.909}	&\textbf{0.888}	&0.861	&0.876
																				 &\textbf{0.931}	&\textbf{0.932}	&\textbf{0.912}	&\textbf{0.880}	&0.879 &\textbf{0.897} (1) \\
\hline
\hline
\textbf{Bayesian RPCA}    &              & & & & & & & & & &         &   \\
\hline
Bayesian RPCA         & Recall      &0.659 &0.626 &0.509 &0.511 & 0.475
    																&0.725 &0.651 &0.569 &0.529 & 0.596 &- \\
Ding et al. \cite{7}  & Precision   &0.828 &0.826 &0.785 &0.739 & 0.593
                                    &0.781 &0.754 &0.761 &0.510 & 0.640 &- \\
                      & F-measure   &0.736 &0.715 &0.623 &0.609 & 0.529    
                        						&0.752 &0.699 &0.654 &0.520 & 0.618 &0.643 \\
\hline
Variational BRPCA        & Recall      &0.676 &0.639 &0.710 &0.618 &0.654 
       																 &0.694 &0.698 &0.671 &0.611 &0.642 &- \\
Babacan et al. \cite{24} & Precision   &0.833 &0.838 &0.757 &0.811 &0.569
						                           &0.775 &0.772 &0.748 &0.758 &0.628 &- \\
                         & F-measure   &0.748 &0.727 &0.733 &0.704 &0.609
																			&0.733 &0.733 &0.707 &0.678 &0.635 &0.700 \\
\hline 
MOG-RPCA                               & Recall      & 0.981	& 0.984	& 0.979	& 0.974	& 0.960	
																										 & 0.980	& 0.982	& 0.978	& 0.920	& 0.977	& - \\ 
Zhao et al. \cite{1024}                & Precision   & 0.637	& 0.674	& 0.620	& 0.589	& 0.537	
																										 & 0.651	& 0.699	& 0.652	& 0.529	& 0.629	& - \\                                                                                        & F-measure   & 0.773	& 0.800	& 0.759	& 0.734	& 0.689	
                                                     & 0.782	& 0.817	& 0.782	& 0.672	& 0.765	& 0.757 \\
\hline
\hline
\textbf{Approximated RPCA}    &              & & & & & & & & & &         &   \\
\hline
GoDec                 & Recall      &0.690 &0.772 &0.750 &0.679 &0.629 
																		&0.724 &0.721 &0.707 &0.613 &0.677 &- \\
Zhou and Tao \cite{6} & Precision   &0.817 &0.724 &0.703 &0.716 &0.536
                                    &0.792 &0.795 &0.748 &0.807 &0.681 &- \\
                      & F-measure   &0.749 &0.747 &0.726 &0.697 &0.580
                                    &0.757 &0.756 &0.727 &0.699 &0.679 &0.711 \\
\hline 
SemiSoft GoDec        & Recall      &0.692 &0.700 &0.717 &0.730 &0.664
                                    &0.726 &0.718 &0.673 &0.642 &0.716 &- \\																			         
Zhou and Tao \cite{6} & Precision   &0.816 &0.818 &0.752 &0.772 &0.601 
                                    &0.792 &0.799 &0.750 &0.804 &0.688 &- \\
                      & F-measure   &0.750 &0.755 &0.734 &0.750 &0.631
                                    &0.758 &0.757 &0.710 &0.715 &0.702 &0.726 \\
\hline
\end{tabular}} 
\caption{BMC dataset: Evaluation Results using the Synthetic Videos for Evaluation Phase (Part 1). The number between parenthesis indicates the rank of the algorithm in terms of F-measure.}
\label{BMC1}
\end{center}
\end{table*}

\begin{table*}
\begin{center}
\scalebox{0.65}{
\begin{tabular}{|l|l||c|c|c|c|r| |c|c|c|c|r| |c|} 
\hline
Algorithms & Measure & \multicolumn{5}{c|}{Street}     & \multicolumn{5}{c|}{Rotary}  & \multicolumn{1}{c|}{Average}     \\
\cline{3-12} 
                       &         & 112 & 212 & 312 & 412 & 512     & 122 & 222 & 322 & 422 & 522  &                                  \\
\hline 
\hline
\textbf{Robust NMF}                    &              & & & & & & & & & &      &      \\
\hline
MahNMF                 & Recall      & \textit{0.982}	& \textit{0.982}	& \textit{0.934}	& \textit{0.909}	& \textit{0.943}	
																		 & \textit{0.979}	& \textit{0.977}	& \textit{0.944}	& \textit{0.894}	& \textit{0.969}	& -  \\																			   
Guan et al. \cite{761} & Precision   & 0.661	& 0.622	& 0.520	& 0.515	& 0.524	
																		 & 0.624	& 0.612	& 0.538	& 0.524	& 0.576	& - \\
                       & F-measure   & 0.790	& 0.762	& 0.668	& 0.657	& 0.674	
																		 & 0.762	& 0.752	& 0.685	& 0.660	& 0.722	& 0.713 \\     
\hline
\hline
\textbf{Robust Subspace Recovery}    &              & & & & & & & & & &       &     \\
\hline
ROSL                   & Recall      & \textit{0.984}	& \textit{0.985}	& \textit{0.948}	& \textit{0.924}	& \textit{0.948}	
																		 & \textit{0.984}	& \textit{0.983}	& \textit{0.950}	& \textit{0.912}	& \textit{0.976}	& - \\																		
Shu et al. \cite{759-1} & Precision  & 0.744	& 0.744	& 0.528	& 0.519	& 0.527	
																		 & 0.753	& 0.753	& 0.545	& 0.533	& 0.619	& - \\
                        & F-measure  & 0.847	& 0.847	& 0.678	& 0.665	& 0.678	
																		 & 0.853	& 0.853	& 0.693	& 0.673	& 0.757	& 0.754 \\
\hline       
\hline
\textbf{Robust Subspace Tracking}    &              & & & & & & & & & &       &     \\
\hline
GRASTA                 & Recall       &0.700 &0.787 &0.695 &0.787 &0.669 
																			&0.680 &0.637 &0.619 &0.623 &0.791 &- \\
He et al. \cite{27}    & Precision    &0.980 &0.847 &0.965 &0.843 &0.960
																			&0.902 &0.548 &0.530 &0.778 &0.714 &- \\
                       & F-measure    &0.817 &0.816 &0.807 &0.814 &0.789 
                       								&0.776 &0.589 &0.571 &0.692 &0.751 &0.618 \\
\hline
pROST                              & Recall       &0.944 &0.878 &0.853 &0.889 &0.785 
       															              &0.819 &0.838 &0.789 &0.738 &0.863 &- \\																	
Hage and Kleinsteuber \cite{69}    & Precision    &0.844 &0.937 &0.968 &0.931 &0.961
                                                  &0.903 &0.847 &0.953 &0.730 &0.815 &- \\
                                   & F-measure    &0.891 &0.906 &0.907 &0.909 &0.864
																									&0.859 &0.842 &0.863 &0.734 &0.838 &0.718 \\  
\hline         
GOSUS                  & Recall      & 0.982	& 0.981	& 0.945	& 0.933	& 0.949	
                                     & 0.978	& 0.977	& 0.958	& 0.906	& 0.969	& -		 \\																       
Xu et al. \cite{85}    & Precision   & 0.617	& 0.609	& 0.526	& 0.526	& 0.528	
																		 & 0.616	& 0.612	& 0.555	& 0.534	& 0.577	& -    \\
                      & F-measure    & 0.758	& 0.751	& 0.676	& 0.672	& 0.678	
																		 & 0.756	& 0.752	& 0.702	& 0.672	& 0.723	& 0.714 \\																		
\hline
\hline
\textbf{Low Rank Minimization}       &              & & & & & & & & & &        &    \\
\hline
DECOLOR                 & Recall      & \textit{0.982}	& \textit{0.985}	& \textit{0.983}	& \textit{0.980}	& \textit{0.978}
																		  & \textit{0.983}	& \textit{0.983}	& \textit{0.981}	& \textit{0.967}	& \textit{0.980} & - \\	
Zhou et al. \cite{25}   & Precision   & 0.778	& 0.748	& 0.747	& 0.729	& 0.599	
																		  & 0.764	& 0.759	& 0.760	& 0.762	& 0.694	& - \\
                        & F-measure   & 0.868	& 0.851	& 0.849	& 0.836	& 0.743
																			& 0.860	& 0.857	& 0.857	& 0.852	& 0.813	& 0.838 (4) \\
\hline																														 
DRMF                    & Recall      &0.857 &0.864 &0.853 &0.864 &0.903
																			&0.880 &0.834 &0.744 &0.805 &0.827 &- \\
Xiong et al. \cite{54}  & Precision   &0.969 &0.950 &0.968 &0.948 &0.855 
																			&0.891 &0.834 &0.924 &0.700 &0.825 &- \\
                        & F-measure   &0.910 &0.905 &0.907 &0.904 &0.878 
																			&0.885 &0.834 &0.824 &0.749 &0.826 &0.710 \\																			
\hline 
PRMF                    & Recall      &0.944 &0.903 &0.918 &0.901 &0.899
																			&0.891 &0.906 &0.867 &0.824 &0.869 &-  \\																			     
Wang et al. \cite{5400} & Precision   &0.819 &0.919 &0.879 &0.922 &0.862
																		  &0.887 &0.849 &0.888 &0.707 &0.821 &-  \\
                        & F-measure   &0.877 &0.911 &0.898 &0.911 &0.880 
																			&0.889 &0.877 &0.878 &0.761 &0.845 &0.727 \\ 
\hline      
PLRMF (RegL1-ALM)        & Recall     & 0.984	& 0.985	& 0.953	& 0.926	& 0.949	
                                      & 0.984	& 0.983	& 0.950	& 0.914	& 0.976	& -	\\																		
Zheng et al. \cite{1025} & Precision  & 0.744	& 0.745	& 0.533	& 0.520	& 0.529	
                                      & 0.756	& 0.756	& 0.545	& 0.535	& 0.619	& - \\	
                         & F-measure  & 0.847	& 0.848	& 0.684	& 0.666	& 0.679	
                                      & 0.855	& 0.855	& 0.693	& 0.675	& 0.758	& 0.756 \\															   
\hline
\hline
\textbf{Mixture of Gaussians}    &              & & & & & & & & & &          &  \\
\hline

Adaptive MOG              & Recall      &0.827 &0.827 &0.797 &0.761 &0.821 
										    							                         &0.823 &0.831 &0.797 &0.743 &0.834 &- \\
Shimada et al. \cite{701} & Precision   &0.766 &0.768 &0.480 &0.426 &0.519
                                                               &0.786 &0.790 &0.526 &0.435 &0.740 &- \\
                          & F-measure   &0.796 &0.796 &0.605 &0.553 &0.640
                                                               &0.804 &0.810 &0.638 &0.555 &0.784 &0.698 \\
\hline
\end{tabular}} 
\caption{BMC dataset: Evaluation Results using the Synthetic Videos for Evaluation Phase (Part 2). The number between parenthesis indicates the rank of the algorithm in terms of F-measure.}
\label{BMC11}
\end{center}
\end{table*}

\begin{table*}
\begin{center}
\scalebox{0.55}{
\begin{tabular}{|l|l||c|c|c|c|r| c|c|c|r| c|} 
\hline
Algorithms & Measure     & \multicolumn{9}{c|}{Real Videos}  & \multicolumn{1}{c|}{Average}   \\                 
\cline{3-12} 
           &             & 001 & 002 & 003 & 004 & 005 & 006 & 007 & 008 & 009  &     \\
\hline
\hline
RSL                           & Recall      &0.800	&0.689	&0.840	&0.872	&0.861	&0.823	&0.658	&0.589	&0.690 &- \\
De La Torre et al. \cite{1}   & Precision   &0.732	&0.808	&0.804	&0.585	&0.598	&0.713	&0.636	&0.526	&0.625  &- \\
                              & F-measure   &0.765	&0.744	&0.821	&0.700	&0.706	&0.764	&0.647	&0.556	&0.656 &0.707 \\ 
\hline                             
\hline
PCA                        & Recall      & 0.818 & 0.672 & 0.890 & 0.824 & 0.799 & 0.793 & 0.580 & 0.753 & 0.861  &- \\
Oliver et al. \cite{310}   & Precision   & 0.740 & 0.861 & 0.953 & 0.833 & 0.635 & 0.783 & 0.826 & 0.785 & 0.953  &- \\
                           & F-measure   & 0.777 & 0.755 & 0.920 & 0.829 & 0.700 & 0.788 & 0.682 & 0.768 & 0.905  &0.791 \\ 
\hline
\hline
\textbf{RPCA-PCP}       & Candes et al. \cite{3}             & & & & & & & & & &           \\
\hline
EALM                  & Recall      &0.574 &0.616 &0.728 &0.690 &0.511 &0.650 &0.589 &0.520 &0.599 &- \\
Lin et al. \cite{18}  & Precision   &0.405 &0.705 &0.809 &0.695 &0.394 &0.580 &0.750 &0.579 &0.815 &- \\
                      & F-measure   &0.478 &0.658 &0.767 &0.692 &0.447 &0.613 &0.662 &0.548 &0.693 &0.547  \\
                                  
\hline 
IALM                  & Recall       &0.697  &0.515  &0.759  &0.691  &0.635  &0.642  &0.433  &0.617  &0.707 &- \\
Lin et al. \cite{18}  & Precision    &0.585  &0.723  &0.798  &0.678  &0.483  &0.643  &0.683  &0.632  &0.807 &- \\
                      & F-measure    &0.637  &0.605  &0.778  &0.684  &0.551  &0.643  &0.536  &0.624  &0.754  &0.644 \\
\hline 
ADM                           & Recall      &0.691 &0.504 &0.736 &0.681 &0.630 &0.641 &0.427 &0.608 &0.714  &- \\
LRSD)Yuan and Yang \cite{21}  & Precision   &0.601 &0.727 &0.797 &0.670 &0.484 &0.633 &0.678 &0.631 &0.803  &- \\
                              & F-measure   &0.644 &0.599 &0.766 &0.676 &0.549 &0.637 &0.530 &0.620 &0.756 &0.641 \\                                
\hline 
LADMAP                & Recall      &0.691 &0.522 &0.737 &0.681 &0.624 &0.644 &0.424 &0.605 &0.714  &- \\
Lin et al. \cite{39}  & Precision   &0.601 &0.716 &0.796 &0.669 &0.485 &0.632 &0.690 &0.642 &0.803  &- \\
                      & F-measure   &0.643 &0.607 &0.766 &0.675 &0.548 &0.638 &0.532 &0.623 &0.756 &0.642 \\
\hline           
LSADM                       & Recall      &0.693  &0.535  &0.784  &0.721  &0.643  &0.656  &0.449  &0.621  &0.701  &- \\
Goldfarb et al. \cite{47}   & Precision   &0.511  &0.724  &0.802  &0.729  &0.475  &0.655  &0.693  &0.633  &0.809  &- \\
                            & F-measure   &0.591  &0.618  &0.793  &0.725  &0.549  &0.656  &0.551  &0.627  &0.752  &0.650 \\ 
\hline            
LADM                              & Recall      &0.639 &0.522 &0.752 &0.684 &0.598 &0.653 &0.431 &0.601 &0.620  &- \\
(LMaFit) Shen et al. \cite{23}    & Precision   &0.445 &0.688 &0.812 &0.723 &0.438 &0.621 &0.669 &0.632 &0.822  &- \\
                                  & F-measure   &0.528 &0.596 &0.781 &0.703 &0.509 &0.637 &0.530 &0.616 &0.709 &0.622 \\
\hline           
BLWS                   & Recall      &0.576 &0.618 &0.692 &0.661 &0.539 &0.656 &0.602 &0.527 &0.555  &- \\
Lin and Wei \cite{57}  & Precision   &0.399 &0.697 &0.814 &0.686 &0.378 &0.565 &0.748 &0.486 &0.810 &-  \\
                       & F-measure   &0.475 &0.656 &0.749 &0.673 &0.448 &0.607 &0.669 &0.506 &0.663 &0.689 \\
\hline           
FAM (Fast PCP)                     & Recall     &0.859	&0.820	&0.895	&0.863	&0.847	&0.822	&0.788	&0.558	&0.723 &- \\
Rodriguez and Wohlberg \cite{26-1} & Precision  &0.639	&0.672	&0.705	&0.539	&0.555	&0.632	&0.676	&0.508	&0.605&-  \\
                                   & F-measure  &0.733	&0.739	&0.789	&0.664	&0.670	&0.714	&0.728	&0.531	&0.659	&0.692 \\   
\hline
\hline
\textbf{RPCA-SPCP}       & Zhou et al. \cite{5}             & & & & & & & & & &           \\
\hline
NSA                     & Recall      &0.688 &0.616 &0.784 &0.725 &0.511 &0.656 &0.450 &0.621 &0.599  &- \\
Aybat et al. \cite{34}  & Precision   &0.514 &0.705 &0.802 &0.728 &0.394 &0.655 &0.694 &0.632 &0.815  &- \\
                        & F-measure   &0.591 &0.658 &0.793 &0.727 &0.447 &0.656 &0.551 &0.626 &0.693  &0.637 \\  
\hline                                                          
PSPG                     & Recall     &0.858	&0.819	&0.904	&0.851	&0.861	&0.823	&0.796	&0.559	&0.726 &- \\
Aybat et al. \cite{34-3} & Precision  &0.679	&0.666	&0.774	&0.597	&0.576	&0.619	&0.676	&0.507	&0.601 &- \\
                         & F-measure  &0.758	&0.735	&0.834	&0.701	&0.690	&0.706	&0.731	&0.532	&0.657	&0.705 \\
\hline          
IAM-MM (R2PCP)                       & Recall     &0.746 &0.593 &0.860 &0.784	&0.655	&0.670	&0.561	&0.540	&0.590 &- \\
Hinterm\"{u}ller and Wu \cite{1036}  & Precision  &0.803& 0.859	&0.821 &0.728	&0.815	&0.820	&0.673	&0.542	&0.678 &- \\
                                     & F-measure  &0.773 &0.701	&0.840 &0.755	&0.726	&0.737	&0.612	&0.541	&0.631 &0.702 \\
\hline                                     
Variational SPCP  (Lag-SPCP-QN))     & Recall      &0.734	&0.528	&0.723	&0.649	&0.662	&0.593	&0.643	&0.528	&0.688 &- \\
Arakvin and Becker \cite{1029}       & Precision   &0.911	&0.791	&0.828	&0.813	&0.799	&0.781	&0.677	&0.564	&0.774 &- \\
                                     & F-measure   &0.812	&0.633	&0.772	&0.722	&0.724	&0.674	&0.659	&0.545	&0.728 &0.696 \\
\hline
\hline
\textbf{RPCA-QPCP}       & Becker et al. \cite{8}              & & & & & & & & & &           \\
\hline
TFOCS                  & Recall      &0.691  &0.506  &0.751  &0.694  &0.644  &0.646  &0.435  &0.614  &0.707  &- \\
Becker et al. \cite{8} & Precision   &0.587  &0.729  &0.798  &0.673  &0.483  &0.642  &0.683  &0.645  &0.807  &-  \\  
                       & F-measure   &0.635  &0.601  &0.774  &0.683  &0.555  &0.644  &0.538  &0.629  &0.754  &0.644 \\  
\hline
\hline
\textbf{RPCA-BPCP}       & Tang and Nehorai \cite{41}             & & & & & & & & & &           \\
\hline
ALM                        & Recall      & 0.573 & 0.615 & 0.728 & 0.689 & 0.510 & 0.650 & 0.599 & 0.520 & 0.598  &-   \\
Tang and Nehorai \cite{41} & Precision   & 0.404 & 0.705 & 0.808 & 0.694 & 0.493 & 0.679 & 0.750 & 0.678 & 0.815  &-   \\
													 & F-measure   & 0.578 & 0.757 & 0.866 & 0.792 & 0.446 & 0.613 & 0.767 & 0.548 & 0.693  &0.607  \\ 												\hline
\hline
\textbf{RPCA via Stochastic Optimization}       &               & & & & & & & & & &           \\
\hline
OR-PCA without MRF         & Recall   &0.756	&0.683	&0.545	&0.779	&0.781	&0.692	&0.669	&0.507	&0.748 &- \\
Javed et al. \cite{1010-1} & Precision   &0.942	&0.864	&0.874	&0.89	&0.789	&0.918	&0.84	&0.557	&0.965 &- \\
                           & F-measure   &0.839	&0.763	&0.671	&0.831	&0.787	&0.789	&0.745	&0.531	&0.843 &\textbf{0.755} (2) \\
\hline 
OR-PCA with MRF            & Recall       &0.776 &0.845	&0.905	&0.799	&0.779	&0.800	&0.806	&0.566	&0.956 &- \\
Javed et al. \cite{1010-2} & Precision    &0.936 &0.781	&0.738	&0.870	&0.860	&0.891	&0.768	&0.558	&0.746 &- \\                                                          & F-measure      &0.848	&0.812 &0.813	&0.834	&0.826	&0.843	&0.786	&0.562	&0.854 &\textbf{0.797}   (1)\\
\hline
\hline
\textbf{Bayesian RPCA}       &               & & & & & & & & & &           \\
\hline
BRPCA                 & Recall      &0.578  &0.625  &0.737  &0.688  &0.545  &0.643  &0.443  &0.512  &0.591 &- \\
Ding et al. \cite{7}  & Precision   &0.404  &0.707  &0.800  &0.674  &0.386  &0.583  &0.689  &0.583  &0.817 &- \\
                      & F-measure   &0.479  &0.664  &0.767  &0.681  &0.456  &0.612  &0.545  &0.546  &0.689  &0.603 \\                     
\hline 
Variational RPCA         & Recall      &0.685 &0.540 &0.785 &0.725 &0.632 &0.667 &0.457 &0.605 &0.698 &- \\
Babacan et al. \cite{24} & Precision   &0.472 &0.712 &0.800 &0.725 &0.464 &0.659 &0.694 &0.631 &0.801 &- \\
                         & F-measure   &0.563 &0.617 &0.792 &0.725 &0.538 &0.663 &0.556 &0.618 &0.746 &0.645 \\
\hline            
MOG-RPCA                 & Recall     &0.841	&0.760	&0.902	&0.861	&0.832	&0.851	&0.654	&0.564	&0.690 &- \\
Zhao et al. \cite{1024}  & Precision  &0.682	&0.712	&0.800	&0.535	&0.662	&0.683	&0.598	&0.511	&0.597 &- \\
                         & F-measure  &0.753	&0.735	&0.848	&0.660	&0.738	&0.758	&0.625	&0.536	&0.640 &0.699 \\                       
\hline
\hline
\textbf{Approximated RPCA} &               & & & & & & & & & &           \\
\hline
GoDec                 & Recall      &0.684 &0.552 &0.761 &0.709 &0.621 &0.670 &0.465 &0.598 &0.700 &- \\
Zhou and Tao \cite{6} & Precision   &0.444 &0.682 &0.808 &0.728 &0.462 &0.636 &0.626 &0.601 &0.747 &- \\
                      & F-measure   &0.544 &0.611 &0.784 &0.718 &0.533 &0.653 &0.536 &0.600 &0.723 &0.632 \\                      
\hline
SemiSoft GoDec        & Recall       &0.666 &0.491 &0.769 &0.681 &0.636 &0.644 &0.438 &0.594 &0.683 &- \\
Zhou and Tao \cite{6} & Precision    &0.548 &0.706 &0.809 &0.694 &0.489 &0.632 &0.642 &0.629 &0.816 &- \\
                      & F-measure    &0.602 &0.583 &0.789 &0.687 &0.555 &0.638 &0.525 &0.611 &0.744 &0.637 \\ 
\hline
\end{tabular}} 
\caption{BMC dataset: Evaluation Results using the Real Videos for Evaluation Phase (Part 1).}
\label{BMC2}
\end{center}
\end{table*}

\begin{table*}
\begin{center}
\scalebox{0.65}{
\begin{tabular}{|l|l||c|c|c|c|r| c|c|c|r| c|} 
\hline
Algorithms & Measure     & \multicolumn{9}{c|}{Real Videos}  & \multicolumn{1}{c|}{Average}   \\                 
\cline{3-11} 
           &             & 001 & 002 & 003 & 004 & 005 & 006 & 007 & 008 & 009  &     \\
\hline
\hline
\textbf{Robust NMF}                    &              & & & & & & & & & &           \\
\hline

MahNMF                                 & Recall       &0.857 &0.822	&0.901 &0.848	&0.802 &0.823	&0.788 &0.536	&0.716 &- \\
Guan et al. \cite{761}                 & Precision    &0.646 &0.671	&0.739 &0.529	&0.520 &0.620	&0.672 &0.505	&0.599 &- \\
                                       & F-measure    &0.737 &0.739	&0.812 &0.651	&0.631 &0.708	&0.726 &0.520	&0.652 &0.686 \\ 
\hline
\hline
\textbf{Robust Subspace Recovery}      &              & & & & & & & & & &           \\
\hline
ROSL                                   & Recall       &0.743 &0.837	&0.912 &0.851	&0.823 &0.843	&0.778 &0.562	&0.768 &- \\
Shu et al. \cite{759-1}                & Precision    &0.865 &0.731	&0.779 &0.531	&0.512 &0.680	&0.684 &0.508	&0.852 &- \\
                                       & F-measure    &0.799 &0.781	&0.840 &0.654	&0.631 &0.753	&0.728 &0.534	&0.808 &\textbf{0.725} (3) \\ 
\hline
\hline
\textbf{Robust Subspace Tracking}      &              & & & & & & & & & &           \\
\hline
GRASTA               & Recall      & 0.719 &0.767 &0.852 &0.823 &0.533 &0.802 &0.751 &0.673 &0.730 &- \\
He et al.  \cite{27} & Precision   & 0.542 &0.845 &0.963 &0.796 &0.516 &0.711 &0.900 &0.696 &0.950 &- \\ 
                     & F-measure   & 0.618 &0.804 &0.904 &0.809 &0.524 &0.754 &0.819 &0.684 &0.826 &0.674 \\
\hline
pROST                            & Recall       &0.824 &0.672 &0.923 &0.835 &0.760 &0.797 &0.596 &0.741 &0.850 &- \\
Hage and Kleinsteuber \cite{69}  & Precision    &0.632 &0.844 &0.958 &0.884 &0.631 &0.787 &0.796 &0.770 &0.874 &- \\
                                 & F-measure    &0.715 &0.749 &0.940 &0.859 &0.689 &0.792 &0.682 &0.755 &0.862 &0.704 \\ 
\hline
\hline
\textbf{Low Rank Minimization}         &              & & & & & & & & & &           \\
\hline
DRMF                    & Recall       &0.828 &0.719 &0.934 &0.874 &0.772 &0.823 &0.617 &0.762 &0.842 &- \\
Xiong et al. \cite{54}  & Precision    &0.600 &0.856 &0.949 &0.877 &0.618 &0.801 &0.811 &0.766 &0.929 &- \\	
                        & F-measure    &0.696 &0.782 &0.941 &0.875 &0.686 &0.812 &0.701 &0.764 &0.883 &\textbf{0.714} (5) \\               
\hline 
PRMF                    & Recall       &0.830 &0.720 &0.937 &0.874 &0.790 &0.805 &0.596 &0.746 &0.839 &- \\
Wang et al. \cite{5400} & Precision    &0.665 &0.853 &0.949 &0.880 &0.627 &0.790 &0.824 &0.779 &0.962 &- \\                         
                        & F-measure    &0.738 &0.781 &0.943 &0.877 &0.699 &0.797 &0.691 &0.762 &0.897 &\textbf{0.718} (4) \\                         
\hline
\hline
\textbf{Mixture of Gaussians}         &              & & & & & & & & & &           \\
\hline
Adaptive MOG                                & Recall      & 0.849 &0.580 &0.859 &0.829 &0.754 &0.780 &0.691 &0.723 &0.828 &- \\
Shimada et al. \cite{701}                   & Precision   & 0.682 &0.546 &0.780 &0.580 &0.435 &0.636 &0.603 &0.495 &0.790 &- \\
                                            & F-measure   & 0.757 &0.562 &0.818 &0.785 &0.558 &0.702 &0.644 &0.591 &0.809 &0.680 \\ 
\hline
\end{tabular}} 
\caption{BMC dataset: Evaluation Results using the Real Videos for Evaluation Phase (Part 2).}
\label{BMC21}
\end{center}
\end{table*}

\section{Implementation and Computational Cost}
All the algorithms are implemented in Matlab and come from the LRSLibrary\protect\footnotemark[39]. The computational cost of the RPCA algorithms is mainly related to the singular value decomposition (SVD). It can be reduced significantly by using a partial SVD because only the first largest few singular values are needed. Practically, the implementation available in PROPACK\protect\footnotemark[40] are used for the IALM, LADMAP, LSADM and LADM. The SVDs and CPU time of each algorithm were computed for each sequence. Table \ref{ICC1}, Table \ref{ICC2}, Table \ref{ICC3} and Table \ref{ICC4} group the average times. The CPU times are reported in the form hh:mm:ss for images of size $144\times 176$ and with 200 frames for the training to allow easy comparison with other RPCA algorithms as the previous publications in this field present these performances on the I2R dataset \cite{203} in this data format. In this paper, the results for the NSA \cite{34} and the PSPG \cite{34-3} come from their authors. \\
\indent We can see that EALM and ADM are very computational expensive due to the fact that these algorithms compute full SVDs. On these problems of extremely low ranks, the partial SVD technique used in IALM, LADMAP, LSADM and LADM become quite effective and reduce significantly the computation time. For the SPCP, the PSPG solver is the most efficient follows by the NSA and the ASALM. The variational BRPCA is less computational expensive than the BRPCA. The GoDec algorithm is the one which requires less time computation time in the RPCA category, and then it makes large videos applications reachable in real-time. ROSL and ROSL+ reach time requirement under $15$ seconds as OR-PCA algorithms. In the RST category, GOSUS achieved less computation time than GRASTA and pROST. DRMF is the best in the category PRMF. Practically, we observed than the computation time of several algorithms increased differently in function of the image size because of their difference in term of complexity and memory requirements. Thus, the rank in term of computation time is not strictly the same at other image sizes, and several algorithms are only suitable in the case of low resolution. Finally, a full visual ranking of both matrix and tensor algorithms is provided at the LRSLibrary website.

\footnotetext[39]{{https://github.com/andrewssobral/lrslibrary}}
\footnotetext[40]{{http://soi.stanford.edu/rmunk/PROPACK/}}

\begin{table*}
\begin{center}
\scalebox{0.70}{
\begin{tabular}{|l|l|l|l|l|l|}
\hline
\multicolumn{6}{|c|}{RPCA-PCP}     \\
\hline
\scriptsize{EALM \cite{18}}       &\scriptsize{IALM \cite{18}}
&\scriptsize{ADM (LRSD) \cite{21}} &\scriptsize{LADMAP \cite{39}} 
&\scriptsize{LSADM \cite{47}}       &\scriptsize{LADM (LMaFit) \cite{23}} \\
\hline
\hline
\scriptsize{SVDs \hspace{0.5cm} CPU} &\scriptsize{SVDs \hspace{0.5cm} CPU}  
&\scriptsize{SVDs \hspace{0.5cm} CPU} &\scriptsize{SVDs \hspace{0.5cm} CPU} 
&\scriptsize{SVDs \hspace{0.5cm} CPU} &\scriptsize{SVDs \hspace{0.5cm} CPU} \\                 
\hline
\scriptsize{550 \hspace{0.5cm} 00:40:15}  &\scriptsize{38 \hspace{0.5cm}  00:03:47}  
&\scriptsize{510 \hspace{0.5cm} 00:35:20}  &\scriptsize{16 \hspace{0.5cm}  00:05:20}  
&\scriptsize{43 \hspace{0.5cm}  00:04:03}  &\scriptsize{35 \hspace{0.5cm} 00:04:55} \\								
\hline
\end{tabular}}
\caption{Time Performance of RPCA Algorithms: PCP.}
\label{ICC1}
\end{center}
\end{table*}

\begin{table*}
\begin{center}
\scalebox{0.70}{
\begin{tabular}{|l|l|l||l|l|}
\hline
\multicolumn{3}{|c|}{RPCA-SPCP} & \multicolumn{2}{c|}{OR-PCA}    \\
\hline
\scriptsize{ASALM \cite{35}} &\scriptsize{NSA \cite{34}} &\scriptsize{PSPG \cite{34-3}} 
&\scriptsize{OR-PCA \cite{1010-1}} &\scriptsize{OR-PCA with MRF \cite{1010-2}} \\
\hline
\hline
\scriptsize{SVDs \hspace{0.5cm} CPU}  &\scriptsize{SVDs \hspace{0.5cm} CPU} &\scriptsize{SVDs \hspace{0.5cm} CPU} 
&\scriptsize{SVDs \hspace{0.5cm} CPU} &\scriptsize{SVDs \hspace{0.5cm} CPU} \\
\hline
\scriptsize{94 \hspace{0.5cm} 00:15:17}  &\scriptsize{19 \hspace{0.5cm} 00:03:07} &\scriptsize{23 \hspace{0.5cm} 00:01:05} 
&\scriptsize{- \hspace{0.5cm} 00:0:12}   &\scriptsize{- \hspace{0.5cm} 00:00:14} \\
\hline
\end{tabular}}
\caption{Time Performance of RPCA Algorithms: SPCP and OR-PCA.}
\label{ICC2}
\end{center}
\end{table*}

\begin{table*}
\begin{center}
\scalebox{0.70}{
\begin{tabular}{|l|l||l|l|}
\hline
\multicolumn{2}{|c|}{Bayesian RPCA}  & \multicolumn{2}{c|}{Approximated RPCA}     \\
\hline
\scriptsize{BRPCA \cite{7}} &\scriptsize{VBRPCA \cite{24}}
&\scriptsize{GoDec \cite{6}} &\scriptsize{Semi Soft GoDec \cite{6}} \\
\hline
\hline
\scriptsize{SVDs \hspace{0.5cm} CPU} &\scriptsize{SVDs \hspace{0.5cm} CPU} 
&\scriptsize{SVDs \hspace{0.5cm} CPU} &\scriptsize{SVDs \hspace{0.5cm} CPU} \\
\hline
\scriptsize{- \hspace{0.5cm} 00:04:01}  &\scriptsize{- \hspace{0.5cm} 00:01:07}
&\scriptsize{- \hspace{0.5cm} 00:00:50}  &\scriptsize{- \hspace{0.5cm} 00:00:55} \\
\hline\end{tabular}}
\caption{Time Performance of RPCA Algorithms: Bayesian RPCA and Approximated RPCA.}
\label{ICC3}
\end{center}
\end{table*}

\begin{table*}
\begin{center}
\scalebox{0.70}{
\begin{tabular}{|l|l||l|l|l||l|l|}
\hline
\multicolumn{2}{|c|}{RSR}  & \multicolumn{3}{c|}{RST}  & \multicolumn{2}{c|}{RLRM}     \\
\hline
\scriptsize{ROSL \cite{759-1}} &\scriptsize{ROSL+ \cite{759-1}} 
&\scriptsize{GRASTA \cite{27}} &\scriptsize{pROST \cite{69}} &\scriptsize{GOSUS \cite{85}}
&\scriptsize{DRMF \cite{54}} &\scriptsize{PRMF \cite{5400}} \\
\hline
\hline
\scriptsize{SVDs \hspace{0.5cm} CPU} &\scriptsize{SVDs \hspace{0.5cm} CPU}
&\scriptsize{SVDs \hspace{0.5cm} CPU} &\scriptsize{SVDs \hspace{0.5cm} CPU}  &\scriptsize{SVDs \hspace{0.5cm} CPU} 
&\scriptsize{SVDs \hspace{0.5cm} CPU} &\scriptsize{SVDs \hspace{0.5cm} CPU} \\
\hline
\scriptsize{- \hspace{0.5cm} 00:00:15}    &\scriptsize{- \hspace{0.5cm} 00:00:06} 
&\scriptsize{- \hspace{0.5cm}  00:01:06}  &\scriptsize{- \hspace{0.5cm} 00:01:05} &\scriptsize{- \hspace{0.5cm} 00:00:23}
&\scriptsize{- \hspace{0.5cm}  00:00:17}  &\scriptsize{- \hspace{0.5cm} 00:00:36} \\
\hline
\end{tabular}}
\caption{Time Performance of RSR, RST and RLRM Algorithms.}
\label{ICC4}
\end{center}
\end{table*}

\section{Experimental Results on I2R dataset and CD.net dataset}

\subsection{Experimental Results on the I2R dataset}
The I2R dataset provided by Lin and Huang \cite{203} consists of nine video sequences, each sequence presenting dynamic backgrounds, illumination changes and bootstrapping issues. The size of the images is $176 \times 144$ pixels. This dataset consists of the following sequences: Curtain, Campus, Lobby, Shopping Mall, Airport, Restaurant, Water Surface and Fountain. A complete description of this dataset can be found in \cite{7500}. We grouped in Table \ref{I2Rdataset} all the results found over different papers \cite{1033}\cite{1532} to allow a quick comparison.  As can be seen robust designed DLAM models with spatial and temporal constraints such as RFDSA and MODSM outperforms \textbf{1)} RCPA-PCP via EALM \cite{18}, GRASTA \cite{27} and DECOLOR \cite{25}, and \textbf{2)} the BGS algorithms MOG \cite{700} and SOBS \cite{210-1}.

\begin{table*}
\begin{center}
\scalebox{0.60}{
\begin{tabular}{|l|l|c|c|c|c|r| c|c|c|r| c|} 
\hline
Algorithms      & \multicolumn{9}{c|}{Videos}  & \multicolumn{1}{c|}{Average}   \\                 
\hline  
& Water Surface & Curtain & Fountain & Restaurant & Shopping Mall & Airport & Lobby & Bootstrapping & Campus  &  \\          
\hline
\hline
\textbf{Robust PCA}                      & & & & & & & & & &           \\
EALM                                     & 0.4137 &0.6193 &0.5679 &0.5917 &0.7234 &0.6989 &0.6728 &0.6582 &0.3406 &0.587 \\
Lin et al. \cite{18}                     & & & & & & & & & &           \\
\hline 
RFDSA                                     &0.8796 &0.8976 &0.7544 &0.6673 &\textbf{0.7407} &\textbf{0.8029} &0.6353 &0.6841 &0.6779 &0.7489 \\
Guo et al. \cite{1033}                    & & & & & & & & & &           \\
\hline 
MODSM                                       &\textbf{0.9404} &\textbf{0.9098} &0.8205 &\textbf{0.6859} &0.7362 &0.5762 
																						&\textbf{0.7553} &\textbf{0.728} &0.7876 &\textbf{0.7711} \\
Pang et al. \cite{1532}                     & & & & & & & & & &           \\
\hline
\hline
\textbf{Robust Subspace Tracking}                             & & & & & & & & & &           \\
GRASTA                                                        &0.7310 &0.6591 &0.3786 &0.5817 &0.7142 &0.5550 &0.4697 &0.6146 &0.2504 &0.5505 \\
He et al.  \cite{27}                                          & & & & & & & & & &           \\
\hline
\hline
\textbf{Low Rank Minimization}                                & & & & & & & & & &          \\
DECOLOR 																				 				   	 &0.8866 &0.8255 &\textbf{0.8598} &0.6424 &0.6525 &0.6149 
																														 &0.6994 &0.5869 &0.8096 &0.7308 \\
Zhou et al. \cite{25}                                         & & & & & & & & & &           \\
\hline
\hline
\textbf{BGS Algorithms}                                       & & & & & & & & & &           \\
MOG   																          		&0.7948  &0.7580  &0.6854  &0.3335  &0.5363  &0.6519  &0.1388  &0.38380  &0.0757  &0.4842 \\
Stauffer and Grimson \cite{700}									    & & & & & & & & & &           \\
\hline 
SOBS                                                &0.8247  &0.8178  &0.6554  &0.5943  &0.6677  &0.6489  &0.5770  &0.6019  &0.6960  &0.6760 \\
Maddalena and Petrosino \cite{210-1}                & & & & & & & & & &           \\  
\hline\end{tabular}} 
\caption{I2Rdataset: Evaluation results on the 9 real videos.}
\label{I2Rdataset}
\end{center}
\end{table*}

\subsection{Experimental Results on the ChangeDetection.net dataset}
The ChangeDetection.net dataset \cite{202} is a realistic, large-scale video dataset for benchmarking background subtraction methods. It consists of nearly 90,000 frames in 31 video sequences representing 6 categories selected to cover a wide range of challenges in 2 modalities (color and thermal IR). A complete description of this dataset can be found in \cite{7500}. We grouped in Table \ref{CDDataset} all the results found over different papers  \cite{1010-10} to allow a quick comparison.  As can be seen, SRPCA with spatial and temporal constraints via graph regularization outperforms  \textbf{1)} the incremental algorithms GoDec and GRASTA, \textbf{2)} the models with spatial constraints DECOLOR \cite{25}, TVRPCA \cite{1900}, MAMC \cite{2002} and SLMC \cite{1010-10}, and  \textbf{3)} the BGS algorithm SOBS \cite{210-1}.

\begin{table*}
\begin{center}
\scalebox{0.60}{
\begin{tabular}{|l|l|c|c|c|c|r| c|c|c|r| c|} 
\hline
Algorithms    & \multicolumn{10}{c|}{Videos}    \\                 
\hline  
              & Lobby & Boats & Canoe & Fall & Fountain1 & Fountain2 & Overpass & Water Surface & Fountain  & Waving Trees \\          
\hline
\hline
\textbf{Robust PCA}                         & & & & & & & & & &           \\
TVRPCA                                      &0.06 &0.52 &0.70 &0.48 &0.12 &0.72 &0.77 &0.84 &0.73 &0.67 \\
Cao et al. \cite{1900}                      & & & & & & & & & &           \\
\hline 
SRPCA                                       &\textbf{0.83} &\textbf{0.82} &\textbf{0.94} &\textbf{0.92} &\textbf{0.82} 																					                                &0.90 	&\textbf{0.92} &\textbf{0.93} &0.76 &\textbf{0.97} \\
Javed et al. \cite{1010-10}                  & & & & & & & & & &           \\
\hline
\hline
\textbf{Approximated RPCA}                   & & & & & & & & & &           \\
GoDec                                        &0.14 &0.18 &0.42 &0.6 &0.11 &0.38 &0.66 &0.72 &0.40 &0.68 \\
Zhou and Tao \cite{6}                        & & & & & & & & & &           \\
\hline
\hline
\textbf{Robust Matrix Completion}                             & & & & & & & & & &           \\
MAMC                                                          &0.27 &0.78 &0.81 &0.75 &0.51 &\textbf{0.96} &0.82 &0.76 &\textbf{0.77} &0.83 \\
Yang et al. \cite{2002}                                       & & & & & & & & & &           \\
\hline 
SLMC                                                          &0.77 &0.46 &0.33 &0.50 &0.52 &0.64 &0.64 &0.66 &0.51 &0.79 \\
Javed et al.  \cite{1010-10}                                  & & & & & & & & & &           \\
\hline
\textbf{Robust Subspace Tracking}                             & & & & & & & & & &           \\
GRASTA                                                       &0.80 &0.66 &0.46 &0.42 &0.72 &0.75 &0.87 &0.85 &0.60 &0.74 \\
He et al.  \cite{27}                                          & & & & & & & & & &           \\
\hline
\hline
\textbf{Low Rank Minimization}                                & & & & & & & & & &          \\
DECOLOR 																				 				   	 &0.29 &0.19 &0.73 &0.61 &0.02 &0.64 &0.81 &0.83 &0.33 &0.88  \\
Zhou et al. \cite{25}                                         & & & & & & & & & &           \\
\hline
\hline
\textbf{BGS Algorithms}                                       & & & & & & & & & &           \\
SOBS                                                          &0.46 &0.07 &0.64 &0.61 &0.32 &0.88 &0.68 &0.75 &0.68 &0.73 \\
Maddalena and Petrosino \cite{210-1}                         & & & & & & & & & &           \\  
\hline\end{tabular}} 
\caption{ChangeDetection.net 2014 dataset: Evaluation results on the dynamic scenarios.}
\label{CDDataset}
\end{center}
\end{table*}

\section{Conclusion}
\label{sec:Conclusion}
In this paper, we have firstly presented a full review of recent advances on problem formulations based on the decomposition into low-rank plus additive matrices which are robust principal component analysis, robust non-negative matrix factorization, robust matrix completion, robust subspace recovery, robust subspace tracking and robust low-rank minimization. Thus, we proposed a unified view of the decomposition into low-rank plus additive matrices that we called DLAM. We evaluated its adequacy to the application of background/foreground separation by investigating how these
methods are solved and if incremental algorithms and real-time implementations can be achieved. Finally, experimental results on the
Background Models Challenge (BMC) dataset show the comparative performance of these recent methods. \\

\indent In summary, this review for a comparative evaluation of robust subspace learning via decomposition into low-rank plus additive algorithms highlights the following points:
\begin{itemize}
\item Decomposition into low-rank plus additive matrices offers a suitable framework for background/foreground separation. Indeed, DLAM models in their fundamental version outperform in term of detection state-of-the-art models such as the MOG \cite{700}\cite{701}\cite{702} and the KDE \cite{710}\cite{715}. Furthermore, robustness against illumination changes and dynamic backgrounds can be improved if spatial and temporal constraints are taken into account in the optimization problem with structured norms \cite{1037}\cite{1901}, structured group sparsity norm \cite{1542}, dynamic tree structured sparsity \cite{1524} or MRF \cite{1010-2}. Moreover, robust methods specifically designed for background/foreground separation like in the works of Javed et al. \cite{1010-2}\cite{1010-3}\cite{1010-4}\cite{1010-5}\cite{1010-7}\cite{1010-10}\cite{1010-11}, Sobral et al. \cite{1711}{1712} and Mansour et al. \cite{2000}\cite{2001}\cite{1075} offer very impressive performance that are largely comparable to SubSENSE \cite{740} and better than SOBS \cite{210-1}, ViBe \cite{720} and PBAS \cite{730}\cite{735}. \\\
\item The main drawbacks of the DLAM models is that their original version used batch algorithms that often need too expensive computation time to reach real-time requirements. Thus, many effort have been done to reach real-time performance and to develop incremental algorithms like in the works of Rodriguez et al. \cite{26-1}\cite{26-3}\cite{26-4}\cite{26-5}, and the works of Vaswani et al. \cite{14}\cite{17-21}\cite{17-33}. \\
\item As images are stored in vectors which are often exploited as is, DLAM models in their original version loss the spatial and temporal constraints. Thus, it is more suitable to use \textbf{(1)} Markov Random Fields \cite{1010-2}, \textbf{(2)} structured norms aiming to preserve the spatial structures of images while being insensitive to outliers and missing data \cite{85}\cite{1037}\cite{1132}\cite{1712}, or \textbf{(3)} a formulation in the two-dimensional case rather than via image to vector conversion, which enables the preservation of the image spatial information with reduced computational time. Several basic two-dimensional subspace formulations can be found in literature such as two-dimensional PCA (2dPCA) \cite{15090}, two-dimensional SVD (2dSVD) \cite{15091}, two-dimensional LDA (2dLDA) \cite{15092}, two-directional two-dimensional PCA ($(2d)^2$PCA) \cite{15093}, and Generalized Low Rank Approximations of Matrices (GLRAM) \cite{15094}\cite{15095}. Robust formulations can be found such as robust two dimensions (R2DRPCA) \cite{1005}\cite{1132} and Robust GLRAM (RGLRAM) \cite{1509}. \\ 
\end{itemize}
Future research may concern free SVD or less computational SVD algorithms such as LMSVD \cite{2301} for batch algorithms, and
DLAM models which would be both incremental and real-time to reach the performance of the state-of-the-art algorithms \cite{6000}\cite{6010} in term of computation time and memory requirements. Finally, DLAM models show a suitable potential for background modeling and foreground detection in video surveillance \cite{45}\cite{45-1}. Furthermore, DLAM can been extended to the measurement domain, rather than the pixel domain, for use in conjunction with compressive sensing. Moreover, other research may concern the extension of DLAM in tensor-wise way to exploit fully spatial and temporal constraints \cite{753}\cite{754}\cite{755}\cite{756-1}\cite{757}\cite{1340}. The interest of the tensor approach over the matrix approach is investigated by Anandkumar et al. \cite{7050}. Furthermore, efficient tensor incremental algorithms have been recently developed for background/foreground separation \cite{7000}\cite{7010}\cite{7020}\cite{7030}\cite{7040} and background initialization \cite{1010-12}.

\section{Acknowledgment}
The authors would like to thank the following researchers: Zhouchen Lin (Visual Computing Group, Microsoft Research Asia) who has kindly provided the solver LADMAP \cite{39} and the $l_1$-filtering \cite{38}, Shiqian Ma (Institute for Mathematics and Its Applications, Univ. of Minnesota, USA) who has kindly provided the solver LSADM \cite{46}, Congguo Tang (Dept. of Electical System Engineering, Washington Univ., USA) who has kindly provided the solver of RPCA-BPCP \cite{41}.

\bibliographystyle{plain}
\bibliography{rpca,sparse}
\end{document}